\documentclass[oneside,11pt]{article}

\usepackage[small,it]{caption}
\usepackage{url,mathrsfs,algorithm}
\usepackage{amsmath,wrapfig,color}
\usepackage{times}
\usepackage{amsfonts}
\usepackage{graphicx}
\usepackage{subfigure}
\newtheorem{theorem}{Theorem}
\newtheorem{lemma}{Lemma}

\begin{document}

\title{\huge Exact  Sparse Recovery with L0 Projections}

\author{ \textbf{Ping Li} \\
         Department of Statistical Science\\
       Cornell University\\
         Ithaca, NY 14853, USA\\
      \texttt{pingli@cornell.edu}
       \and
 \textbf{Cun-Hui Zhang} \\
         Department of Statistics and Biostatistics\\
       Rutgers University\\
         New Brunswick, NJ 08901, USA\\
         \texttt{cunhui@stat.rutgers.edu}
        }
\date{}
\maketitle

\begin{abstract}
Many\footnote{The results were presented in several seminars in 2012 and in the Third Conference
of Tsinghua Sanya International Mathematics Forum (Jan. 2013) with no published proceedings.}  applications concern sparse signals, for example, detecting anomalies from the differences between consecutive images taken by surveillance cameras. This paper focuses on the problem of  recovering a  $K$-sparse signal $\mathbf{x}\in\mathbb{R}^{1\times N}$, i.e., $K\ll N$ and $\sum_{i=1}^{N} 1\{x_i\neq 0\} = K$. In the  mainstream framework of compressed sensing (CS), the vector $\mathbf{x}$ is recovered from $M$ non-adaptive linear measurements $\mathbf{y} = \mathbf{xS}\in\mathbb{R}^{1\times M}$, where $\mathbf{S}\in\mathbb{R}^{N\times M}$ is typically a Gaussian (or Gaussian-like) design matrix, through some optimization procedure such as linear programming (LP).

In our proposed method, the design matrix $\mathbf{S}$ is generated from an $\alpha$-stable distribution with $\alpha\approx 0$. Our decoding algorithm  mainly requires  one linear scan of the coordinates, followed by a few iterations on a small number of coordinates which are ``undetermined'' in the previous iteration. Our practical algorithm consists of two estimators. In the first iteration, the {\em (absolute) minimum estimator} is able to  filter out a majority of the zero coordinates. The {\em gap estimator}, which is applied in each iteration, can accurately recover the magnitudes of the nonzero coordinates.  Comparisons with two strong baselines, linear programming (LP) and orthogonal matching pursuit (OMP), demonstrate that our algorithm can be significantly faster in  decoding speed and more accurate in recovery quality, for the task of exact spare recovery. Our procedure is robust against measurement noise. Even when there are no sufficient measurements,  our algorithm can still reliably recover a significant portion of the nonzero coordinates.

To provide the intuition for  understanding our method, we also analyze the procedure by assuming an idealistic setting. Interestingly, when $K=2$, the ``idealized'' algorithm achieves exact recovery with merely $3$ measurements, regardless of $N$. For general $K$, the required sample size  of the ``idealized'' algorithm is about $5K$. The gap estimator is a practical surrogate for the ``idealized'' algorithm.

\end{abstract}

\section{Introduction }

The goal of {\em Compressed Sensing (CS)}~\cite{Article:Donoho_CS_JIT06,Article:Candes_Robust_JIT06} is to recover a sparse signal $\mathbf{x}\in\mathbb{R}^{1\times N}$ from a small number of non-adaptive linear measurements $\mathbf{y} = \mathbf{xS}$, (typically) by convex optimization (e.g., linear programming). Here, $\mathbf{y}\in\mathbb{R}^{1\times M}$ is the vector of measurements and $\mathbf{S}\in\mathbb{R}^{N\times M}$ is the design matrix (also called the measurement matrix). In classical settings, entries of $\mathbf{S}$ are i.i.d. samples from the Gaussian distribution $N(0,1)$, or a Gaussian-like distribution (e.g., a distribution with finite variance).

In this paper, we  sample  $\mathbf{S}$ from a heavy-tailed distribution which only has the $\lambda$-th moment with $\lambda<\alpha$ and we will choose $\alpha\approx 0$. Strikingly, using such a design matrix  turns out to result in a simple and powerful solution to the problem of exact $K$-sparse recovery, i.e., $\sum_{i=1}^N 1\{x_i\neq 0\} = K$.

\subsection{Compressed Sensing}

Sparse recovery (compressed sensing), which has been an  active area of research, can be naturally suitable for: (i) the ``single pixel camera'' type of applications; and (ii) the ''data streams'' type of applications. The idea of compressed sensing may be traced back to  many prior papers such as~\cite{Article:Stark_89,Article:Huo_JIT01,Article:Cormode_05}.

\begin{figure}[h!]
\begin{center}
\mbox{\includegraphics[width=2.4in]{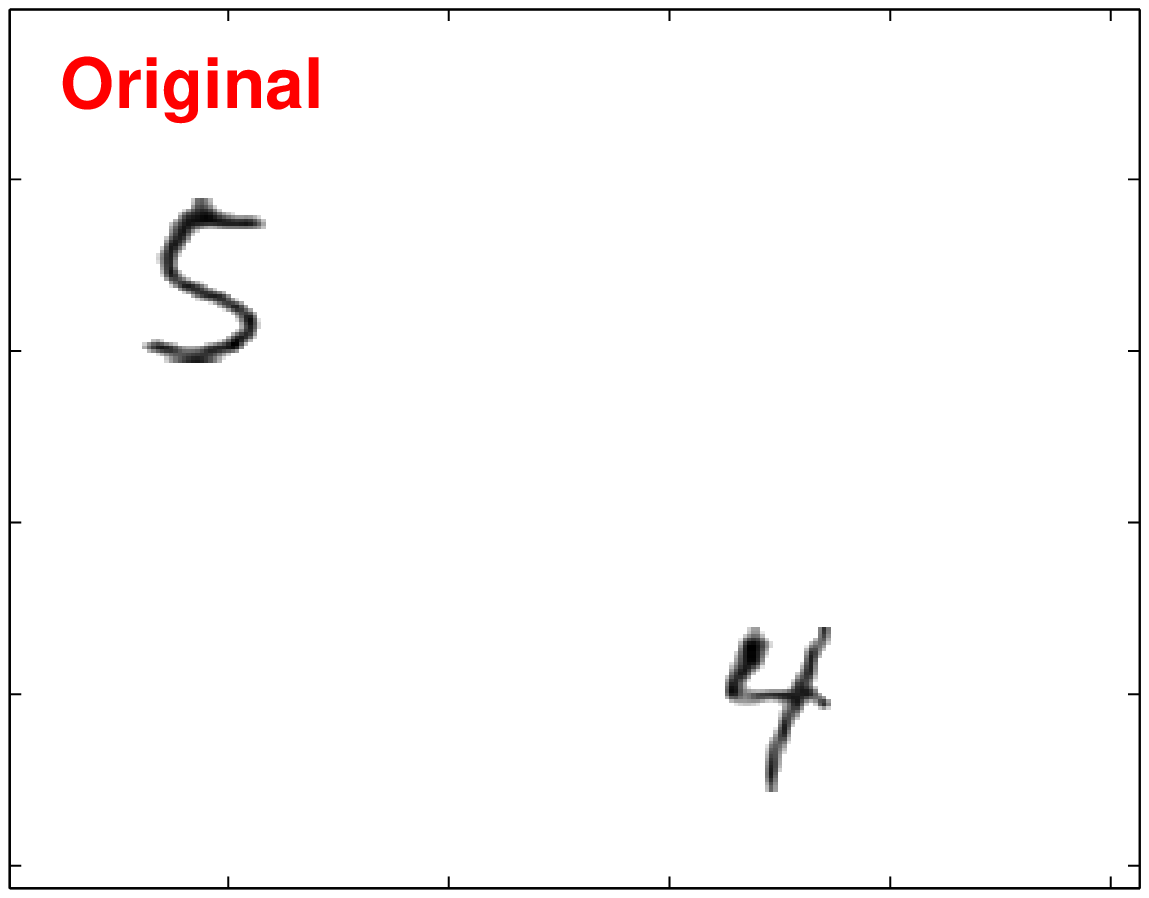}\hspace{-0.25in}
\includegraphics[width=2.4in]{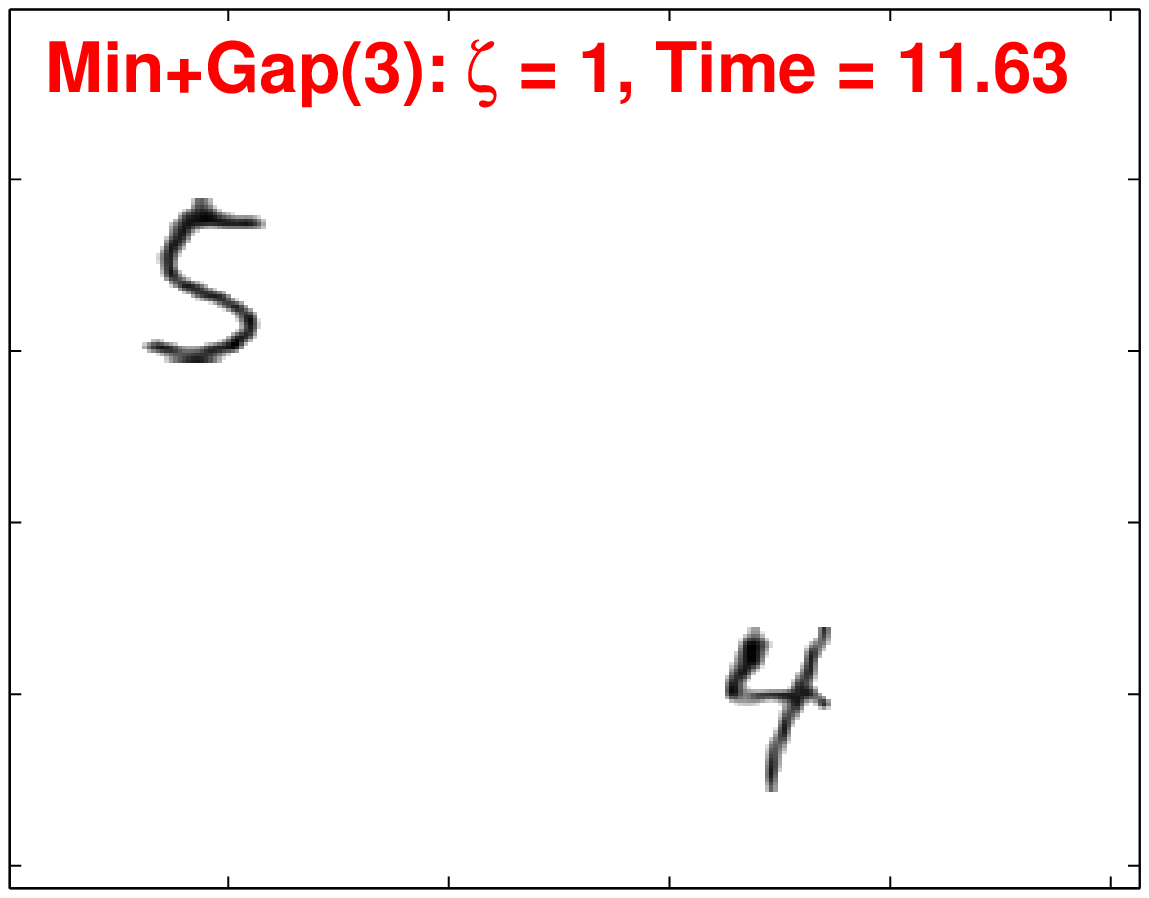}\hspace{-0.25in}
\includegraphics[width=2.4in]{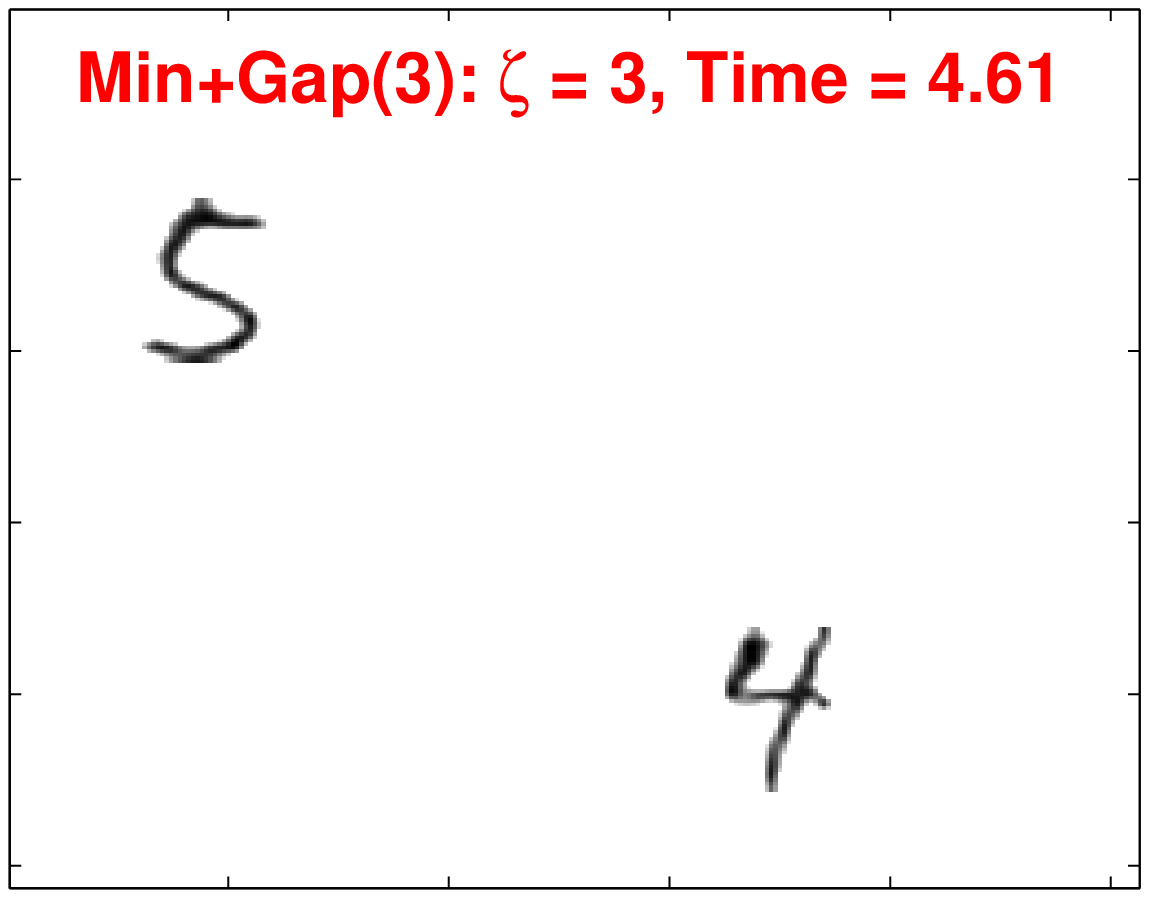}
}
\mbox{
\includegraphics[width=2.4in]{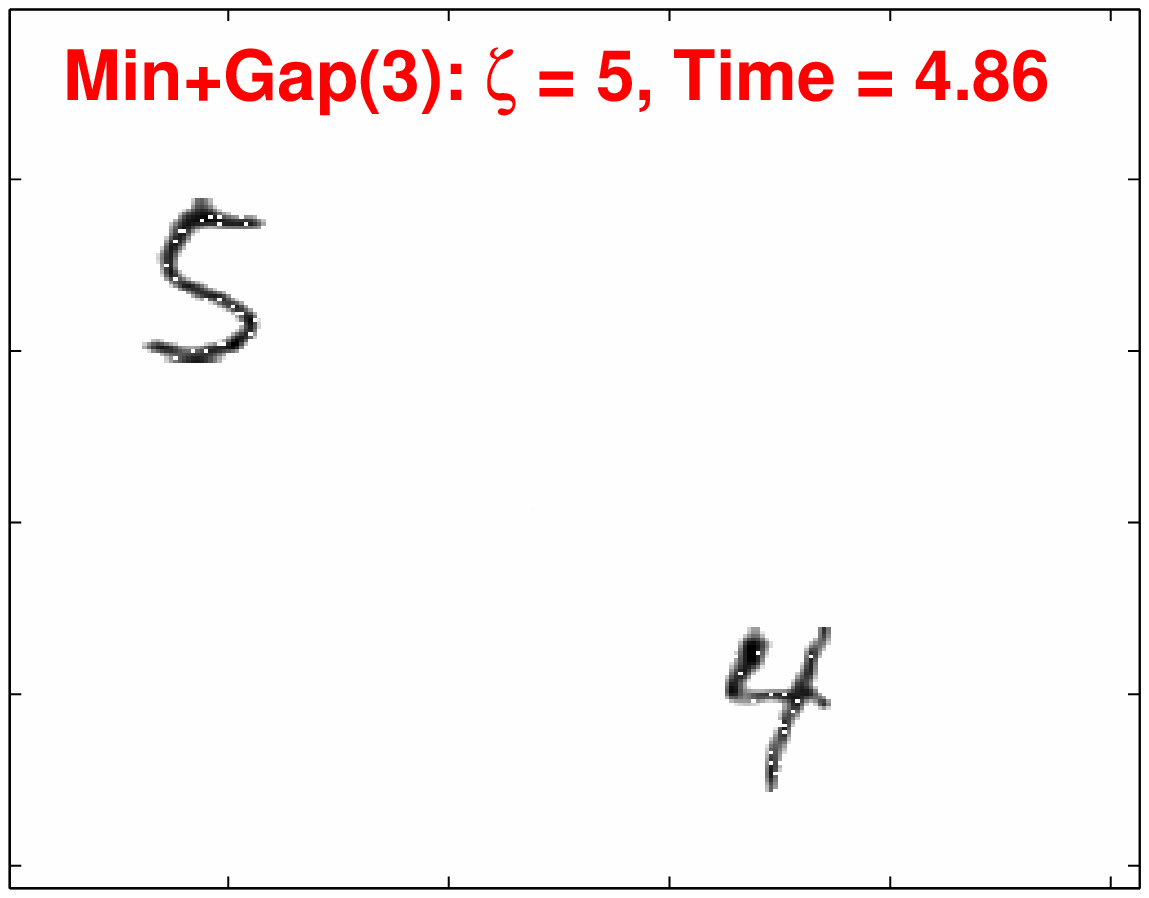}\hspace{-0.25in}
\includegraphics[width=2.4in]{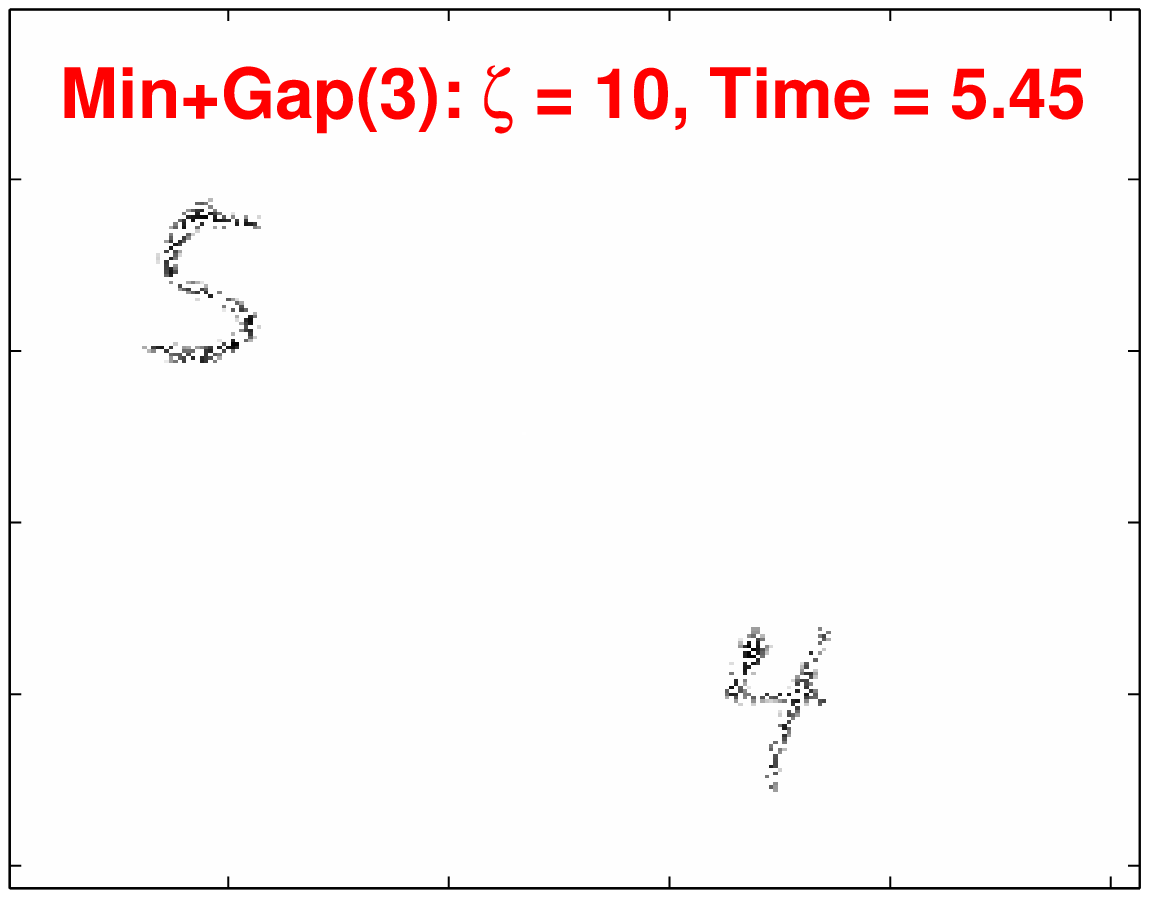}\hspace{-0.25in}
\includegraphics[width=2.4in]{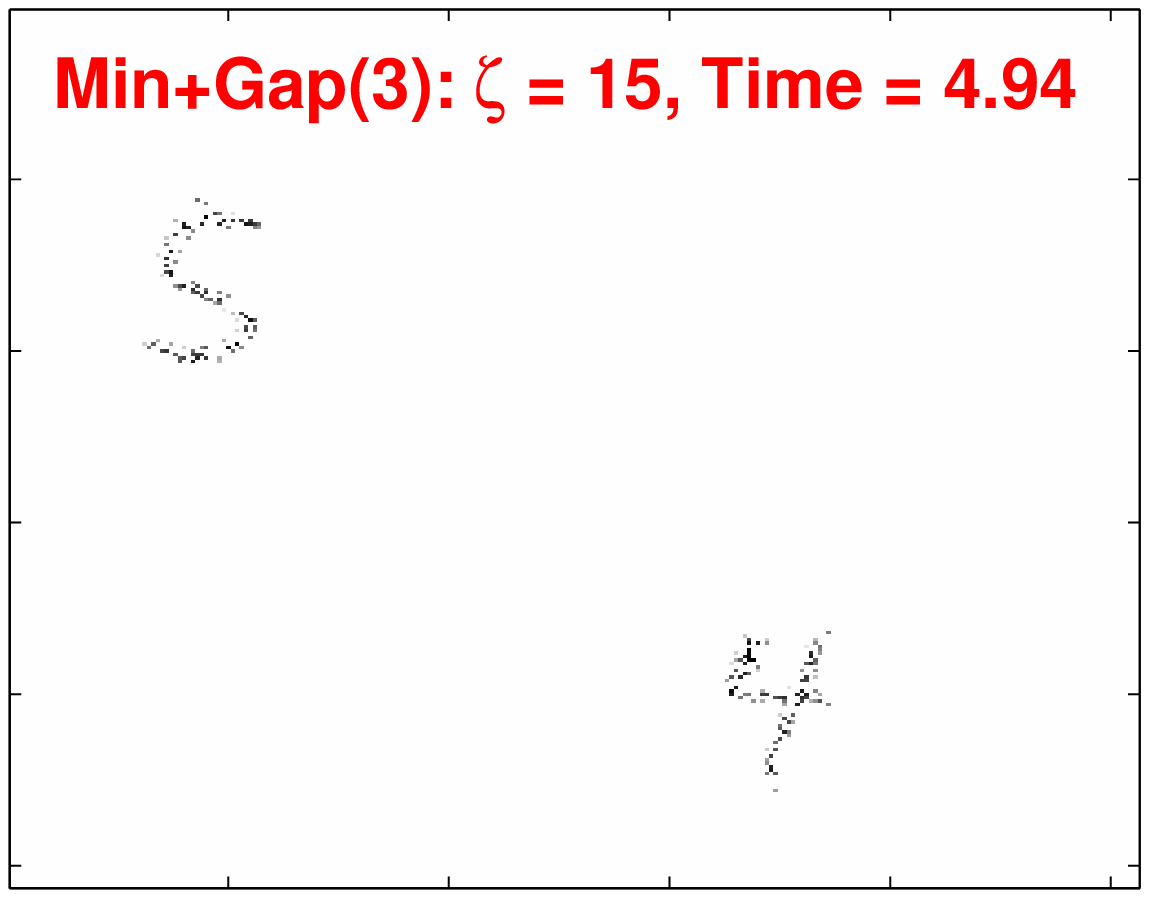}
}
\end{center}
\vspace{-0.2in}
\caption{The task is to reconstruct a $256\times 256$ image (i.e., $N=65536$) with  $K=852$ nonzero pixels, using our proposed method with $M = K\log( (N-K)/0.01)/\zeta)$ measurements, for $\zeta = 1, 3, 5, 10, 15$. Our method, which is named ``Min+Gap(3)'' will be explained later in the paper. For $\zeta=1$, our method is able to exactly reconstruct the image, using 11.63 seconds. Strikingly,  even with $\zeta = 15$ (i.e., $M=891$ measurements only), the reconstructed  image by our method could still be  informative. Note that, although natural images are often not as sparse as this example, sparse images are possible in important applications, e.g., the differences between consecutive image/video frames taken by surveillance cameras. }\label{fig_Zip54}
\end{figure}

It has been realized (and implemented by hardware)    that collecting a linear combination of a sparse vector, i.e., $\mathbf{y =xS}$, can be more  advantageous than sampling the vector itself. This is the foundation of the  ``single pixel camera'' proposal. See the site \url{https://sites.google.com/site/igorcarron2/}\\\url{compressedsensinghardware} for a list of  implementations of single-pixel-camera type of applications.  Figure~\ref{fig_Zip54} provides an illustrative example.

Natural images are in general not as sparse as the example in Figure~\ref{fig_Zip54}. We nevertheless expect that in many practical scenarios, the sparsity assumption can  be reasonable. For example, the differences between consecutive image/video frames taken by surveillance cameras are usually very sparse because the background remains still. In general, anomaly detection problems are often very sparse.

Another line of applications concerns data streams, which can be conceptually viewed as a long sparse dynamic vector with entries rapidly varying over time. Because of the dynamic nature, there may not be an easy way of  knowing where   the nonzero coordinates are, since the history of streaming is usually not stored. Perhaps surprisingly, many problems can be  formulated as sparse data streams. For example, video data are naturally streaming. A common task in databases is to find the ``heavy-hitters''~\cite{Article:Muthukrishnan_05}, e.g., finding  which product items have the highest total sales. In networks~\cite{Proc:Zhao_IMC07}, this is often referred to as the ``elephant detection'' problem; see some recent papers on compressed sensing for network applications~\cite{Proc:Lin_Globecom12,Proc:Wang_SIGCOMM12,Proc:Wang_Infocom12}.

 For data stream applications, entries of the signals $\mathbf{x}$ are (rapidly) updated over time (by addition and deletion). At a time $t$, the $i_t$-th entry is updated by $I_t$, i.e., $x_{i_t} \rightarrow x_{i_t} +I_t$. This is often referred to as the {\em turnstile model}~\cite{Article:Muthukrishnan_05}.   As the projection operation is linear, i.e., $\mathbf{y  = x S}$, we can (re)generate corresponding entries of $\mathbf{S}$ on-demand whenever one entry of $\mathbf{x}$ is altered, to update all  entries of the measurement vector $\mathbf{y}$. The use of stable random projections for estimating the $\alpha$-th frequency moment $\sum_{i=1}^N |x_i|^\alpha$ (instead of the individual terms $x_i$) was studied in~\cite{Article:Indyk_JACM06}. \cite{Proc:Li_SODA08} proposed  the use of geometric mean estimator for stable random projections, for estimating $\sum_{i=1}^N |x_i|^\alpha$ as well as the harmonic mean estimator  for estimating $\sum_{i=1}^N |x_i|^\alpha$  when $\alpha\approx 0$.  When the streaming model is not the turnstile model (for example, the update mechanism may be nonlinear), \cite{Proc:Li_Church_Hastie_NIPS08} developed the method named {\em conditional random sampling (CRS)}, which has been used in network applications~\cite{Proc:Zhao_NPC11}. At this point, our work  focuses on the turnstile data stream model.

 In this paper, our goal is to use stable random projections to recover the individual entries $x_i$'s, not just the summary statistics such as the frequency moments. We first provide a review of $\alpha$-stable distributions.

\subsection{Review of  $\alpha$-Stable Distribution}

A random variable $Z$ follows an $\alpha$-stable distribution with unit scale, denoted by $S(\alpha,1)$, if its characteristic function can be written as~\cite{Book:Samorodnitsky_94}
\begin{align}
E\left(e^{\sqrt{-1} Z t}\right) = e^{-|t|^\alpha},\hspace{0.3in} 0<\alpha\leq 2
\end{align}
When $\alpha=2$, $S(2,1)$ is equivalent to  the normal distribution with variance 2, i.e.,  $N(0,2)$. When $\alpha=1$, $S(1,1)$ is the standard Cauchy distribution centered at zero with unit scale.

To sample from $S(\alpha,1)$, we use the  CMS procedure~\cite{Article:Chambers_JASA76}. That is, we sample independent exponential  $w\sim exp(1)$ and uniform  $u\sim unif(-\pi/2,\pi/2)$  variables, and then compute $Z\sim S(\alpha,1)$ by
\begin{align}\label{eqn_stable_sample}
Z = \frac{\sin(\alpha u)}{(\cos u)^{1/\alpha}}
\Big[\frac{\cos(u-\alpha u)}{w}\Big]^{(1-\alpha)/\alpha}\sim S(\alpha,1)
\end{align}
If $S_1, S_2 \sim S(\alpha,1)$ i.i.d., then for any constants $C_1, C_2$, we have $C_1S_1 + C_2S_2= S\times (|C_1|^\alpha+|C_2|^\alpha)^{1/\alpha}$, where $S\sim S(\alpha,1)$. More generally, $\sum_{i=1}^N x_i S_i = S\times (\sum_{i=1}^N |x_i|^\alpha)^{1/\alpha}$.

In this paper, we propose using $\alpha\approx 0$. In our numerical experiments with Matlab, the value of  $\alpha$ is taken to be $0.03$ and  no special data storage structure is needed. While the precise theoretical  analysis   based on a particular choice of (small) $\alpha$ is technically nontrivial, our algorithm  is intuitive, as illustrated by a simpler analysis of an ``idealized'' algorithm using the limit of $\alpha$-stable distributions as $\alpha\rightarrow0$.

\subsection{The Proposed Practical Recovery Algorithm}

\begin{algorithm}{\small
\textbf{Input:} $K$-sparse signal $\mathbf{x}\in\mathbb{R}^{1\times N}$, threshold $\epsilon>0$ (e.g., $10^{-5}$), design matrix $\mathbf{S}\in\mathbb{R}^{N\times M}$ sampled from $S(\alpha,1)$ with $\alpha\approx 0$ (e.g., 0.03).  $\mathbf{S}$ can be generated on-demand when the data arrive at a streaming fashion. \\

\textbf{Output:} The recovered signal, denoted by $\hat{x}_i$, $i = 1$ to $N$. \\

\textbf{Linear measurements:} $ \mathbf{y} = \mathbf{xS}$, which can be conducted incrementally if entries of $\mathbf{x}$ arrive in a streaming fashion.   \\

\textbf{Detection:} For $i=1$ to $N$, compute $\hat{x}_{i,min} = y_{t}/s_{it}$, where $t = \text{argmin}_j |y_j/s_{ij}|$. If $|\hat{x}_{i,min}| \leq \epsilon$, set $\hat{x}_{i}=0$. \\

\textbf{Estimation:} If $|\hat{x}_{i,min}|>\epsilon$, compute the gaps for the sorted observations $y_{j}/s_{ij}$ and estimate $x_i$ using the gap estimator $\hat{x}_{i,gap}$. Let $\hat{x}_i = \hat{x}_{i,gap}$. See the details below. \\

\textbf{Iterations:}   If $|\hat{x}_{i,min}|>\epsilon$ and the minimum gap length  $>\epsilon$, we call this $i$ an ``undetermined'' coordinate and set $\hat{x}_{i} = 0$. Compute the residuals: $\mathbf{r} = \mathbf{y}- \mathbf{\hat{x}S}$, and apply the gap estimator using the residual  $\mathbf{r}$, only  on the set of ``undetermined'' coordinates. Repeat the iterations a number of times (e.g., 2 to 4) until no changes are observed. This iteration step  is particularly helpful when $M\ll M_0 =  K\log((N-K)/\delta)$  where $\delta = 0.05$ or 0.01.

}\caption{The proposed recovery algorithm. }
\label{alg_recovery}
\end{algorithm}

We assume $\mathbf{x}\in\mathbb{R}^{1\times N}$ is $K$-sparse and we do not know where the nonzero coordinates are.  We obtain $M$ linear measurements $\mathbf{y}=\mathbf{xS}\in\mathbb{R}^{1\times M}$, where entries of $\mathbf{S}\in\mathbb{R}^{N\times M}$, denoted by $s_{ij}$, are i.i.d. samples from $S(\alpha,1)$ with a small $\alpha$ (e.g., 0.03). That is,  each measurement is $y_j = \sum_{i=1}^N x_i s_{ij}$.  Our algorithm, which consists of two estimators, utilizes the ratio statistics $z_{i,j} = y_j/s_{ij}$, $j = 1, 2, ..., M$, to recover $x_i$.

The \textbf{absolute minimum estimator} is defined as
\begin{align}
\hat{x}_{i,min} = z_{i,t}, \ \ \text{ where } t = \underset{1\leq j\leq M}{\text{argmin}}\  |z_{i,j}|,\hspace{0.2in} z_{ij} = \frac{y_j}{s_{ij}}
\end{align}
which is  effective for detecting whether any $x_i = 0$. In fact we prove  that essentially $M_0 = K\log \left((N-K)/\delta\right)$ measurements are sufficient for detecting all zeros with at least probability $1-\delta$. The actual required number of measurements will be significantly lower than $M_0$ if we use the minimum algorithm together with the gap estimator and the iterative process.

When $|\hat{x}_{i,min}|>\epsilon$, in order to estimate the magnitude of $x_i$ , we resort to the \textbf{gap estimator} defined as follows. First we sort $z_{i,j}$'s and write them as order statistics: $z_{i,(1)} \leq z_{i,(2)} \leq  ... \leq z_{i,(M)}$.
Then we compute the gaps: $g_j = z_{i,(j+1)} - z_{i,(j)},\  1\leq j\leq M-1$. The gap estimator is simply
\begin{align}
\hat{x}_{i,gap} = \frac{1}{2}\left\{z_{i,(j_i)} +z_{i,(j_i+1)}\right\}, \hspace{0.3in}
j_i = \underset{1\leq j\leq M-1}{\text{argmin}}\ g_j = z_{i,(j+1)} - z_{i,(j)}
\end{align}
We  have also derived  theoretical error probability bound for $\mathbf{Pr}\left(|\hat{x}_{i,gap} - x_i|>\epsilon\right)$. When $M<M_0$, we discover that it is  better to apply the gap estimator a number of times, each time using the residual measurements only on the ``undetermined'' coordinates; see Alg.~\ref{alg_recovery}. The iteration procedure will become intuitive after we explain the ``idealized algorithm'' in Sec.~\ref{sec_intuition} and Sec.~\ref{sec_ideal}. \\

Note that our algorithm does not directly utilize the information of $K$. In fact, for small $\alpha$, the quantity $\theta^\alpha$ (which is very close to $K$) can be reliably estimated by the following harmonic mean estimator~\cite{Proc:Li_SODA08}:
\begin{align}
\hat{\theta}^\alpha = \frac{-\frac{2}{\pi}\Gamma(-\alpha)\sin\frac{\pi}{2}\alpha}{\sum_{j=1}^M\frac{1}{|y_j|^\alpha}}
\left(M-\left(\frac{-\pi\Gamma(-2\alpha)\sin(\pi\alpha)}{\left[\Gamma(-\alpha)\sin\frac{\pi}{2}\alpha\right]^2}-1\right)\right)
\end{align}

\section{Intuition}\label{sec_intuition}

While a precise analysis of our proposed method is technical, our procedure is   intuitive from the ratio of two independent $\alpha$-stable random variables, in the limit  when $\alpha\rightarrow0$.

Recall that, for each coordinate $i$, our observations are $(y_j, s_{ij})$, $j = 1$ to $M$. Naturally our first attempt was to use the joint likelihood of $(y_j, s_{ij})$. However, in our proposed method, the observations are utilized only through the ratio statistics $y_j/s_{ij}$. We first explain why.

\subsection{Why Using the Ratio Statistics $y_j/s_{ij}$?}

For convenience, we first define
\begin{align}
&\theta = \left(\sum_{i=1}^N |x_i|^\alpha\right)^{1/\alpha},\hspace{0.2in}
\theta_i = \left(\theta^\alpha- |x_i|^\alpha\right)^{1/\alpha}
\end{align}

Denote the density function of $S(\alpha,1)$ by $f_S(s)$. By a conditional probability argument, the joint density of $(y_j,s_{ij})$ can be shown to be $\frac{1}{\theta_i}f_S(s_{ij})f_S\left(\frac{y_j-x_is_{ij}}{\theta_i}\right)$, from which we can derive the joint log-likelihood of $(y_j, s_{ij})$, $j =1$ to $M$, as
\begin{align}
l(x_i,\theta_i) = \sum_{j=1}^M \log f_S(s_{ij}) + \sum_{j=1}^M \log f_S\left(\frac{y_j-x_is_{ij}}{\theta_i}\right) - M\log\theta_i
\end{align}
Here we can treat $x_i$ and $\theta_i$ as the parameters to be estimated. Closed-form density functions of $f_S$ are in general not available (except for $\alpha=2$ and $\alpha =1$). Interestingly, when $\alpha\approx 0$, we can obtain a  convenient approximation. Recall, for two independent variables $w\sim exp(1)$ and  $u\sim unif(-\pi/2,\pi/2)$, we have
\begin{align}\notag
Z = \frac{\sin(\alpha u)}{(\cos u)^{1/\alpha}}
\Big[\frac{\cos(u-\alpha u)}{w}\Big]^{(1-\alpha)/\alpha}\sim S(\alpha,1),
\end{align}
Thus, it is intuitive that $1/|Z|^\alpha$ is approximately $w\sim exp(1)$ when $\alpha\approx0$. As rigorously shown by \cite{Article:Cressie_75}, $1/|Z|^\alpha\rightarrow exp(1)$ in distribution. Using this limit, the density function $f_S(s)$ is approximately $\frac{\alpha}{2}\frac{e^{-|s|^{-\alpha}}}{|s|^{\alpha+1}}$, and hence the joint log-likelihood $l(x_i,\theta_i)$ is  approximately
\begin{align}\notag
&l\left(x_i,\theta_i\right)
\approx \sum_{j=1}^M \log f_S(s_{ij}) + \sum_{j=1}^M\left\{-\frac{\theta_i^\alpha}{|y_j-x_is_{ij}|^\alpha}
-(\alpha+1)\log|y_j-x_is_{ij}|\right\} + \alpha M\log\theta_i+M\log \left(\frac{\alpha}{2}\right)
\end{align}
which approaches infinity (i.e., the maximum likelihood) at the poles: $y_j - x_i s_{ij} = 0$, $j = 1$ to $M$. This is the reason why we use only the ratio statistics $z_{i,j} = y_j/s_{ij}$ for recovering $x_i$.

\subsection{The Approximate Distribution of $y_j/s_{ij}$}

Since our procedure utilizes the statistic $y_j/s_{ij}$, we need to know its distribution, at least approximately.  Note that $\frac{y_j}{s_{ij}}  = \frac{\sum_{t=1}^N x_t s_{tj}}{s_{ij}} = \frac{\sum_{t\neq i} x_t s_{tj}}{s_{ij}} + x_i = \theta_i\frac{S_2}{S_1} + x_i$,  where $S_1$ and $S_2$ are i.i.d. $S(\alpha,1)$ variables. Recall the definition $\theta_i = \left(\sum_{t\neq i}|x_t|^\alpha\right)^{1/\alpha}$. Thus, $\mathbf{Pr}\left( \frac{y_j}{s_{ij}} < t\right) = \mathbf{Pr}\left( \frac{S_2}{S_1} < \frac{t-x_i}{\theta_i}\right)$ and the problem boils down to finding the distribution of the ratio of two $\alpha$-random variables with $\alpha\approx 0$. Using the limits: $1/|S_1|^{\alpha} \rightarrow exp(1)$ and  $1/|S_2|^{\alpha} \rightarrow exp(1)$, as $\alpha\rightarrow0$, it is not difficult  to show an approximate cumulative distribution function (CDF) of $y_j/s_{ij}$:
\begin{align}\label{eqn_appr_CDF}
\mathbf{Pr}\left( \frac{y_j}{s_{ij}} < t\right) = \mathbf{Pr}\left( \frac{S_2}{S_1} < \frac{t-x_i}{\theta_i}\right)
\approx\left\{ \begin{array}{cc}
\frac{1}{2\left(1+\left|\frac{t-x_i}{\theta_i}\right|^\alpha\right)} & t<x_i\\\\
1-\frac{1}{2\left(1+\left|\frac{t-x_i}{\theta_i}\right|^\alpha\right)}&t\geq x_i
\end{array}\right.
\end{align}
The CDF of $S_2/S_1$ is the also given by (\ref{eqn_appr_CDF}) by letting $x_i=0$ and $\theta_i=1$.\\

\begin{figure}[h!]
\begin{center}
\mbox{
\includegraphics[width=3.2in]{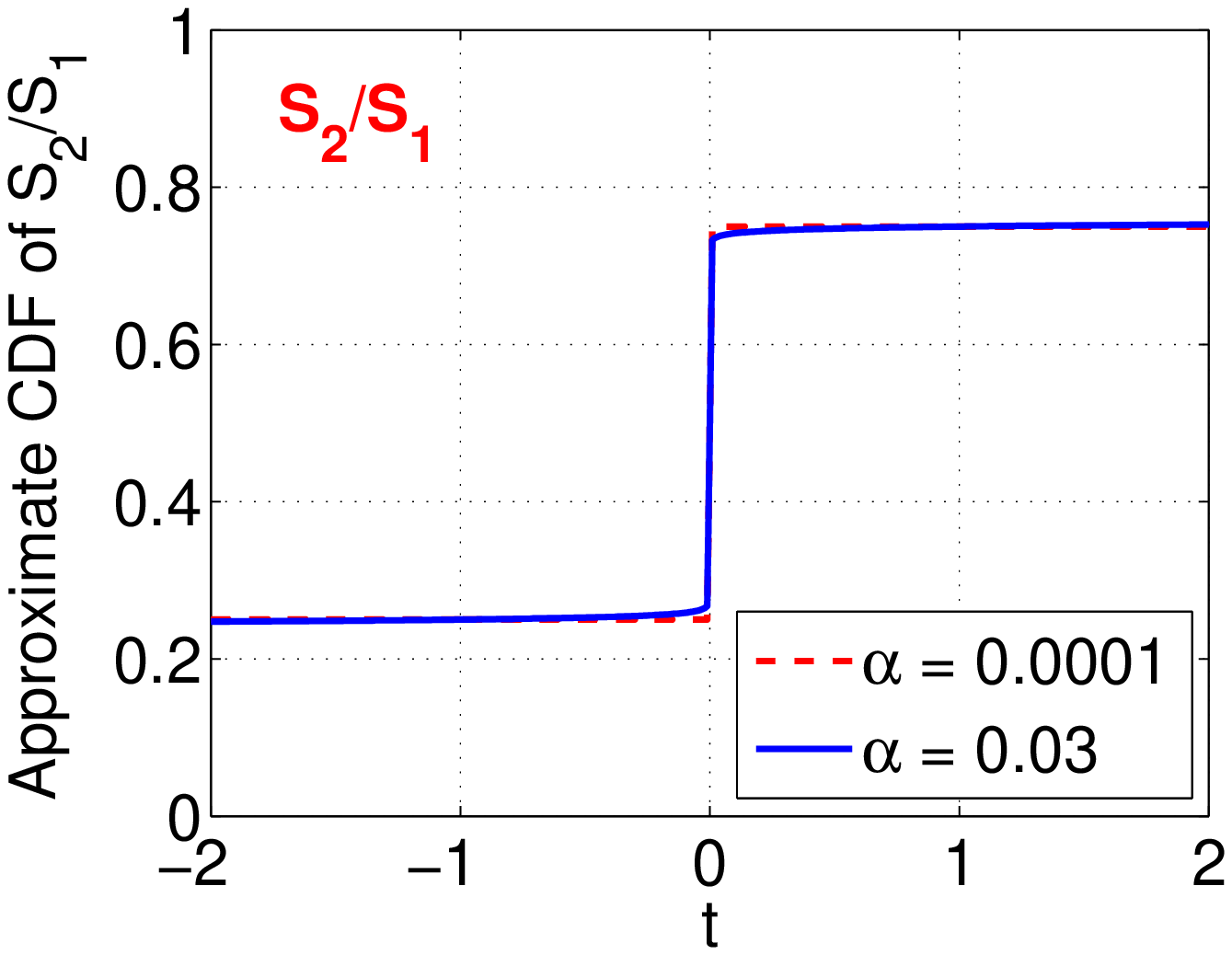}
\includegraphics[width=3.2in]{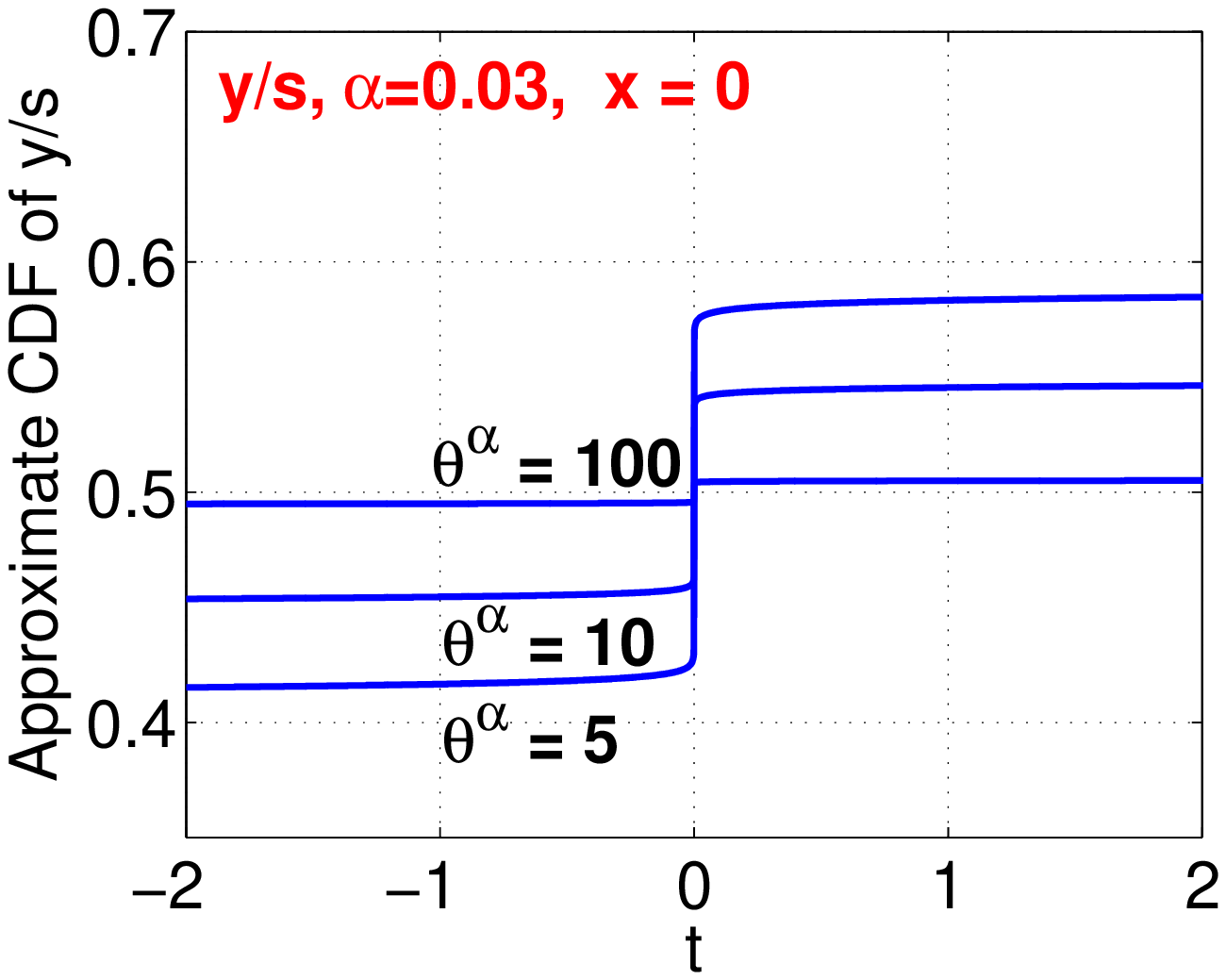}
}
\end{center}
\vspace{-0.20in}
\caption{Approximate CDFs of $S_2/S_1$ (left panel) and $\frac{y_j}{s_{ij}}$ (right panel) as given by (\ref{eqn_appr_CDF}). The CDF of $S_2/S_1$ is very heavy-tailed with an essentially vertical jump around 0. This means, samples of $S_2/S_1$ are likely to be either very close 0 or very large (approaching $\pm \infty$).
The CDF of $y_j/s_{ij}$ is a scaled (and shifted) version of the CDF of $S_2/S_1$, with a almost vertical jump
around  the true $x_i$. This special structure motivates us to develop the gap estimator for $x_i$. Suppose we have $M$ observations of $y_j/s_{ij}$, $j = 1$ to $M$. Observations outside $(x_i-e, x_i+e)$ for very small $e$ will   likely be far away from each other because the distribution is extremely heavy-tailed. Observations within $(x_i-e, x_i+e)$ are extremely close to each other, which will be used for identifying the true $x_i$ because these observations cluster around $x_i$.}\label{fig_appr_CDF}\vspace{0.0in}
\end{figure}

\begin{figure}[h!]
\begin{center}
\mbox{\includegraphics[width=2.8in]{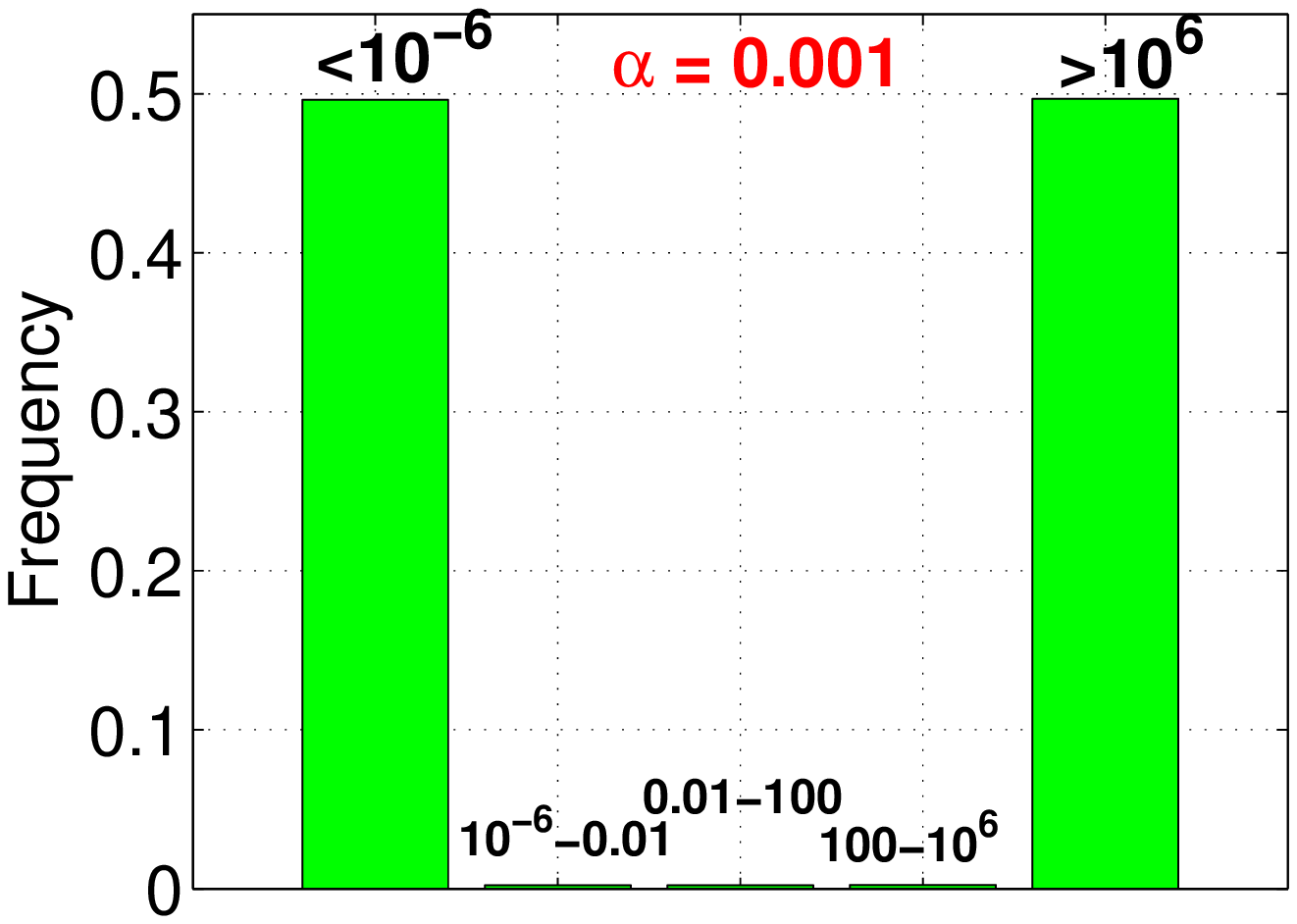}
\includegraphics[width=2.8in]{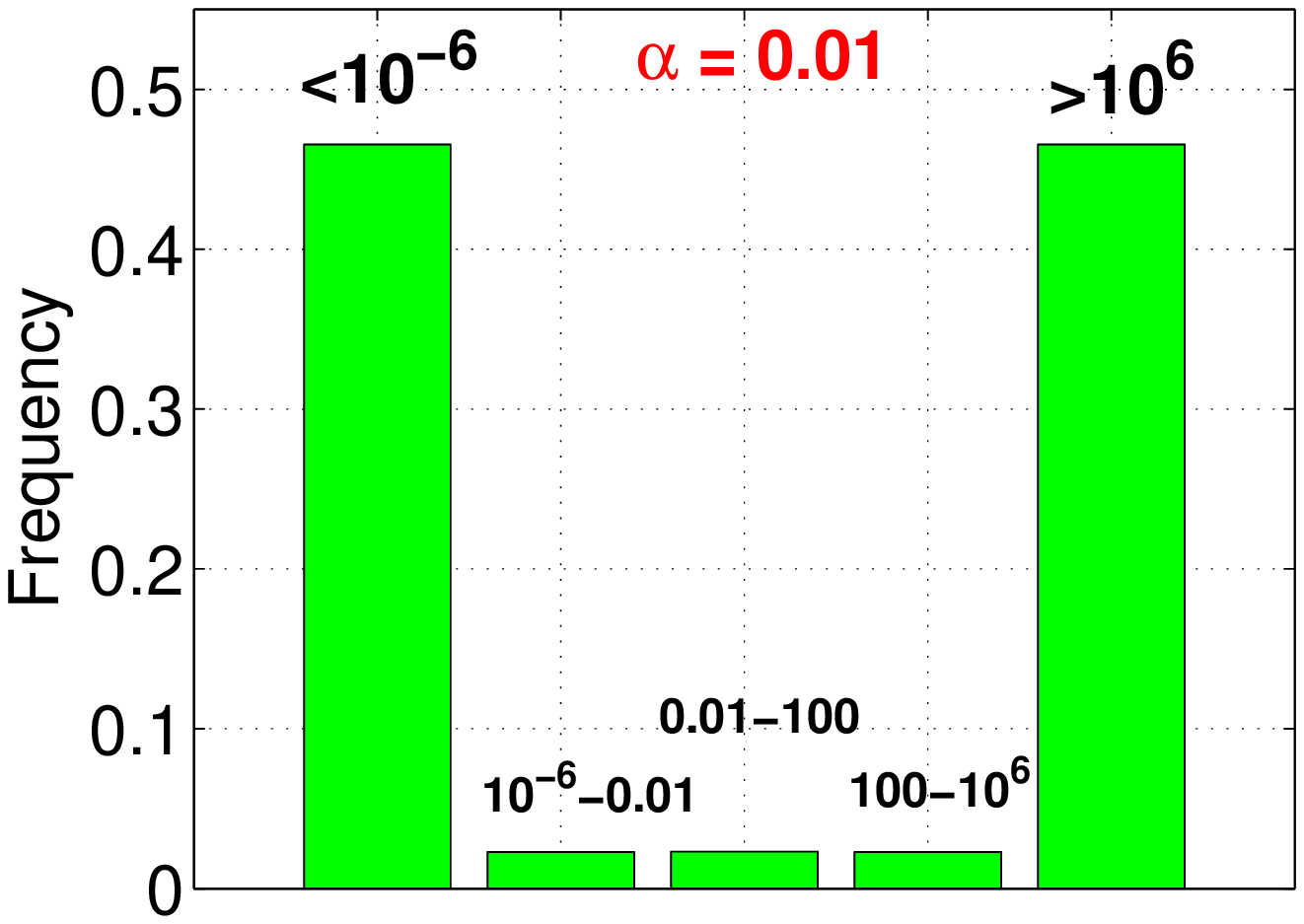}}
\mbox{\includegraphics[width=2.8in]{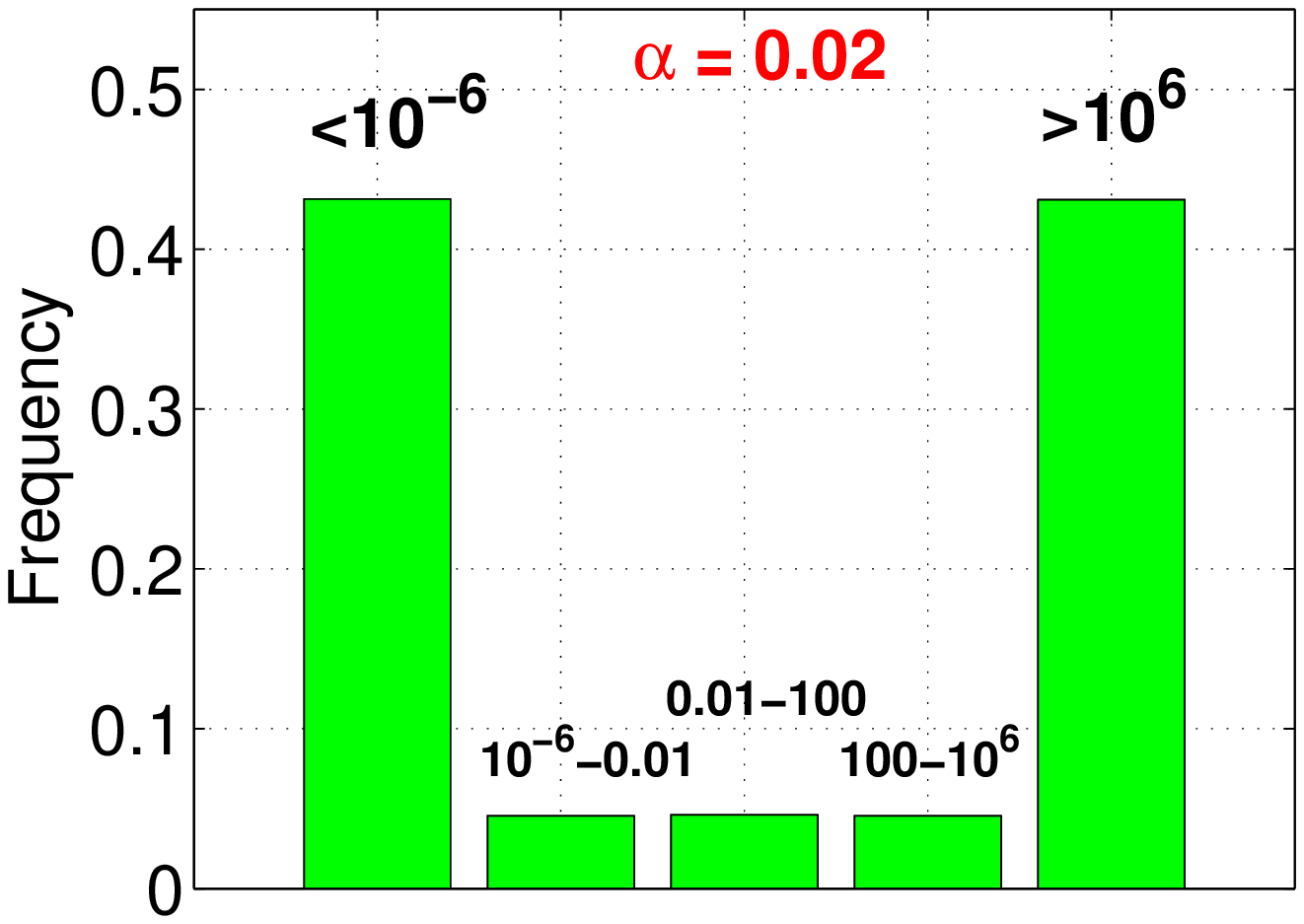}
\includegraphics[width=2.8in]{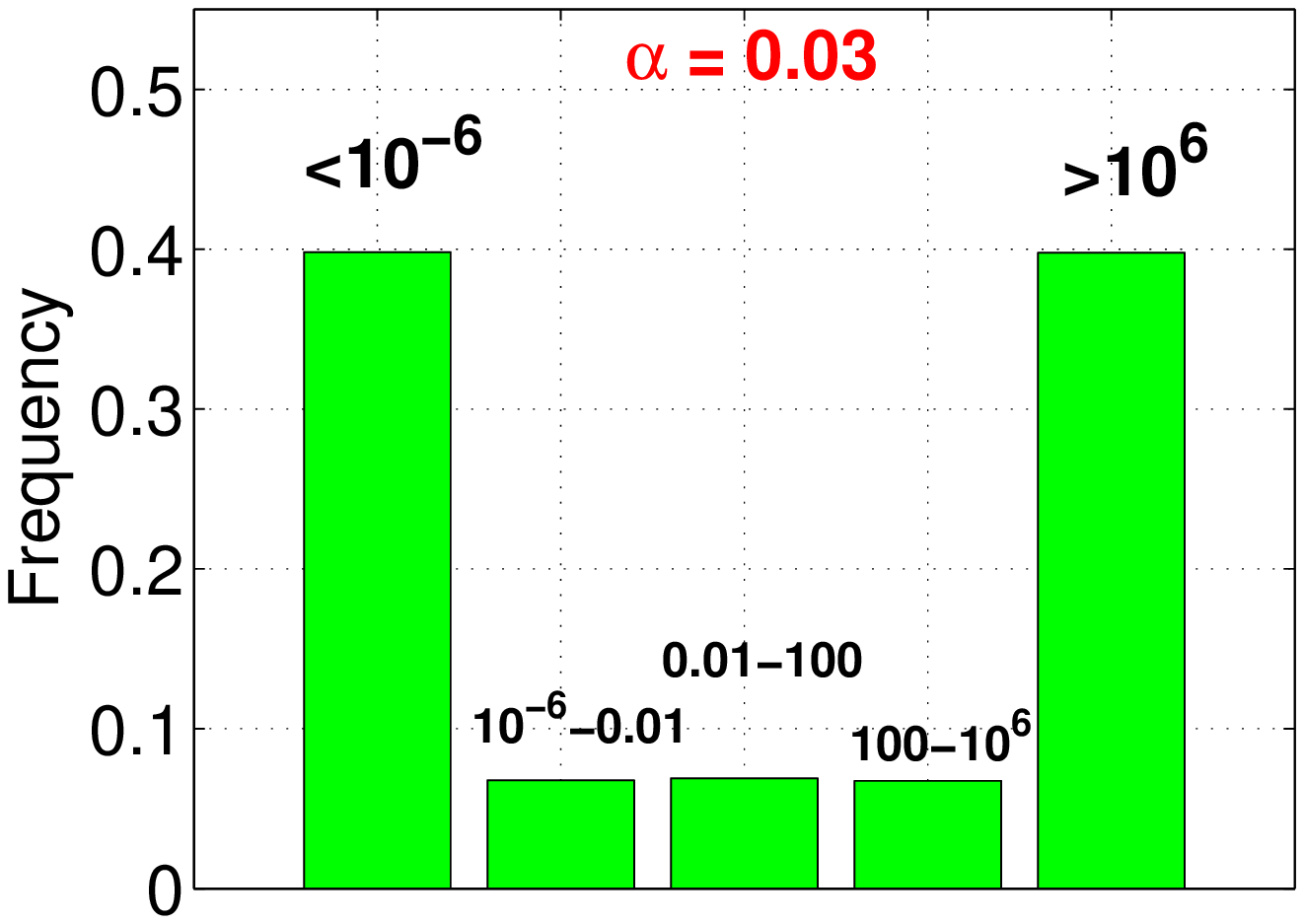}}
\end{center}
\vspace{-0.2in}
\caption{Simulations of $|S_2/S_1|$ directly using the formula (\ref{eqn_stable_sample}) for generating two independent $\alpha$-stable variables $S_1$ and $S_2$. With $\alpha\rightarrow0$, it is clear that most of the samples are either very close to 0  ($<10^{-6}$) or very large ($>10^6$).}\label{fig_PMFS2S1}
\end{figure}

Figure~\ref{fig_appr_CDF} plots the approximate CDFs (\ref{eqn_appr_CDF}) for $S_2/S_1$ (left panel) and $y_j/s_{ij}$ (right panel, with $x_i=0$ and three values of $\theta^\alpha$). While the distribution of $S_2/S_1$ is extremely heavy-tailed, about half of the probability mass concentrated near 0. This means, as $\alpha\rightarrow0$, samples of $|S_2/S_1|$ are  equal likely to be either very close to zero or very large. Since (\ref{eqn_appr_CDF}) is only approximate, we also provide the simulations of $S_2/S_1$ in Figure~\ref{fig_PMFS2S1} to help verify the approximate CDF in Figure~\ref{fig_appr_CDF}.

\subsection{An Example with $K=2$ to Illustrate the ``Idealized'' Algorithm}

To illustrate our algorithm, in particular the iterative procedure in Alg.~\ref{alg_recovery}, we consider the simplest example of $K=2$. Without loss of generality, we let $x_1 = x_2 = 1$ and $x_i=0, \forall \ 3\leq i\leq N$. This way, our observations are $y_j = x_1 s_{1j} + x_2 s_{2j} = s_{1j} + s_{2j}$ for $j =1$ to $M$. The ratio statistics are
\begin{align}\notag
&z_{1,j} = y_j/s_{1j} = 1 + \frac{s_{2j}}{s_{1j}}\\\notag
&z_{2,j} = y_j/s_{2j} = 1 + \frac{s_{1j}}{s_{2j}}\\\notag
&z_{i,j} = y_j/s_{ij} = \frac{s_{1j}}{s_{ij}}+ \frac{s_{2j}}{s_{ij}}, \ i \geq 3
\end{align}

We assume an ``idealized'' algorithm, which allows us to use an extremely small $\alpha$.  As $\alpha\rightarrow0$, $\frac{s_{2j}}{s_{1j}}$ is either (virtually) 0 or $\pm\infty$. Note that $\frac{s_{1j}}{s_{2j}}$ is the reciprocal of $\frac{s_{2j}}{s_{1j}}$, i.e., $\frac{s_{1j}}{s_{2j}}\approx0\Longleftrightarrow \frac{s_{2j}}{s_{1j}}\approx\pm\infty$.

Suppose, with $M=3$ observations,  the ratio statistics, for $i=1, 2$, are:
\begin{align}\notag
(z_{1,1}, z_{2,1}) = (1,\pm\infty),\hspace{0.2in}  (z_{1,2}, z_{2,2}) = (\pm\infty, \ 1),\hspace{0.2in} (z_{1,3}, z_{2,3}) = (1,\pm\infty)
\end{align}
Then we have seen $z_{1,j} =1$ twice and we assume this ``idealized'' algorithm is able to correctly  estimate $\hat{x}_1 = 1$, because there is a ``cluster'' of 1's. After we have estimated $x_1$, we compute the residual $r_j = y_j - \hat{x}_1s_{1j} = s_{2j}$. In the second iteration, the ratio statistics become
\begin{align}\notag
&r_j/s_{2j} = 1\\\notag
&r_j/s_{ij} =  \frac{s_{2j}}{s_{ij}}, \ i \geq 3
\end{align}
This means we know $x_2=1$. Next, we again update the residuals, which become zero. Therefore, in the third iteration, all zero coordinates can be recovered.  The most exciting part of this  example  is that, with $M=3$ measurements, we can recovery a signal with $K=2$, regardless of $N$. When $K>2$, the analysis of the ``idealized'' algorithm requires a bit more work, which we present in Sec.~\ref{sec_ideal}.

We summarize the ``idealized'' algorithm (see more details in Sec.~\ref{sec_ideal})  as follows:
\begin{enumerate}
\item The algorithm assumes $\alpha\rightarrow0$, or as small as necessary.
\item As long as there are two observations $y_j/s_{ij}$ in the extremely narrow interval $(x_i-e, x_i+e)$ with $e$ very close to 0, the algorithm  is able to correctly recover $x_i$. We assume $e$ is so small that it is outside the required precision range of $x_i$. Here we purposely use $e$ instead of $\epsilon$ to differentiate it from the parameter $\epsilon$ used in our practical procedure, i.e., Alg.~\ref{alg_recovery}.
\end{enumerate}

Clearly, this ``idealized'' algorithm can not be strictly faithfully implemented. If we have to use a small $\alpha$ instead of $\alpha=0$, the observations $|y_j/s_{ij}|$ will be  between 0 and $\infty$, and we will not be able to identify the true $x_i$ with high confidence unless we see two essentially identical  observations. This is why we specify that an algorithm must see at least two repeats in order to exactly recover $x_i$. As analyzed in Sec.~\ref{sec_theory},  the proposed gap estimator is a practical surrogate, which converges to the ``idealized'' algorithm when $\alpha\rightarrow0$.

\subsection{The Intuition for the Minimum Estimator and the Gap Estimator}

As shown in Figure~\ref{fig_appr_CDF} (right panel), while the distribution of $y_j/s_{ij}$ is  heavy-tailed, its CDF has a significant jump  very near the true $x_i$. This means more than one observations (among $M$ observations) will likely lie in the extremely narrow interval around $x_i$, depending on the value of $\theta^\alpha$ (which is essentially $K$).  We are able to detect whether $x_i=0$ if there is just one observation near zero. To estimate the magnitude of $x_i$, however, we need to see  a ``cluster'' of observations, e.g., two or more observations which are  essentially identical. This is the intuition for the minimum estimator and the gap estimator. Also, as one would expect, Figure~\ref{fig_appr_CDF} shows that the performance  will degrade (i.e., more observations are needed) as $\theta_i^\alpha$ increases.

The gap estimator is a practical surrogate for the ``idealized'' algorithm. Basically, for each $i$, if we sort the observations: $z_{i,(1)}\leq z_{i,(2)}\leq ...\leq z_{i,(M)}$,  the two neighboring observations corresponding to the minimum gap will be likely lying in a narrow neighborhood of $x_i$, provided that the length of the minimum gap is very small, due to the heavy concentration of the probability mass about $x_i$.

 If the observed minimum gap is not small, we give up estimating this (``undermined'') coordinate in the current iteration. After we have removed the (reliably) estimated coordinates by computing the residuals,  we may have a better chance to successfully recover some of these undermined coordinates because  the effective ``$K$''  and the effective ``$N$'' are significantly reduced.\\

Finally, to better understand the difference between the minimum estimator and the gap estimator, we compute  (assuming $x_i>0$) the probability $\mathbf{Pr}\left( |y_j/s_{ij}|<x_i-\epsilon\right)$, which is related to the probability that the two estimators differ, i.e.,  $|\hat{x}_{i,gap}|>|\hat{x}_{i,min}|$. Using the approximate CDF of $y_j/s_{ij}$ (\ref{eqn_appr_CDF}), we have
\begin{align}\notag
\mathbf{Pr}\left( |y_j/s_{ij}|<x_i-\epsilon\right) \approx \frac{1/2}{1+\epsilon^\alpha/K} - \frac{1/2}{1+(2x_i)^\alpha/K}
\end{align}
When $\alpha = 0.03$, $x_i = 10$, $\epsilon = 10^{-5}$, $K=100$, this probability is about 0.002, which is  not small considering that we normally have to use at least $M=5K$ measurements. Given enough measurements, almost certainly the minimum estimator will be smaller than the gap estimator in absolute values, as verified in the simulations in Sec.~\ref{sec_simu}. In other words, the minimum estimator will not be reliable for exact recovery. This also explains why we need to see a clustered observations instead of relying on the absolute minimum observation to estimate the magnitude.

\subsection{What about the Sample Median Estimator?}

Since the distribution of $y_j/s_{ij}$ is symmetric about the true $x_i$, it is also natural to consider using the sample median to estimate $x_i$. We have found, however, at least empirically, that the median estimator will require significantly more measurements, for example, $M> 5M_0$ or more. This is why we develop the gap estimator to better exploit the special structure of the distribution as illustrated in Figure~\ref{fig_appr_CDF}.

\section{Two Baselines: LP and OMP}

Both linear programming (LP) and orthogonal matching pursuit (OMP) utilize a design matrix sampled from Gaussian (i.e., $\alpha$-stable with $\alpha=2$) or  Gaussian-like distribution (e.g., a two-point distribution on $\{-1,+1\}$ with equal probabilities). Here, we use $\mathbf{S}_{(2)}$ to denote such a design matrix.

The well-known LP algorithm recovers the signal $\mathbf{x}$ by solving the following $l_1$ optimization problem:
\begin{align}
\min_{\mathbf{x}}\|\mathbf{x}\|_1 \ \ \ \text{ subject to } \ \ \mathbf{y=xS_{(2)}}
\end{align}
which is also commonly known as  {\em Basis Pursuit}~\cite{Article:Chen98}. It has been proved that LP can  recover $\mathbf{x}$ using $M =O\left(K\log (N/K)\right)$ measurements~\cite{Article:Donoho09}, although the exact constant is   unknown. This procedure  is  computationally prohibitive for large $N$ (e.g., $N=10^9$). When there are measurement noises, the LP algorithm  can be modified as  other convex optimization problems, for example,  the {\em Lasso} algorithm~\cite{Article:Tibshirani_96}.

The orthogonal matching pursuit (OMP) algorithm~\cite{Article:Mallat93} is a  popular greedy iterative procedure. It typically proceeds with $K$ iterations. At each iteration,  it conducts univariate least squares for all the coordinates on the residuals, and chooses the coordinate which maximally reduces the overall square errors. At the end of each iteration, all the chosen coordinates are used to update the residuals via a multivariate least square. The algorithm can be coded  efficiently (e.g., in Matlab) (But we find OMP is still significantly slower than our  method especially when $K$ is not small.) \cite{Article:Zhang_RIP11,Proc:OMP_NIPS09} showed that, under appropriate conditions, the required number of measurements of OMP is essentially $O(K\log(N-K))$, which improved the prior result in~\cite{Article:Tropp_JIT04}. There are also modified OMP algorithms, e.g., {\em CoSaMP}~\cite{Article:CoSaMP_09}.

In this paper, our experimental  study will focus on the comparisons with  OMP and LP,  as these two methods are  the most basic and still strong baselines. Of course, we recognize that compressed sensing is a rapidly developing area of research and we are aware that there are other promising sparse recovery methods such as the ``message-passing'' algorithm~\cite{Article:Donoho_PNAS09} and the ``sparse matrix'' algorithm~\cite{Article:Gilbert_IEEE10}. We plan to compare our algorithm with those methods   in separate future papers.

\section{Simulations}\label{sec_simu}

To validate the  procedure in Alg.~\ref{alg_recovery}, we  provide some simulations (and comparisons with LP and OMP), before presenting the theory.  In each simulation, we randomly select $K$ coordinates from a total of $N$ coordinates. We  set the magnitudes (and signs) of these $K$ coordinates according to one of the two mechanisms.  (i) \textbf{Gaussian signals}: the values are sampled from  $Normal(0,5^2)$. (ii) \textbf{Sign signals}: we simply take the signs, i.e., $\{-1, 0, 1\}$,  of the generated Gaussian signals. The number of measurements $M$ is chosen by
\begin{align}
M = M_0/\zeta,\hspace{0.2in} M_0 = K\log\left((N-K)/\delta\right)
\end{align}
where $\delta = 0.01$ and $\zeta \in \{1, 1.3, 2, 3, 4, 5\}$. When $M=M_0$, all methods perform  well in terms of accuracies; and hence it is more interesting to examine the results when $M<M_0$. Also, we simply fix $\epsilon = 10^{-5}$ in Alg~\ref{alg_recovery}. Our method is not sensitive to $\epsilon$ (as long as it is small). We will explain the reason in the theory section.

\subsection{Sample Instances of Simulations}

\begin{figure}[h!]
\begin{center}
\mbox{
\includegraphics[width=2.5in]{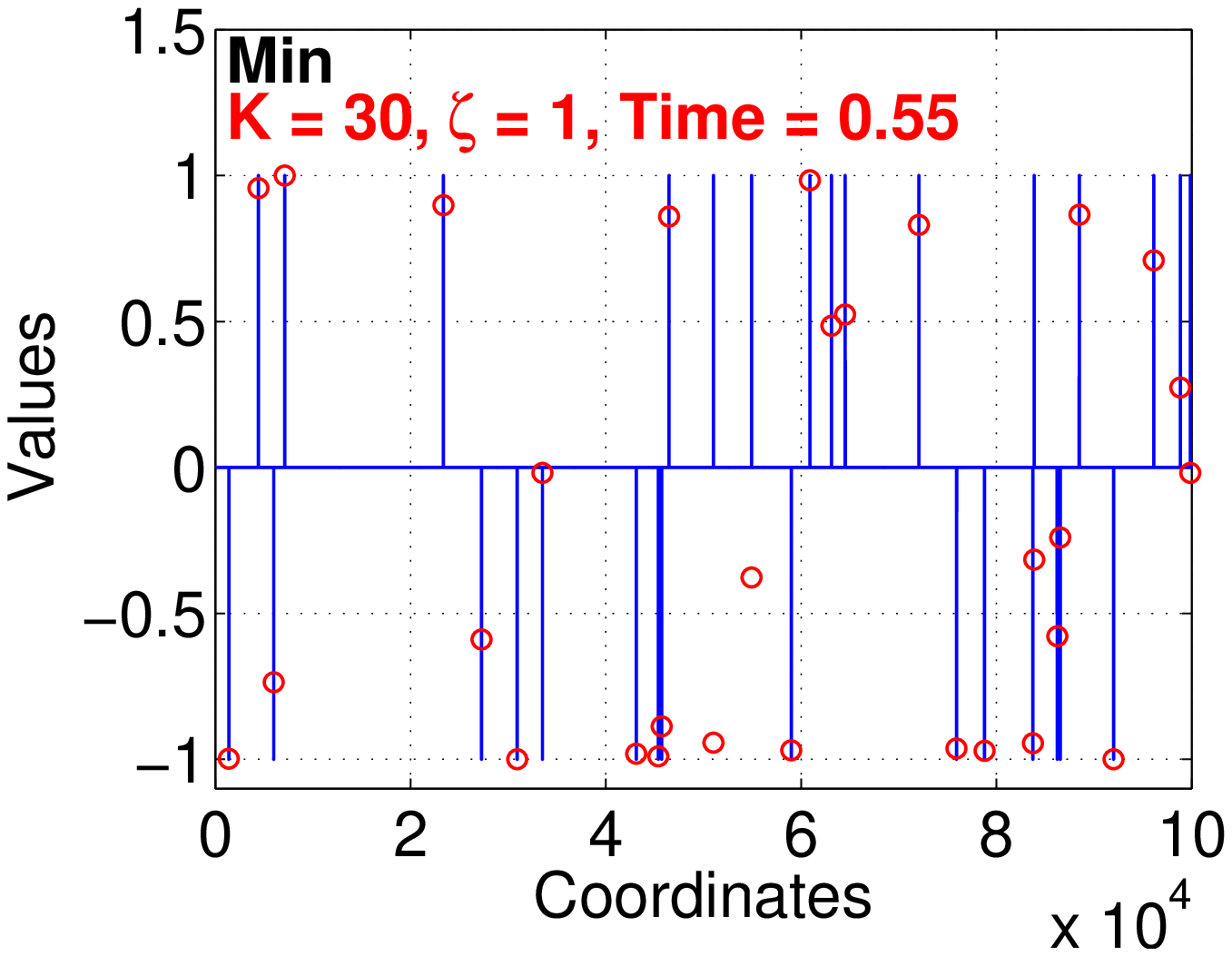}\hspace{0in}
\includegraphics[width=2.5in]{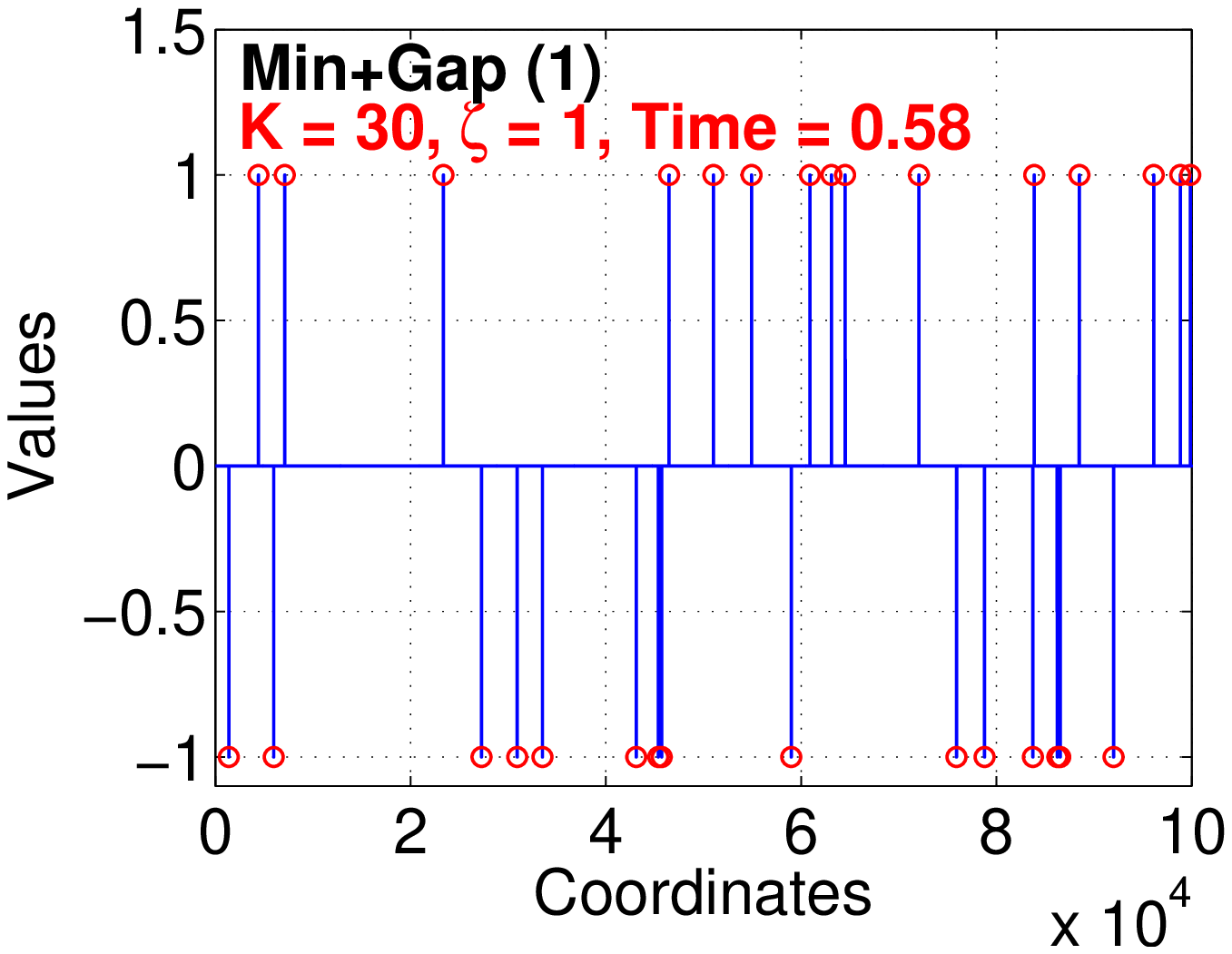}}

\mbox{
\includegraphics[width=2.5in]{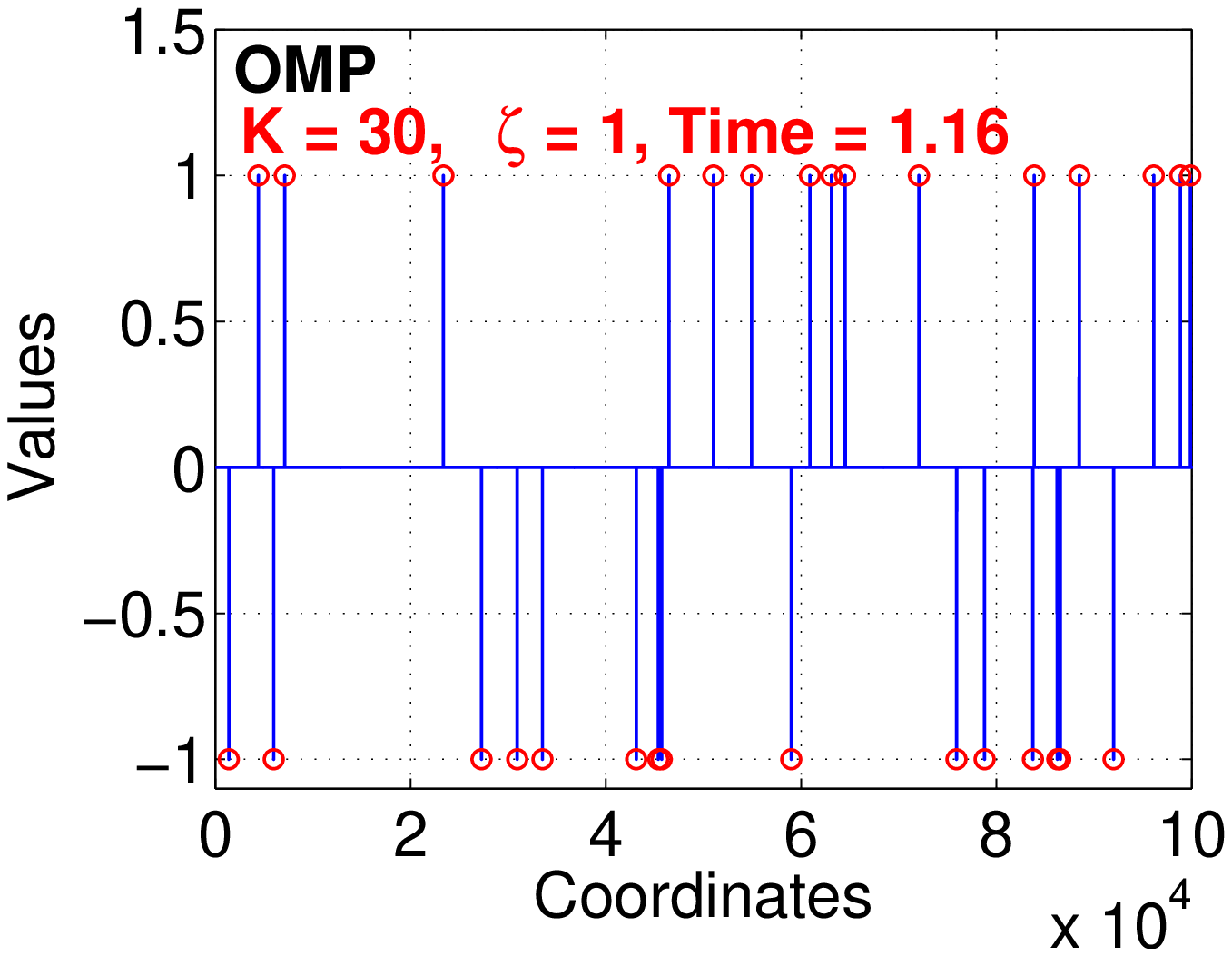}\hspace{0in}
\includegraphics[width=2.5in]{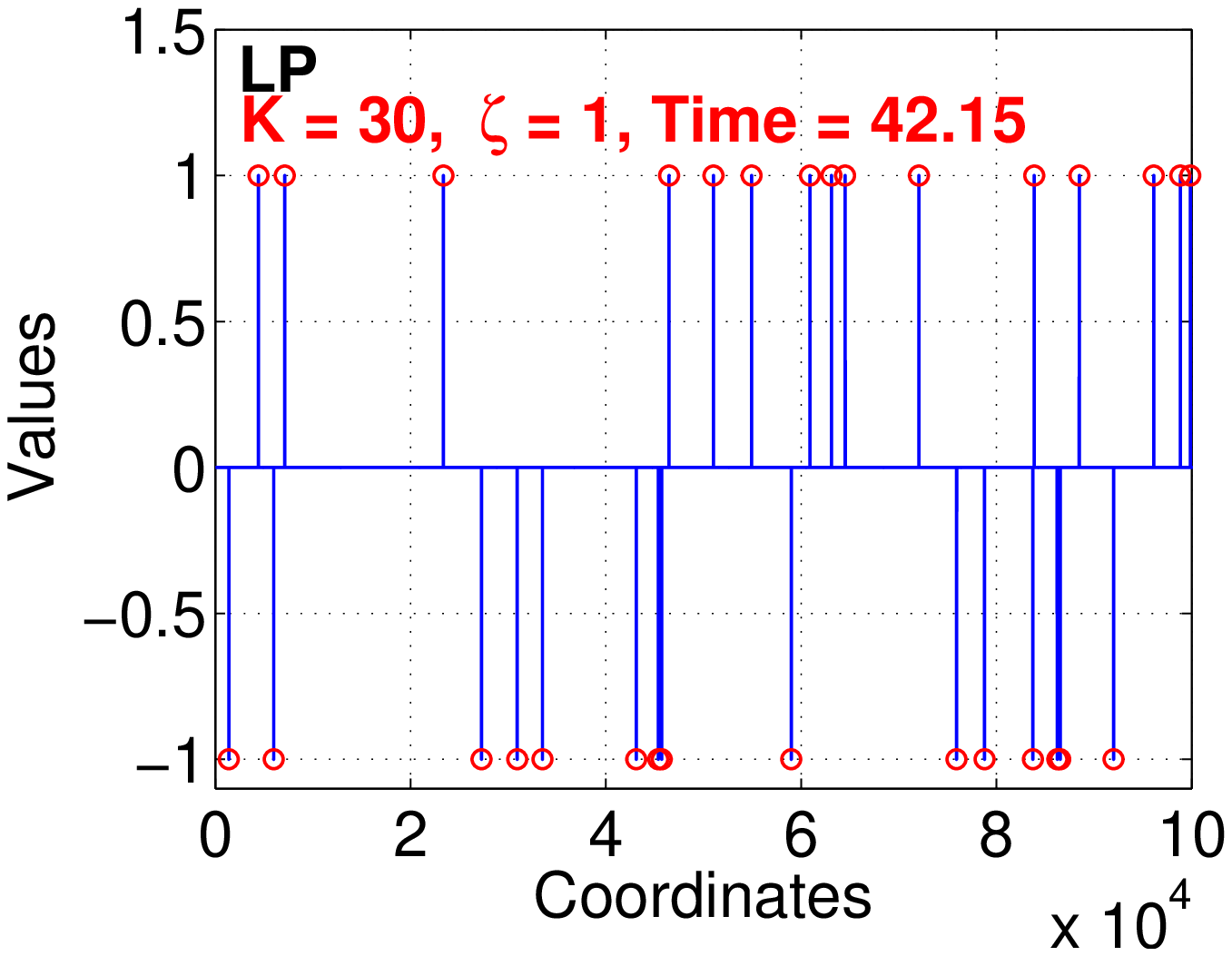}}
\end{center}
\vspace{-0.2in}
\caption{\small Reconstruction results from one simulation, using $N=100000$, $K=30$, $M=M_0$ (i.e., $\zeta=1$), and sign signals. The reconstructed signals are denoted by (red) circles. The minimum estimator (left-top) is able to  identify all nonzero coordinates with no false positives, using only 0.55 seconds. With one iteration of the gap estimator (right-top), we can  perfectly reconstruct the signal using additional 0.03 seconds (so the total time is 0.58 seconds).  The OMP and LP algorithms (bottom panels) also perform well at significantly higher computational costs.}\label{fig_RecSignB1}
\end{figure}

\begin{figure}[h!]
\begin{center}
\mbox{
\includegraphics[width=2.5in]{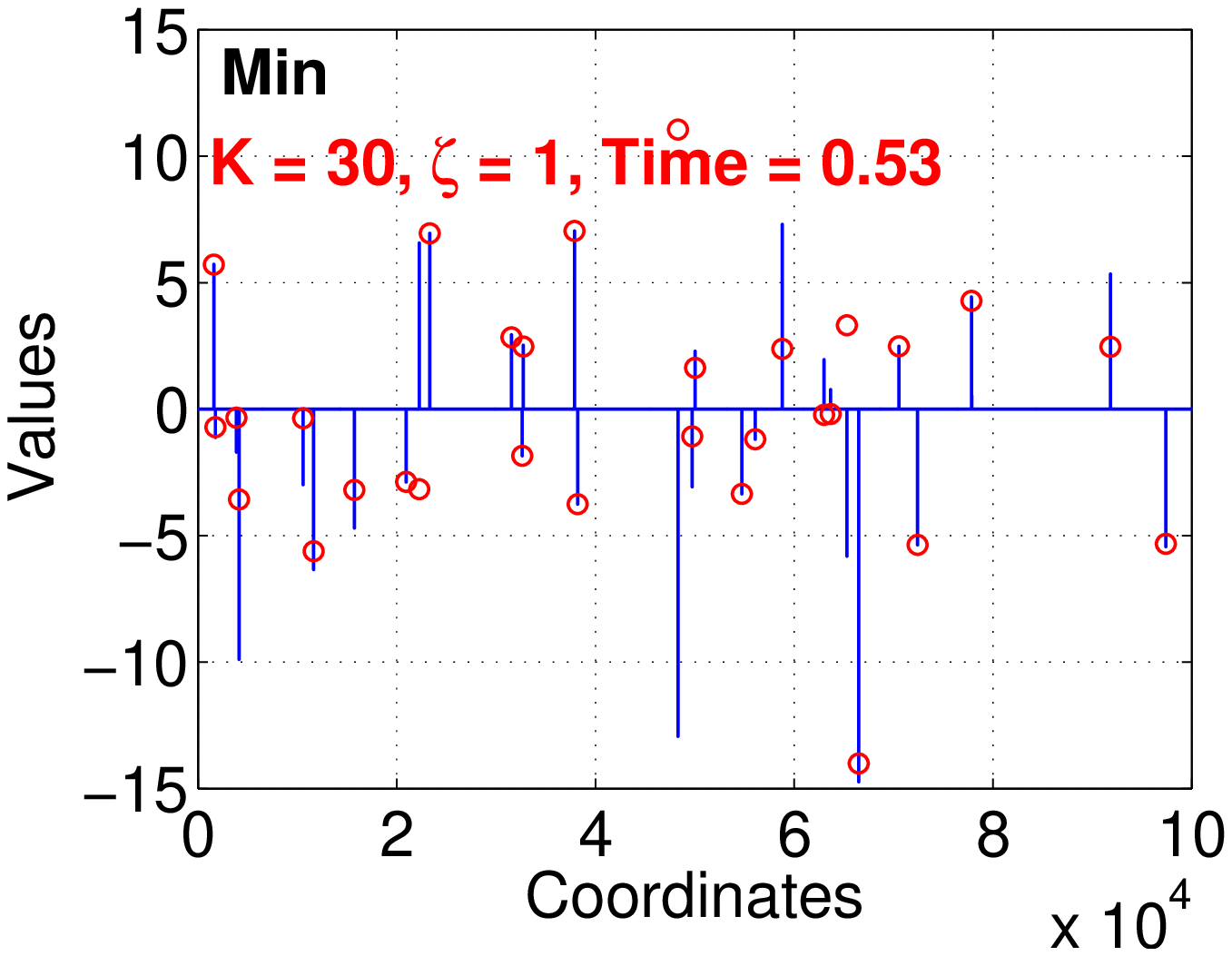}\hspace{0in}
\includegraphics[width=2.5in]{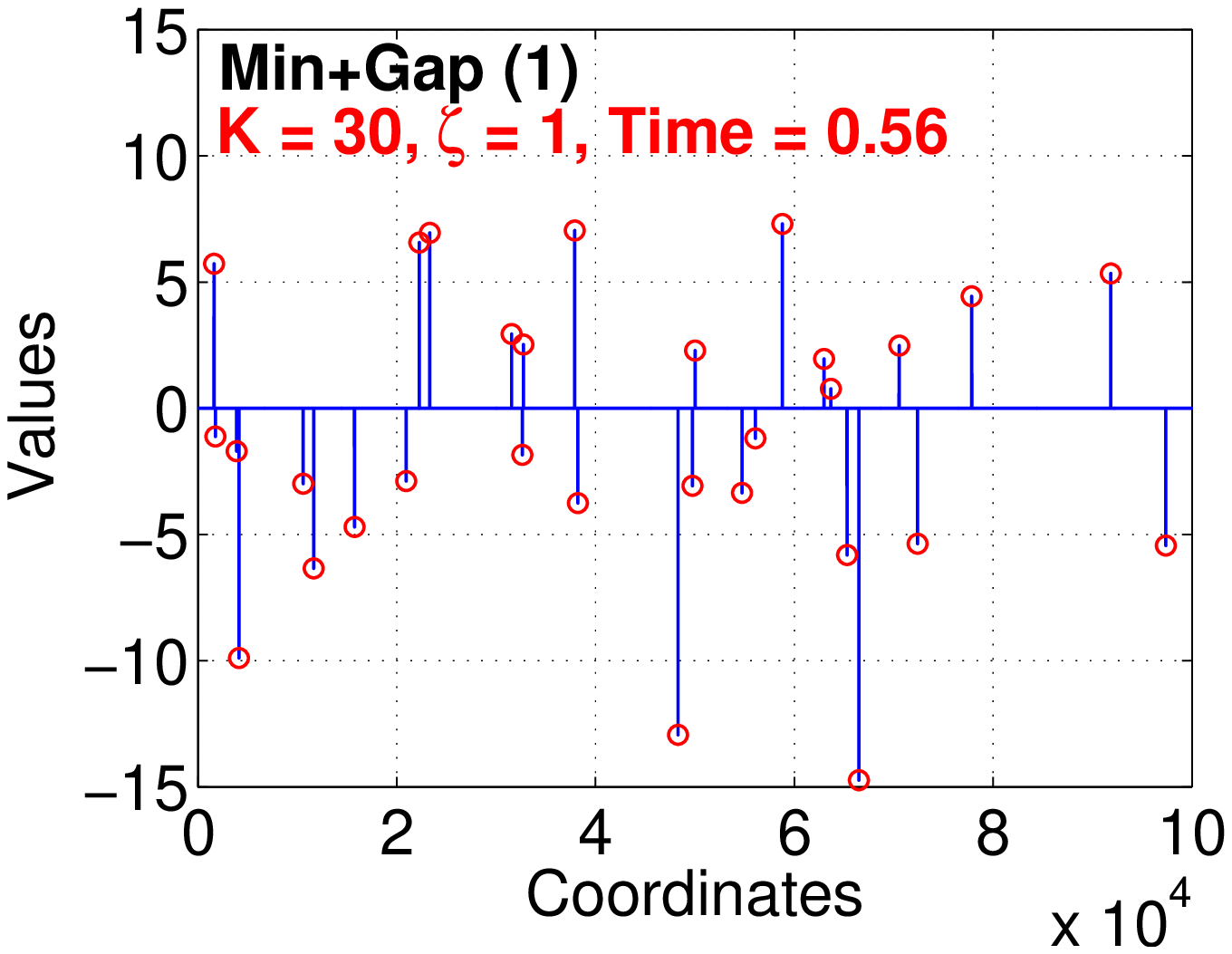}}

\mbox{
\includegraphics[width=2.5in]{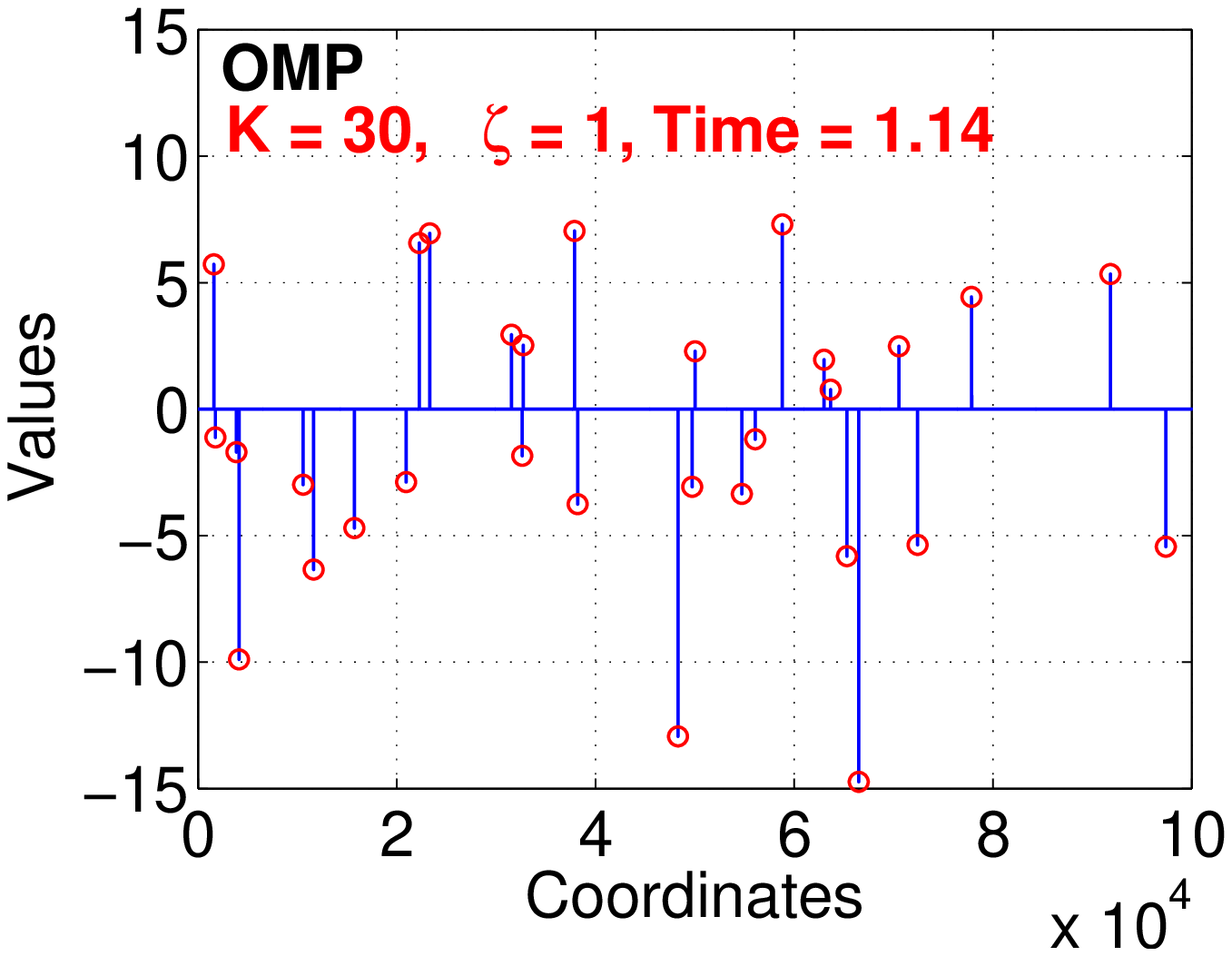}\hspace{0in}
\includegraphics[width=2.5in]{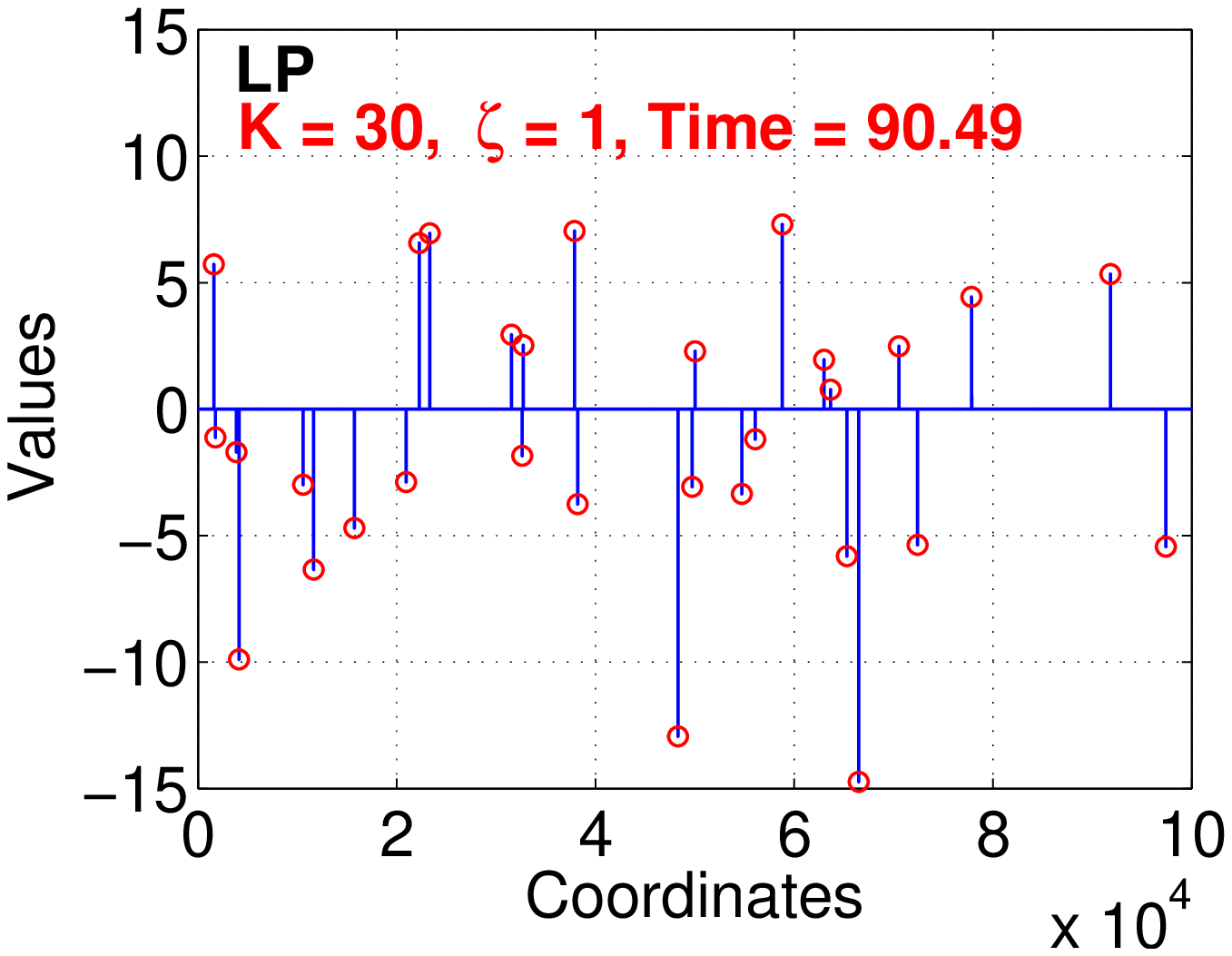}}
\end{center}
\vspace{-0.2in}
\caption{\small Reconstruction results from one simulation, using $N=100000$, $K=30$, $M=M_0$, and Gaussian signals.}\label{fig_RecGausB1}\vspace{0in}

\end{figure}

\begin{figure}[h!]
\begin{center}
\mbox{
\includegraphics[width=2.5in]{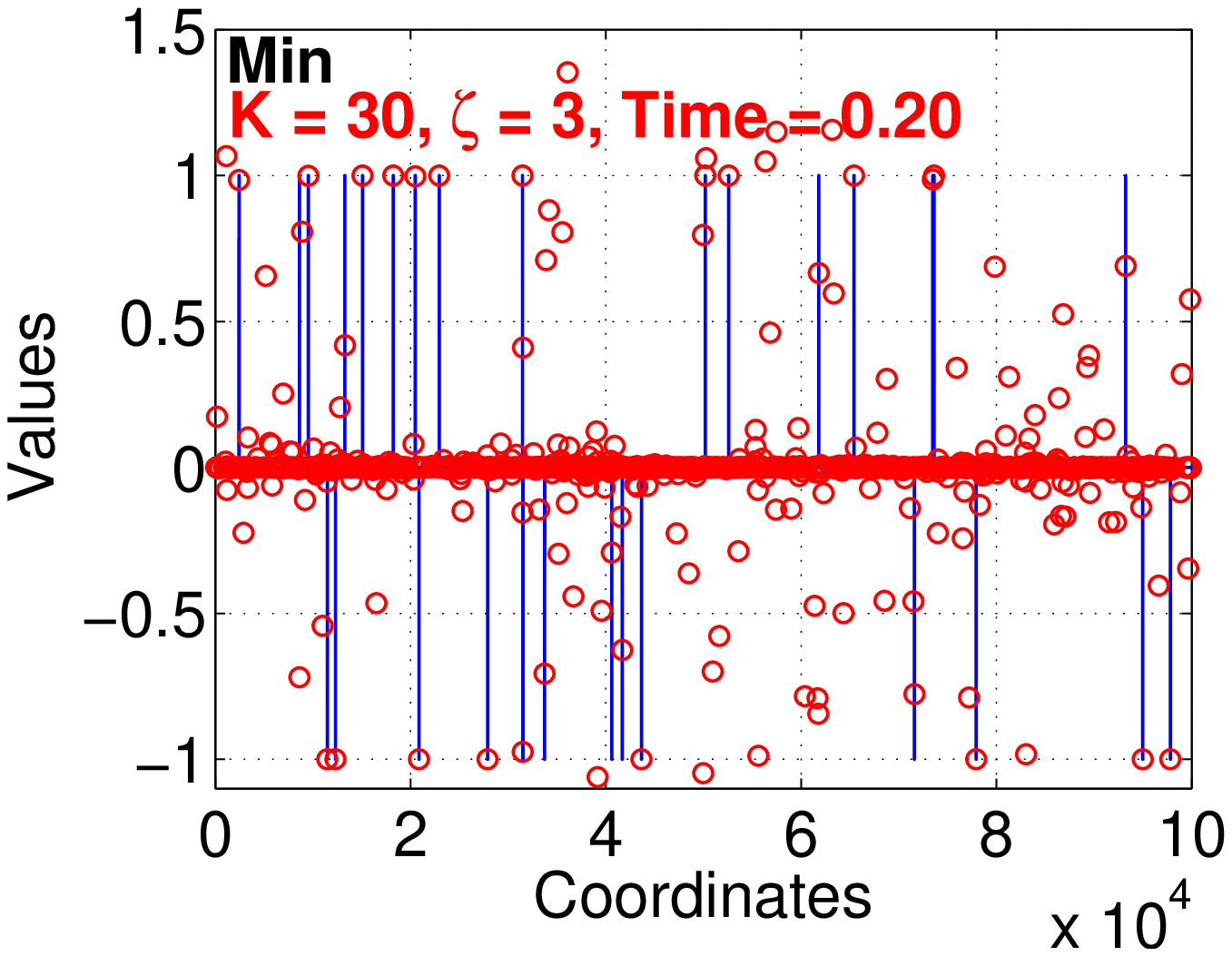}\hspace{0in}
\includegraphics[width=2.5in]{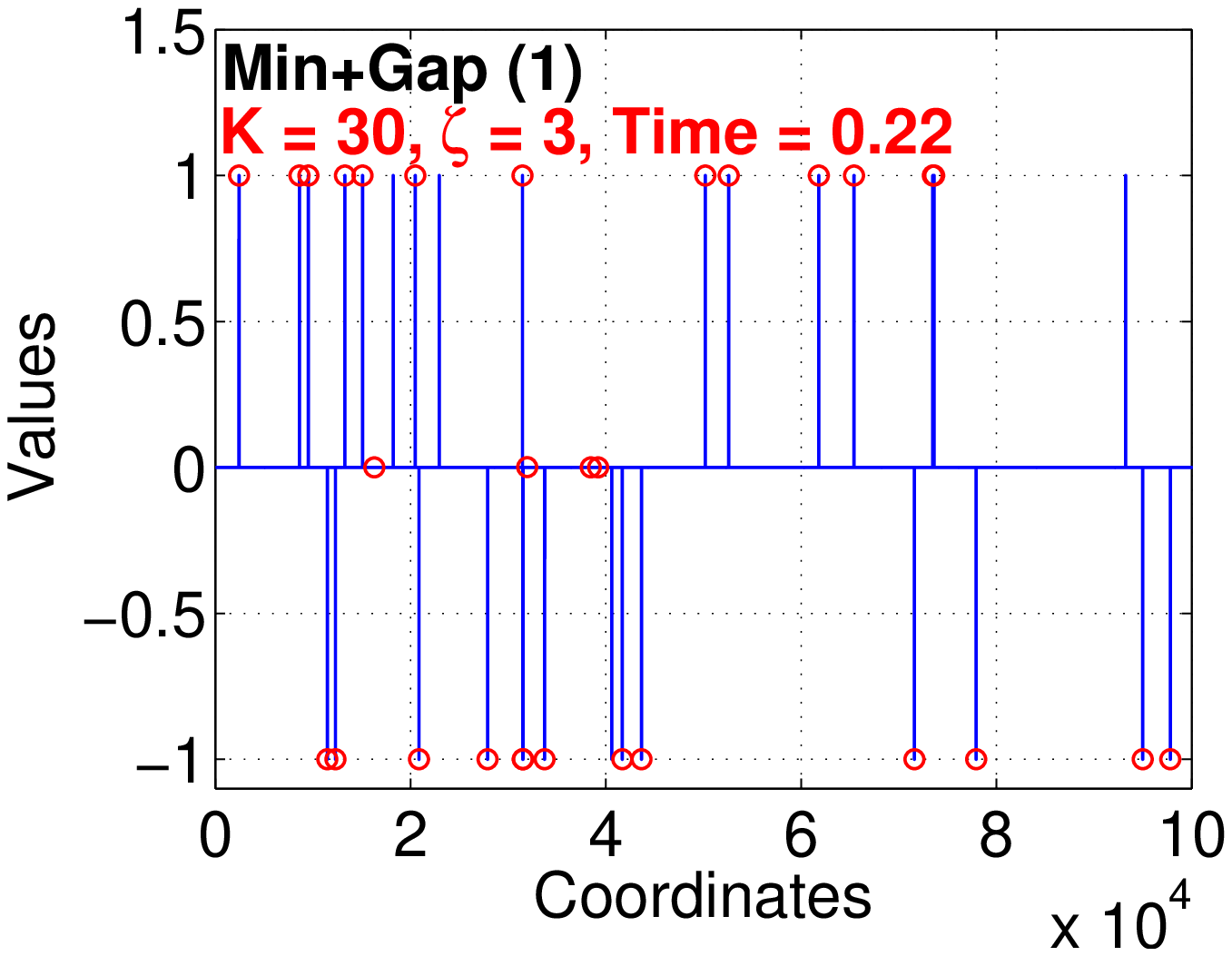}}

\mbox{
\includegraphics[width=2.5in]{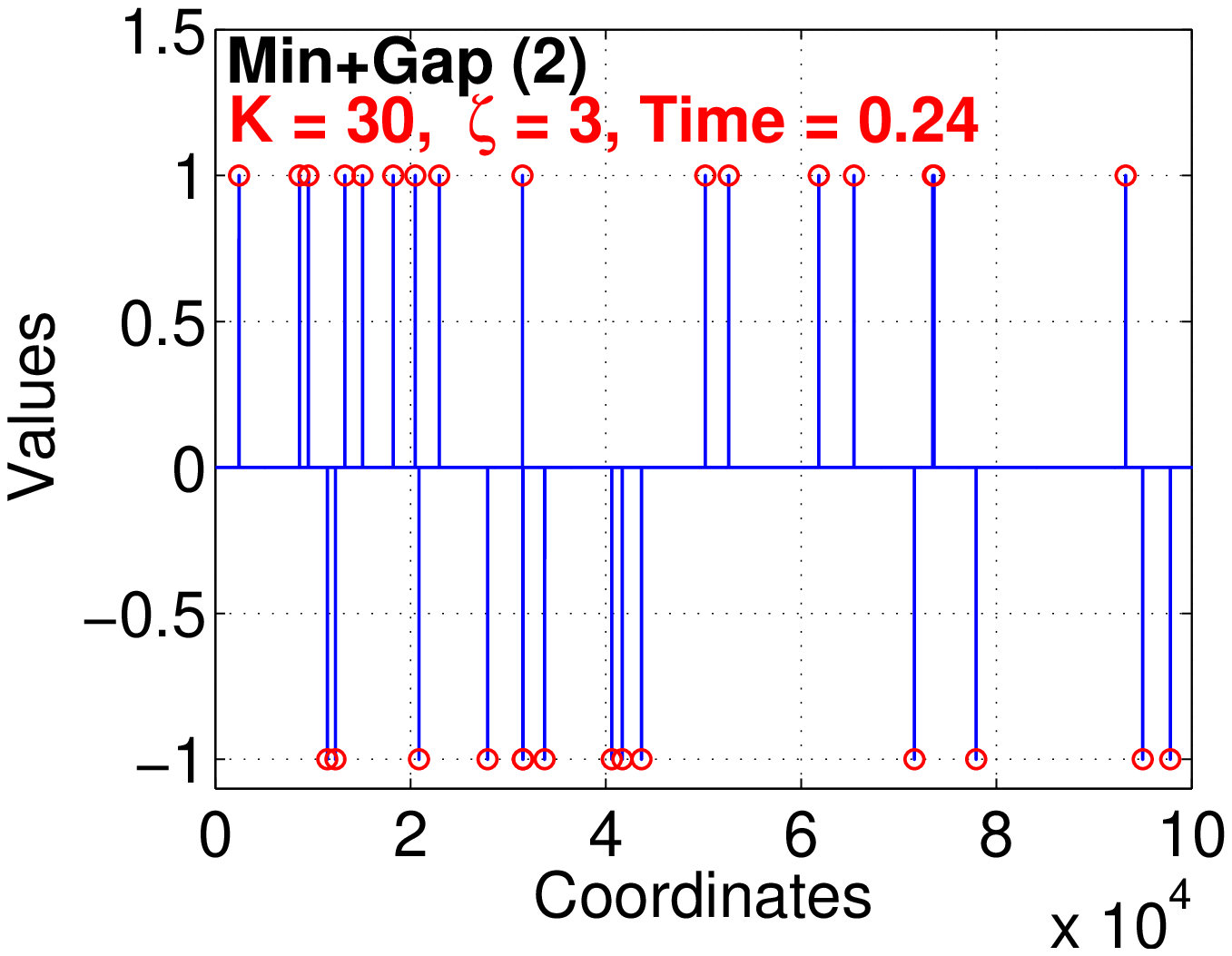}\hspace{0in}
\includegraphics[width=2.5in]{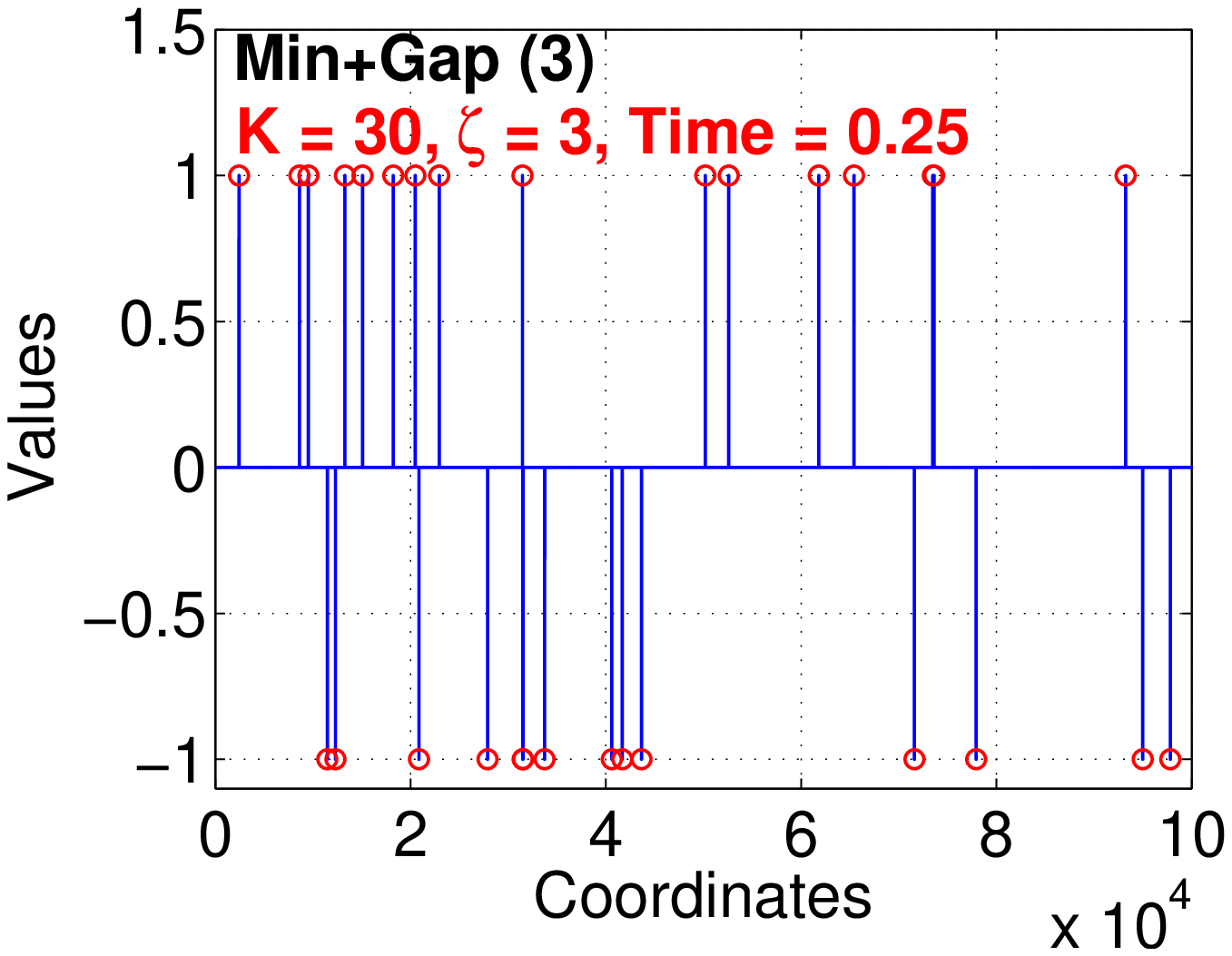}}

\mbox{
\includegraphics[width=2.5in]{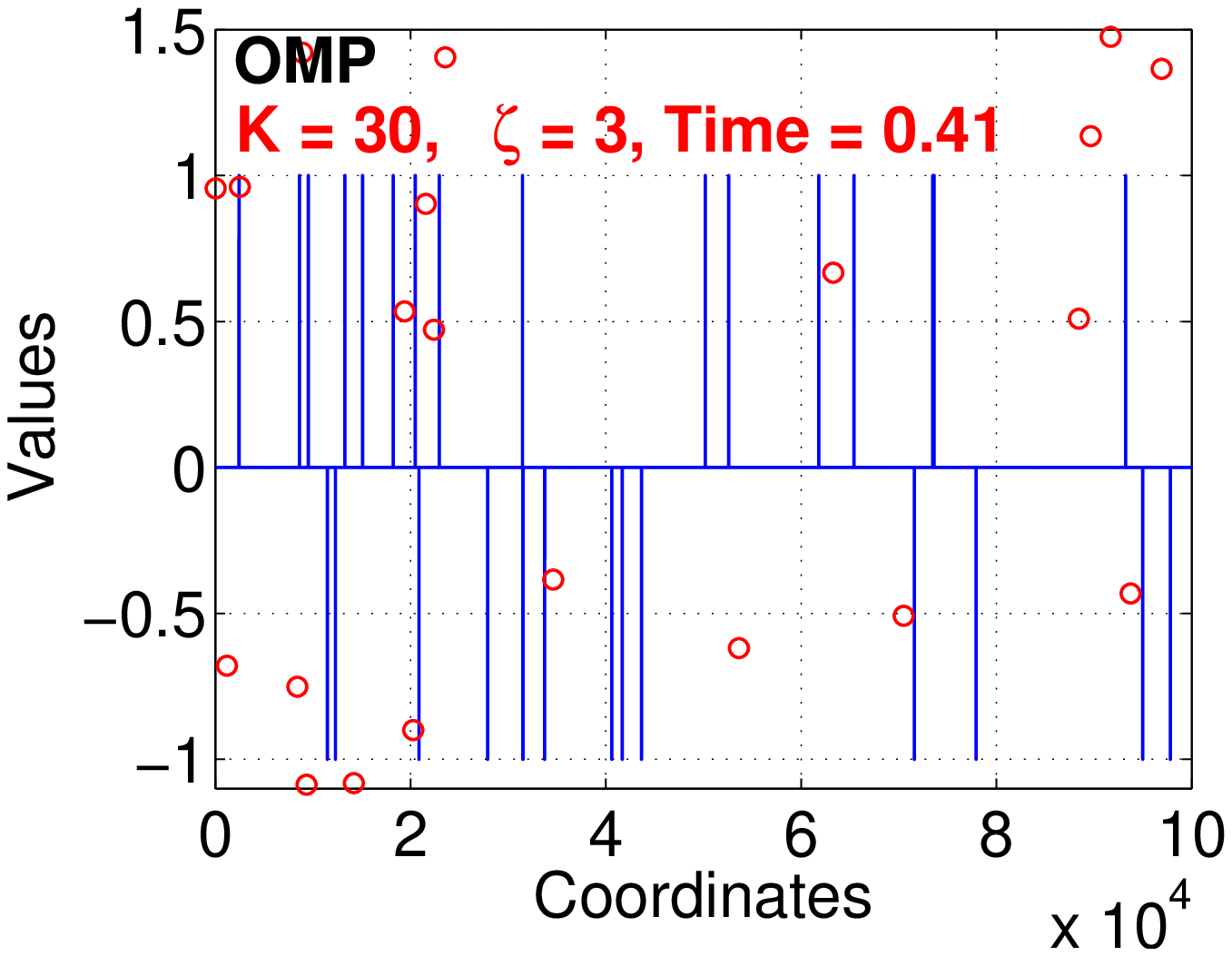}\hspace{0in}
\includegraphics[width=2.5in]{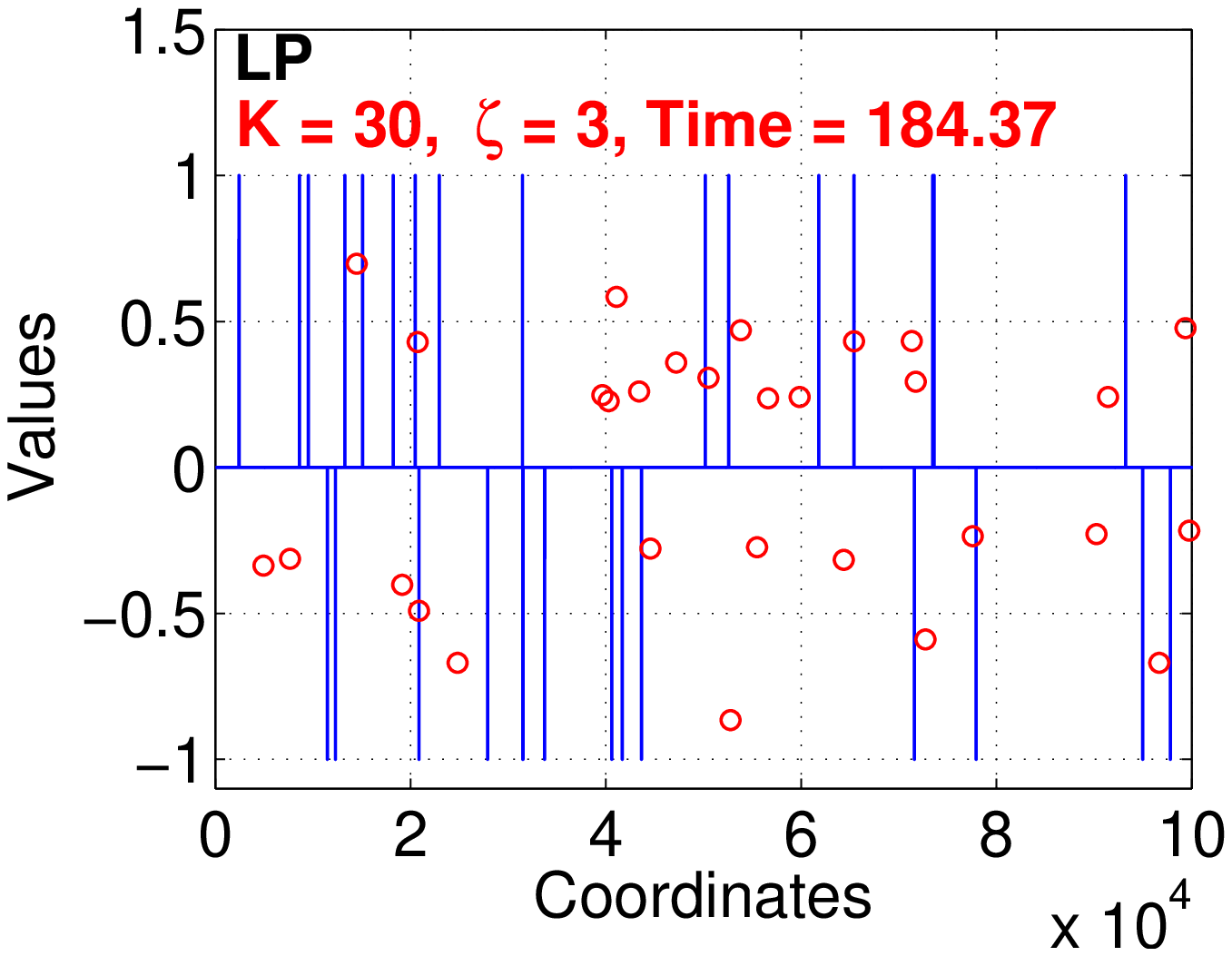}}
\end{center}
\vspace{-0.2in}
\caption{\small Reconstruction results from one simulation, with $N=100000$, $K=30$, $M=M_0/3$ (i.e., $\zeta=3$), and sign signals. Many of the false positives produced by the min estimator are removed by the gap estimator after  1 iteration. The signal is perfectly reconstructed after the second iteration. In comparisons, both OMP and LP perform poorly.}\label{fig_RecSignB3}
\end{figure}

\begin{figure}[h!]
\begin{center}
\mbox{
\includegraphics[width=2.5in]{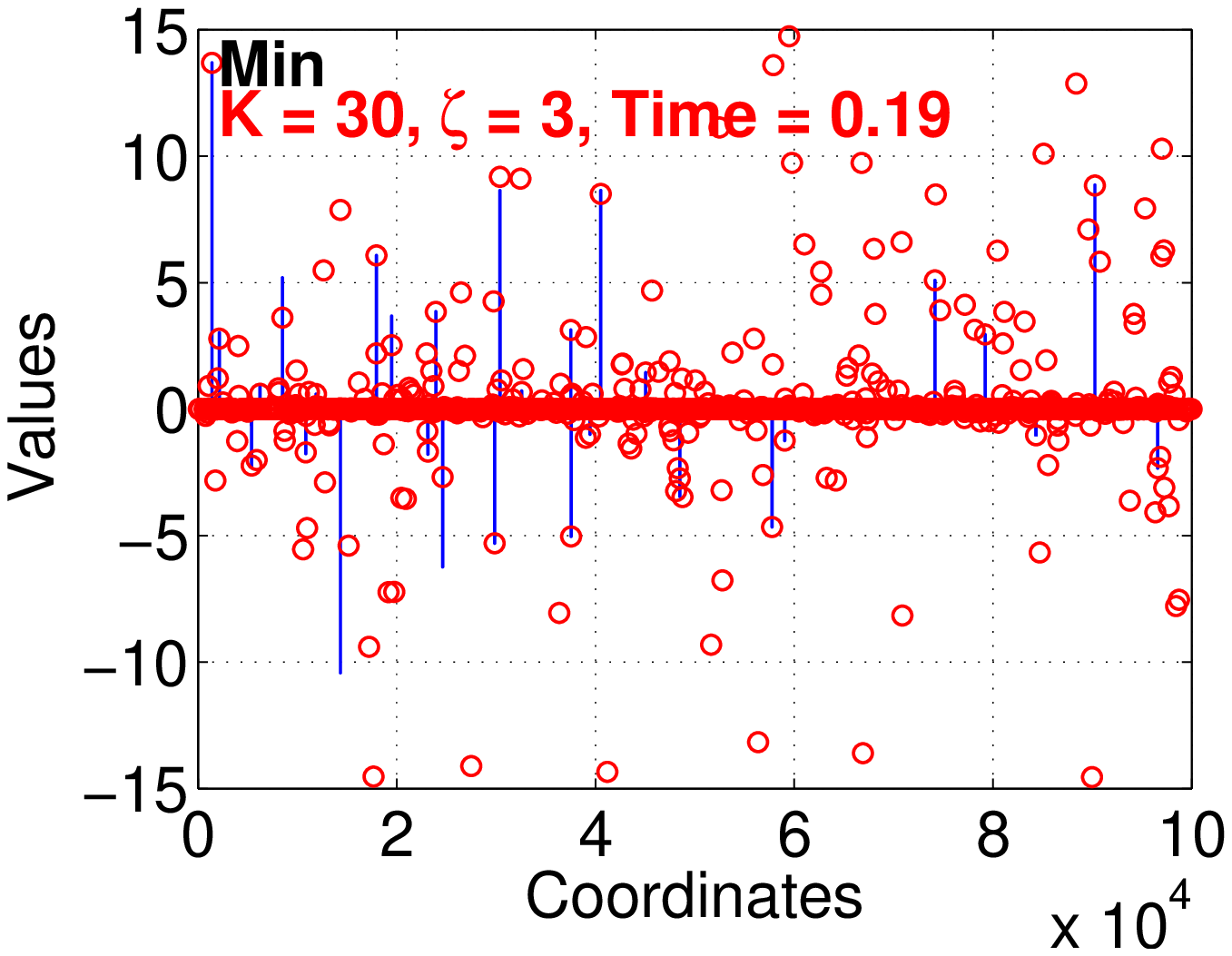}\hspace{0in}
\includegraphics[width=2.5in]{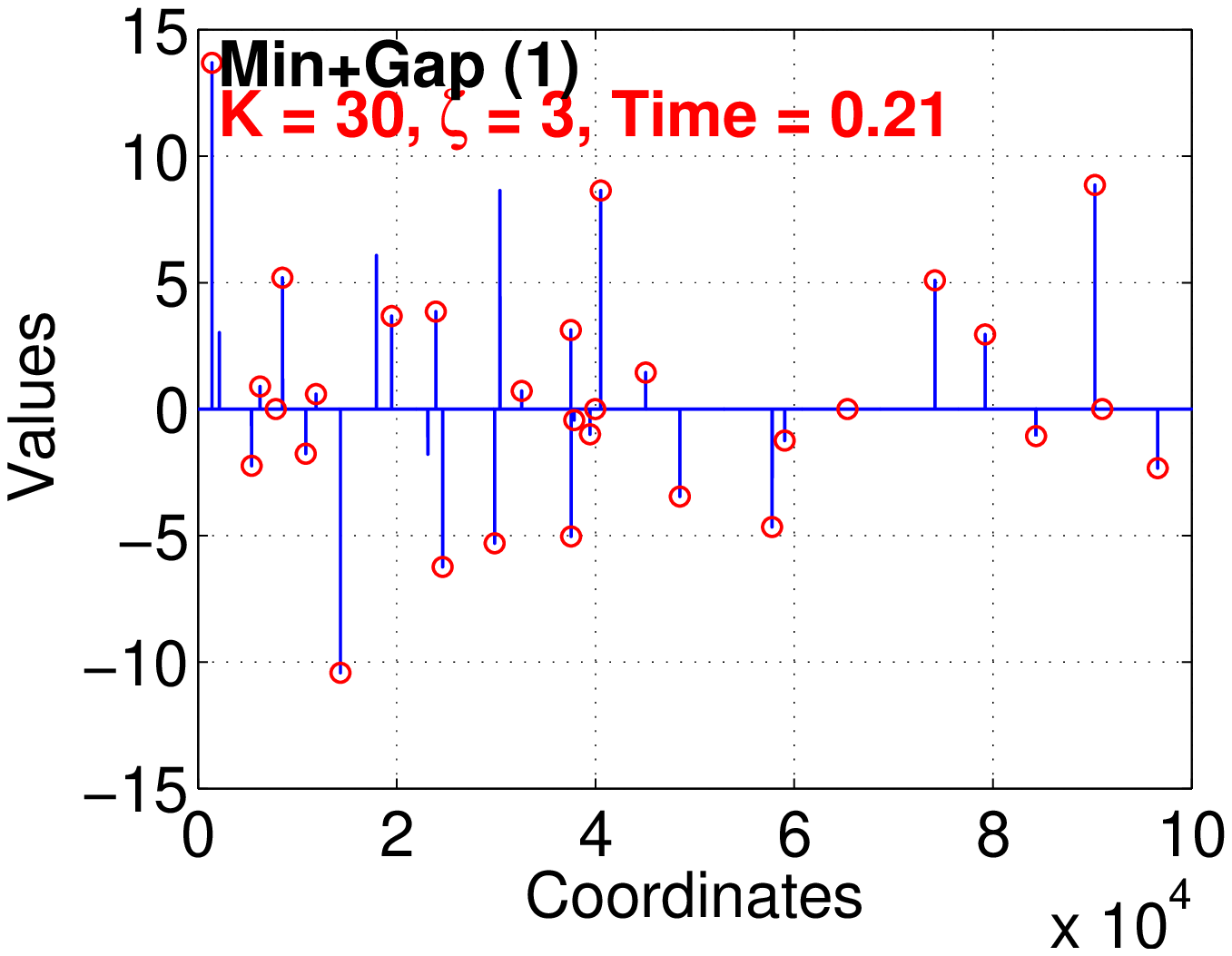}}

\mbox{
\includegraphics[width=2.5in]{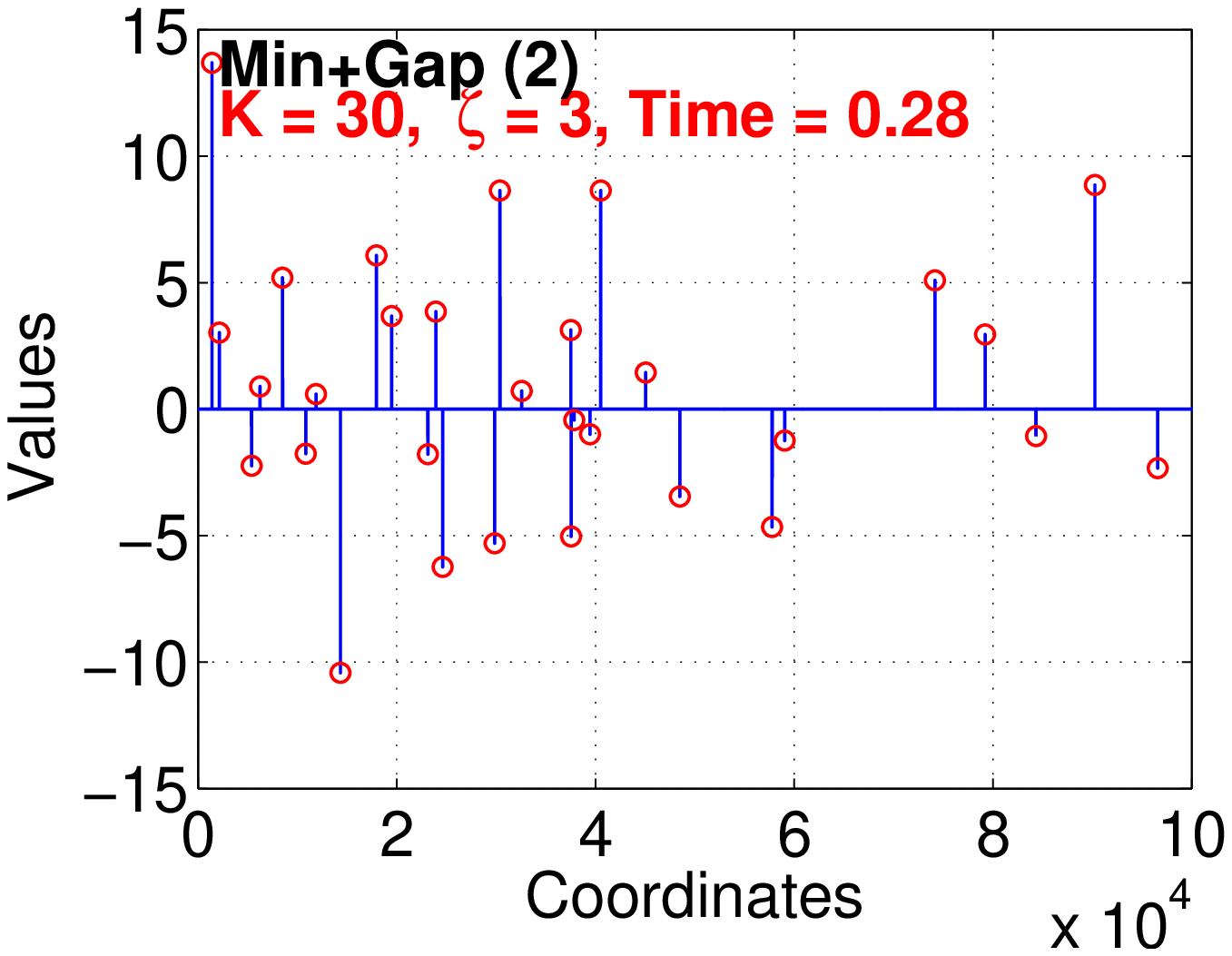}\hspace{0in}
\includegraphics[width=2.5in]{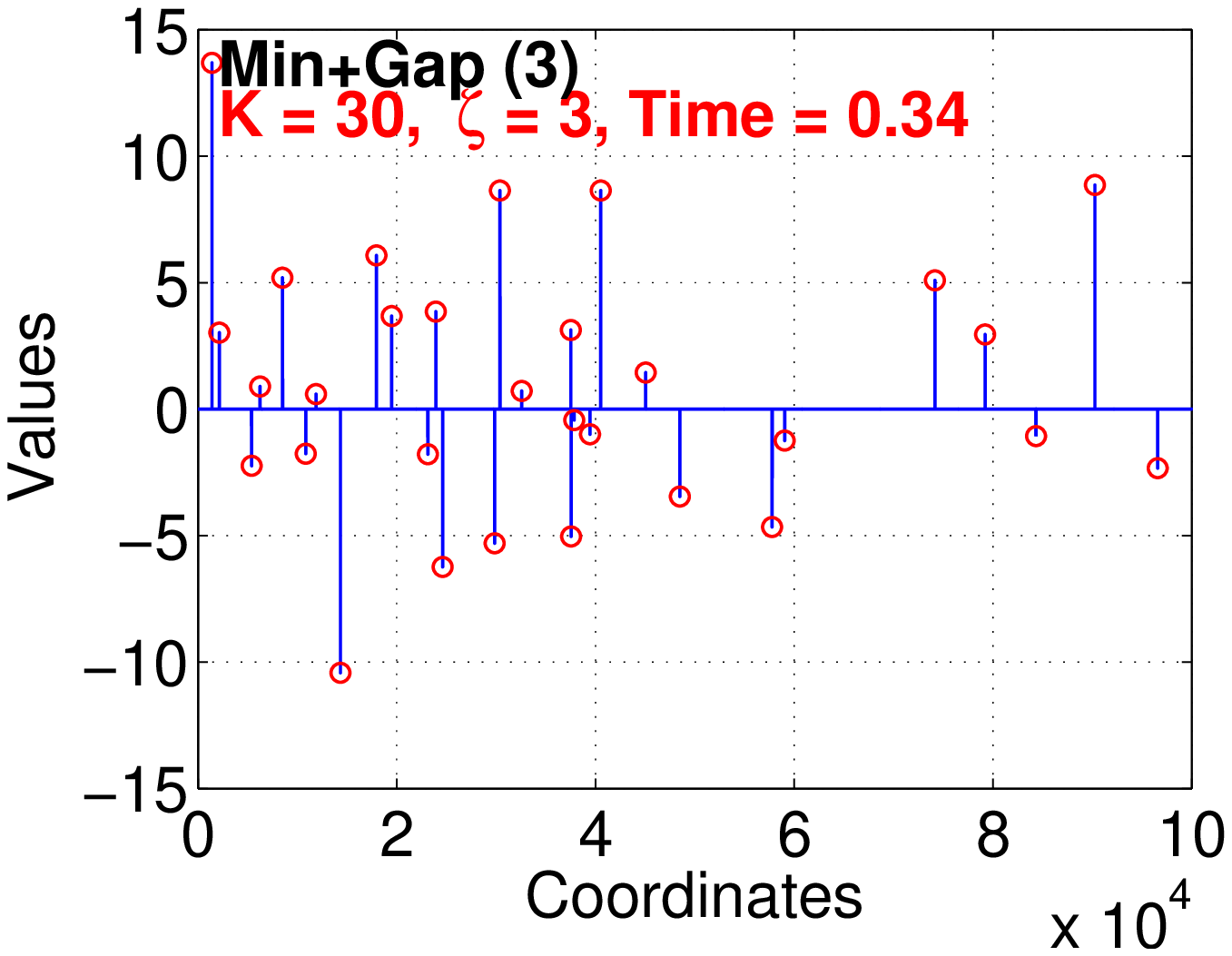}}

\mbox{
\includegraphics[width=2.5in]{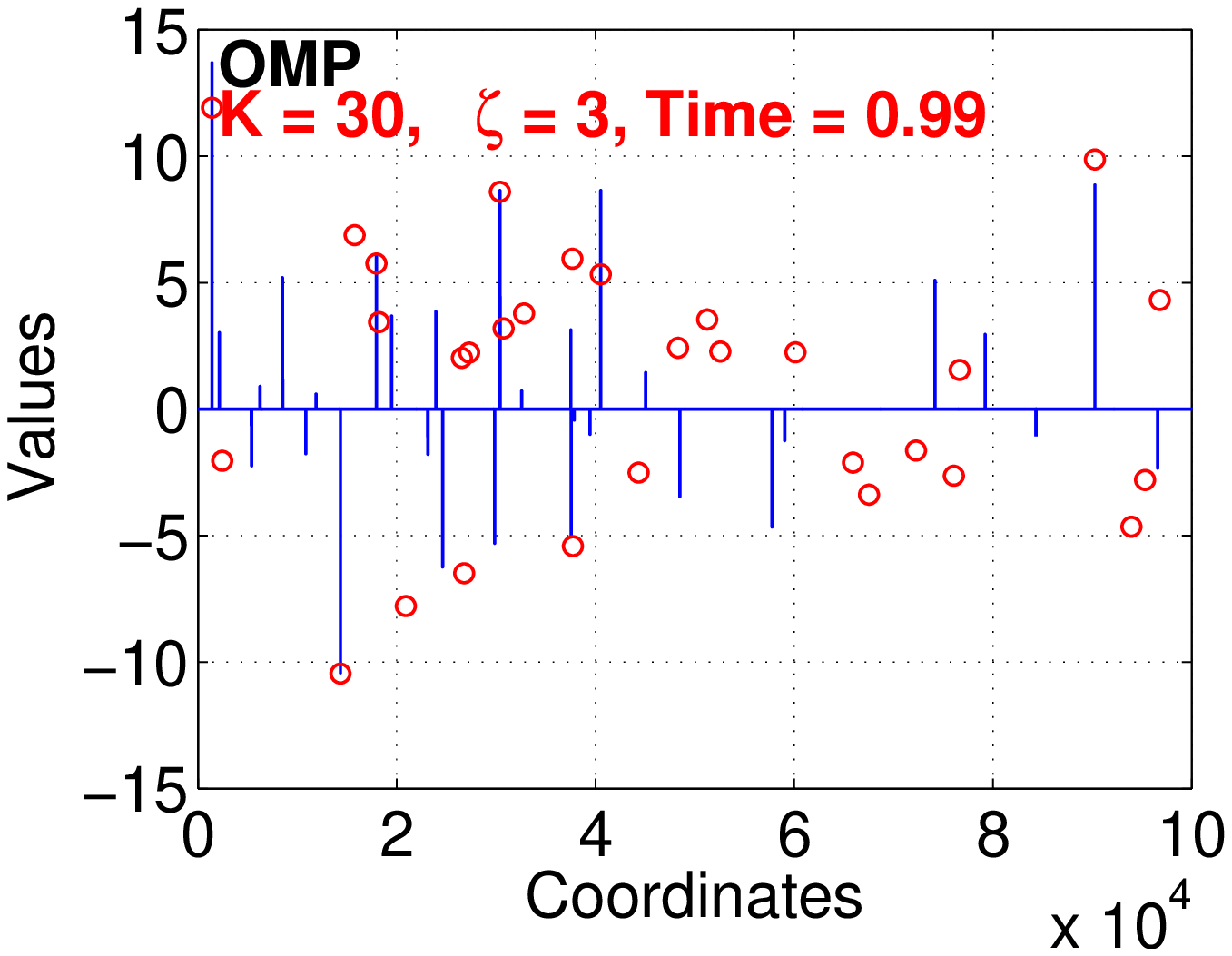}\hspace{0in}
\includegraphics[width=2.5in]{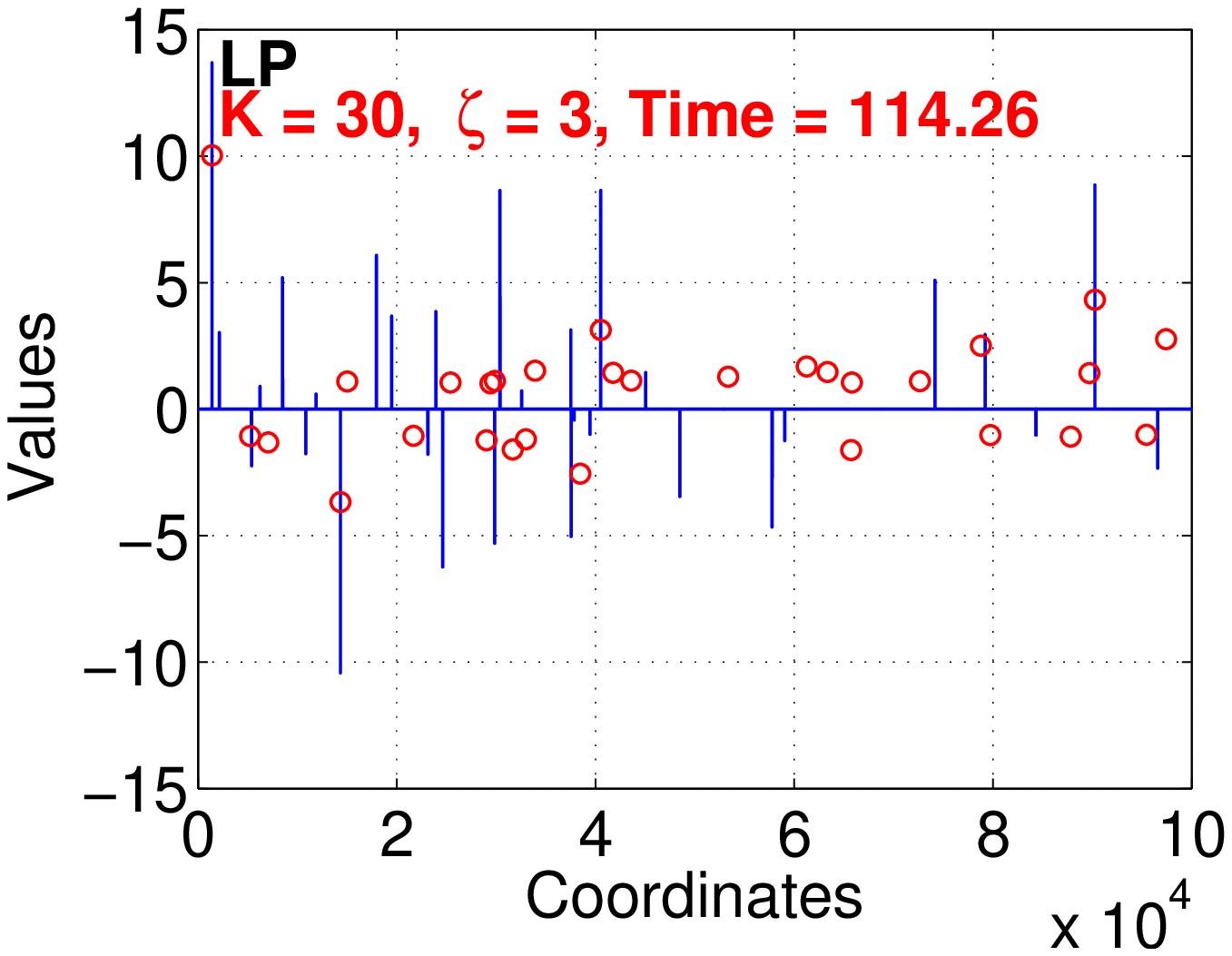}}
\end{center}
\vspace{-0.2in}
\caption{\small Reconstruction results from one simulation, using $N=100000$, $K=30$, $M=M_0/3$ (i.e., $\zeta=3$), and Gaussian signals. Our method (using two iterations) can still perfectly reconstruct the signal.}\label{fig_RecGausB3}
\end{figure}

\begin{figure}[h!]
\begin{center}
\mbox{
\includegraphics[width=2.5in]{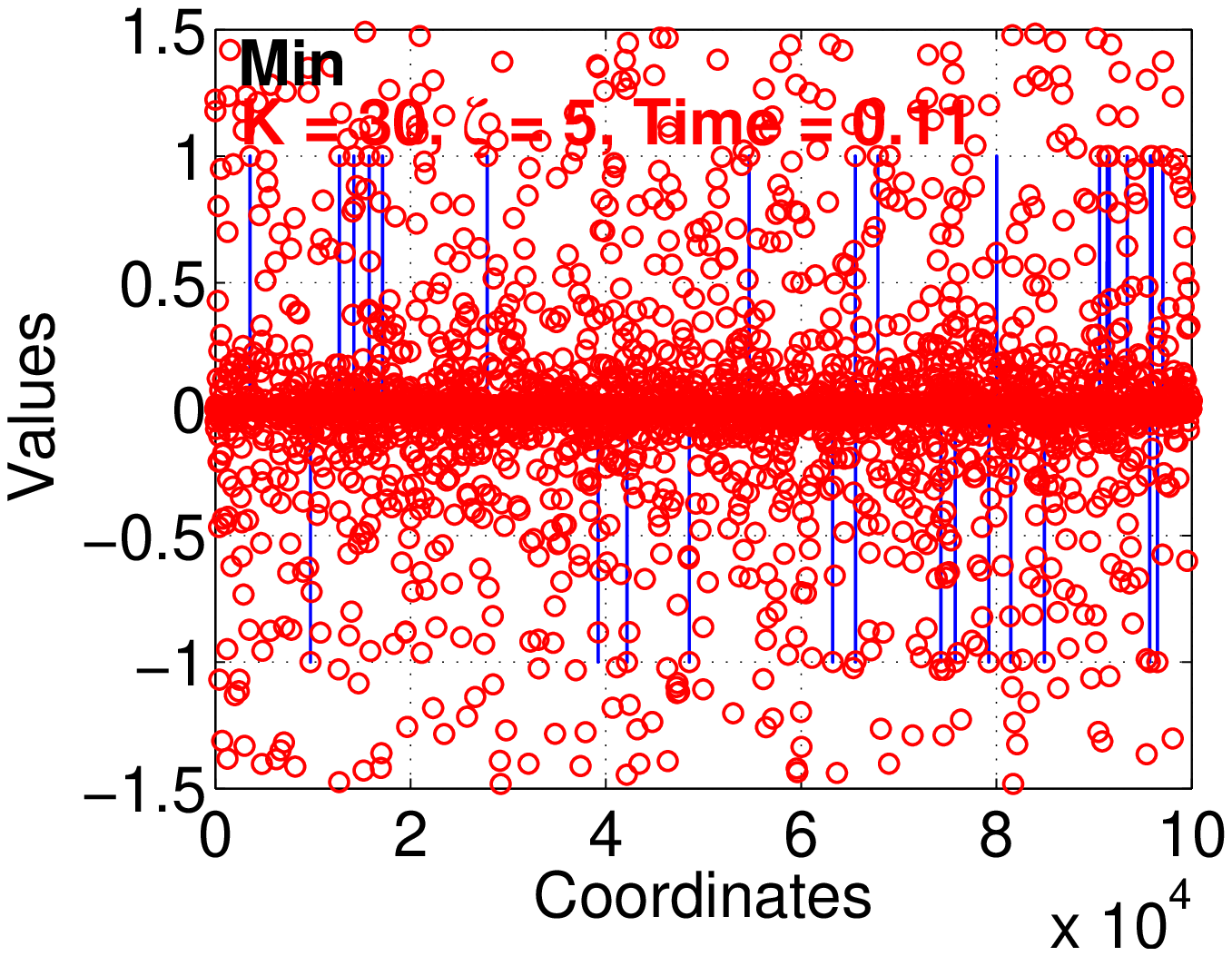}\hspace{0in}
\includegraphics[width=2.5in]{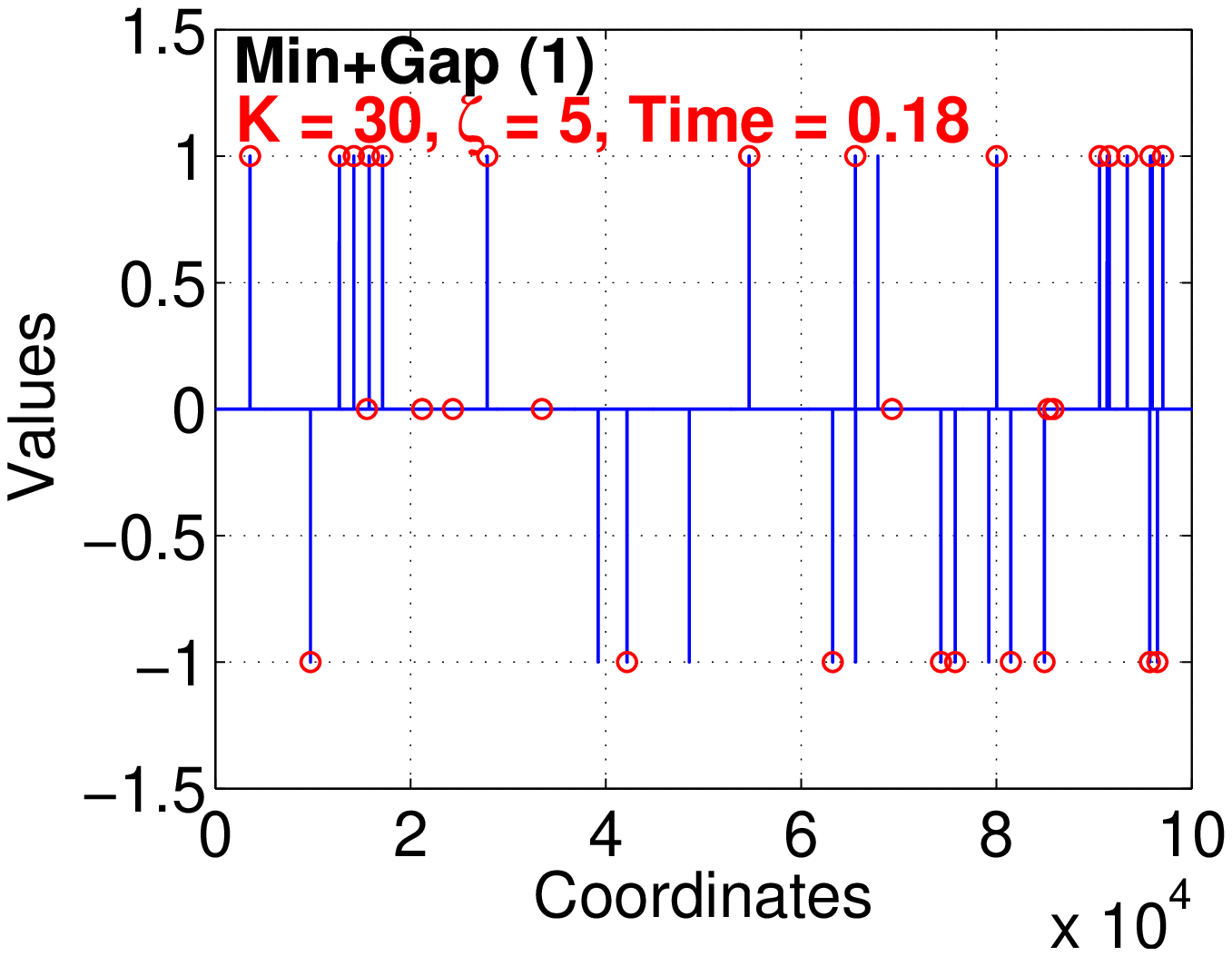}}

\mbox{
\includegraphics[width=2.5in]{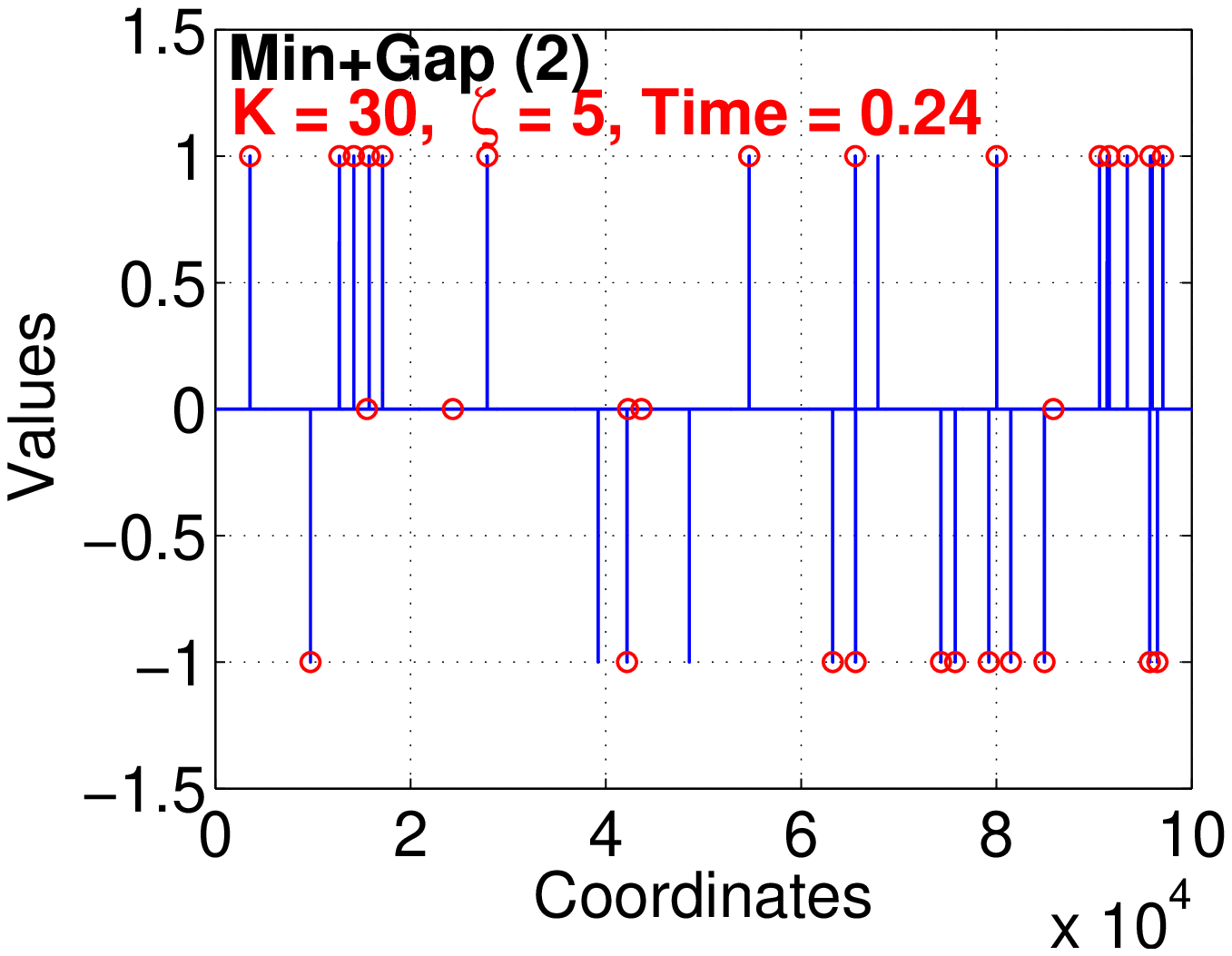}\hspace{0in}
\includegraphics[width=2.5in]{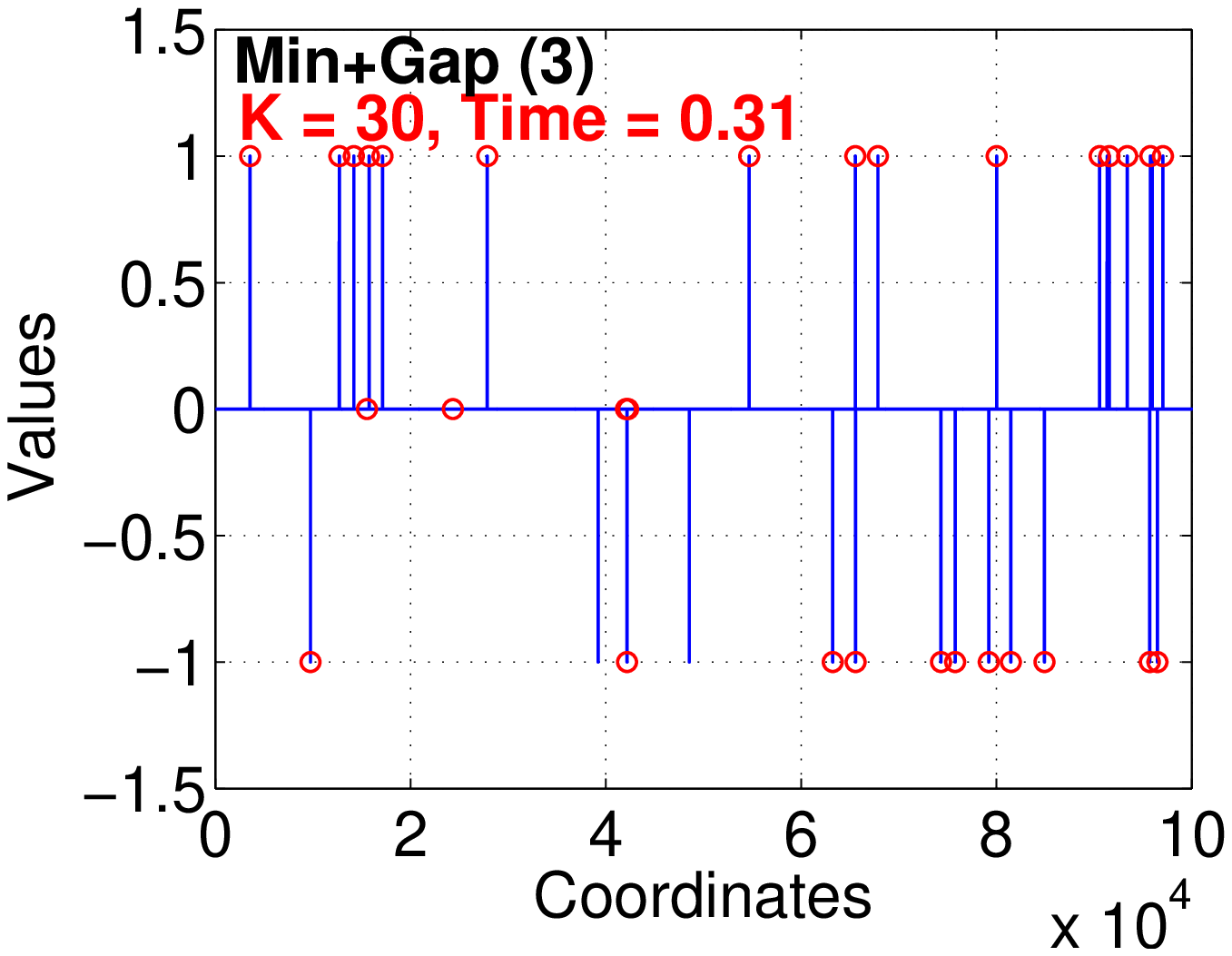}}

\mbox{
\includegraphics[width=2.5in]{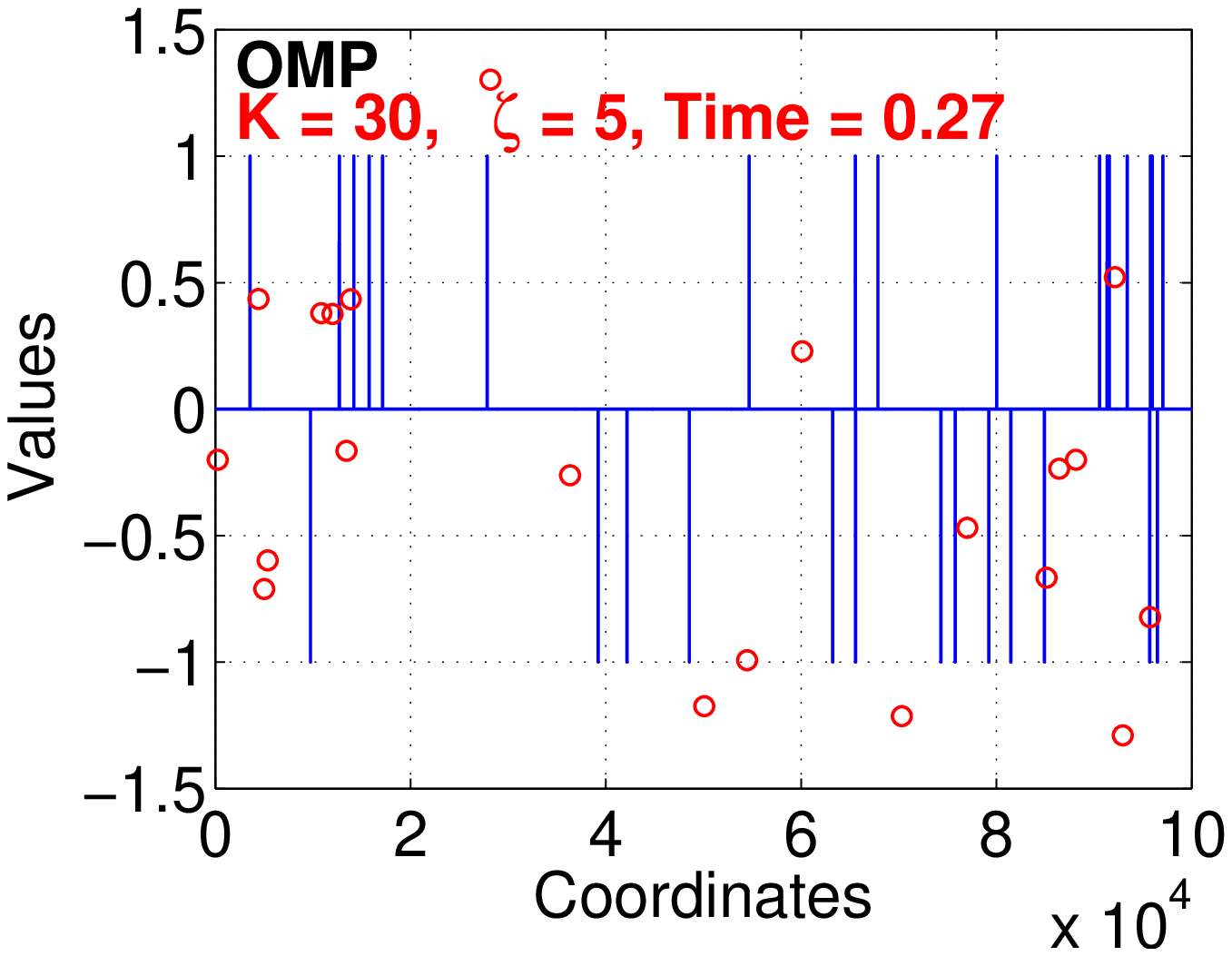}\hspace{0in}
\includegraphics[width=2.5in]{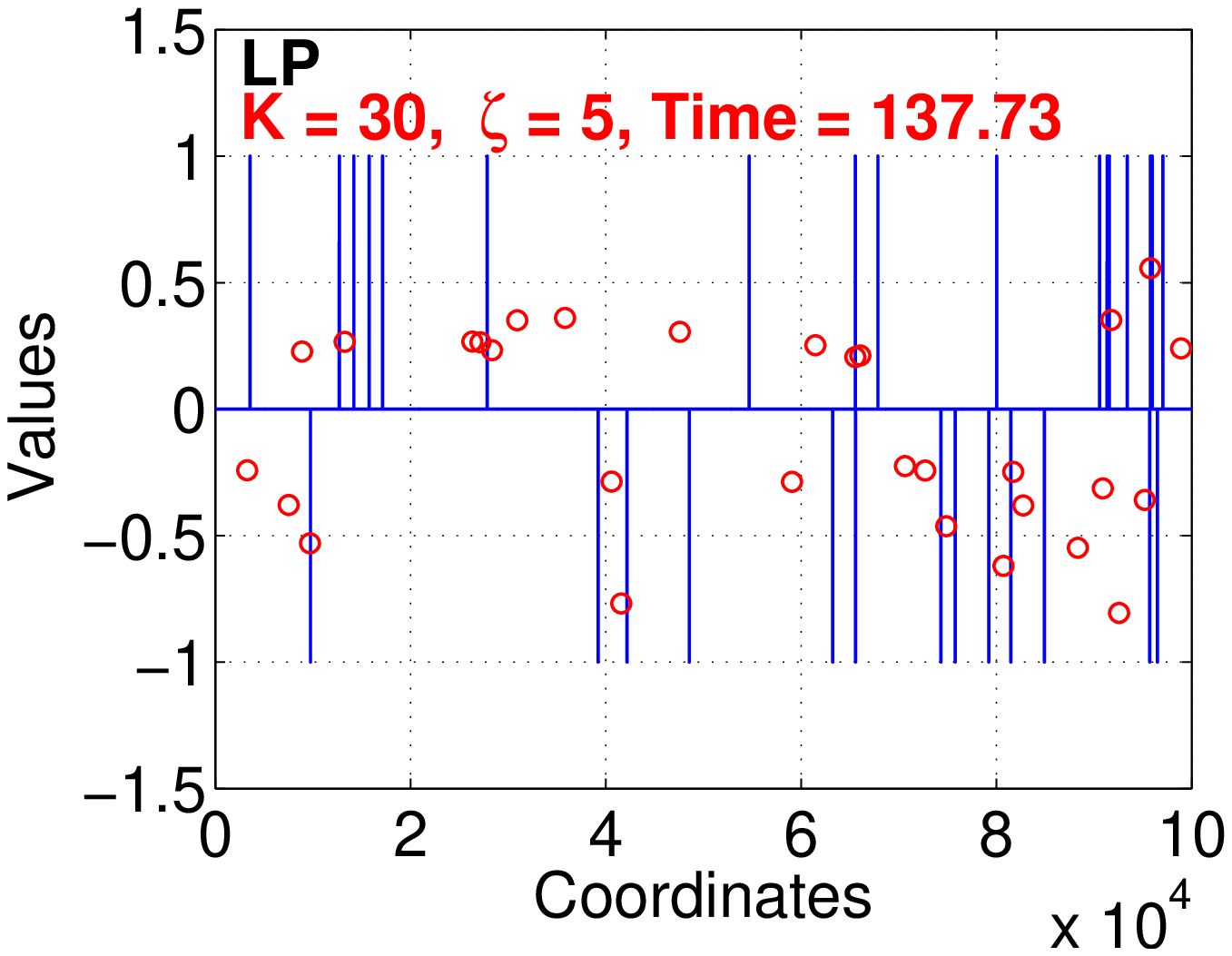}}
\end{center}
\vspace{-0.2in}
\caption{\small Reconstruction results from one simulation, with $N=100000$, $K=30$, $M=M_0/5$ (i.e., $\zeta=5$), and sign signals. Since $M$ is not large enough, a small fraction of  the nonzero coordinates are not  reconstructed by our method. In comparisons, both OMP and LP perform very poorly in that none of the reported nonzero coordinates is correct.}\label{fig_RecSignB5}
\end{figure}

\begin{figure}[h!]
\begin{center}
\mbox{
\includegraphics[width=2.5in]{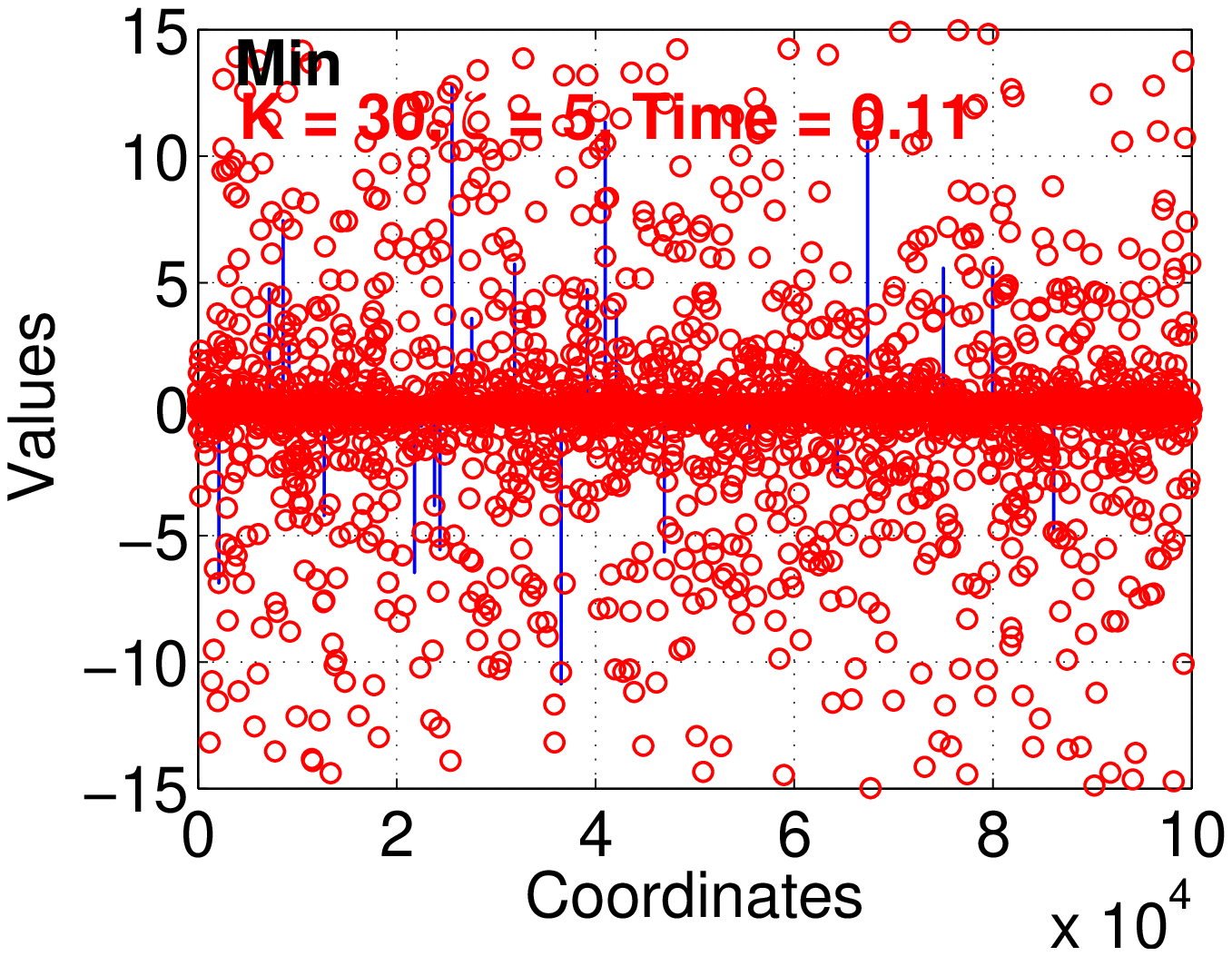}\hspace{0in}
\includegraphics[width=2.5in]{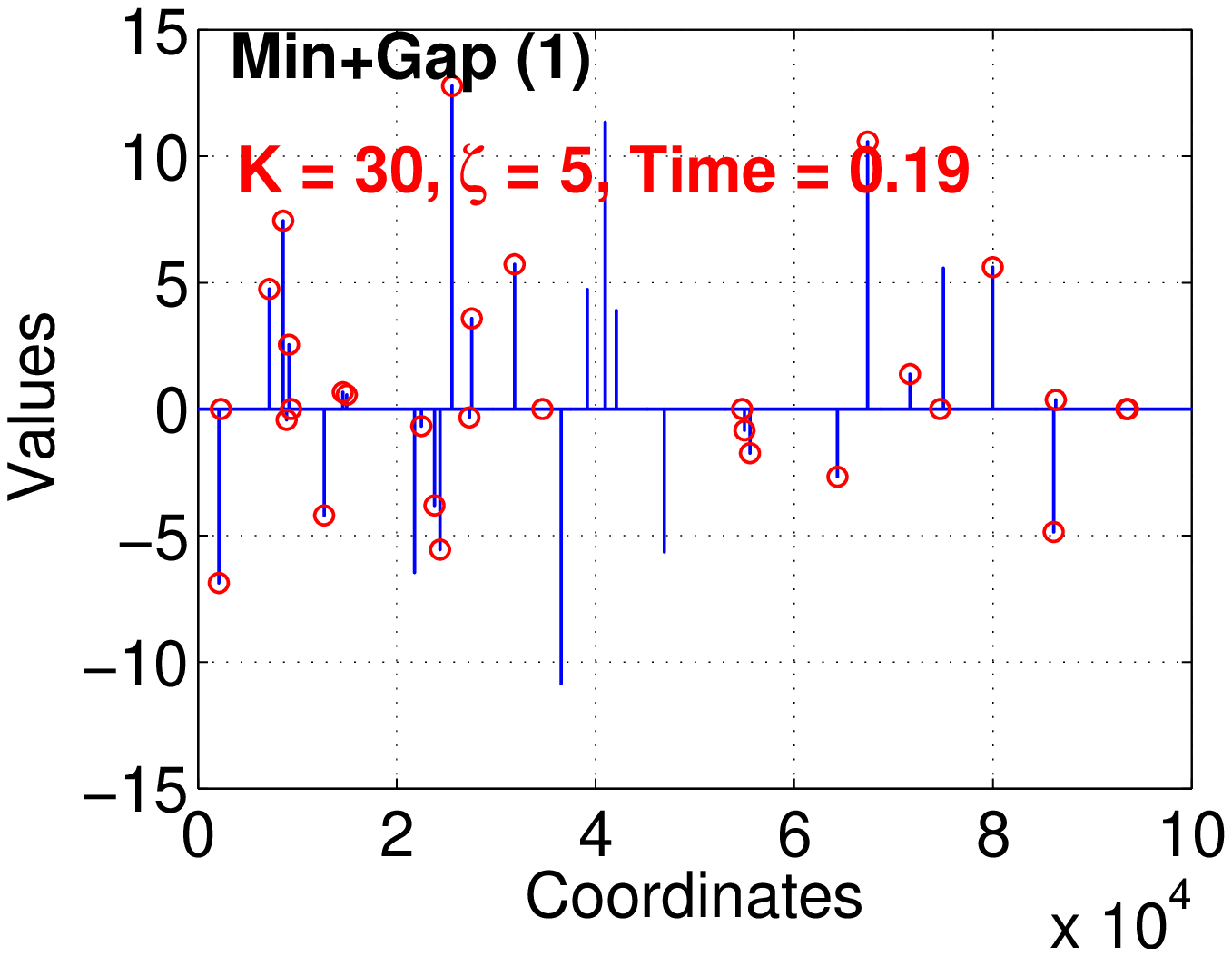}}

\mbox{
\includegraphics[width=2.5in]{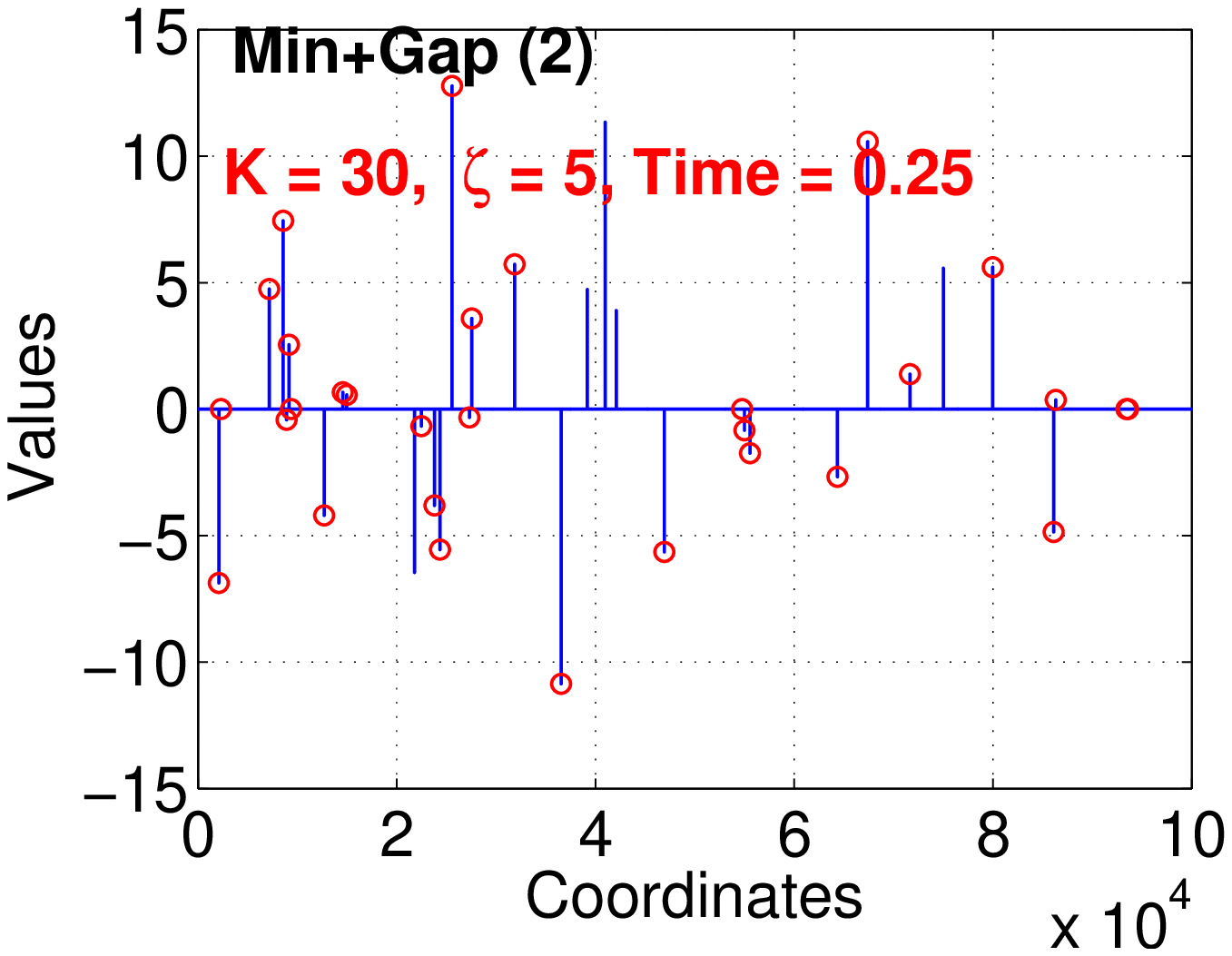}\hspace{0in}
\includegraphics[width=2.5in]{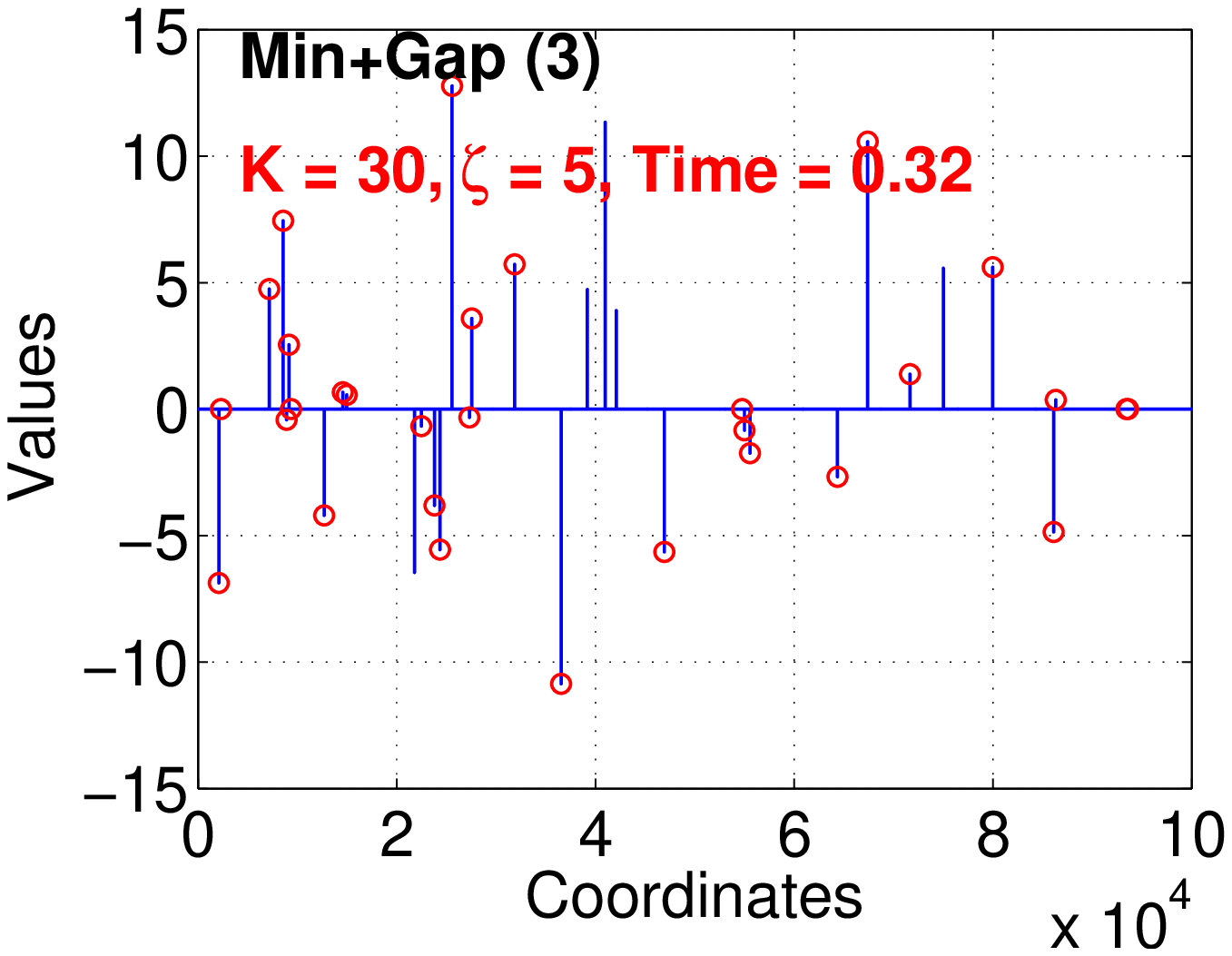}}

\mbox{
\includegraphics[width=2.5in]{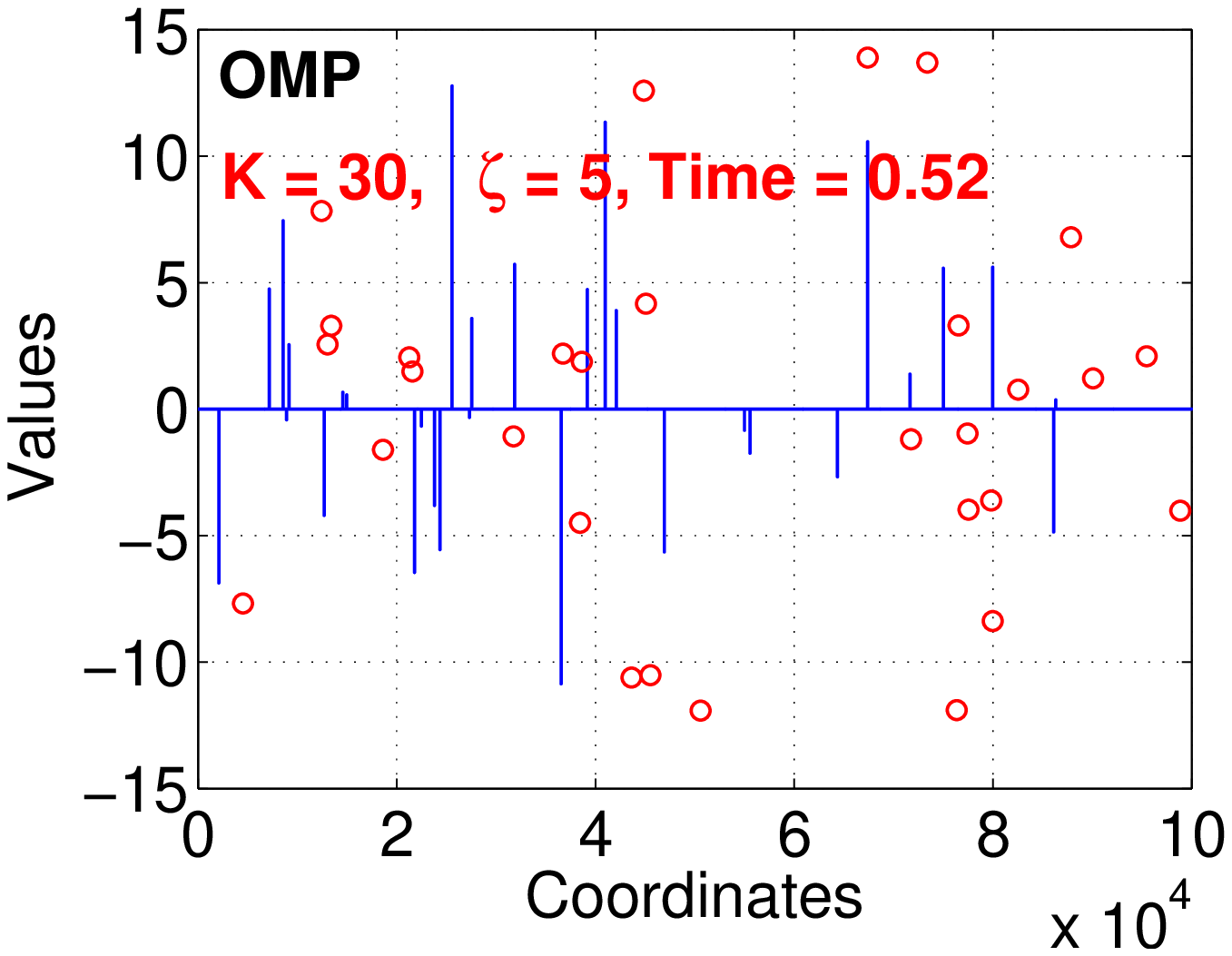}\hspace{0in}
\includegraphics[width=2.5in]{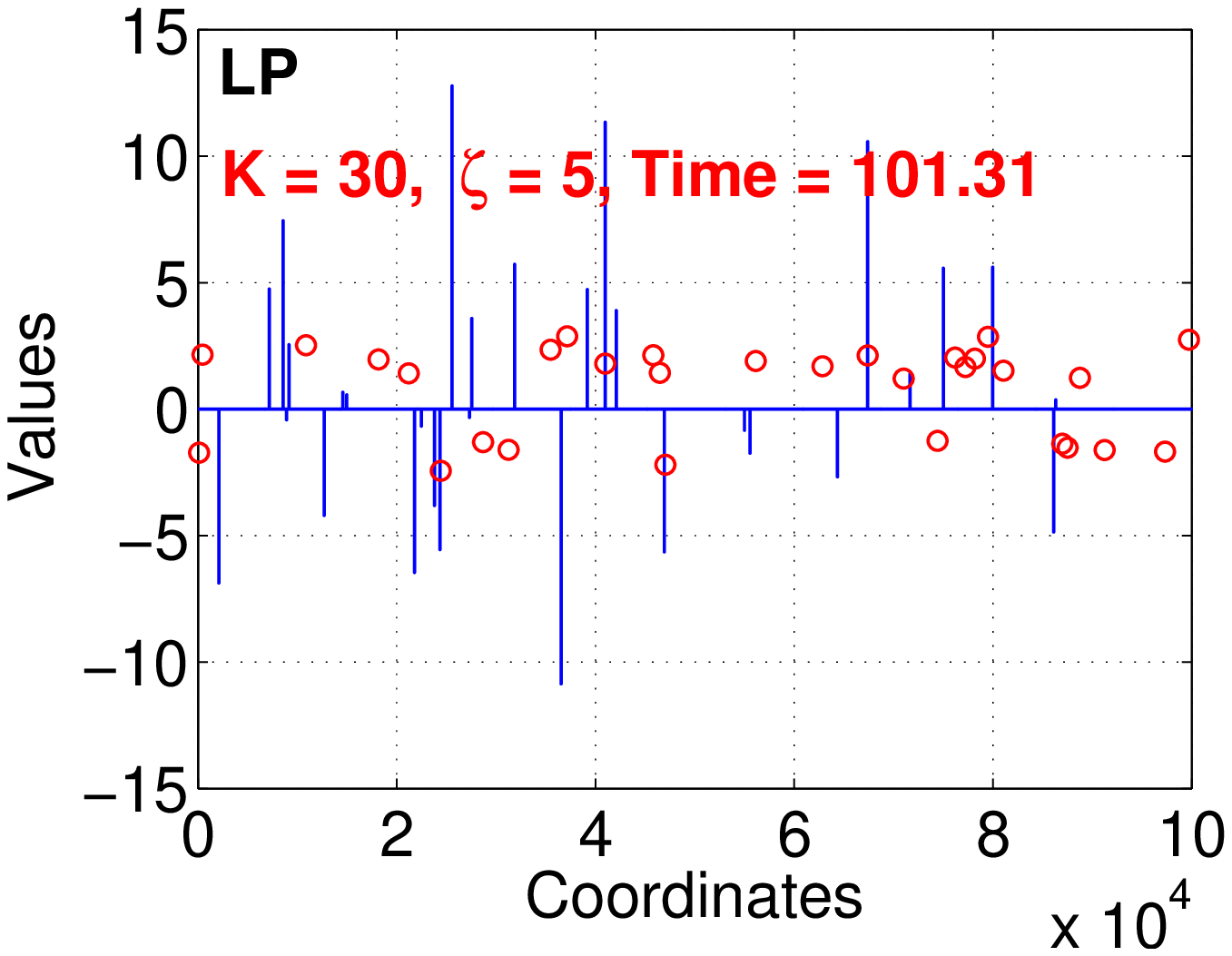}}
\end{center}
\vspace{-0.2in}
\caption{\small Reconstruction results from one simulation, using $N=100000$, $K=30$, $M=M_0/5$ (i.e., $\zeta=5$), and Gaussian signals. Again, since $M$ is not large enough, a small fraction of  the nonzero coordinates are not reconstructed by our method. In comparisons, both OMP and LP perform very poorly.}\label{fig_RecGausB5}
\end{figure}

Figures~\ref{fig_RecSignB1} to~\ref{fig_RecGausB5} present several instances of  simulations, for $N=100000$ and $K=30$. In each simulation (each figure), we generate the heavy-tailed design matrix $\mathbf{S}$ (with $\alpha = 0.03$) and the Gaussian design matrix $\mathbf{S}_{(2)}$ (with $\alpha = 2$), using the same random variables ($w$'s and $u$'s) as  in (\ref{eqn_stable_sample}). This  provides shoulder-by-shoulder comparisons of our method with LP and OMP. We use the   $l_1$-magic package~\cite{Report:L1Magic} for LP, as we find that $l_1$-magic produces very similar recovery results as the Matlab build-in LP solver and is noticeably faster. Since $l_1$-magic is popular, this should facilitate reproducing our work by interested readers.\\

In Figures~\ref{fig_RecSignB1} and~\ref{fig_RecGausB1}, we let $M = M_0$ (i.e., $\zeta = 1$). For this $M$,  all  methods perform well, for both {sign signal} and {Gaussian signal}. The \textbf{left-top} panels of Figures~\ref{fig_RecSignB1} and~\ref{fig_RecGausB1} show that the {\em  minimum estimator} $\hat{x}_{i,min}$ can precisely identify all the nonzero coordinates. The \textbf{right-top} panels show that the {\em gap estimator} $\hat{x}_{i,gap}$ applied on the coordinates identified by $\hat{x}_{i,min}$, can accurately estimate the magnitudes. The label ``\textbf{min+gap(1)}'' means only one iteration is performed (which is good enough for $\zeta=1$).

The \textbf{bottom} panels of Figures~\ref{fig_RecSignB1} and~\ref{fig_RecGausB1}  show that both OMP and LP also perform well when $\zeta=1$.  OMP is noticeably more costly than our method (even though $K$ is small) while LP is significantly much more expensive than all other methods.\\

We believe these plots of sample instances provide  useful information, especially when $M\ll M_0$. If $M$ is too small, then all methods will ultimately fail, but the failure patterns are important, for example, a ``catastrophic'' failure such that none of the reported nonzeros is correct will be very undesirable.  Figure~\ref{fig_RecSignB5} and Figure~\ref{fig_RecGausB5} will show that our method does not fail catastrophically even with $M = M_0/5$.

\clearpage

Simulations in Figures~\ref{fig_RecSignB3} and~\ref{fig_RecGausB3} use $M = M_0/3$ (i.e., $\zeta = 3$). The minimum estimator $\hat{x}_{i,min}$ outputs a significant number of false positives but our method can still perfectly reconstruct signal using the gap estimator with one additional iteration (i.e., \textbf{Min+Gap(2)}). In comparisons,  both LP and OMP perform  poorly. Furthermore,  Figures~\ref{fig_RecSignB5} and~\ref{fig_RecGausB5} use $M = M_0/5$ (i.e., $\zeta=5$) to demonstrate the robustness of our algorithm. As $M$ is not large enough, a small fraction of nonzero coordinates are not  recovered by our method, but there are no catastrophic failures. This point is of course already illustrated  in Figure~\ref{fig_Zip54}. In comparison, when $M=M_0/5$, both LP and OMP perform very poorly.

Note that the decoding times for our method and OMP are fairly consistent in that smaller $M$ results in faster decoding. However, the run times of LP can vary substantially, which should have to do with the quality of measurements and implementations. Also, note that we only display the top-$K$ (in absolute values) reconstructed coordinates for all methods.

\clearpage

\begin{figure}[h!]
\begin{center}
\mbox{
\includegraphics[width=2.4in]{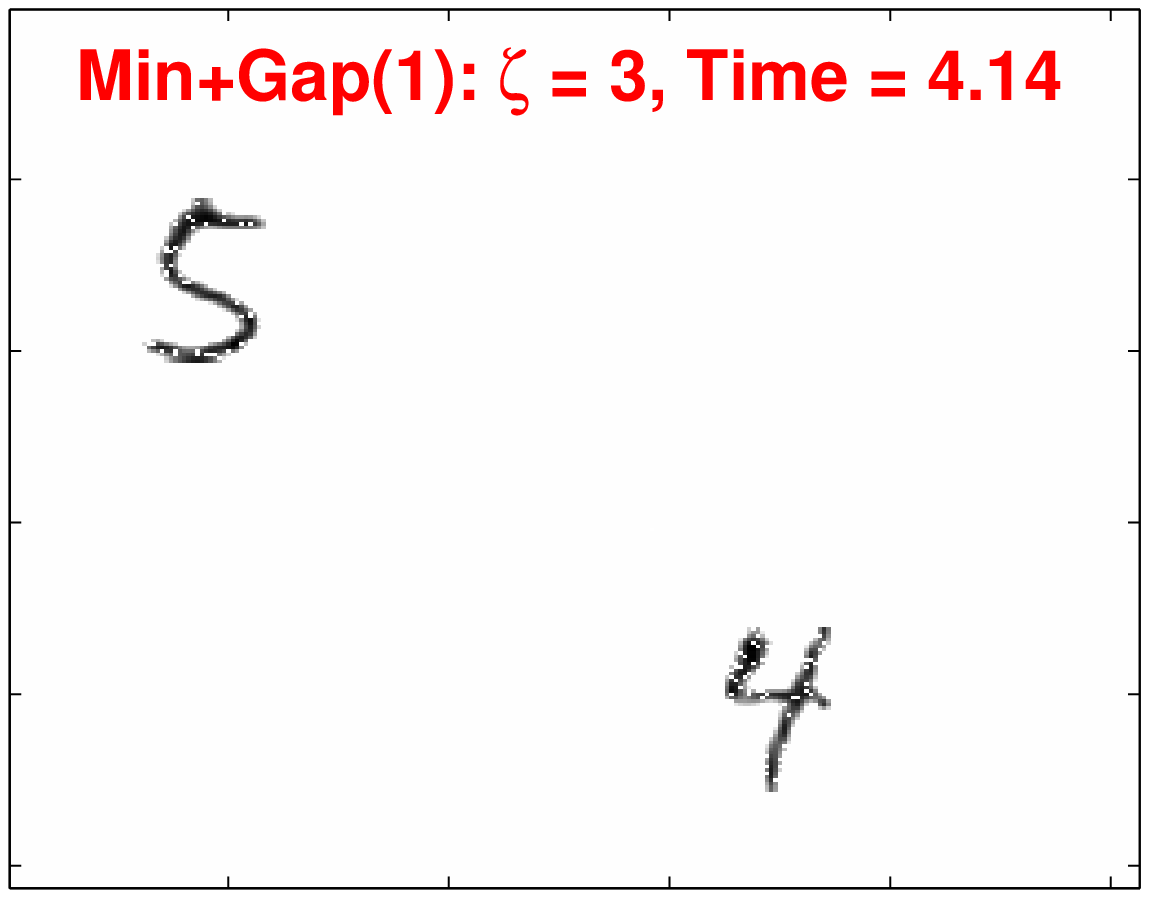}\hspace{-0.25in}
\includegraphics[width=2.4in]{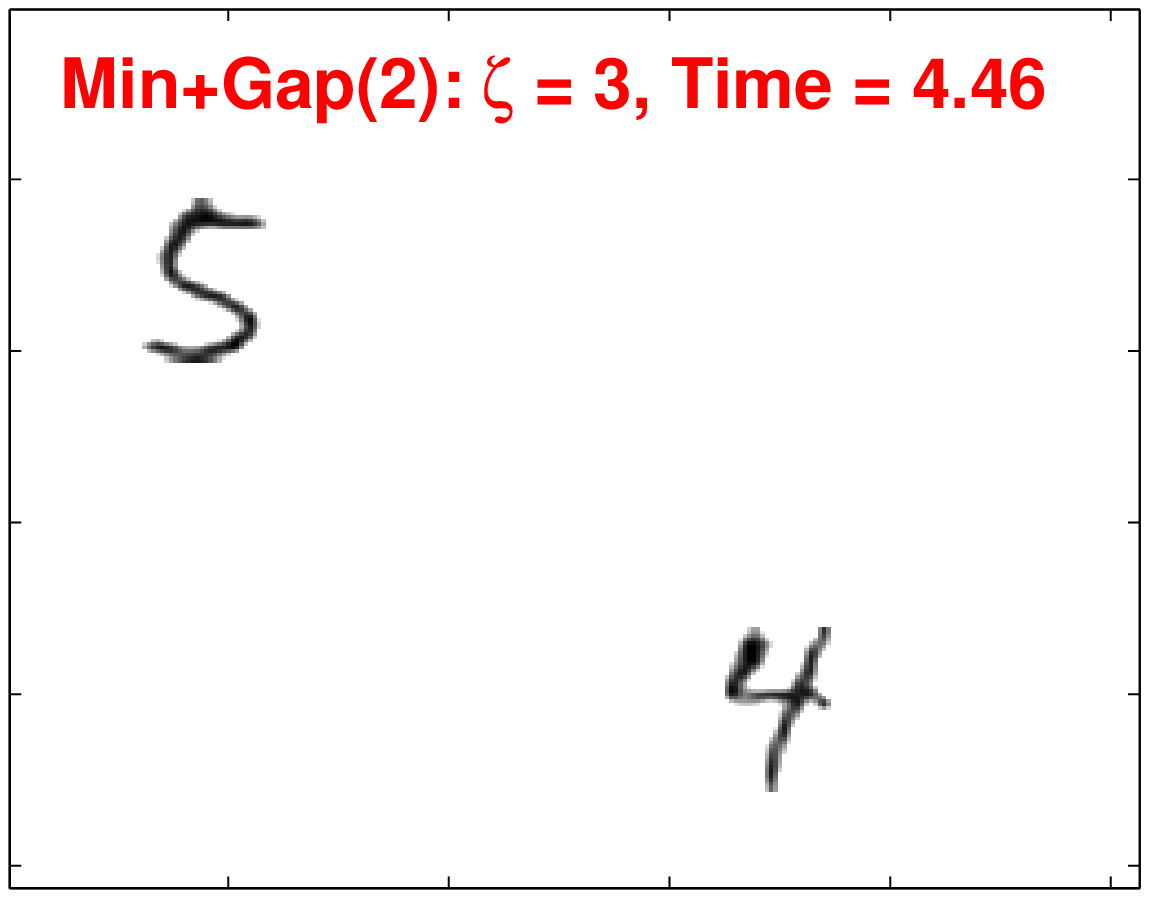}\hspace{-0.25in}
\includegraphics[width=2.4in]{Zip54B3Gap3.eps}
}
\mbox{
\includegraphics[width=2.4in]{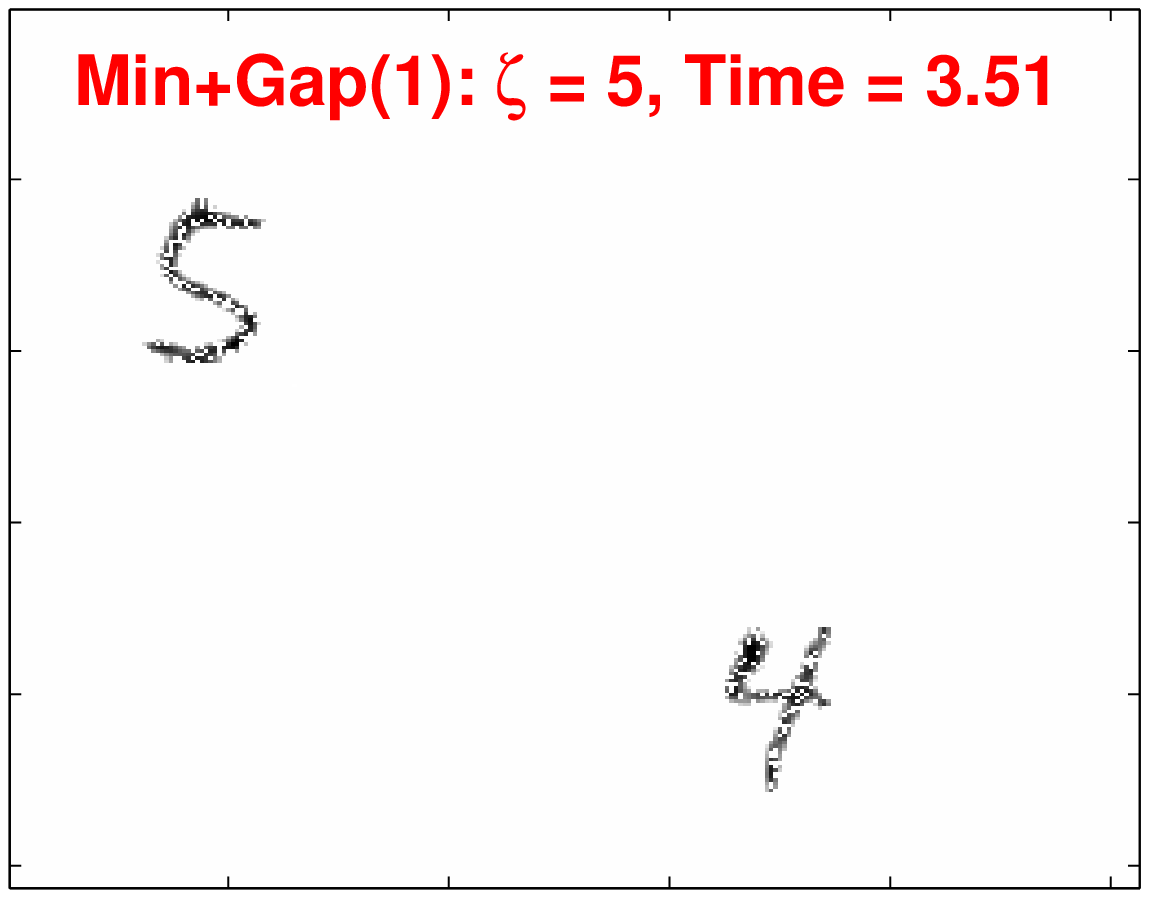}\hspace{-0.25in}
\includegraphics[width=2.4in]{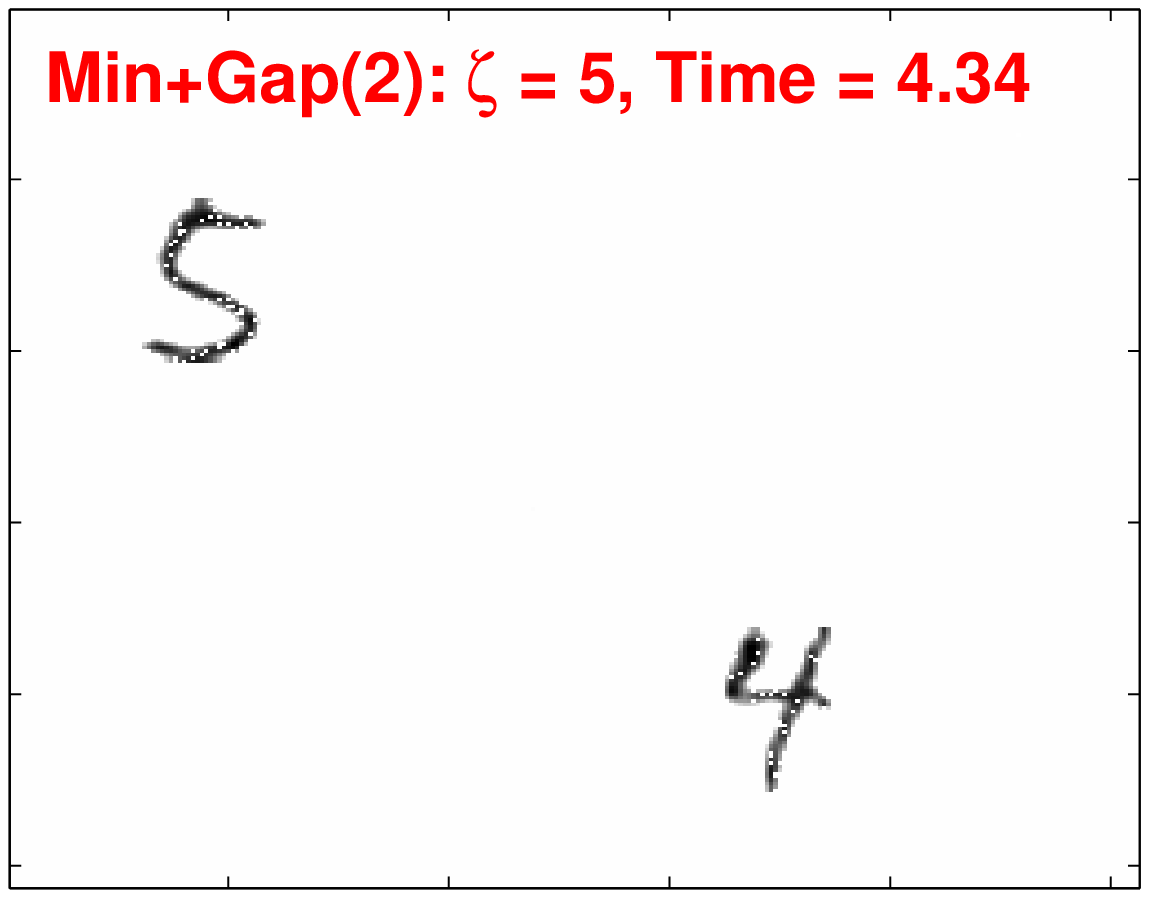}\hspace{-0.25in}
\includegraphics[width=2.4in]{Zip54B5Gap3.eps}
}
\end{center}
\vspace{-0.2in}
\caption{This is the continuation of the example in Figure~\ref{fig_Zip54}, for $M=M_0/3$ (top panels) and $M=M_0/5$ (bottom panels), to demonstrate that the iterative procedure improves the quality of signal reconstructions.}\label{fig_Zip54Gap123}
\end{figure}

Finally, Figure~\ref{fig_Zip54Gap123} supplements the example in Figure~\ref{fig_Zip54} (which only displays ``Min+Gap(3)'')  by presenting the results of ``Min+Gap(1)'' and ``Min+Gap(2)'', to illustrate that our proposed iterative procedure improves the quality of reconstructions.

\subsection{Summary Statistics from Simulations}

We repeat the simulations many times to compare the aggregated  reconstructed errors and run times. In this set of experiments, we choose $(N,\ K)$ from $\{(5000,50), (10000,50), (10000,100), (100000,100)\}$ combinations. We choose $M = M_0/\zeta$ with $\zeta \in \{1, 1.3, 2, 3, 4, 5\}$. We again experiment with both Gaussian $Normal(0,5^2)$ signals and sign signals.

For each setting, we repeat the simulations 1000 times, except  $(N,K) = (100000,100)$, for which  we only repeat 100 times  as the LP experiments take too long.

\subsubsection{Precision and Recall}

For sparse recovery, it is crucial to correctly recover the nonzero locations. Here we  borrow the concept of precision and recall from the literature of information retrieval (IR):
\begin{align}\notag
&\text{Precision = } \frac{\text{\# True Nonzeros}}{\text{\# Returned Nonzeros}} = \frac{tp}{tp+fp},\hspace{0.4in}
\text{Recall = } \frac{\text{\# True Nonzeros}}{\text{\# Total True Nonzeros}} = \frac{tp}{tp+fn}
\end{align}
to compare the proposed absolute minimum estimator with LP decoding.  Here, we view   nonzero coordinates as ``positives'' (p) and zero coordinates as ``negatives'' (n). Ideally, we hope to maximize ``true positives'' (tp) and minimize ``false positives'' (fp) and ``false negatives'' (fn).  In reality, we usually hope to achieve at least  perfect recalls so that the retrieved set of  coordinates contain all the true nonzeros.

\begin{figure}[h!]
\begin{center}
\mbox{
\includegraphics[width=2.5in]{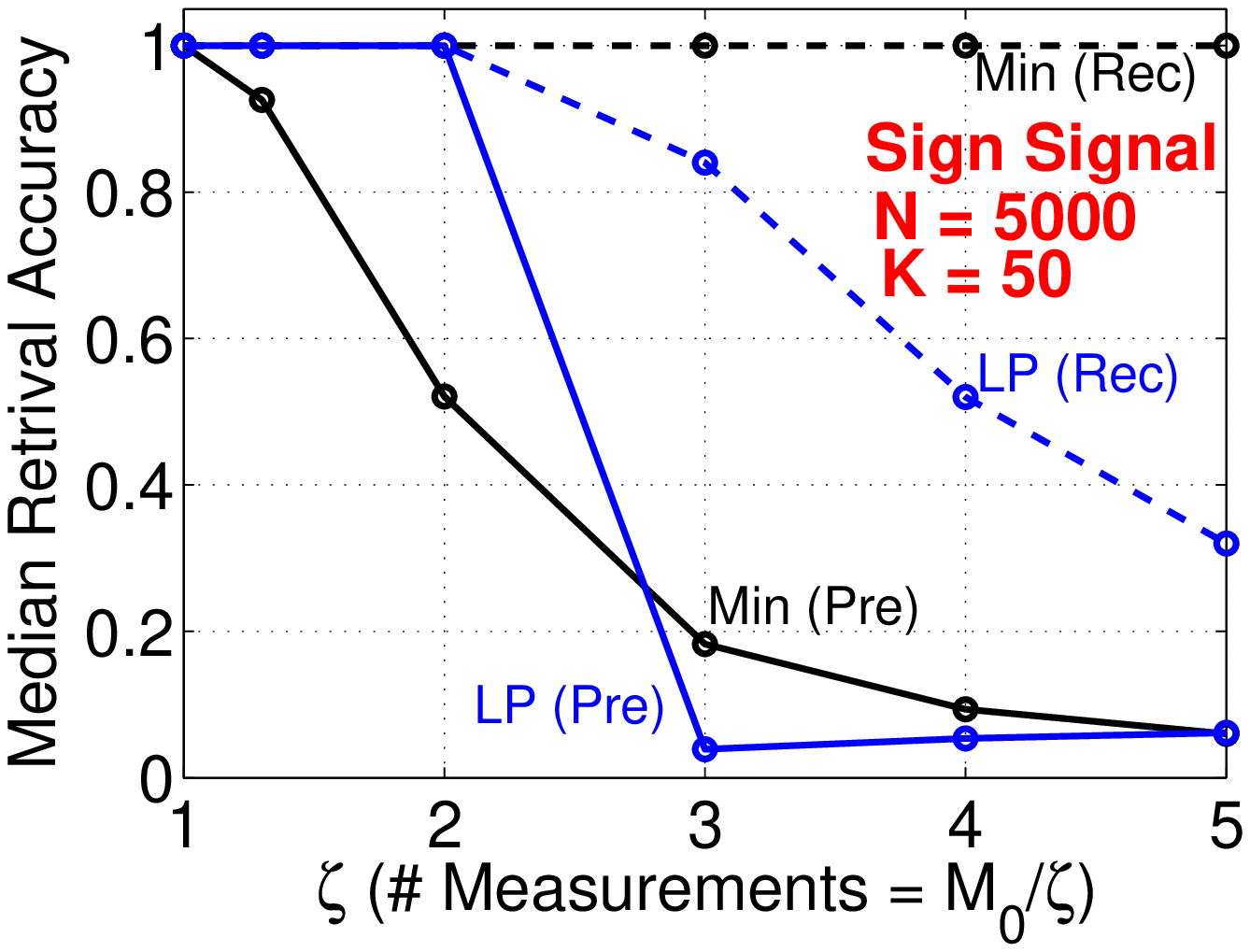}\hspace{0in}
\includegraphics[width=2.5in]{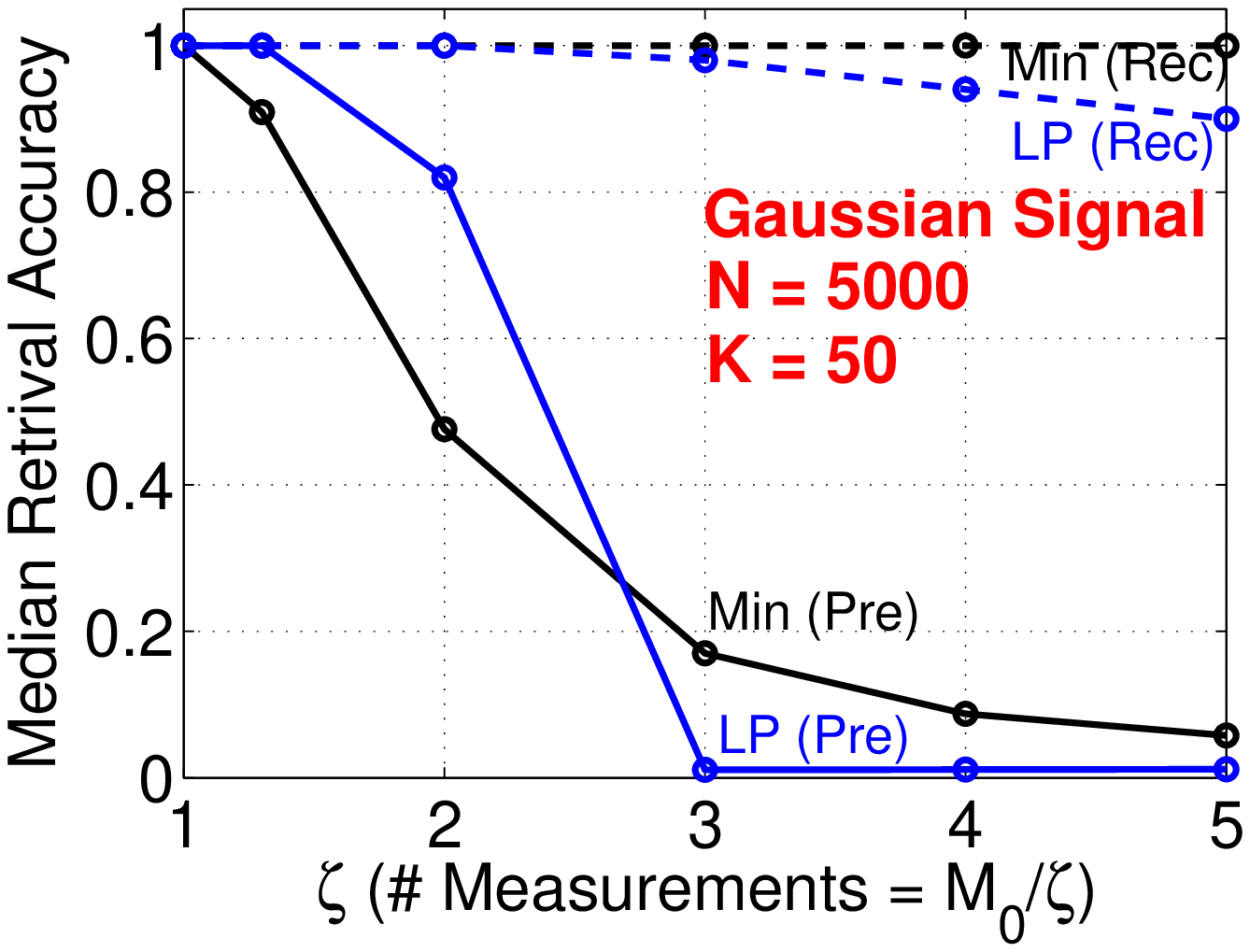}}

\mbox{
\includegraphics[width=2.5in]{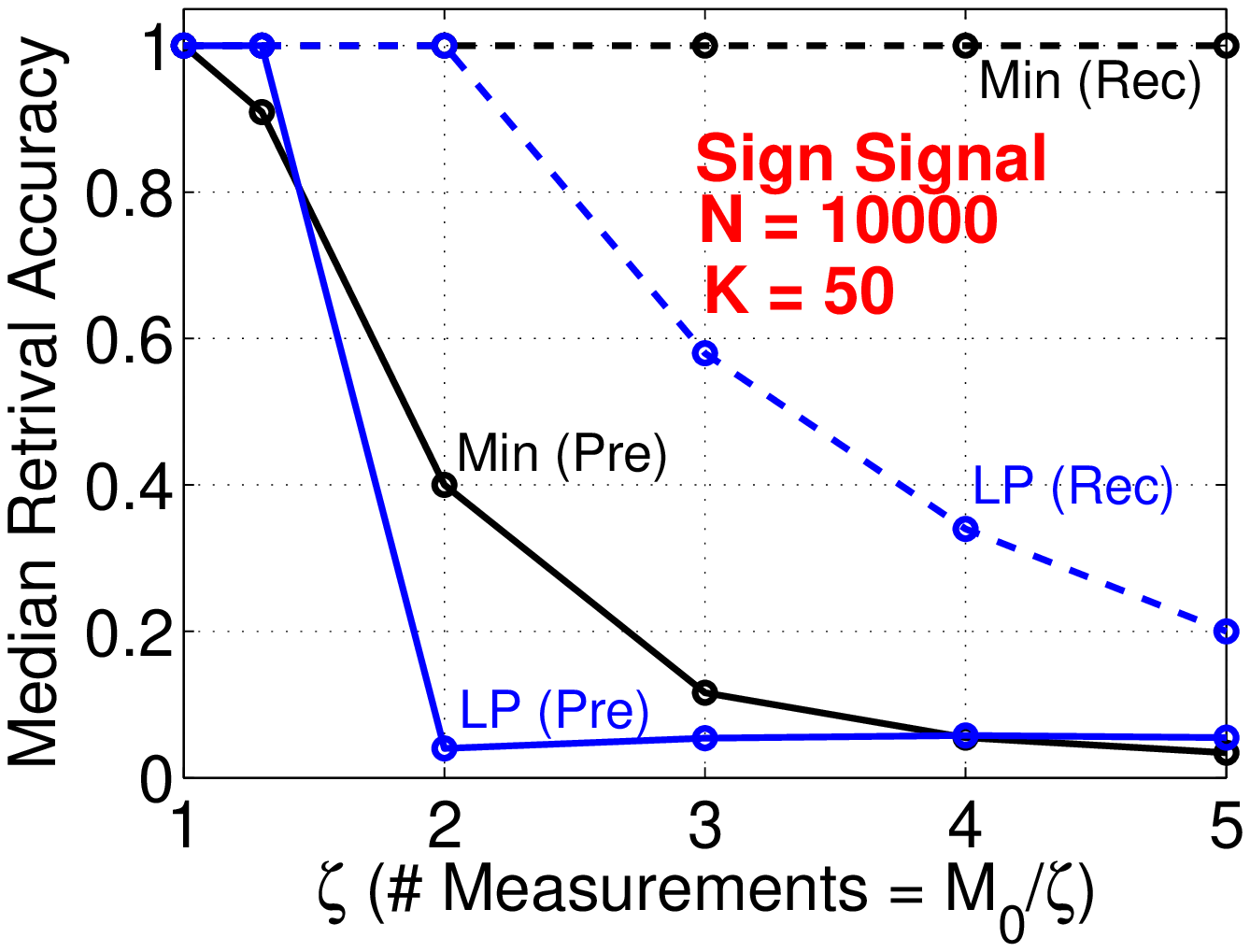}\hspace{0in}
\includegraphics[width=2.5in]{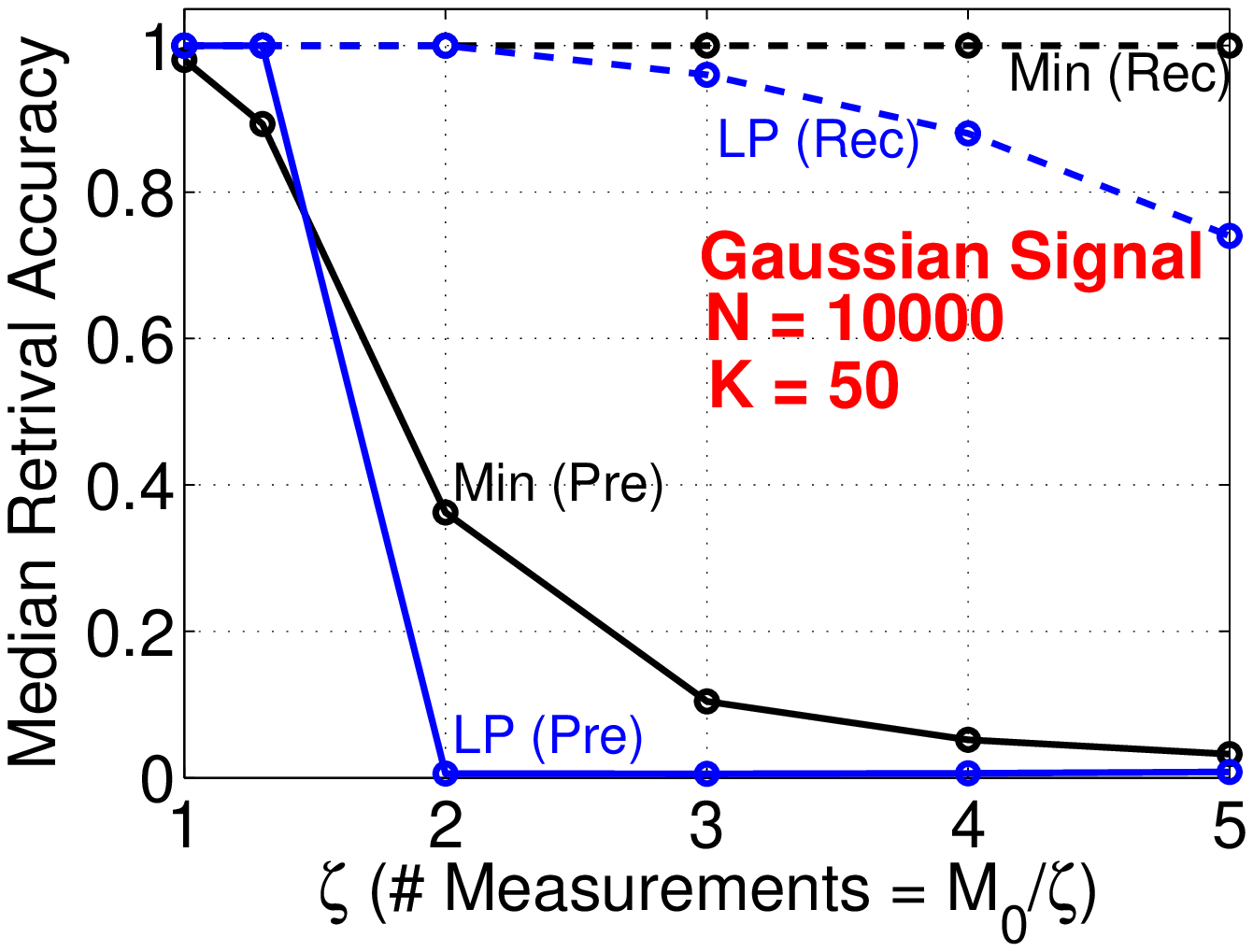}}

\mbox{
\includegraphics[width=2.5in]{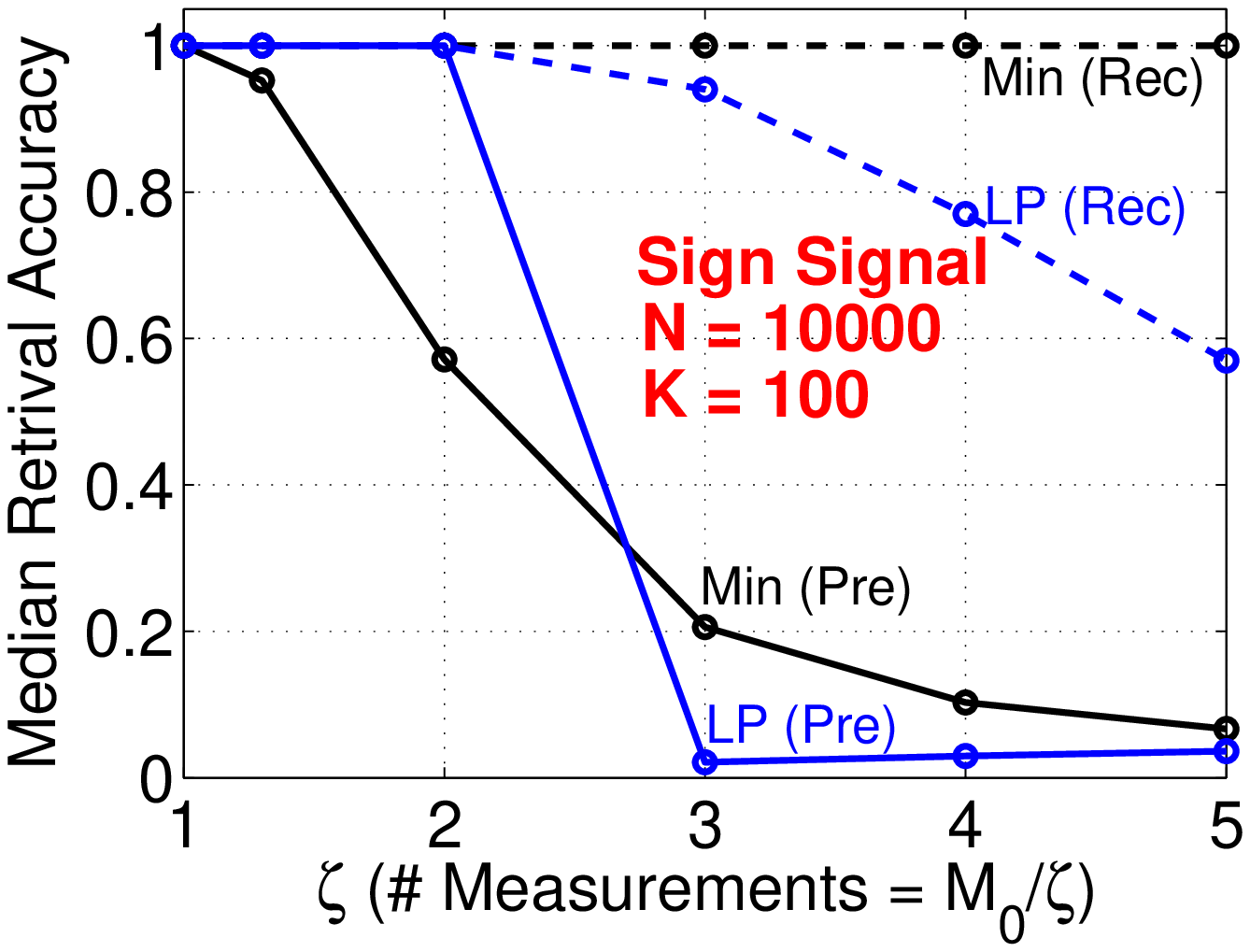}\hspace{0in}
\includegraphics[width=2.5in]{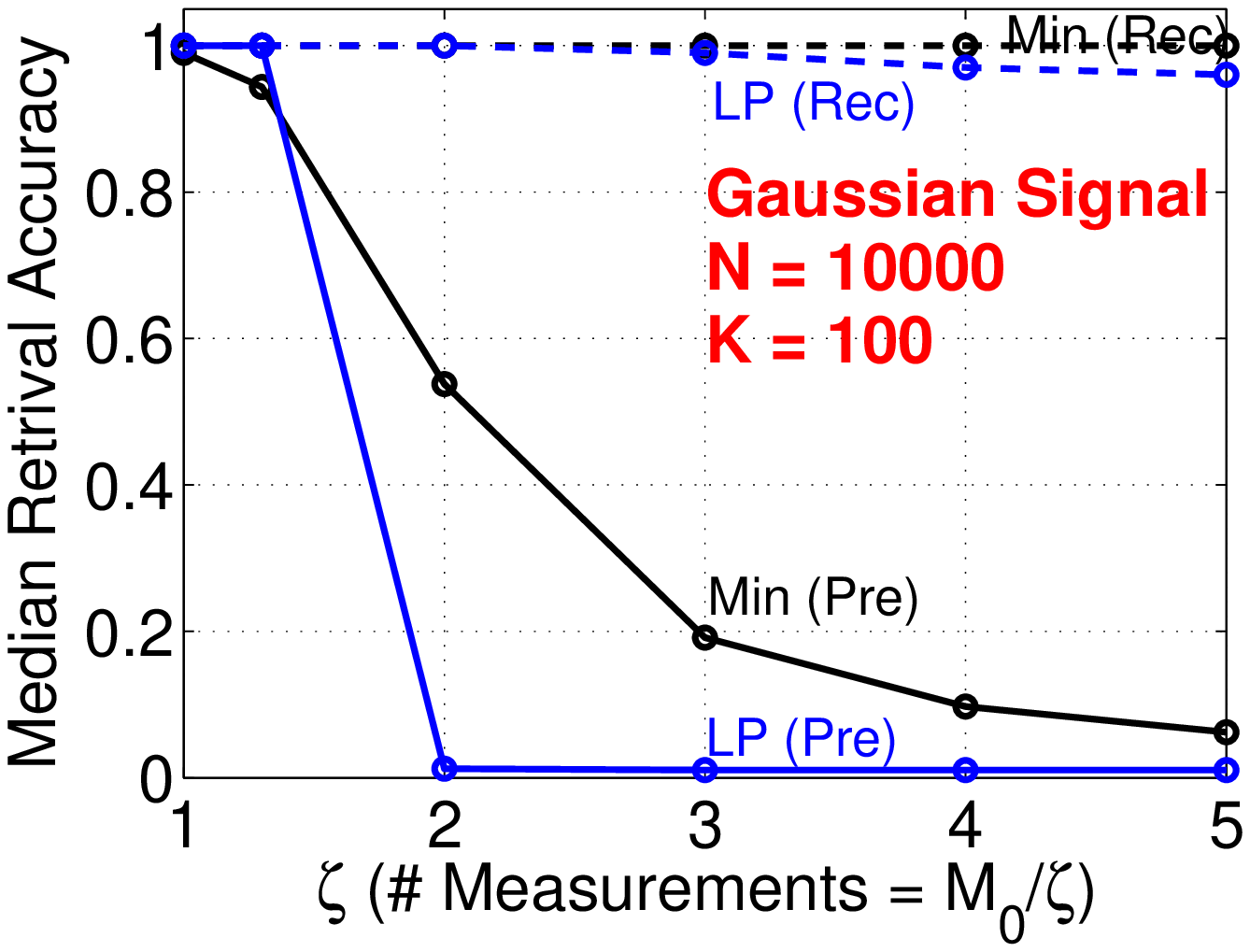}}

\mbox{
\includegraphics[width=2.5in]{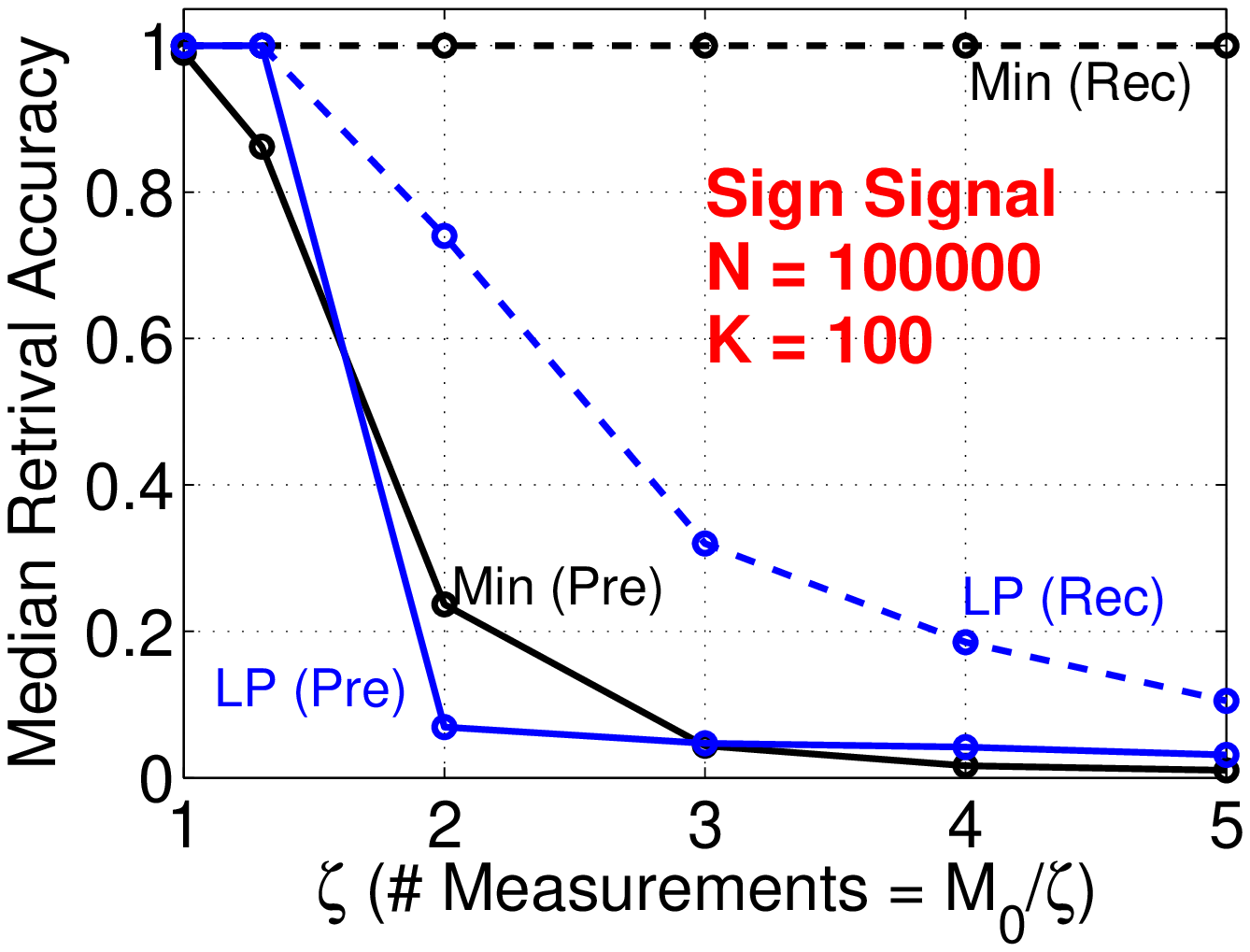}\hspace{0in}
\includegraphics[width=2.5in]{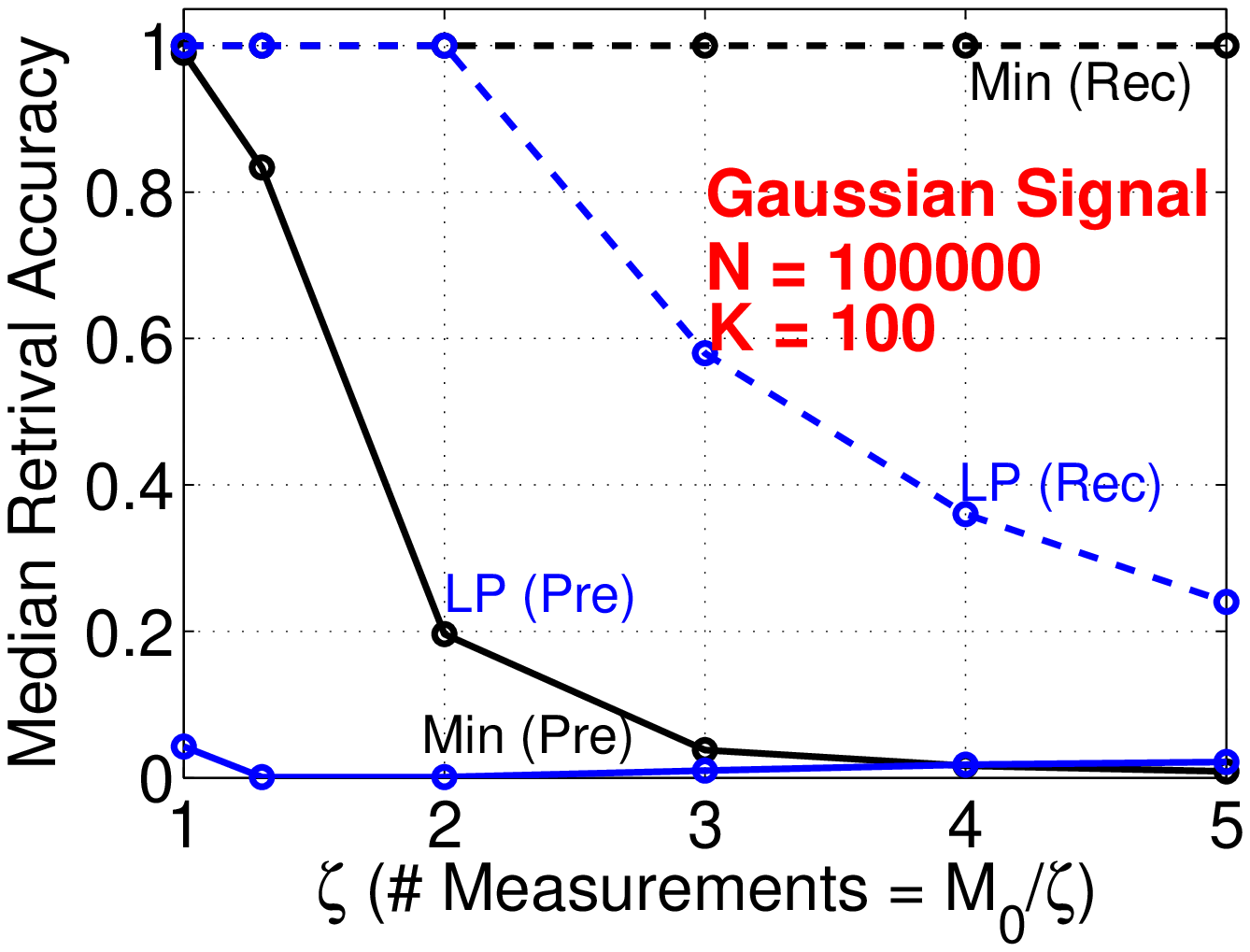}}
\end{center}
\vspace{-0.2in}
\caption{Median precision and recall curves, for comparing our proposed minimum estimator with LP decoding. The minimum estimator produces essentially $100\%$ recalls even for $M$ as small as $M_0/5$. }\label{fig_PR}
\end{figure}

Figure~\ref{fig_PR} presents the (median) precision-recall curves. Our minimum estimator always produces essentially $100\%$ recalls, meaning that the true positives are always included for the next stage of reconstruction. In comparison, as $M$ decreases, the recalls of LP decreases significantly.

\clearpage\newpage

\subsubsection{\textbf{Reconstruction Accuracy}}

The reconstruction accuracy is another useful measure of quality. We define the reconstruction error as
\begin{align}\label{eqn_error}
\text{Error} = \sqrt{\frac{\sum_{i=1}^N\left(x_i - \text{estimated } x_i\right)^2}{\sum_{i=1}^N x_i^2} }
\end{align}
Note that, since errors are normalized, a value $>1$ should indicate a very bad reconstruction outcome.

\begin{figure}[h!]
\begin{center}
\mbox{
\includegraphics[width=2.5in]{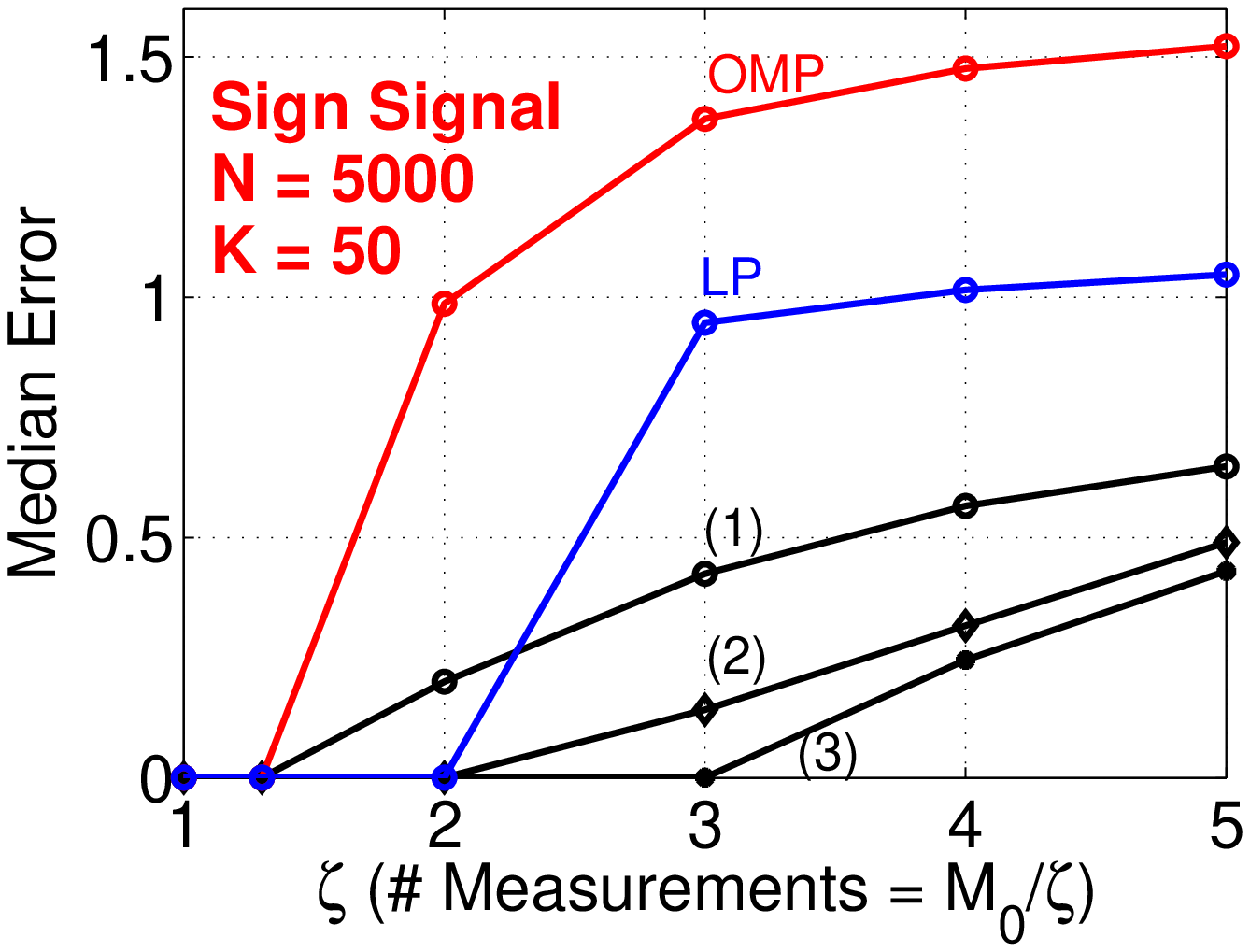}\hspace{0in}
\includegraphics[width=2.5in]{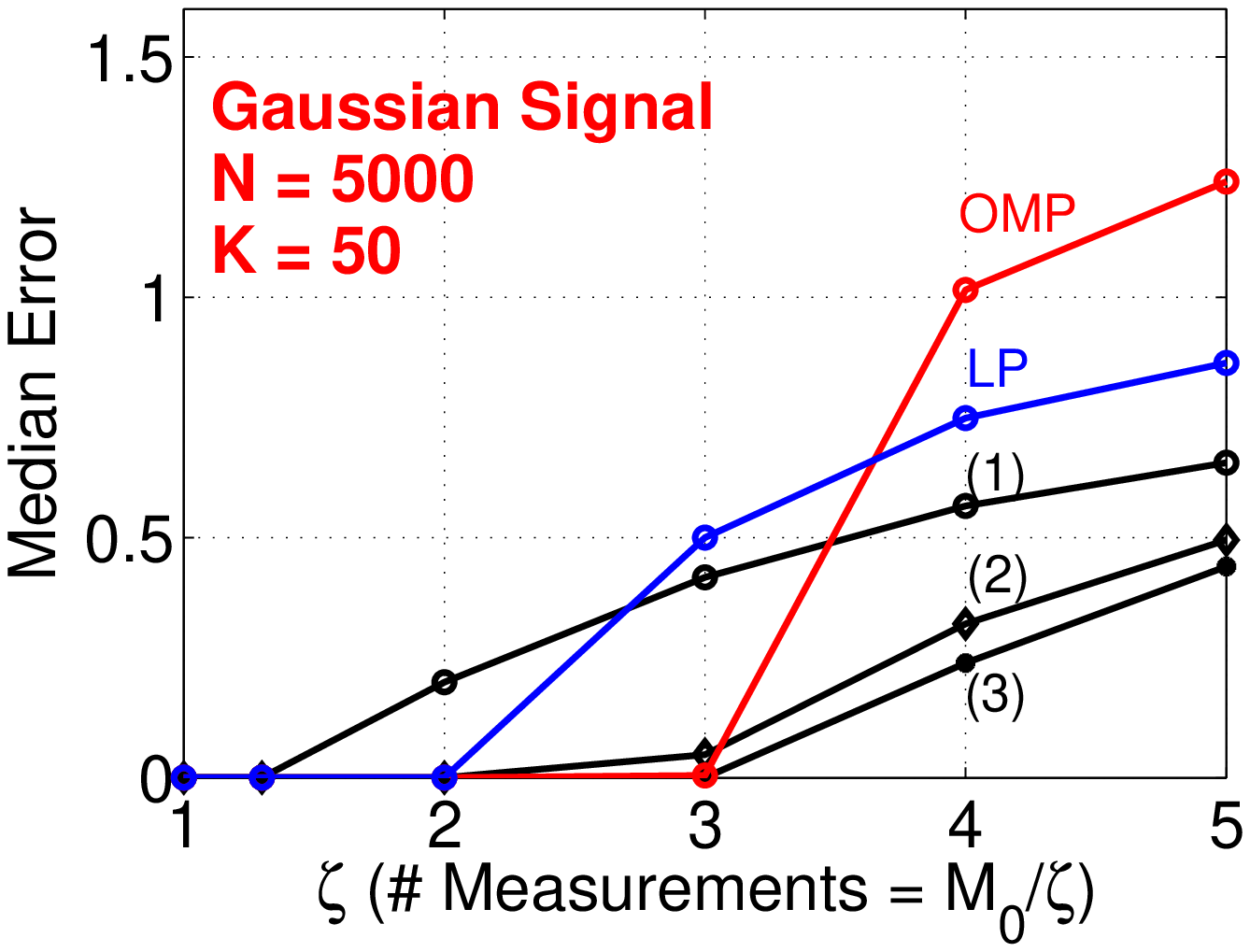}}

\mbox{
\includegraphics[width=2.5in]{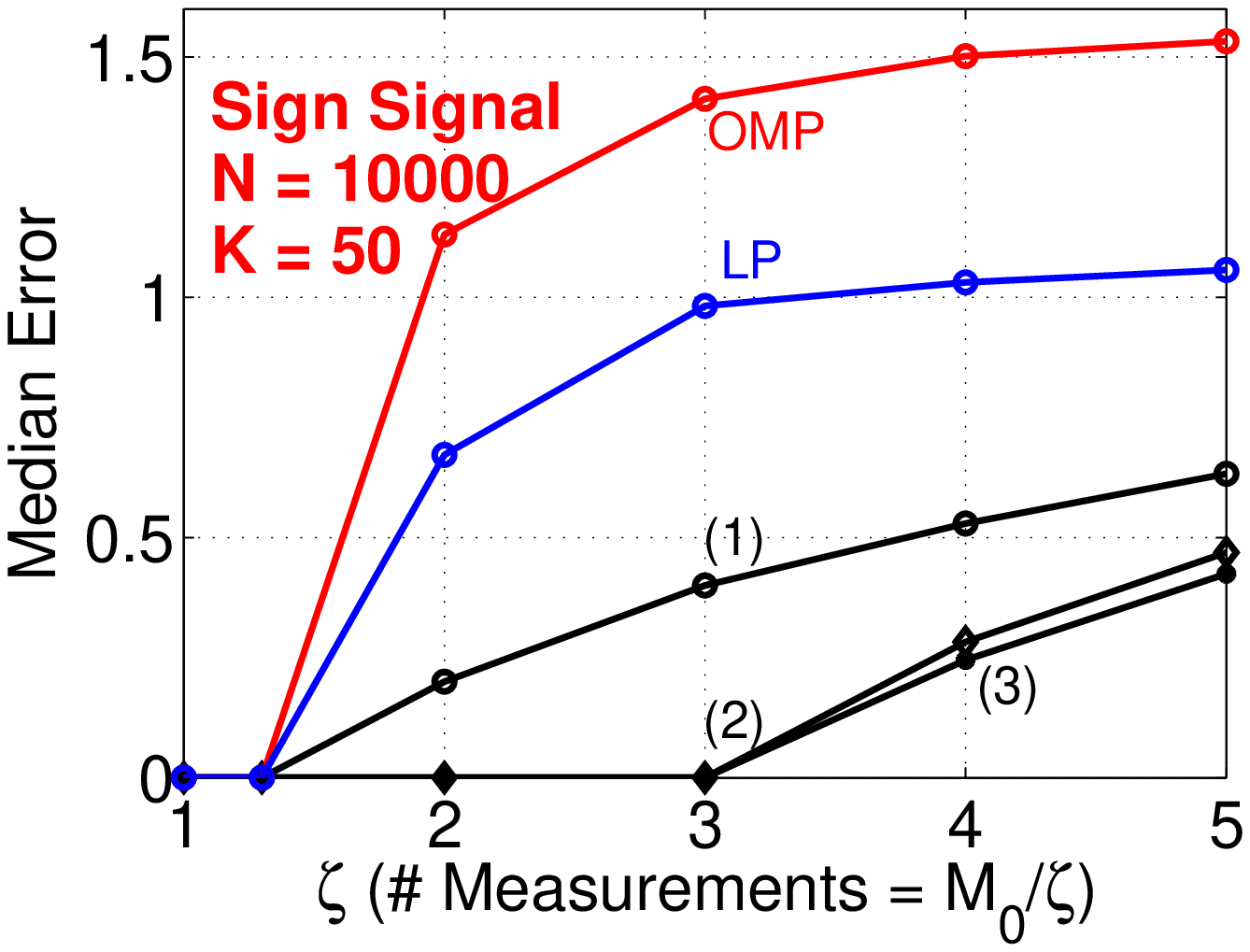}\hspace{0in}
\includegraphics[width=2.5in]{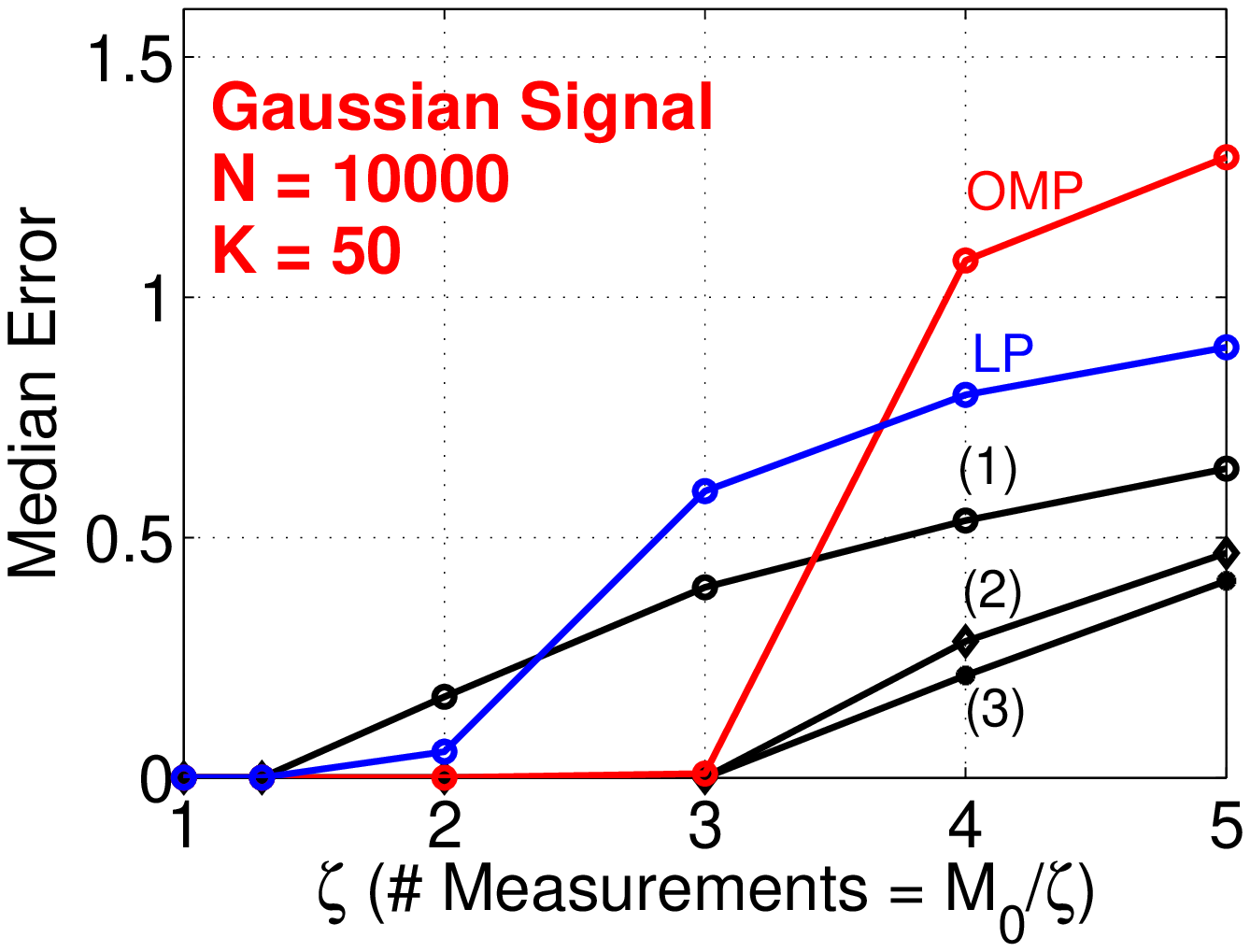}}

\mbox{
\includegraphics[width=2.5in]{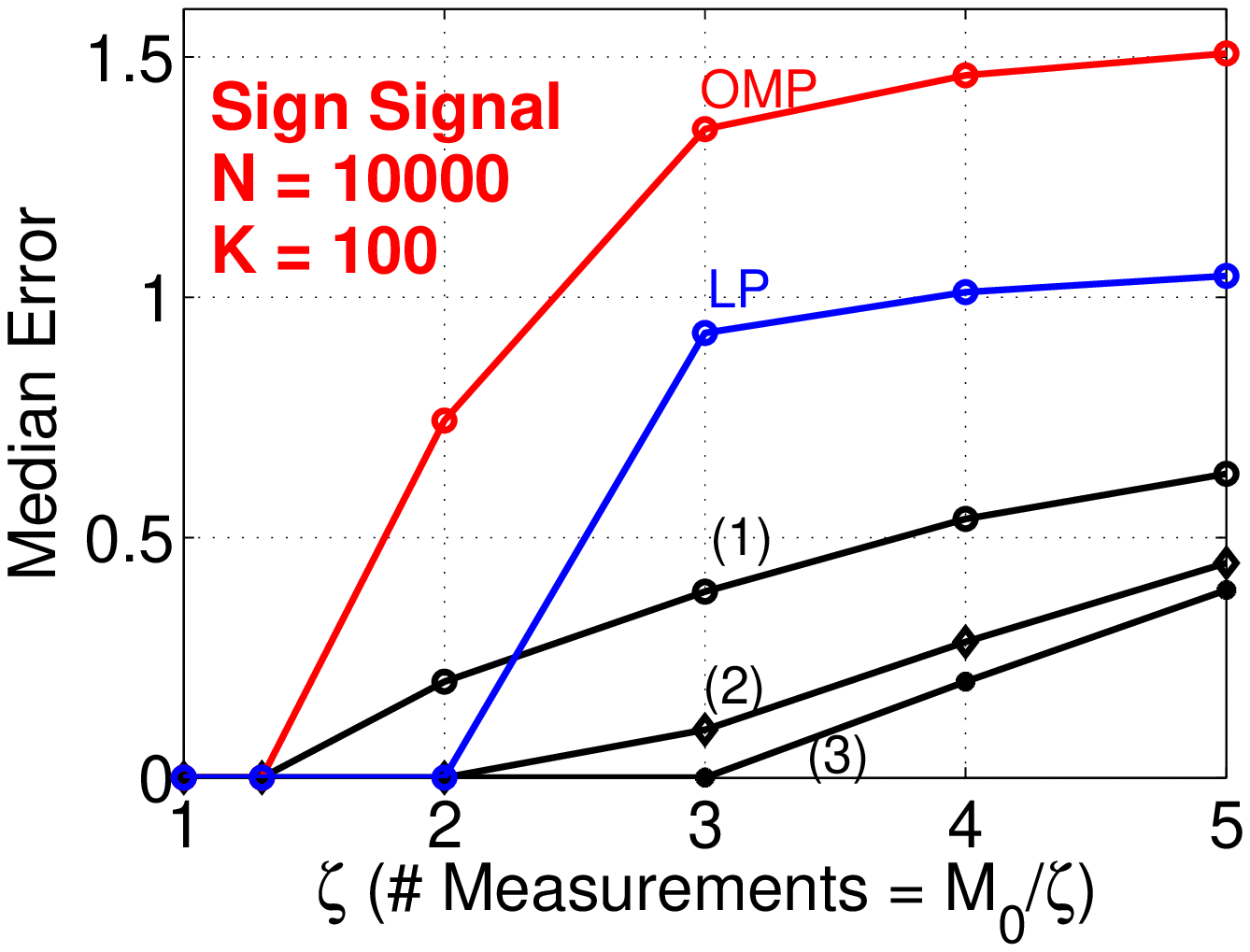}\hspace{0in}
\includegraphics[width=2.5in]{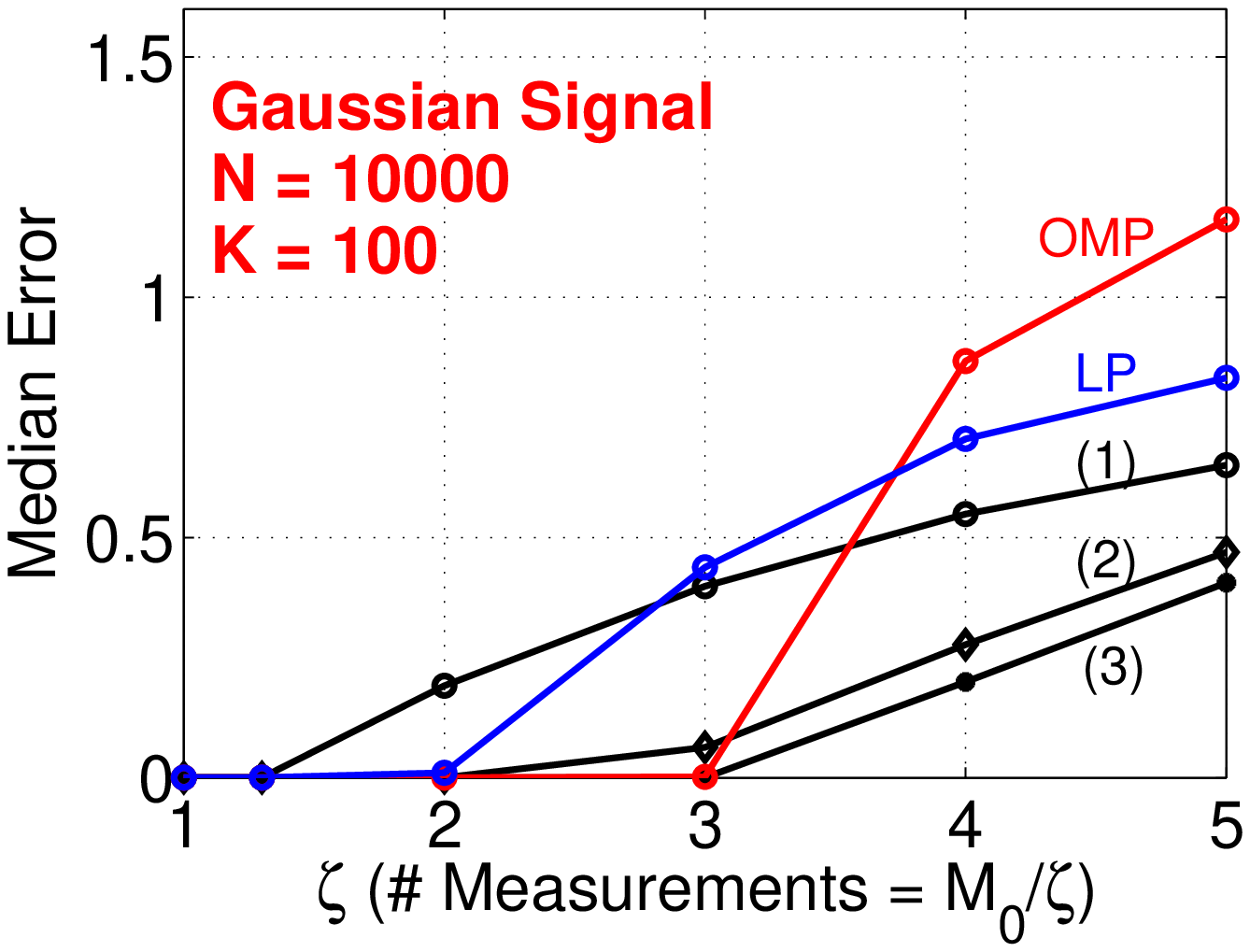}}

\mbox{
\includegraphics[width=2.5in]{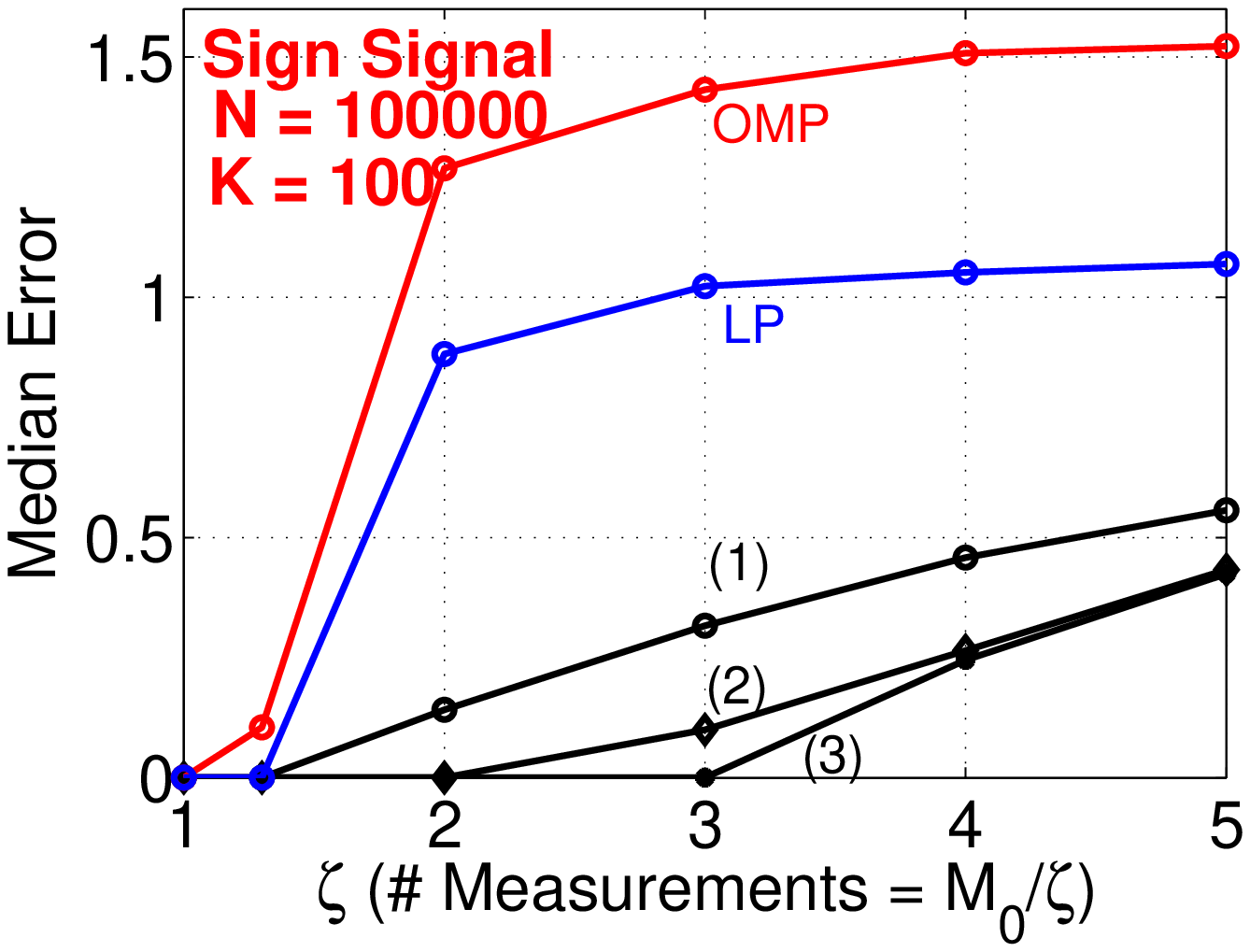}\hspace{0in}
\includegraphics[width=2.5in]{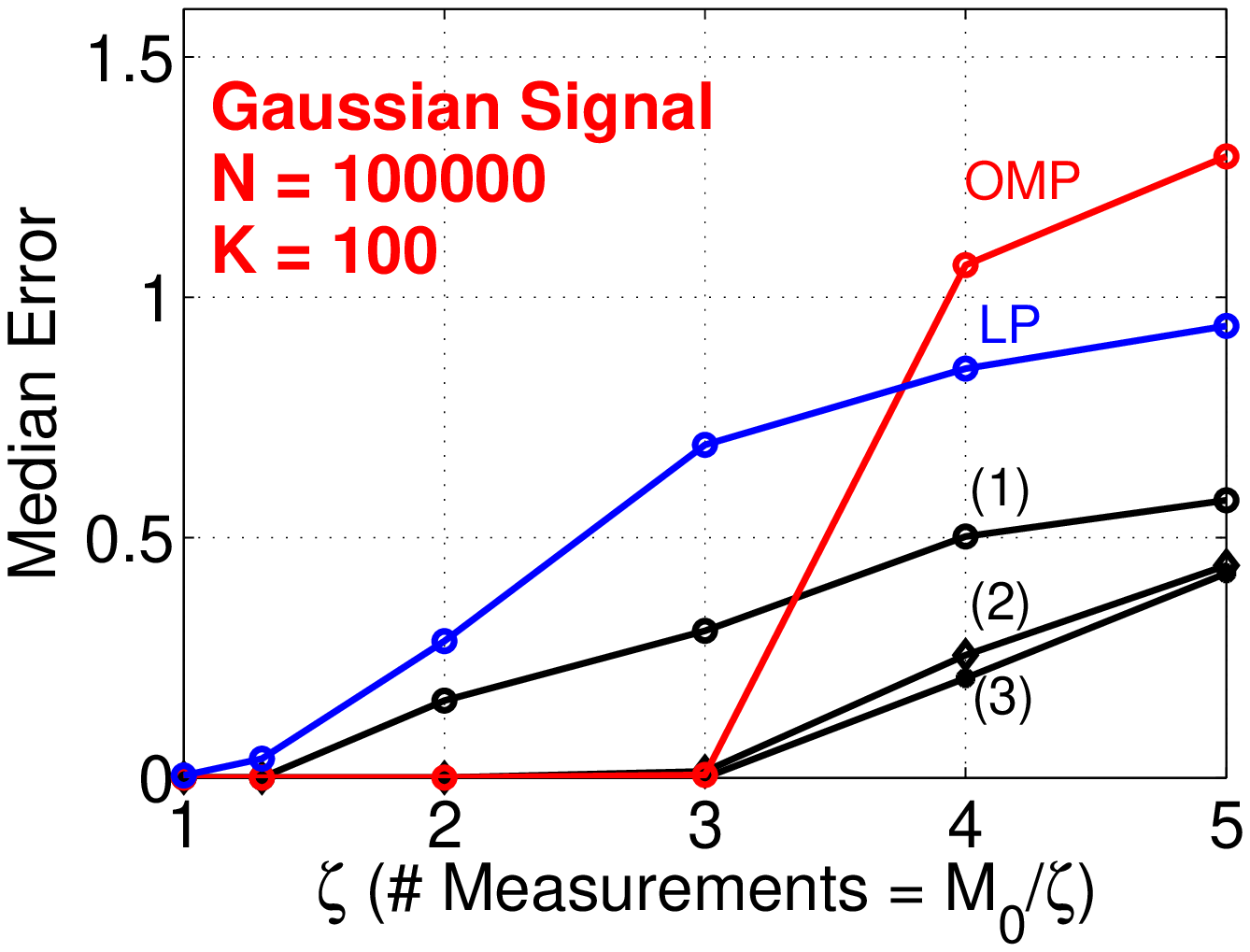}}
\end{center}
\vspace{-0.2in}
\caption{Median reconstruction errors (\ref{eqn_error}), for comparing our proposed algorithms with LP and OMP. When the number of measurements $M = M_0$, all methods perform well. As $M$ decreases, the advantage of our proposed algorithm becomes more obvious, especially with 2 or 3 iterations.  Note that, with $M=M_0/5$, even though the error of our method in terms of (\ref{eqn_error}) is quite large, the error comes from the small fraction of coordinates which our method ``gives up''. The reported nonzero coordinates by our method are still very accurate. See Figure~\ref{fig_RecSignB5} and Figure~\ref{fig_RecGausB5}.   }\label{fig_Err}\vspace{0in}
\end{figure}

Figure~\ref{fig_Err} presents the median reconstruction errors. At $M=M_0$ (i.e., $\zeta=1$), all methods perform well. For sign signals, both OMP and LP perform poorly as soon as $\zeta> 1.3$ or 2 and OMP results are particularly bad. For Gaussian signals, OMP can produce good results even when $\zeta = 3$.\\

Our method performs well, and 2 or 3 iterations of the gap estimation procedure help noticeably. One should keep in mind that errors defined by (\ref{eqn_error}) may not always be as informative. For example, with $M=M_0/5$,  Figures~\ref{fig_RecSignB5} and~\ref{fig_RecGausB5} show that, even though our method fails to recover a small fraction of nonzero coordinates, the recovered coordinates are accurate. In comparison, for OMP and LP, essentially none of the nonzero coordinates in Figures~\ref{fig_RecSignB5} and~\ref{fig_RecGausB5} could be accurately identified when $M = M_0/5$. This confirms that our method is   stable and reliable. Of course, we have already seen this behavior in the example in Figure~\ref{fig_Zip54}. In that example, even with $M\approx K$, the reconstructed signal by our method is still quite informative.

\subsubsection{\textbf{Reconstruction Time}}

\begin{figure}[h!]
\begin{center}
\mbox{
\includegraphics[width=2.5in]{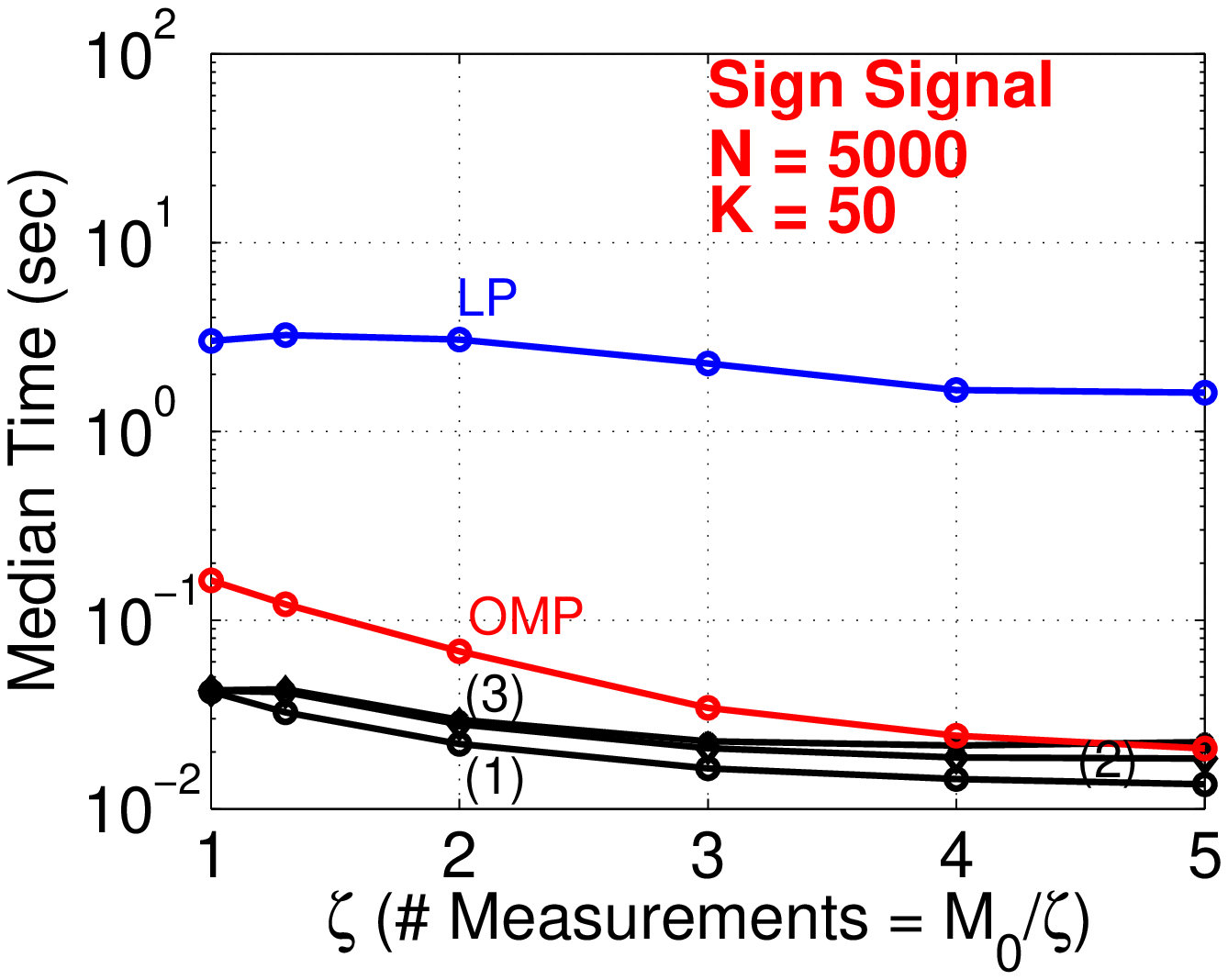}\hspace{0in}
\includegraphics[width=2.5in]{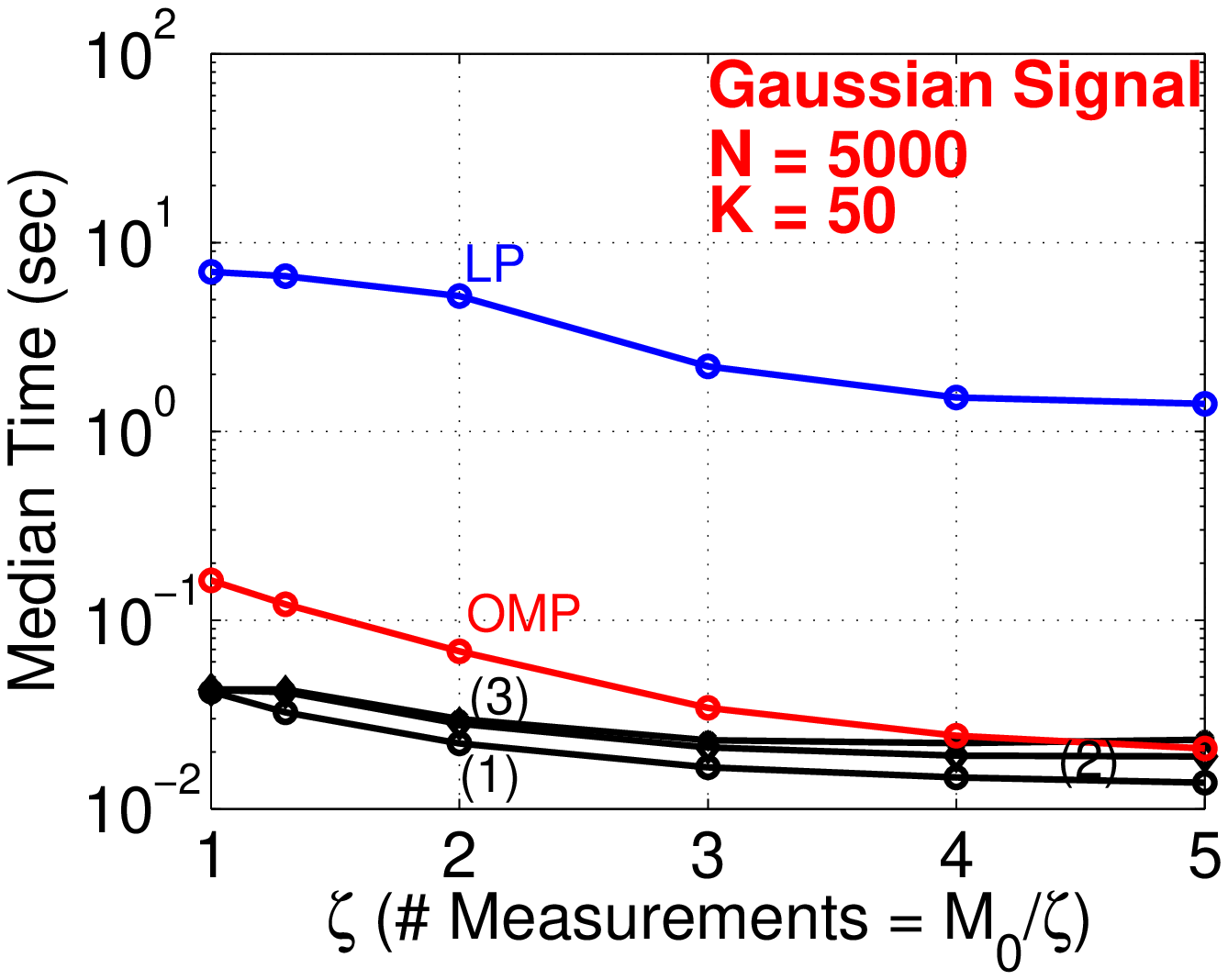}}

\mbox{
\includegraphics[width=2.5in]{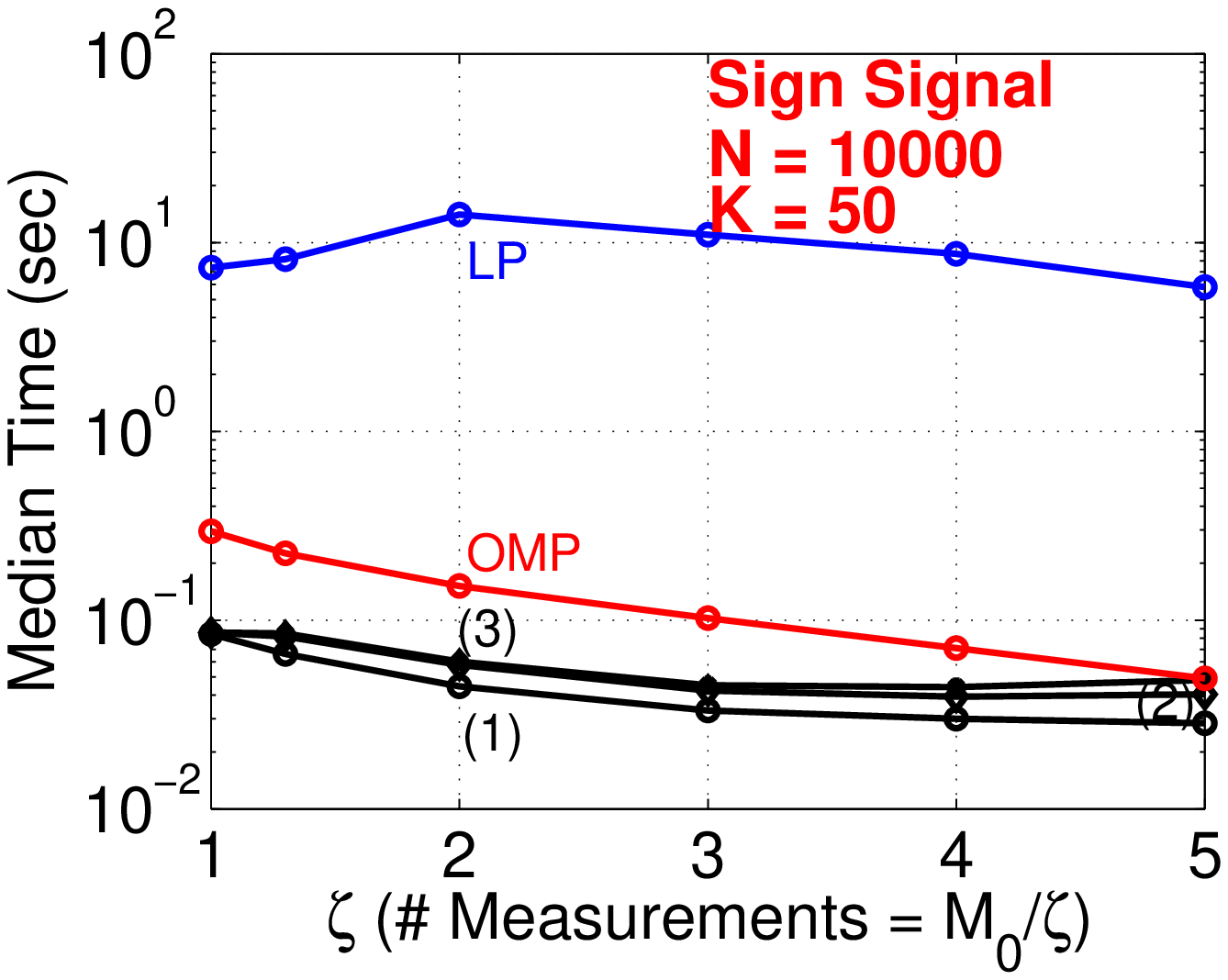}\hspace{0in}
\includegraphics[width=2.5in]{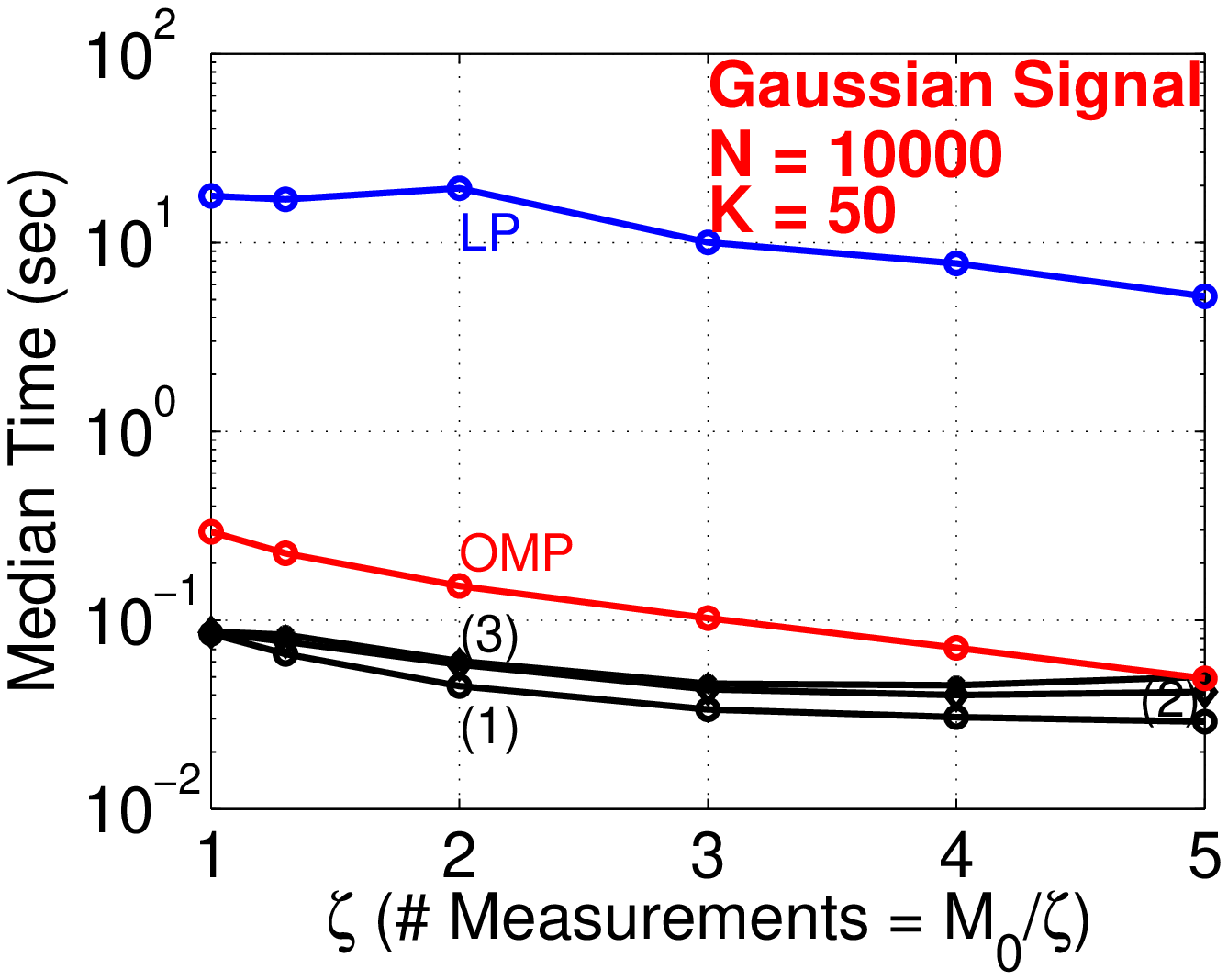}}

\mbox{
\includegraphics[width=2.5in]{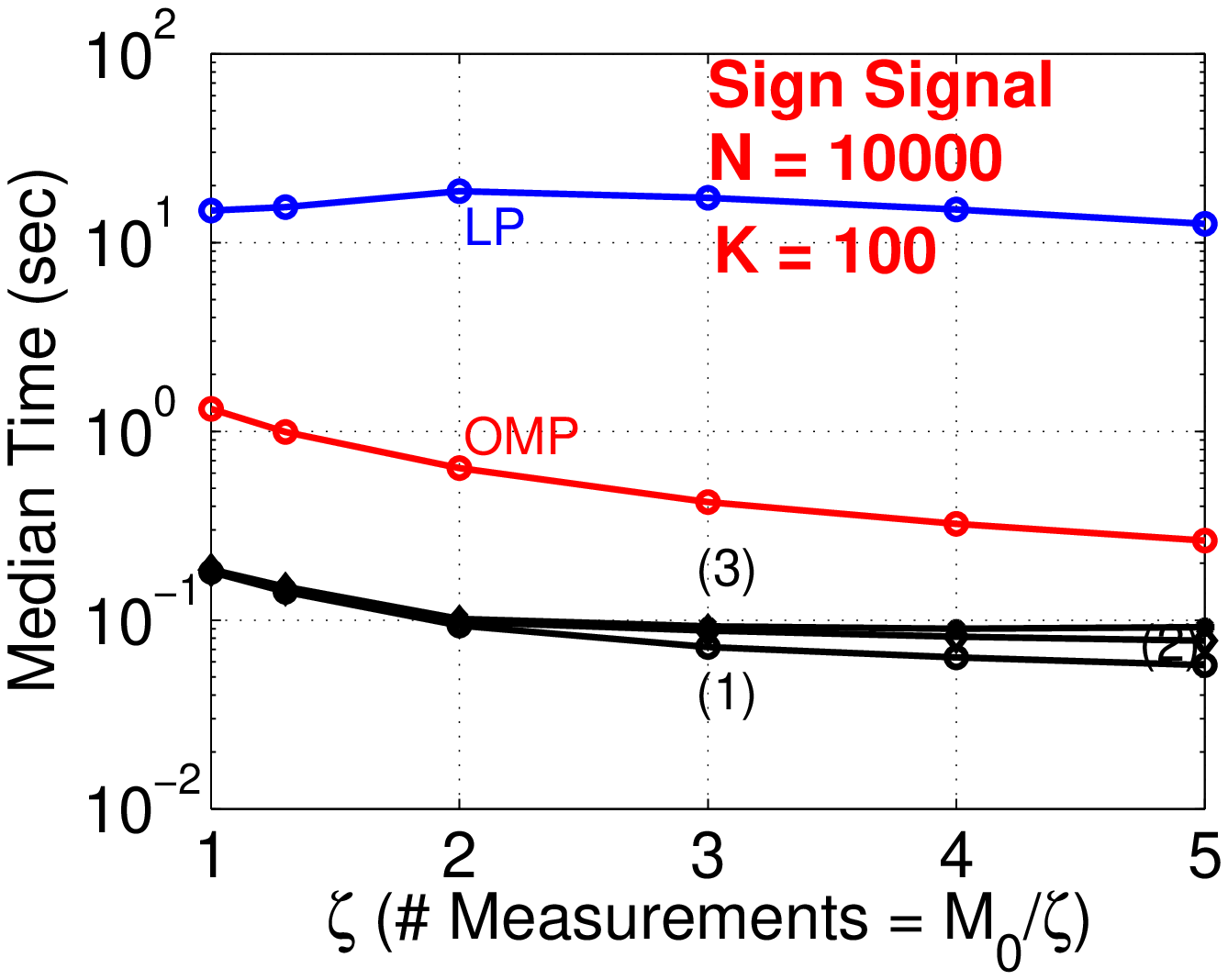}\hspace{0in}
\includegraphics[width=2.5in]{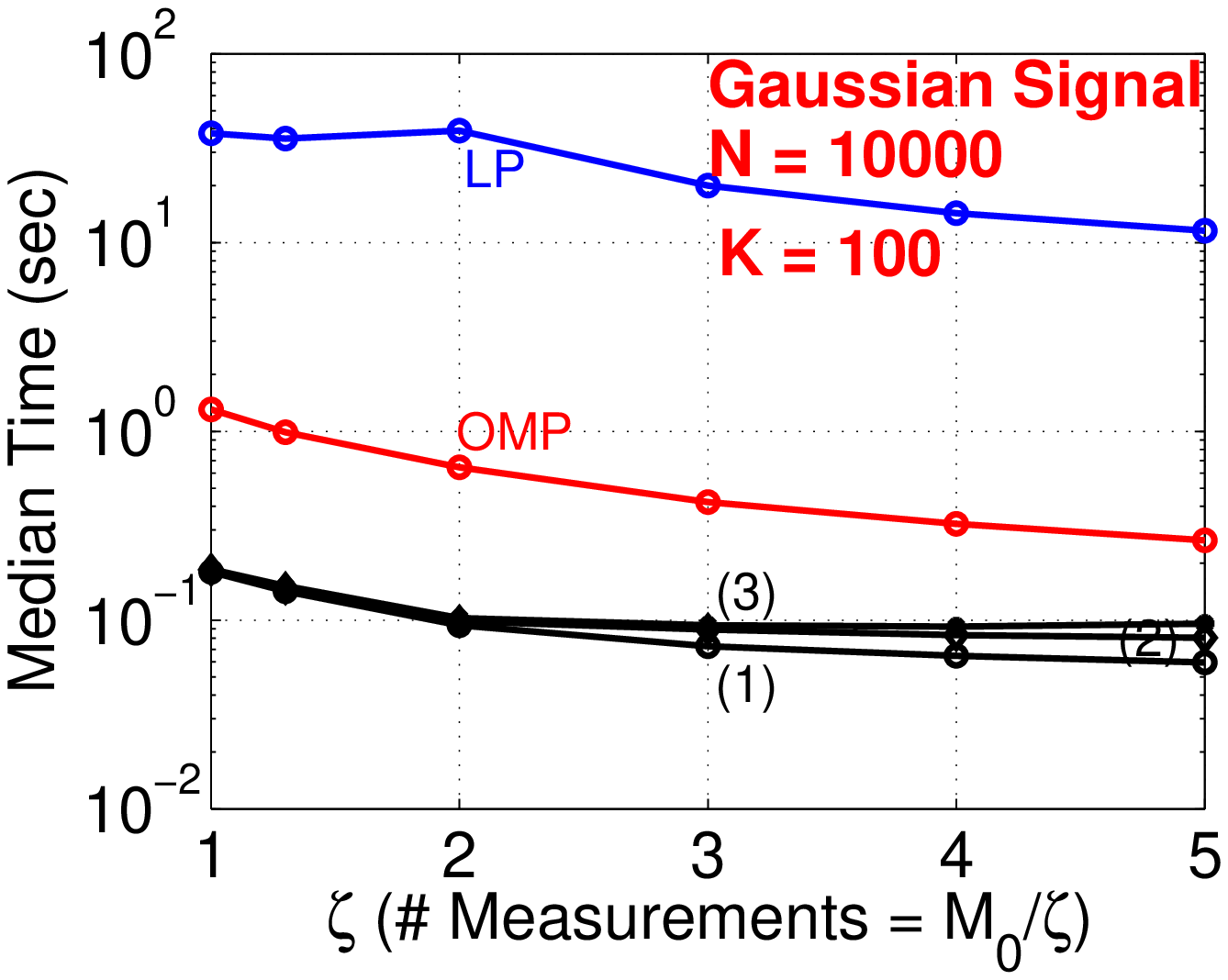}}

\mbox{
\includegraphics[width=2.5in]{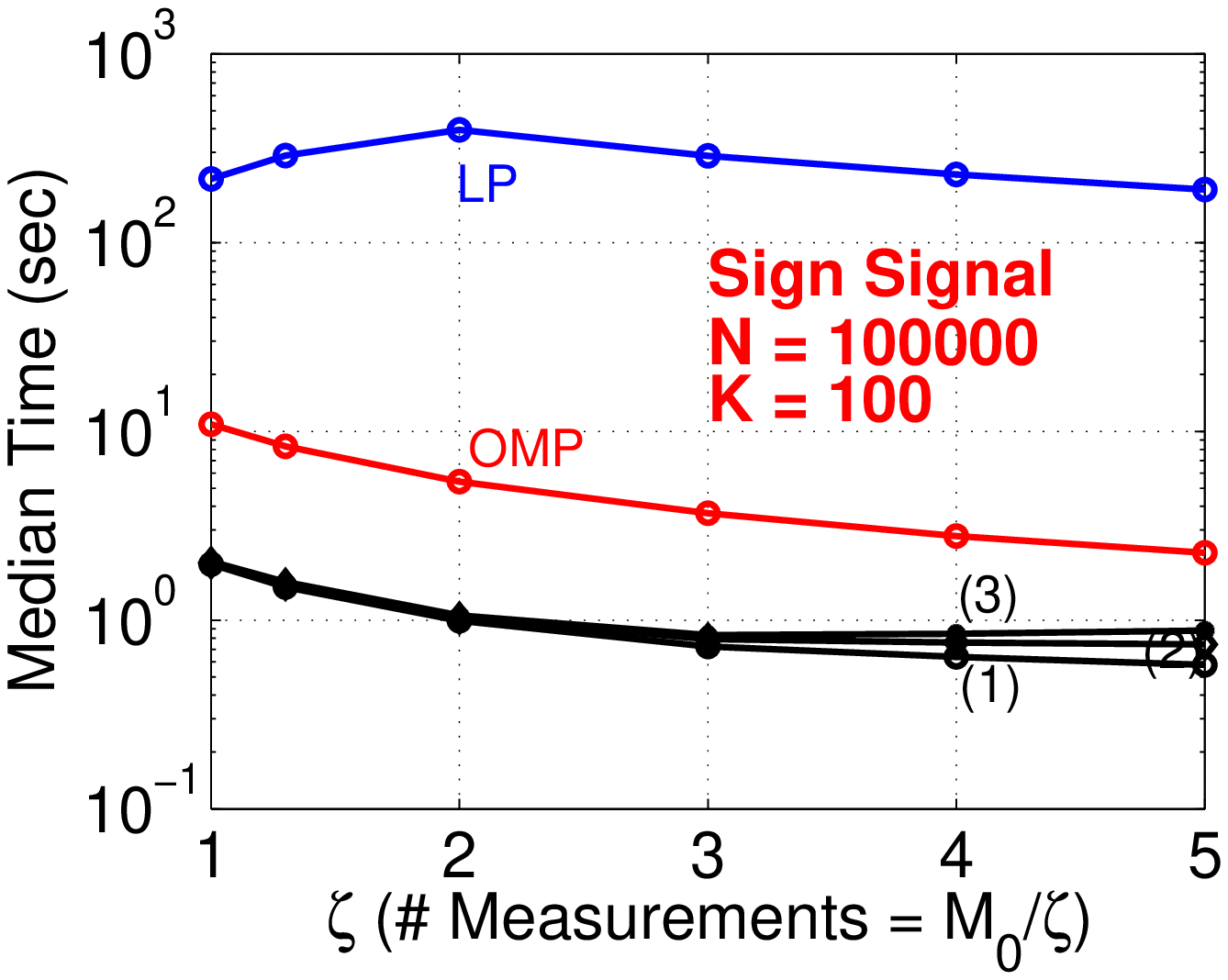}\hspace{0in}
\includegraphics[width=2.5in]{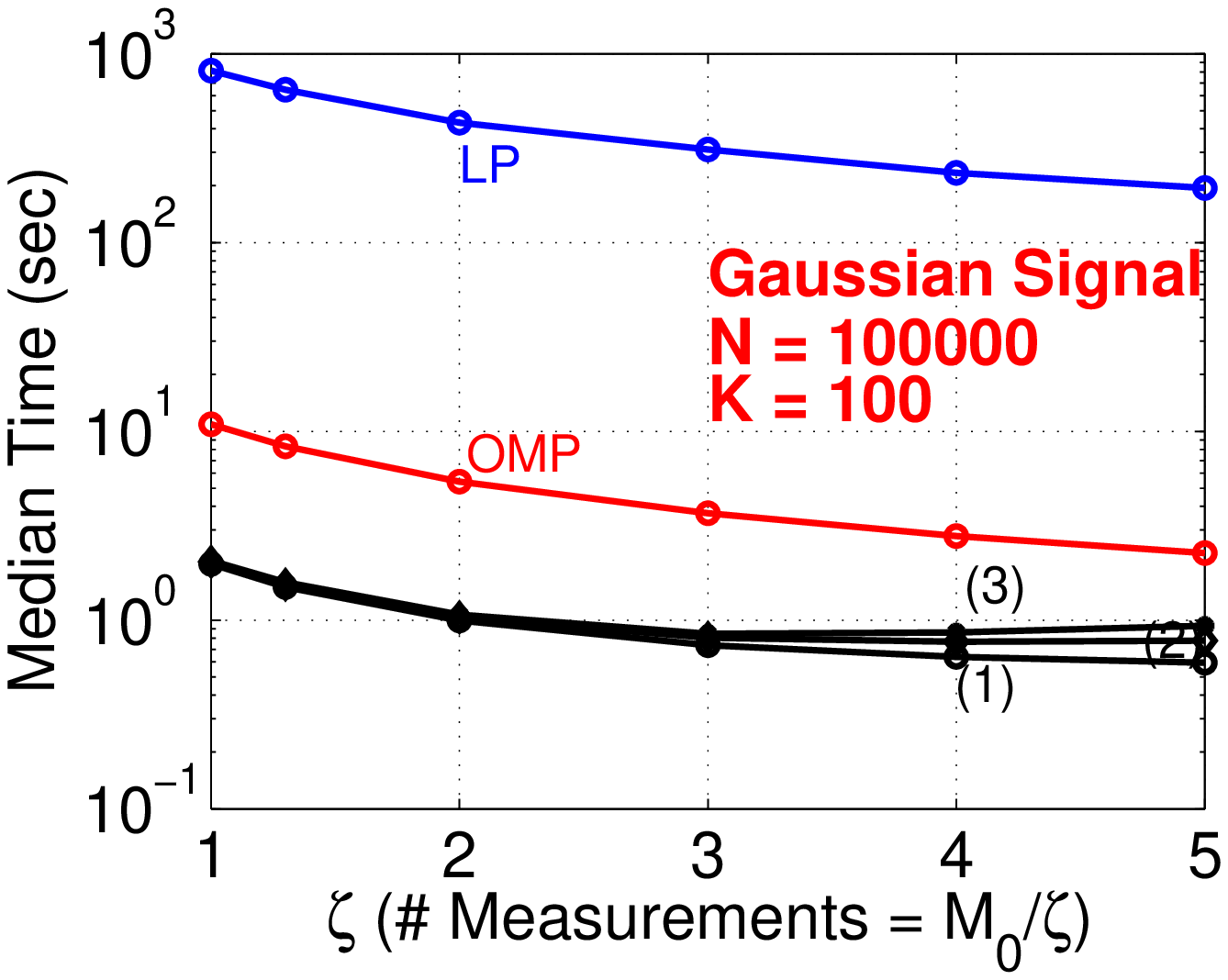}}
\end{center}
\vspace{-0.2in}
\caption{Median reconstruction times, for comparing our proposed algorithm with OMP and LP. When $K$ is not small, our method can be significantly more efficient than OMP. LP is very expensive. Note that in these experiments, $K$ is not too large, otherwise the computational advantages over LP and OMP will be even more significant. }\label{fig_Time}
\end{figure}

Figure~\ref{fig_Time} confirms that LP is computationally  expensive, using the   $l_1$-magic package~\cite{Report:L1Magic}. We have also found that the Matlab build-in L1 solver can take significantly more time (and consume more memory) than the  $l_1$-magic package. In comparison, OMP is substantially more efficient than LP, although it is still much more costly than our proposed algorithm, especially when $K$ is not small. In our experiments with the data for generating Figure~\ref{fig_Zip54}, OMP was more than   100 times  more expensive than our method.

\subsection{Measurement Noise}

Our proposed algorithm is robust against measurement noise. Recall $z_{i,j} = y_j/s_{ij} = x_i + \theta_i \frac{S_2}{S_1}$. With additive noise $n$, we have $\frac{y_j+n}{s_{ij}} =x_i + \theta_iS_2/S_1 + n/S_1$. Since the observations are only useful when $S_2/S_1$ is essentially 0, i.e., $S_1$ is large, $n/S_1$ does not really matter for our procedure. We will provide more results about measurement noise in Sec.~\ref{sec_noise}.

\clearpage\newpage
\section{Analysis of the ``Idealized'' Algorithm}\label{sec_ideal}

The analysis of our practical algorithm, i.e., the gap estimator and iterative procedure, is technically nontrivial. To better understand our method, we first provide the analysis of the ``idealized'' algorithm.

\subsection{Assumptions}

The ``idealized'' algorithm makes the following three major assumptions:
\begin{enumerate}
\item The coordinate $x_i$ is perfectly estimated (i.e., effectively zero error) if the estimate $\hat{x}_i$ satisfies $|\hat{x}_i - x_i|\leq e$, for small $e$. This assumption turns out to be  realistic in practice. For example, if  $x_i$ can only be either 0 or integers, then  $e<0.1$ (or even $e<0.5$) is good enough. If we know that the signal comes from a system with only 15 effective digits, then $e<10^{-16}$ is good enough (for example, in Matlab ``(1+1e-16)-1==0''). Here, we do not try to specify how $e$ is determined; we can just think of it as a  small number which  physically exists in every application.

\item The algorithm is able to use $\alpha\rightarrow0$ for generating stable random variables and no numerical errors occur during calculations. This convenient assumption is of course strong and not truly necessary. Basically, we just need  $\alpha$ to be small enough so that  $|\hat{x}_i - x_i|\leq e$. For the sake of simplifying the analysis, we just assume $\alpha\rightarrow0$.
\item As long as there are at least two observations $y_j/s_{ij}$ in the   interval $(x_i-e, x_i+e)$, the algorithm  is able to perfectly recover $x_i$. This step is subtly different from the gap estimator in our practical procedure. Because we do not know $x_i$ in advance, we have to rely on gaps to guess $x_i$. Although the distribution of $y_j/s_{ij}$ is extremely heavy-tailed (as shown in Figure~\ref{fig_appr_CDF}), there is always an extremely small chance that two nearly identical observations reside outside $(x_i-e, x_i+e)$. Of course, with $\alpha$ decreases, the difference between the gap estimator  and the ``idealized'' algorithm diminishes.
\end{enumerate}

\subsection{Success Probability of Each Observation}

With the above assumptions, we are able to analyze the ``idealized'' algorithm. First, we define the success probability for each observation:
\begin{align}\label{eqn_pe}
p_e = \lim_{\alpha\rightarrow0} \mathbf{Pr}\left(|y_j/s_{ij}-x_i| < e\right)
\end{align}
As shown in Sec.~\ref{sec_intuition},   $y_j/s_{ij} = x_i + \theta_iS_2/S_1$, where $S_1, S_2 \sim S(\alpha,1)$ i.i.d. and $\theta_i = \left(\sum_{t\neq i} |x_t|^\alpha\right)^{1/\alpha}$. Thus, $p_e = \lim_{\alpha\rightarrow0} \mathbf{Pr}\left(|S_2/S_1| < e/\theta_i\right)$. Recall $1/|S_1|^\alpha\rightarrow w_1 \sim exp(1)$ and $1/|S_2|^\alpha\rightarrow w_2 \sim exp(1)$, as $\alpha\rightarrow 0$. In this section, for notational convenience, we simply write $1/|S_1|^\alpha = w_1$ and $1/|S_2|^\alpha= w_2$. Therefore
\begin{align}\notag
p_e =& \lim_{\alpha\rightarrow0} \mathbf{Pr}\left(|S_2/S_1| < e/\theta_i\right) =  \lim_{\alpha\rightarrow0} \mathbf{Pr}\left(w_1/w_2 < e^\alpha/\theta_i^\alpha\right) = \lim_{\alpha\rightarrow0} \frac{1}{1+1/\left(e^\alpha/\theta_i^\alpha\right)}\\
=&\left\{\begin{array}{cc}
\frac{1}{K}, &\text{ if } \ x_i\neq 0\\
\frac{1}{K+1}, &\text{ if } \ x_i = 0
\end{array}
\right.
\end{align}
This means the number of observations falling in $(x_i-e, x_i+e)$ follows a binomial distribution $(M,p_e)$. We can bound the failure probability (i.e., the probability of having at most one success) by $\delta$, as
\begin{align}
\left(1-p_e\right)^M + M\left(1-p_e\right)^{M-1} p_e \leq \delta
\end{align}
To make sure that $x_i$ can be perfectly recovered, we need to choose $M$ large enough such that
\begin{align}
\left(1-1/K\right)^M + \left(1-1/K\right)^{M-1} M/K \leq \delta
\end{align}
Clearly, $M = 2K\log1/\delta$ is sufficient for any $\delta\in(0,1)$. But we can do  better for small $\delta$:
\begin{align}\label{eqn_5K}
&M \geq 1.60K\log1/\delta, \hspace{0.2in}\text{ if } \ \delta \leq 0.05\\
&M \geq 1.45K \log 1/\delta, \hspace{0.2in} \text{ if } \ \delta \leq 0.01
\end{align}
which is  $4.8K$ when $\delta = 0.05$, and about $6.7K$ when $\delta = 0.01$.

\subsection{The Iterative Procedure}

There are $N$ coordinates. If we stop the algorithm with only one iteration, then we have to use the union bound which will result in an additional $\log N$ multiplicative term. However, under our idealistic setting, this $\log N$ term is actually not needed if we perform iterations based on residuals. Each time, after we remove a nonzero coordinate which is perfectly recovered, the remaining problem only becomes easier, i.e., the success probability becomes $\geq \frac{1}{K}$ and $M$ remains the same.\\

Suppose, $x_1 = x_2 = ... = x_K = 1$ and $x_i = 0$ if $ K<i\leq N$. The ratio statistics are, for $j = 1$ to $M$,
\begin{align}\notag
&z_{1,j} = 1 + \frac{s_{2j}}{s_{1j}} + \frac{s_{3j}}{s_{1j}}  + ... +  \frac{s_{Kj}}{s_{1j}} \\\notag
&z_{2,j} = 1 + \frac{s_{1j}}{s_{2j}} + \frac{s_{3j}}{s_{2j}}  + ... +  \frac{s_{Kj}}{s_{2j}} \\\notag
&z_{3,j} = 1 + \frac{s_{1j}}{s_{3j}} + \frac{s_{2j}}{s_{3j}}  + ... +  \frac{s_{Kj}}{s_{3j}} \\\notag
&...\\\notag
&z_{K,j} = 1 + \frac{s_{1j}}{s_{Kj}} + \frac{s_{2j}}{s_{Kj}}  + ... +  \frac{s_{K-1,j}}{s_{Kj}} \\\notag
\end{align}

Suppose in the first iteration, $x_1$ is perfectly recovered. That is,  among $M$ observations, for at least two observations, $z_{1,j}$ is very close 1, i.e.,  $\left|\frac{s_{2j}}{s_{1j}} + \frac{s_{3j}}{s_{1j}}  + ... +  \frac{s_{Kj}}{s_{1j}}\right|<e$, for at least $j$ values.

After we recover  $x_1$, we compute the residual and move to the second iteration:
\begin{align}\notag
&z_{2,j} = 1  + \frac{s_{3j}}{s_{2j}}  + ... +  \frac{s_{Kj}}{s_{2j}} \\\notag
&z_{3,j} = 1  + \frac{s_{2j}}{s_{3j}}  + ... +  \frac{s_{Kj}}{s_{3j}} \\\notag
&...\\\notag
&z_{K,j} = 1 +  \frac{s_{2j}}{s_{Kj}}  + ... +  \frac{s_{K-1,j}}{s_{Kj}} \\\notag
\end{align}
We need to check to make sure that we have an easier problem in that the success probability is at least $1/K$. It turns out this probability is at least $1/(K-1)$, which is even better of course. To see this, we consider two cases, depending on whether in the first iteration $z_{1,j}$ is a success. \\

\noindent\textbf{Case 1: } In the first iteration, the  observation $z_{1,j}$ is a success.
\begin{align}\notag
&\mathbf{Pr}\left(\left. \left|\frac{s_{3j}}{s_{2j}}  + ... +  \frac{s_{Kj}}{s_{2j}} \right|<e \right|\left|\frac{s_{2j}}{s_{1j}} + \frac{s_{3j}}{s_{1j}}  + ... +  \frac{s_{Kj}}{s_{1j}}\right|<e\right) \\\notag
=& \mathbf{Pr}\left(\left. \left|\frac{S(K-2)^{1/\alpha}}{s_{2j}} \right|<e \right|\left|\frac{s_{2j}+S(K-2)^{1/\alpha}}{s_{1j}}\right|<e\right),\hspace{0.3in}S(K-2)^{1/\alpha} = s_{3j}+...+s_{Kj} \\\notag
=&
\mathbf{Pr}\left(\left|\frac{S(K-2)^{1/\alpha}}{s_{2j}} \right|<e, \left|\frac{s_{2j}+S(K-2)^{1/\alpha}}{s_{1j}}\right|<e\right)/(1/K)\\\notag
\geq&K\mathbf{Pr}\left(\left|\frac{S}{s_{2j}}\right|<e/(K-2)^{1/\alpha},\ \left|\frac{s_{2j}}{s_{1j}} \right|<e/(1+e)\right)\\\notag
=&K\mathbf{Pr}\left(\left|\frac{S}{s_{2j}}\right|<e/(K-2)^{1/\alpha},\ \left|\frac{s_{1j}}{s_{2j}} \right|>e/(1+e)\right)\\\notag
=&K\left(\mathbf{Pr}\left(\left|\frac{S}{s_{2j}}\right|<e/(K-2)^{1/\alpha}\right)-\mathbf{Pr}\left(\left|\frac{S}{s_{2j}}\right|<e/(K-2)^{1/\alpha},\ \left|\frac{s_{1j}}{s_{2j}} \right|<e/(1+e)\right)\right)\\\notag
=&K\left(\frac{1}{K-1} - \frac{1}{K}\right) = \frac{1}{K-1} > \frac{1}{K}
\end{align}

\noindent\textbf{Case 2: } In the first iteration, the  observation $z_{1,j}$ is not a success.
\begin{align}\notag
&\mathbf{Pr}\left(\left. \left|\frac{s_{3j}}{s_{2j}}  + ... +  \frac{s_{Kj}}{s_{2j}} \right|<e \right|\left|\frac{s_{2j}}{s_{1j}} + \frac{s_{3j}}{s_{1j}}  + ... +  \frac{s_{Kj}}{s_{1j}}\right|>1/e\right) \\\notag
=& \mathbf{Pr}\left(\left. \left|\frac{S(K-2)^{1/\alpha}}{s_{2j}} \right|<e \right|\left|\frac{s_{2j}+S(K-2)^{1/\alpha}}{s_{1j}}\right|>1/e\right)\\\notag
=&
\mathbf{Pr}\left(\left|\frac{S(K-2)^{1/\alpha}}{s_{2j}} \right|<e, \left|\frac{s_{2j}+S(K-2)^{1/\alpha}}{s_{1j}}\right|>1/e\right)/(1-1/K)\\\notag
\geq&\mathbf{Pr}\left(\left|\frac{S}{s_{2j}}\right|<e/(K-2)^{1/\alpha},\ \left|\frac{s_{2j}}{s_{1j}} \right|(1-e)>1/e\right)/(1-1/K)\\\notag
=&\mathbf{Pr}\left(\left|\frac{S}{s_{2j}}\right|<e/(K-2)^{1/\alpha},\ \left|\frac{s_{1j}}{s_{2j}} \right|<(e-e^2)\right)/(1-1/K)\\\notag
=&\left(\frac{1}{1+(K-2)+1}\right)/(1-1/K) = \frac{1}{K-1} > \frac{1}{K}
\end{align}

In both cases, the success probability in the second iteration increases (i.e., larger than $1/K$).  Note that, assuming $x_1 = x_2 = ... = x_K=1$ is  for notational convenience. As long as $\alpha\rightarrow0$, the same result holds.

\subsection{A Simplified Analysis for the Total Required Measurements}

The above analysis provides the good intuition but it is not the complete analysis for the iterative procedure.  Note that the total number of iterations $m$ satisfies $K\delta^m \leq 1$, i.e., $m \geq \log K/\log 1/\delta$. A precise analysis of the iterative procedure involves calculating the conditional probability. For example, after the first iteration, $n$ (out of $N$) nonzero coordinates are recovered. A rigorous analysis of the success probability for the second iteration will require conditioning on all these $n$ recovered coordinates and the remaining $N-n$ unrecovered coordinates. Such an analysis may add  complication for understanding our method.

Here, for simplicity, we assume that in each iteration, we use ``fresh'' projections so that the iterations become independent. This way, the total number of required measurements becomes (assuming $\delta\leq 0.05$):
\begin{align}
1.6K \log 1/\delta + \delta 1.6K \log 1/\delta +  \delta^2 1.6K \log 1/\delta + ... < \frac{1}{1-\delta}1.6K\log1/\delta
\end{align}

The additional multiplicative factor $\frac{1}{1-\alpha}$ has little impact on the overall complexity. When $\delta = 0.05$, the required number of measurements becomes $5.0K$ instead of $4.8K$.  The result is still highly encouraging.

\section{Analysis of the Practical Algorithm}\label{sec_theory}

This section will develop a more formal theoretical analysis of our procedure in Alg.~\ref{alg_recovery}, including the  minimum estimator and the gap estimator. The gap estimator is a surrogate for  the ``idealized'' algorithm.

The minimum estimator is  not  crucial once we have the gap estimator and the iterative process. We keep it in our proposed procedure for two reasons. Firstly, it is faster than the gap estimator and is able to identify a majority of the zero coordinates in the first iteration. Secondly, even if we just use one iteration, the required sample size for the minimum estimator $M$ is essentially $K\log N/\delta$, which already matches the known complexity bounds in the compressed sensing literature.

\subsection{Probability Bounds}

Our analysis uses the distribution of the ratio of two independent stable random  variables, $S_1, S_2\sim S(\alpha,1)$. As a closed-form expression is not available,  we compute the lower and upper bounds. First,  we define
\begin{align}
F_\alpha(t) = \mathbf{Pr}\left(\left|{S_2}/{S_1}\right|^{\alpha/(1-\alpha)}\leq t\right), \ \ t\geq 0
\end{align}
where
\begin{align}\notag
&\left|{S_2}/{S_1}\right|^{\alpha/(1-\alpha)} = Q_\alpha\frac{w_1}{w_2},\hspace{0.05in} Q_\alpha = Q_\alpha(u_1,u_2) = \left|\frac{q_\alpha(u_2)}{q_\alpha(u_1)}\right|^{\alpha/(1-\alpha)},\hspace{0.05in}
q_\alpha(u)  = \frac{\sin(\alpha u)}{\cos^{1/\alpha} u}\left[\cos(u-\alpha u)\right]^{(1-\alpha)/\alpha}
\end{align}
based on the CMS procedure (\ref{eqn_stable_sample}) for generating $\alpha$-stable random variables.  The following lemmas provide several useful bounds for $F_\alpha(t)$.
\begin{lemma}\label{lem_F_lower}
For all $t\geq 0$,
\begin{align}\label{eqn_F_lower}
&F_\alpha(t) = E\left(\frac{1}{1+Q_\alpha/t}\right)
 \geq \max\left\{\frac{1/2}{1+1/t},\ \frac{1+(1/t-3)\mathbf{Pr}(Q_\alpha\leq t)/2}{1+1/t}
\right\}
\end{align}
In particular, when $t\leq 1/3$, we have
\begin{align}
F_\alpha(t) \geq \frac{1}{1+1/t}, \hspace{0.2in} t\leq1/3\hspace{0.5in}
\end{align}
In addition, for any fixed $t\geq0$,
\begin{align}
\lim_{\alpha\rightarrow 0} F_\alpha(t) = \frac{1}{1+1/t}
\end{align}

\noindent\textbf{Proof:}\ See Appendix~\ref{app_F_lower}.\hspace{0.1in} $\hfill\square$\\
\end{lemma}

\begin{lemma}\label{lem_F_upper}
If $0<\alpha<1/3$, then
\begin{align}\label{eqn_F_upper}
F_\alpha(t) \leq&C_\alpha  t^{\frac{1-\alpha}{1+\alpha}}\max\{1, t^{\frac{2\alpha}{1+\alpha}}\}
\end{align}\vspace{-0.3in}
\begin{align}\label{eqn_ConstC}
\text{where}\hspace{0.5in}&C_\alpha = \mu_1\mu_2 + \frac{1}{\pi}\left(\mu_2(1-\alpha)\right)^{\frac{1-\alpha}{1+\alpha}}
 \left(\frac{1-\alpha}{\alpha}\right)^{\frac{2\alpha}{1+\alpha}} \left(\frac{1+\alpha}{1-\alpha}\right)\\
&\mu_1 =\frac{1}{\pi} \frac{\Gamma\left(1/(2-2\alpha)\right)\Gamma\left((1-3\alpha)/(2-2\alpha)\right)}{\Gamma\left((2-3\alpha)/(2-2\alpha)\right)}\\
&\mu_2 = 1/\cos\left(\pi\alpha/(2-2\alpha)\right)
\end{align}
$C_\alpha\rightarrow 1+1/\pi$ as $\alpha\rightarrow 0$,  $C_\alpha<1.5$ if $\alpha\leq 0.05$, and $C_\alpha<2$ if $\alpha\leq0.16$.\\

\noindent\textbf{Proof:}\ See Appendix~\ref{app_F_upper}. Figure~\ref{fig_ConstC} plots $C_\alpha$ for $\alpha\in[0,0.3]$.\hspace{0.1in}$\hfill\square$
\end{lemma}

\begin{figure}[h!]
\begin{center}
\includegraphics[width=2.5in]{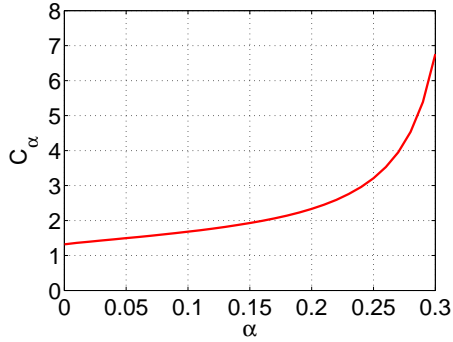}
\end{center}
\vspace{-0.25in}
\caption{The constant $C_\alpha$ as defined in (\ref{eqn_ConstC}).}\label{fig_ConstC}\vspace{-0.1in}
\end{figure}
%

\begin{lemma}
For all $0<s<t$,
\begin{align}
F_\alpha(t)-F_\alpha(s) \leq (1-s/t)F_\alpha(t)\leq (t/s-1)F_\alpha(s)
\end{align}
\textbf{Proof:}\ For all $0<s<t$, we have $F_\alpha(t)/t =E\left(\frac{1}{t+Q_\alpha}\right)\leq F_\alpha(s)/s =E\left(\frac{1}{s+Q_\alpha}\right)$.\hspace{0.1in}  $\hfill\square$
\end{lemma}

\begin{figure}[h!]
\begin{center}
\includegraphics[width=2.5in]{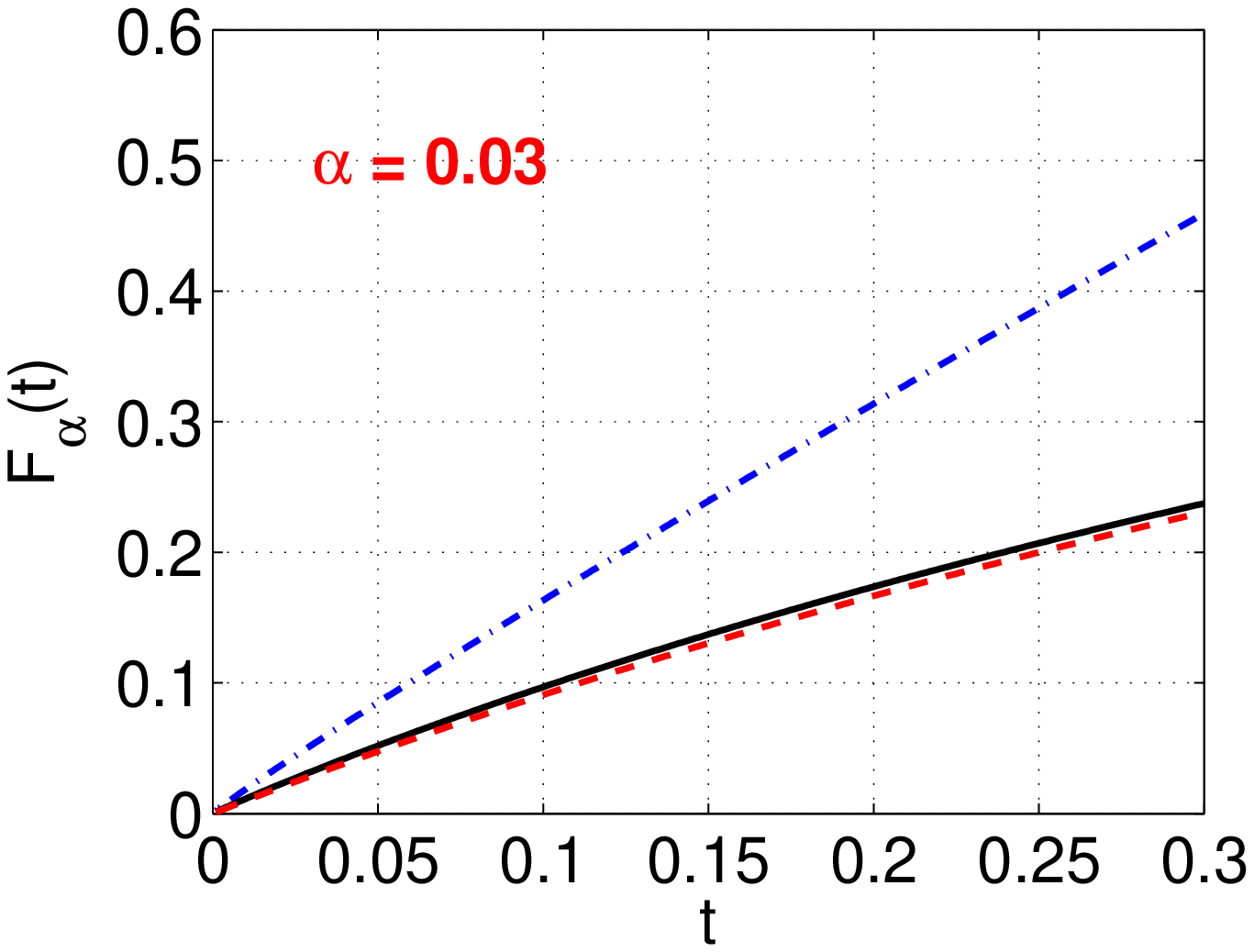}
\includegraphics[width=2.5in]{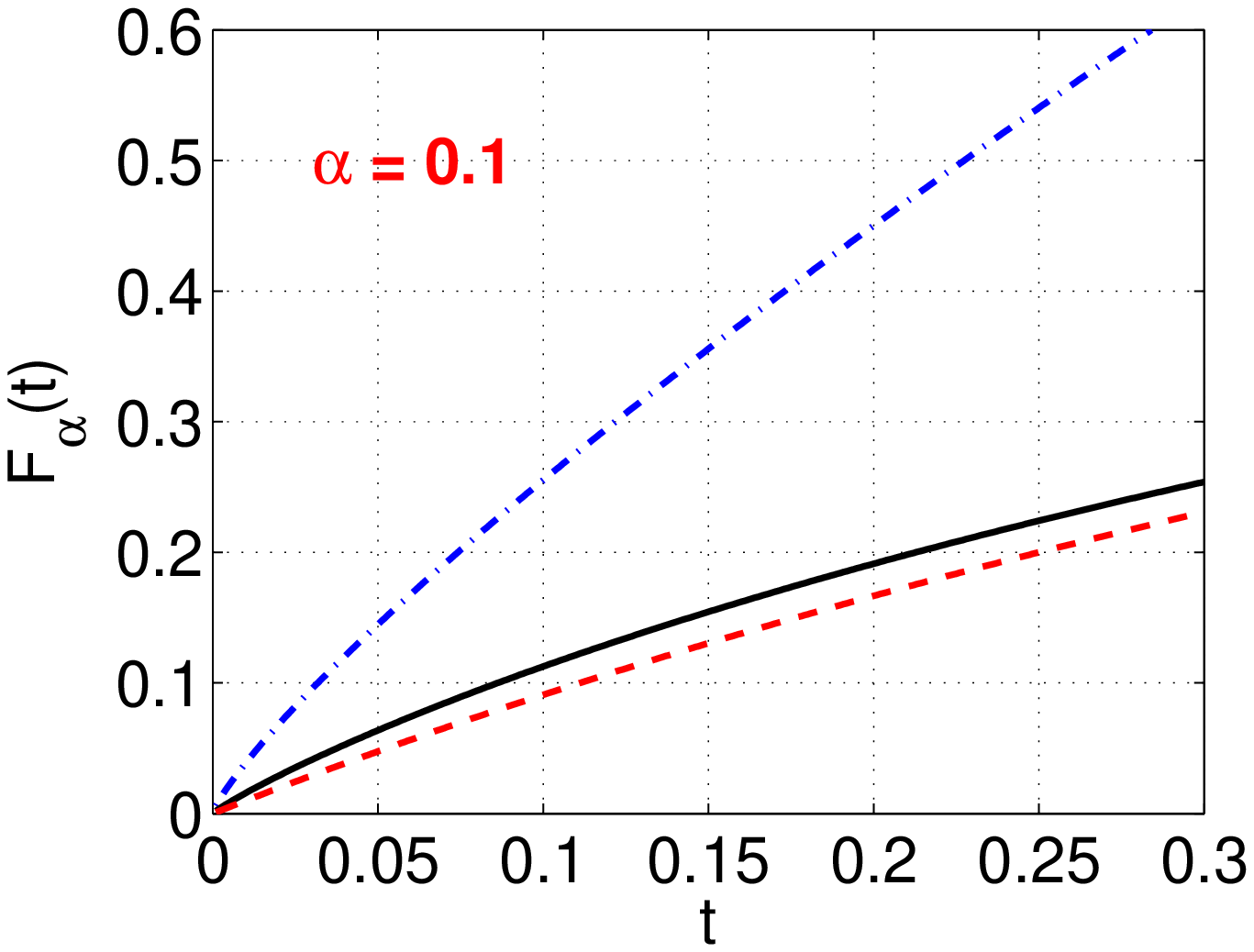}
\end{center}
\vspace{-0.25in}
\caption{$F_\alpha(t)$ obtained by simulations (solid curves), for $\alpha = 0.03$ and $\alpha = 0.1$. In each panel, the bottom and top curves are the lower bound (\ref{eqn_F_lower}) and upper bound (\ref{eqn_F_upper}), respectively. The lower bound is  sharp, especially for small $\alpha$.}\label{fig_F}
\end{figure}

Figure~\ref{fig_F} plots the simulated $F_\alpha(t)$ curves together with the upper and lower bounds.

\newpage

\subsection{Analysis of the Absolute Minimum Estimator}
Recall the definition of the absolute min estimator:
\begin{align}
\hat{x}_{i,min} = \frac{y_{t}}{s_{it}},\hspace{0.2in}\text{ where } \ t = \underset{1\leq j\leq M}{\text{argmin}}\ \ \left|\frac{y_j}{s_{ij}}\right|
\end{align}
If $|\hat{x}_{i,min}|>\epsilon$, then we consider the $i$-th coordinate is a (candidate of) nonzero entry. Our task is to analyze the probability of false positive, i.e., $\mathbf{Pr}\left(|\hat{x}_{i,min}| > \epsilon,x_i=0\right)$, and  the probability of false negative, i.e., $\mathbf{Pr}\left(|\hat{x}_{i,min}| \leq \epsilon,|x_i|>\epsilon\right)$. Again, we should keep in mind that, in the proposed method, i.e., Alg.~\ref{alg_recovery}, the minimum estimator is merely the  first crude step for filtering out many true zero coordinates. False positives will have chance to be removed by the gap estimator and iterative process.

\subsubsection{\textbf{Analysis of False Positives}}

\begin{theorem}\label{thm_fp}
Assume $\psi = \left(\frac{\epsilon}{\theta}\right)^{\frac{\alpha}{1-\alpha}} \leq 1/3$, where $\theta^\alpha = \sum_{i=1}^N |x_i|^\alpha$.  Then
\begin{align}
\mathbf{Pr}\left(|\hat{x}_{i,min}| > \epsilon,x_i=0\right) \leq \frac{1}{(1+\psi)^M}.
\end{align}
\textbf{Proof:}\ $\frac{y_j}{s_{ij}}  = \frac{\sum_{t=1}^N x_t s_{tj}}{s_{ij}} =\theta_i\frac{S_2}{S_1} + x_i$,  where $S_1$ and $S_2$ are i.i.d. $S(\alpha,1)$ variables. When $x_i=0$,  $\frac{y_j}{s_{ij}} = \theta\frac{S_2}{S_1}$. Using the probability bound in  Lemma~\ref{lem_F_lower}, we obtain
\begin{align}\notag
&\mathbf{Pr}\left(|\hat{x}_{i,min}| > \epsilon,x_i=0\right)= \left[\mathbf{Pr}\left(\left|\frac{y_j}{s_{ij}}\right| > \epsilon,x_i=0\right)\right]^M
=  \left[\mathbf{Pr}\left(|S_2/S_1|> \epsilon/\theta\right)\right]^M\\\notag
=&\left[1-\mathbf{Pr}\left(|S_2/S_1|^{\alpha/(1-\alpha)}\leq \left(\epsilon/\theta\right)^{\alpha/(1-\alpha)}\right)\right]^M
= \left(1-F_\alpha(\psi)\right)^M \leq
\left(1-\frac{1}{1+1/\psi}\right)^M = \frac{1}{(1+\psi)^M}
\end{align}$\hfill\square$
\end{theorem}

The assumption  $\psi = \left(\frac{\epsilon}{\theta}\right)^{\frac{\alpha}{1-\alpha}}\leq 1/3$ is reasonable for small $\alpha$ because $\psi \approx \frac{\epsilon^\alpha}{K}\approx 1/K$.

\subsubsection{\textbf{Required Number of Measurements}}

We  derive the required $M$, number of measurements, based on the false positive probability in Theorem~\ref{thm_fp}. This complexity result is useful if we just use one iteration, which matches the known complexity bounds in the compressed sensing literature.
\begin{theorem}
To ensure that the total number of false positives is bounded by $\delta$, it suffices to let
\begin{align}
M \geq \frac{\log\left( (N-K)/\delta\right)}{\log(1+\psi)}\ \
\end{align}$\hfill\square$
\end{theorem}
Since $\psi = \left(\frac{\epsilon}{\theta}\right)^{\frac{\alpha}{1-\alpha}} \approx 1/K$ and $1/\log(1+\psi)\approx K$,  we define
\begin{align}
M_0 = K\log\left((N-K)/\delta\right)
\end{align}
as a convenient approximation. Note that the parameter $\epsilon$ affects the required $M$ only through $\epsilon^\alpha$. This means our algorithm is not sensitive to the choice of $\epsilon$. For example, when $\alpha = 0.03$, then $(10^{-3})^{\alpha} = 0.8128$,  $(10^{-4})^{\alpha} = 0.7586$. If we can afford to use very small $\alpha$ like 0.001, then $(10^{-10})^{0.001} = 0.9772$ and $(10^{-40})^{0.001} = 0.912$.

\subsubsection{Analysis of False Negatives}

\begin{theorem}\label{thm_fn}
If $\alpha\leq0.05$, $\left(\frac{|x_i|+\epsilon}{\theta_i}\right)^{\alpha/(1-\alpha)}<1/3$, then
\begin{align}
&\mathbf{Pr}\left(|\hat{x}_{i,min}| \leq \epsilon,|x_i|>\epsilon\right)\leq\left\{1-\left[1-\frac{3}{4}\left|\frac{|x_i|+\epsilon}{\theta_i}\right|^{\frac{\alpha}{1+\alpha}}
\left(1-\left|\frac{|x_i|-\epsilon}{|x_i|+\epsilon}\right|^{\alpha/(1-\alpha)}\right)\right]^M\right\}
\end{align}
\textbf{Proof:}\ \ \
\begin{align}\notag
&\mathbf{Pr}\left(|\hat{x}_{i,min}| \leq \epsilon,|x_i|>\epsilon\right)
= 1-\left[1-\mathbf{Pr}\left(\left|\frac{y_j}{s_{ij}}\right| \leq \epsilon,|x_i|>\epsilon\right)\right]^M
\end{align}
Again, we can write $\frac{y_j}{s_{ij}}  =  \theta_i\frac{S_2}{S_1} + x_i$. By symmetry,
\begin{align}\notag
&\mathbf{Pr}\left(\left|\frac{y_j}{s_{ij}}\right| \leq \epsilon,|x_i|>\epsilon\right)
= \mathbf{Pr}\left((|x_i|-\epsilon)/\theta_i\leq S_2/S_1\leq (|x_i|+\epsilon)/\theta_i\right)\\\notag
=& \frac{1}{2}\left(F_\alpha\left(\left|\frac{|x_i|+\epsilon}{\theta_i}\right|^{\alpha/(1-\alpha)}\right)-F_\alpha\left(\left|\frac{|x_i|-\epsilon}{\theta_i}\right|^{\alpha/(1-\alpha)}\right)\right)
\end{align}
The  result follows from the probability bounds, $F_\alpha(t)\leq 1.5t^{\frac{1-\alpha}{1+\alpha}}$ and $F_\alpha(t)-F_\alpha(s) \leq (1-s/t)F_\alpha(t)$. $\hfill\square$
\end{theorem}

\subsubsection{\textbf{The Choice of Threshold $\epsilon$}}

We can better understand the choice of $\epsilon$ from the false negative probability as shown in Theorem~\ref{thm_fn}. Assume $x_i\neq 0$ and $|x_i|/\epsilon = H_i\gg1$, the probability $\mathbf{Pr}\left(|\hat{x}_{i,min}| \leq \epsilon,|x_i|>\epsilon\right)$ upper bound is roughly
\begin{align}\notag
&1-\left[1-\frac{3/4}{K}
\left(1-\left|\frac{H_i-1}{H_i+1}\right|^{\alpha}\right)\right]^M\approx 1-\left[1-\frac{3/4}{K}\frac{2\alpha}{H_i}\right]^M
\approx1- e^{-\frac{3/2\alpha M}{KH_i}} \approx \frac{3/2\alpha M}{KH_i}
\end{align}
As we usually choose $M\leq M_0 = K\log ((N-K)/\delta)$, we have
$\frac{\alpha M}{KH_i} < \frac{\alpha \log ((N-K)/\delta)}{H_i}$. To ensure that all the $K$ nonzero coordinates can be safely detected by the minimum estimator (i.e., the total false negatives should be less than $\delta$), we need to choose
\begin{align}
\sum_{x_i\neq 0}\frac{1}{H_i} < \frac{\delta}{1.5\alpha \log \frac{N-K}{\delta}}.
\end{align}
For sign signals, i.e., $|x_i|=1$ if $x_i\neq 0$, we need to have  $H_i > 1.5\alpha K\log \frac{N-K}{\delta}/\delta$, or equivalently $\epsilon < \frac{\delta}{1.5\alpha K\log \frac{N-K}{\delta}}$. If $K=100$ (or 1000), it is sufficient to let $\epsilon = 10^{-4}$ (or $10^{-5}$). Note that even with $N=2^{32}$ (and $\delta = 0.01$), $\log(N/\delta) =  26.8$ is still not large.

For general signals, when the smallest $H_i$ dominate $\sum_{x_i\neq 0}\frac{1}{H_i}$, we essentially just need the smallest $H_i > 1.5\alpha \log \frac{N-K}{\delta}/\delta$, without the $K$ term. In our experiments, for simplicity, we let $\epsilon =10^{-5}$, for both sign signals and Gaussian signals.\\

Again, we emphasize that this threshold analysis is based  on the minimum estimator, for the first iteration only. With the gap estimator and the iterative process, we find the performance is not sensitive to $\epsilon$ as long as it is small, e.g., $10^{-6}$ to $10^{-4}$.

\subsection{Analysis of the Gap Estimator }

The absolute minimum estimator only detects the locations of nonzero coordinate (in the first iteration). To estimate the magnitudes of these detected coordinates, we resort to the gap estimator, defined as follows:
\begin{align}
&z_{i,j} = y_j/s_{ij}, \ \ \ z_{i,(1)}\leq z_{i,(2)}\leq ... \leq z_{i,(M)}\\
&j_i = \underset{1\leq j\leq M-1}{\text{argmin}}\ \{z_{i,(j+1)}-z_{i,(j)}\}\\
&\hat{x}_{i,gap} = \frac{z_{i,(j_i)} + z_{i,(j_i+1)}}{2}
\end{align}

Although the gap estimator is intuitive from Figure~\ref{fig_appr_CDF}, the precise analysis is not trivial. To analyze the recovery error probability bound of $\mathbf{Pr}\left(|\hat{x}_{i,gap} - x_i|>\epsilon\right)$, we first need a bound of the gap probability.

\subsubsection{\textbf{The Gap Probability Bound}}

\begin{lemma}\label{lem_eta}
Let $k>1$, $\gamma = (1-\alpha)/\alpha$, $1\leq c_0\leq 2$, $z_{i,j} = y_j/s_{ij}$, $t_{i,j} = \left(|z_{i,j} -x_i|/\theta_i\right)^{1/\gamma}$, and $\{[1], [2], ..., [M]\}$ a permutation of $\{1, 2, ..., M\}$ giving $t_{i,[1]} \leq t_{i,[2]}\leq ... \leq t_{i,[M]}$. Then
\begin{align}
&\mathbf{Pr}\left(\frac{|z_{i,[k+2]}| - |z_{i,[k+1]}|}{|z_{i,[2]} - z_{i,[1]}|}\leq 1, \ \frac{F_\alpha(t_{i,[1]})}{F_\alpha(t_{i,[2]})}\leq (c_0-1)^{1/\gamma}\right)
\leq \eta_{k,\gamma,c_0}\left(1 + \frac{1}{2k}\right)\\\label{eqn_eta}
&\eta_{k,\gamma,c_0}= \min\left\{u\in(0,1): c_0\left(1-\left(\frac{u}{2k}\right)^{1/k}\right)^{\gamma} + \left(1-\frac{u}{2k}\right)^{\gamma}\leq 1\right\}
\end{align}
\textbf{Proof:}\ \ See Appendix~\ref{app_lem_eta}.\hspace{0.1in}$\hfill\square$
\end{lemma}

Although in this paper we  only use $\eta_{k,\gamma,2}$ (i.e., $c_0=2$ and   $\frac{F_\alpha(t_{i,[1]})}{F_\alpha(t_{i,[2]})}\leq (c_0-1)^{1/\gamma}$ always holds), we  keep a more general bound which might   improve the analysis of the gap estimator (or other estimators) in future study. Also, we should notice that as $\gamma\rightarrow \infty$ (i.e., $\alpha\rightarrow0$), $\eta_{k,\gamma,c0}\rightarrow 0$.

\begin{figure}[h!]
\begin{center}
\mbox{
\includegraphics[width=3.2in]{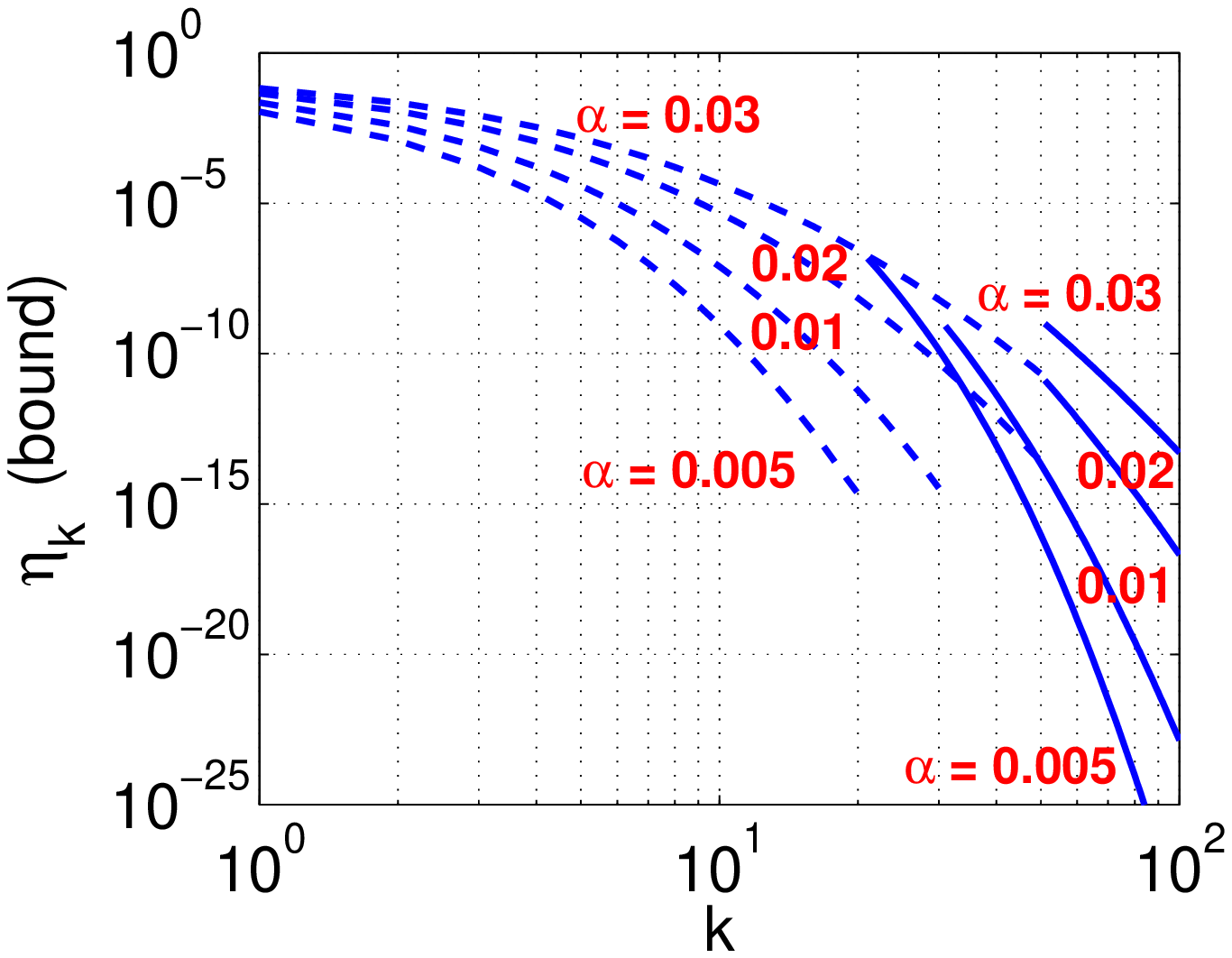}\hspace{0in}
\includegraphics[width=3.2in]{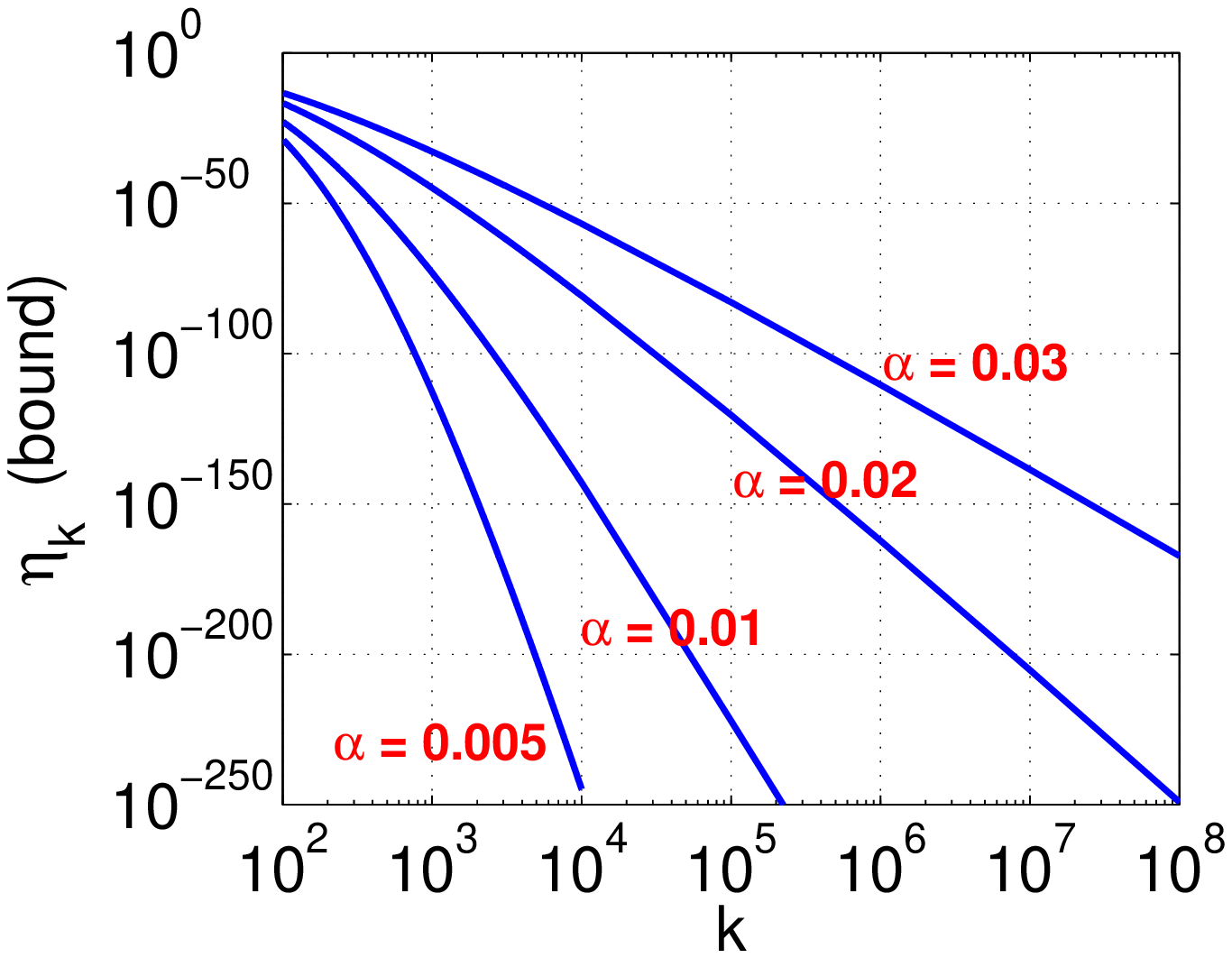}}

\end{center}

\vspace{-0.2in}
\caption{$\eta_{k,\gamma,2}$ or its upper bound for $\alpha  = 0.005$, 0.01, 0.02, 0.03. For $k\leq 20$ ($\alpha = 0.005$), $k\leq 30$ ($\alpha=0.01$), and $k\leq 50$  ($\alpha = 0.02, 0.03$), we numerically solve for $\eta_{k,\gamma,2}$ from (\ref{eqn_eta}), presented as dashed curves in the left panel. For larger $k$ values, we use the upper bound in (\ref{eqn_eta_bound}), presented as solid curves. }\label{fig_eta}
\end{figure}

As shown in Figure~\ref{fig_eta}, for small $k$, the constant $\eta_{k,\gamma,c_0}$ can be numerically evaluated. For larger $k$ values, we resort to an upper bound which is numerically stable for any $k$, in the next Lemma.
\begin{lemma}
\begin{align}\label{eqn_eta_bound}
&\eta_{k,\gamma,c_0} \leq \min\left\{u\in(0,1): \log c_0 + \gamma\log\log\frac{2k}{u} - \gamma\log k+\log\left(1+\frac{2k}{u\gamma}\right)\leq 0\right\}
\end{align}
\textbf{Proof:}\ Note that $\left(u/(2k)\right)^{1/k}$ is close to 1 and $u/(2k)$  close to 0. We  apply the inequalities: $1-t \leq -\log t$, $1-(1-t)^\gamma \leq 1/(1+1/(\gamma t))$, $\forall 0<t<1$, to obtain
\begin{align}\notag
&c_0\left(1-\left(\frac{u}{2k}\right)^{1/k}\right)^{\gamma} + \left(1-\frac{u}{2k}\right)^{\gamma}\leq 1
\Leftarrow c_0\left(\frac{1}{k}\log\frac{u}{2k} \right)^\gamma \leq \frac{1}{1+1/(\gamma u/2k)}\ \
\end{align}$\hfill\square$
\end{lemma}

\subsubsection{\textbf{Reconstruction Error Probability}}

\begin{theorem}\label{thm_gap}
Let $\gamma = (1-\alpha)/\alpha$. Suppose the existence of $a_0>1$ and positive integer $k_0$ satisfying
$\theta\left(a_0k_0/(M+a_0k_0)\right)^\gamma\leq \epsilon$, i.e., $a_0k_0\leq M/ {K}^*_{\epsilon,\theta,\alpha}$, where ${K}^*_{\epsilon,\theta,\alpha} = 1/\left(\epsilon/\theta\right)^{\alpha/(1-\alpha)}-1$.
Then
\begin{align}\label{eqn_upper_G}
&\mathbf{Pr}\left(|\hat{x}_{i,gap} - x_i|> \epsilon\right) \leq {G_{M,{K}^*_{\epsilon,\theta,\alpha}}}
=
\underset{a_0,k_0}{\min} \left\{ B(M,a_0k_0/M)+\sum_{k=k_0}^{M-2}\left(1+\frac{1}{2k}\right)\eta_{k,\gamma,2}\right\}
\end{align}
where ${\small B(M,a_0k_0/M) = \mathbf{Pr}\left(Binomial(M,a_0k_0/M)< k_0\right)}$ is the binomial CDF.\\

\noindent\textbf{Proof:}\ \ The  idea is that we can start counting the gaps from $k_0$-th gap, because we only need to ensure that the estimated $x_i$ is within a small neighborhood of the true $x_i$, not necessarily from the first gap.

We choose $q = a_0k_0/M$. If  $F_\alpha\left(t_{i,[k_0+1]}\right) <q$,  then $1/\left(1+1/t_{i,[k_0+1]}\right) <q$, or equivalently, $t_{i,[k_0+1]} < q/(1-q)$. Thus, $|\hat{x}_{i,gap} - x_i|\leq \theta_it_{i,[k_0+1]}^\gamma < \theta \left(q/(1-q)\right)^\gamma\leq \epsilon$ when $F_\alpha\left(t_{i,[k_0+1]}\right)<q$ and\\ $\min_{k_0\leq k\leq M-2}\left(|z_{i,[k+2]} - z_{i,[k+1]}|\right) > |z_{i,[2]} - z_{i,[1]}|$.  Thus, the result follows from   Lemma~\ref{lem_eta} and the fact that  $\mathbf{Pr}\left(F_\alpha\left(t_{i,[k_0+1]}\right)\geq q\right) \leq \mathbf{Pr}\left(Binomial(M,q)< k_0\right)$. $\hfill\square$\\

\end{theorem}

Since $k_0$ in Theorem~\ref{thm_gap} only takes finite values, we can basically numerically evaluate $G_{M,{K}^*_{\epsilon,\theta,\alpha}}$ to obtain the upper bound for $\mathbf{Pr}\left(|\hat{x}_{i,gap} - x_i|> \epsilon\right)$. It turns out that,  once $\alpha$ and $\epsilon$ are fixed,  $G$ is only a function of ${K}^* = {K}^*_{\epsilon,\theta,\alpha}$ and the ratio $\frac{M}{{K}^*_{\epsilon,\theta,\alpha}}$. Also, note that ${K}^* \approx K/\epsilon^\alpha$.

\begin{figure}[h!]
\begin{center}
\includegraphics[width=2.5in]{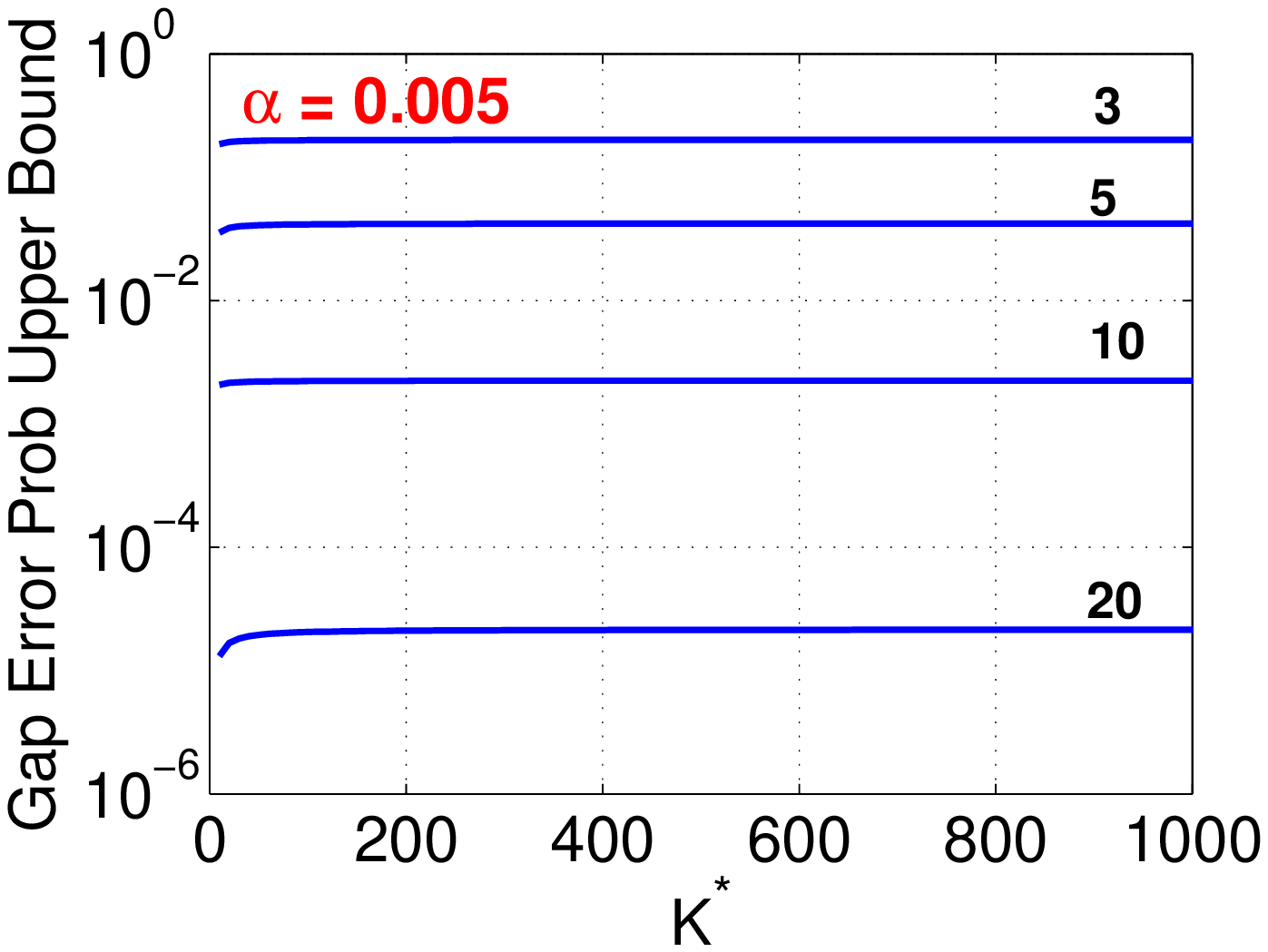}
\includegraphics[width=2.5in]{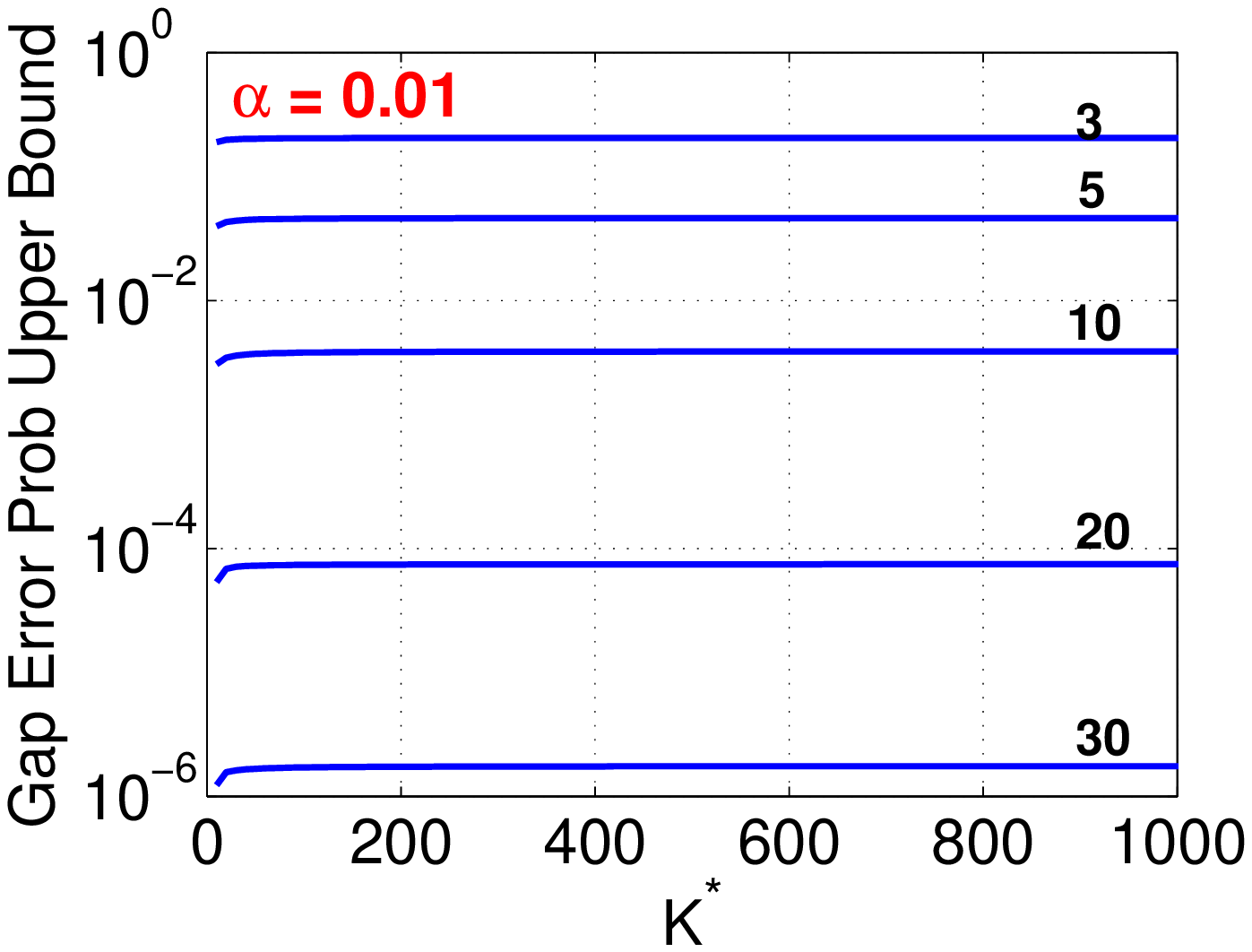}
\includegraphics[width=2.5in]{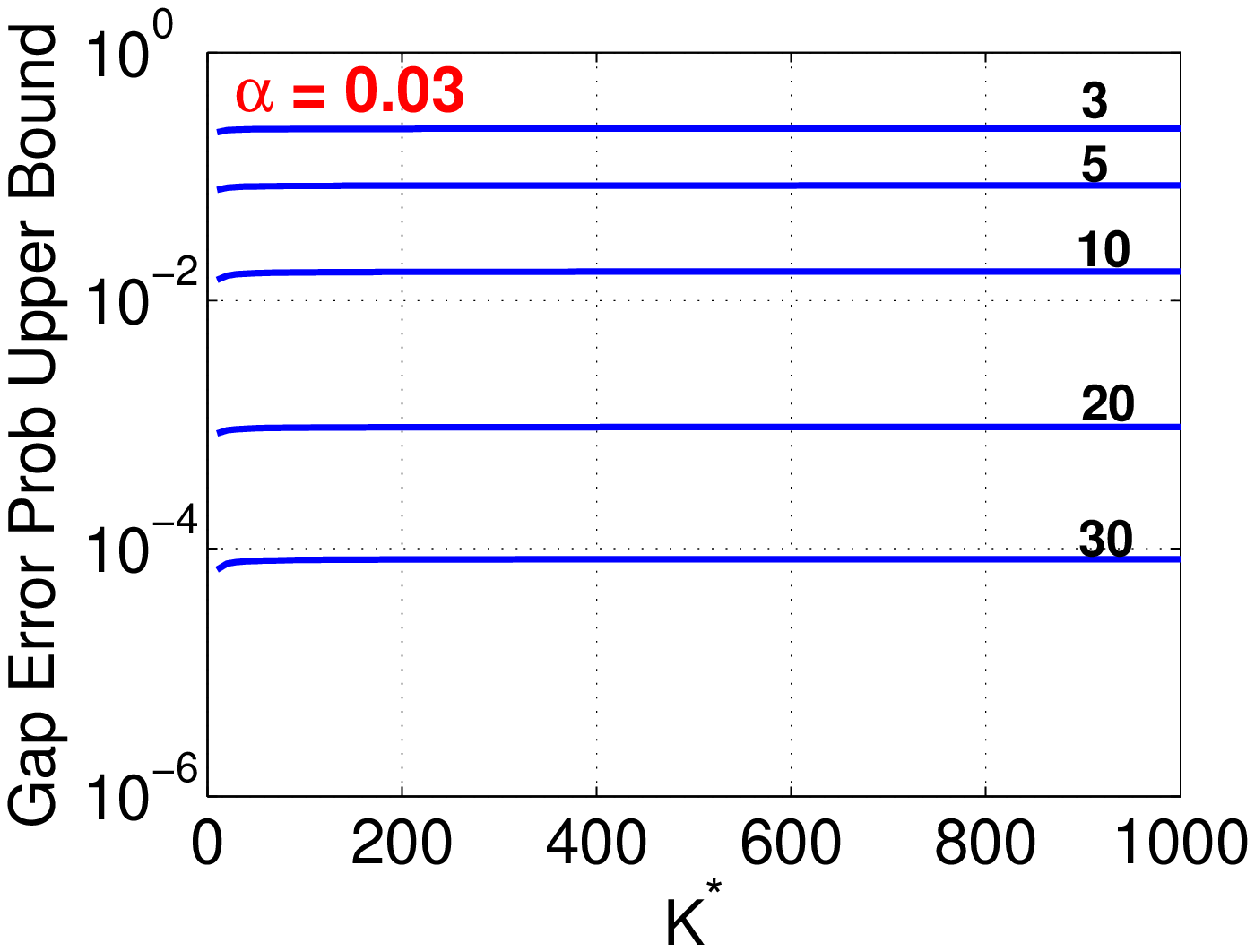}
\includegraphics[width=2.5in]{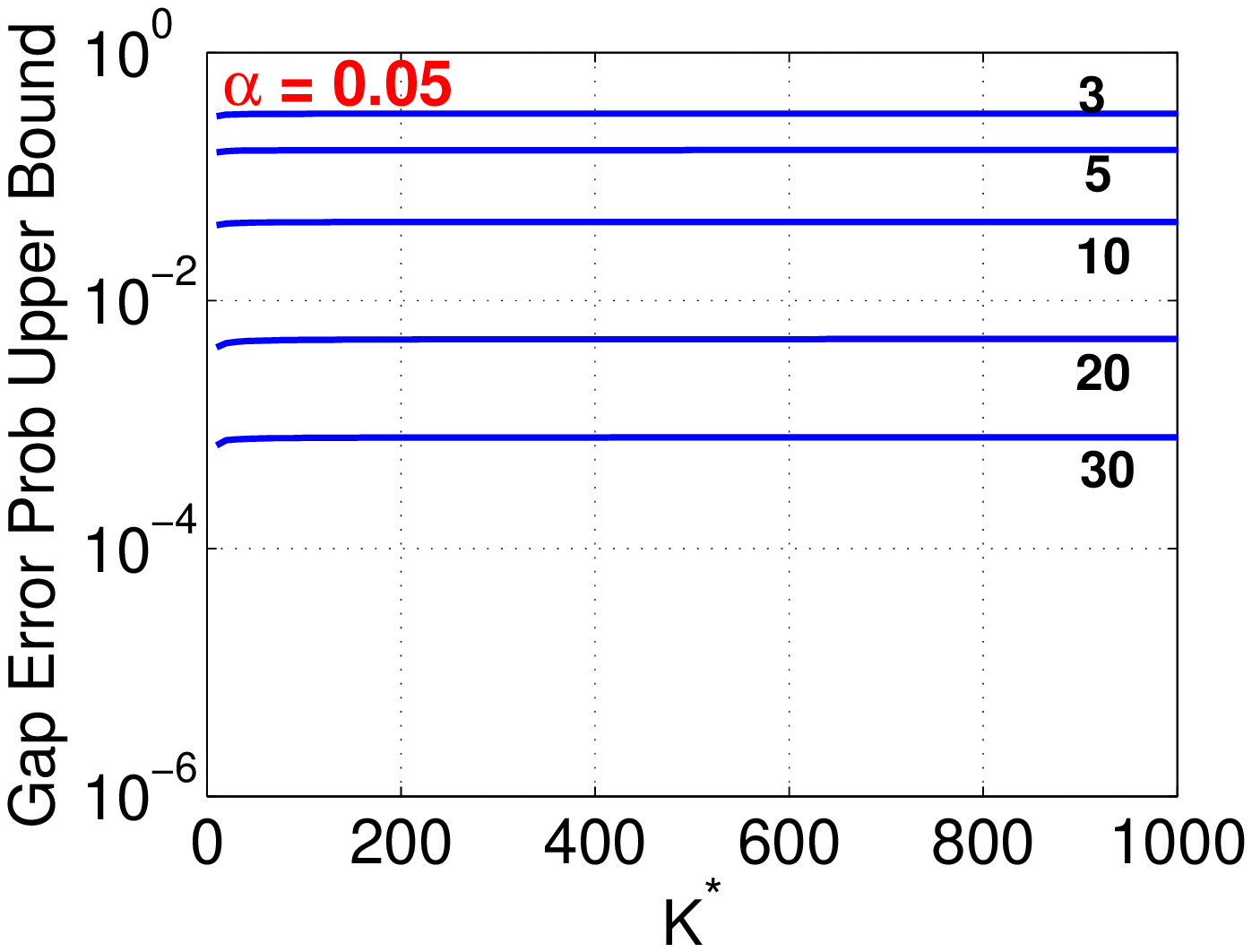}
\end{center}
\vspace{-0.2in}
\caption{Numerical values of the upper bound of the error probability  $\mathbf{Pr}\left(|\hat{x}_{i,gap} -x_i|>\epsilon\right)\leq G_{M,K^*_{\epsilon,\theta,\alpha}}$ as computed in (\ref{eqn_upper_G}). The labels on the curves are the values of $M/K^*_{\epsilon,\theta,\alpha}$. For example, when using $M = 5K^*$ and $\alpha =$ 0.005 / 0.01 / 0.03 / 0.05,   the error probabilities  are 0.042 / 0.046 / 0.084 / 0.163.  In other word, in order for the error probability to be $\leq 0.05$, it suffices to use $M = 5K^*$  if $\alpha = 0.005$ or 0.01. However, if $\alpha = 0.03$ or $0.05$, we will have to use respectively $M = 7K^*$ and $M=10K^*$ measurements in order to achieve error probability $<0.05$. }\label{fig_G}
\end{figure}

Figure~\ref{fig_G} plots the upper bound $G_{M,{K}^*_{\epsilon,\theta,\alpha}}$ for $\alpha = 0.005$, 0.01, 0.03, and 0.05, in terms of ${K}^*$ and $\frac{M}{{K}^*}$. For example, when using $M = 5K^*$ and $\alpha =$ 0.005 / 0.01 / 0.03 / 0.05,   the error probabilities  are 0.042 / 0.046 / 0.084 / 0.163.  In other word, in order for the error probability to be $\leq 0.05$, it suffices to use $M = 5K^*$  if $\alpha = 0.005$ or 0.01. When $\alpha = 0.03$ or $0.05$, we will have to use respectively $M = 7K^*$ and $M=10K^*$ measurements in order to achieve error probability $<0.05$.

This way, the required sample size can be at least numerically computed from Theorem~\ref{thm_gap}. Of course, we should keep in mind that the values are merely the (possibly conservative) upper bounds.

\subsubsection{Connection to the ``Idealized'' Algorithm}

The ``idealized'' algorithm analyzed in Sec.~\ref{sec_ideal} assumes $\alpha\rightarrow0$ and that, as long as there are two or more observations within $(x_i-e, x_i+e)$, there will be an algorithm which could perfectly recover $x_i$ (provided $e$ is small enough). The gap estimator is a surrogate for implementing the ``idealized'' algorithm.

In the ``idealized'' algorithm, the  probability that $x_i$ can not be recovered is
\begin{align}\notag
p_{ideal} = \left(1-1/K\right)^M + \left(1-1/K\right)^{M-1}M/K
\end{align}
which is basically the limit of  $\mathbf{Pr}\left(|\hat{x}_{i,gap} - x_i|> \epsilon\right)$ in Theorem~\ref{thm_gap}, as $\alpha\rightarrow0$.  Recall  $\eta_{k,\gamma,c_0} \rightarrow0$ if $\alpha\rightarrow0$. We can see  from (\ref{eqn_upper_G}) that $G_{M,K^*_{\epsilon,\theta,\alpha}}\rightarrow \mathbf{Pr}\left(Binomial(M,1/K)< 2\right) $,  which is exactly $p_{ideal}$.

\subsubsection{Practicality of the Gap Estimator}

The gap estimator is  practical in that the error probability bound (\ref{eqn_upper_G}) holds for any $\alpha$ and $\epsilon$. This property allows us to use a finite $\alpha$ (e.g., 0.03) and very small $\epsilon$. Note that $K^*$ is basically $K/\epsilon^\alpha$. Even if we have to choose $\epsilon=10^{-10}$, i.e., $(10^{-10})^{0.03} = 0.5$, we can still recover $x_i$ by using twice as many examples compared to using $\alpha\rightarrow0$.

The analysis of the ``idealized'' algorithm reveals that $M=5K$ to $7K$ might be sufficient for achieving perfect recovery. In our simulation study in Sec.~\ref{sec_simu}, we find $M = M_0/3$ is good enough for our practical algorithm for a range of $(M,K)$ values. Perhaps not surprisingly, one can verify that the values of $M_0/3$ roughly fall in the $5K \sim  7K$ range for those values of $(M,K)$. This, to an extent, implies that the performance of our practical procedure, i.e., Alg.~\ref{alg_recovery} can be close to what the ``idealized'' algorithm could achieve, despite that the theoretical probability upper bound $G_{M,K^*}$ might be  too conservative when $\alpha$ is away from zero.

\newpage

\section{Measurement Noise}\label{sec_noise}

It is intuitive that our method is robust against measurement noise.
In this paper, we focus on exact sparse recovery. In the compressed sensing literature, the common model is to assume additive measurement noise $\mathbf{y = xS + n}$, where each component $n_j$ is the random noise, which is  typically assumed to be $n_j \sim Normal\left(0, \sigma^2N\right)$. A precise analysis will involve a complicated calculation of convolution.

\subsection{Additive Noise}

\begin{figure}[h!]
\begin{center}
\mbox{
\includegraphics[width=2.5in]{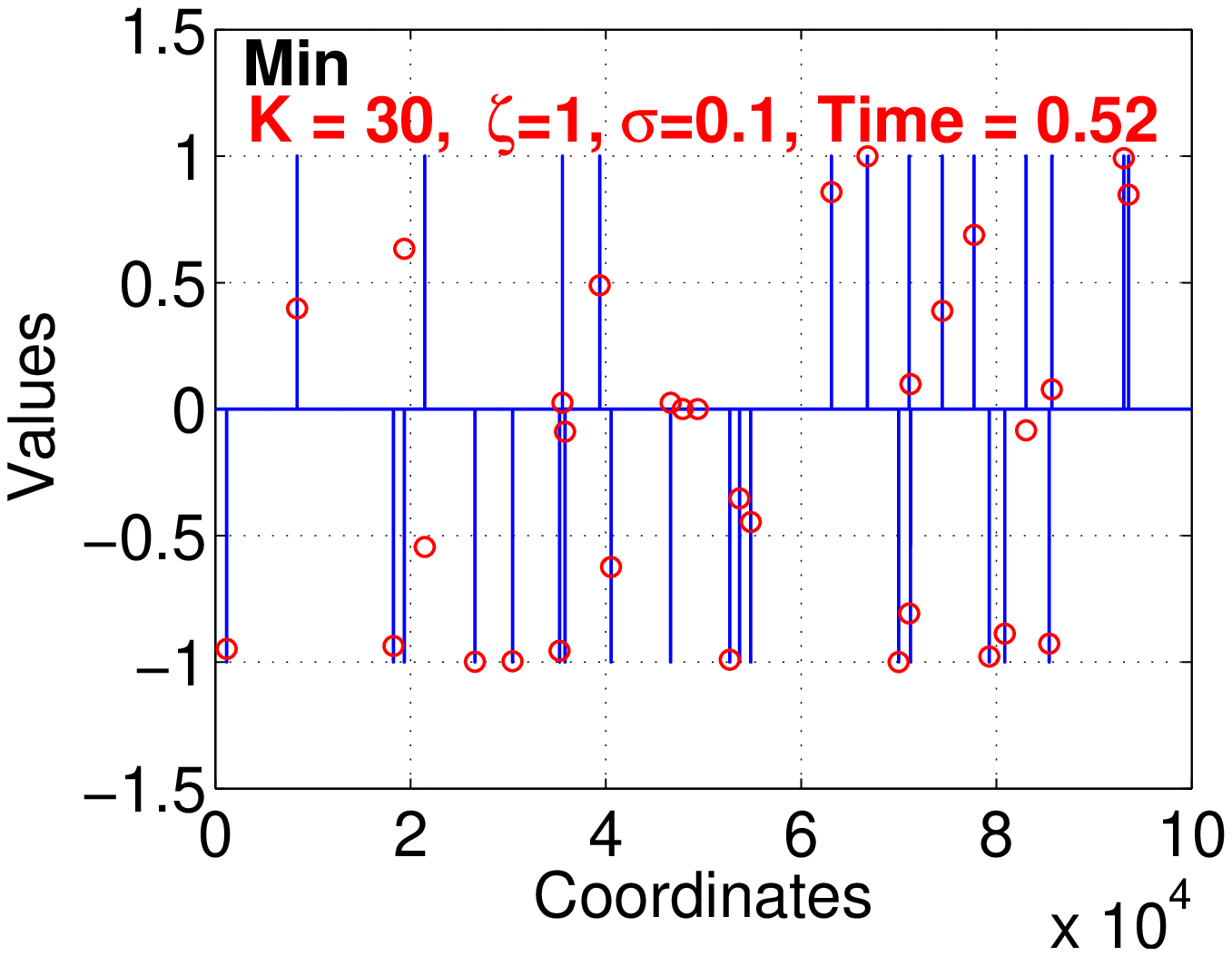}\hspace{0in}
\includegraphics[width=2.5in]{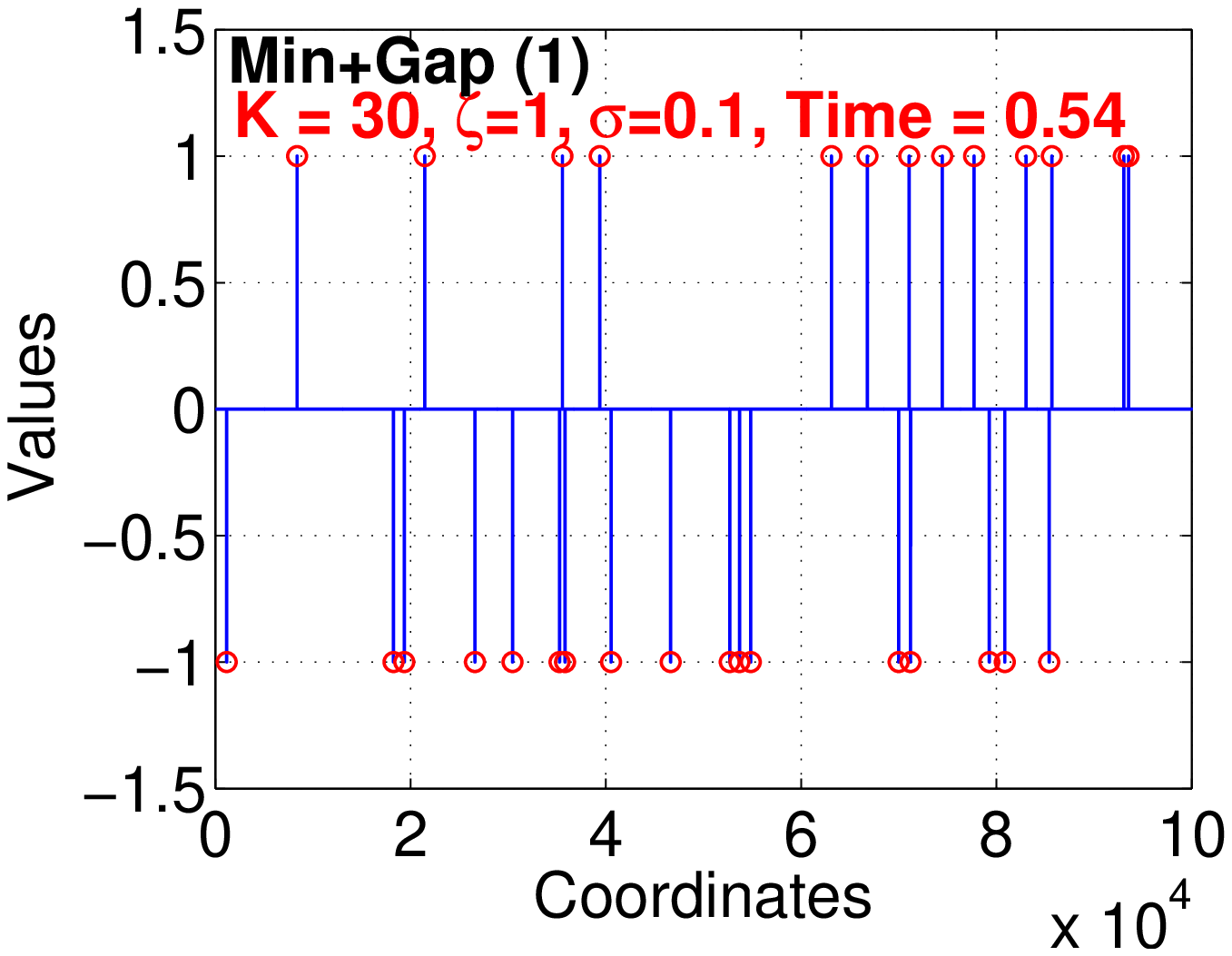}}

\mbox{
\includegraphics[width=2.5in]{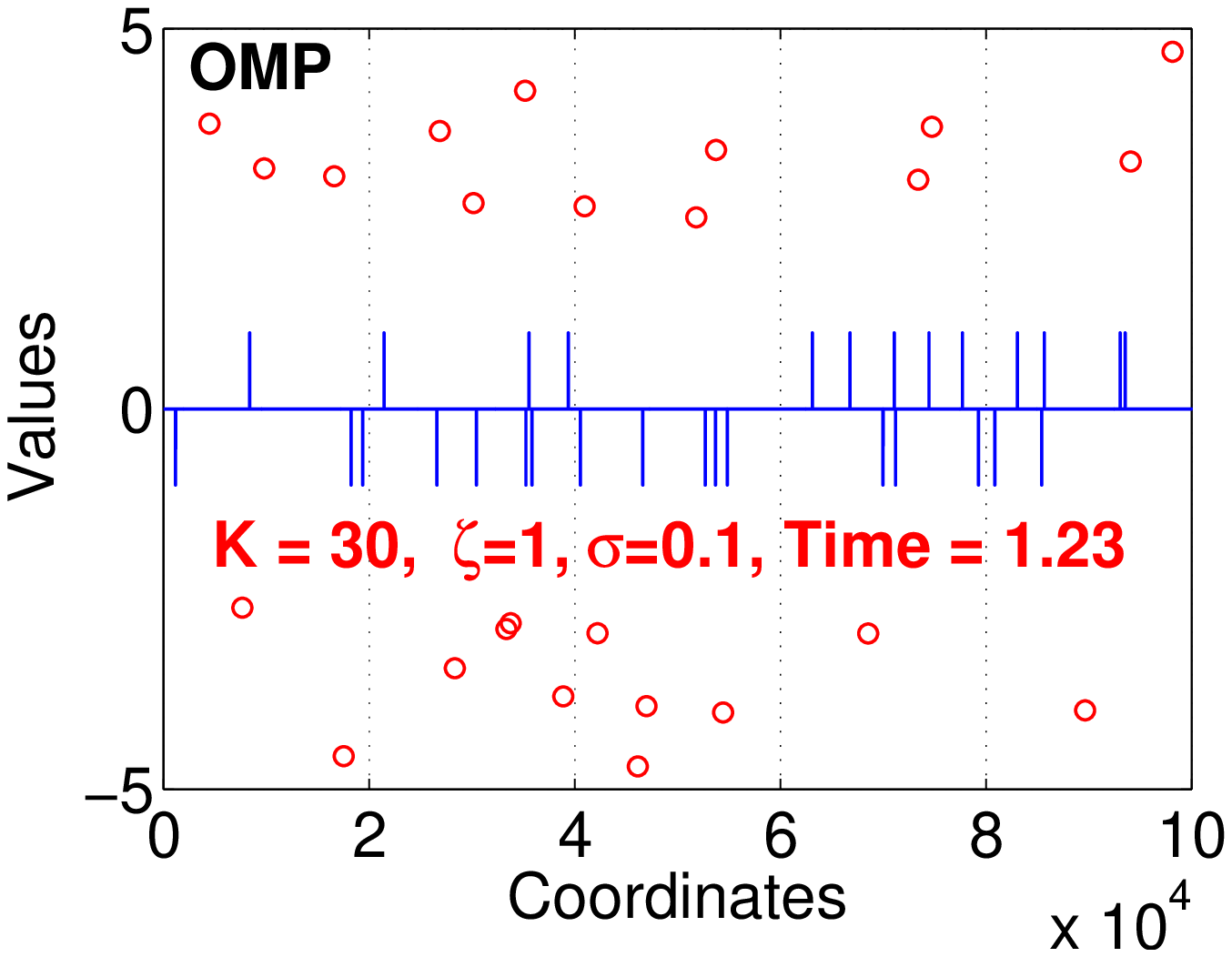}\hspace{0in}
\includegraphics[width=2.5in]{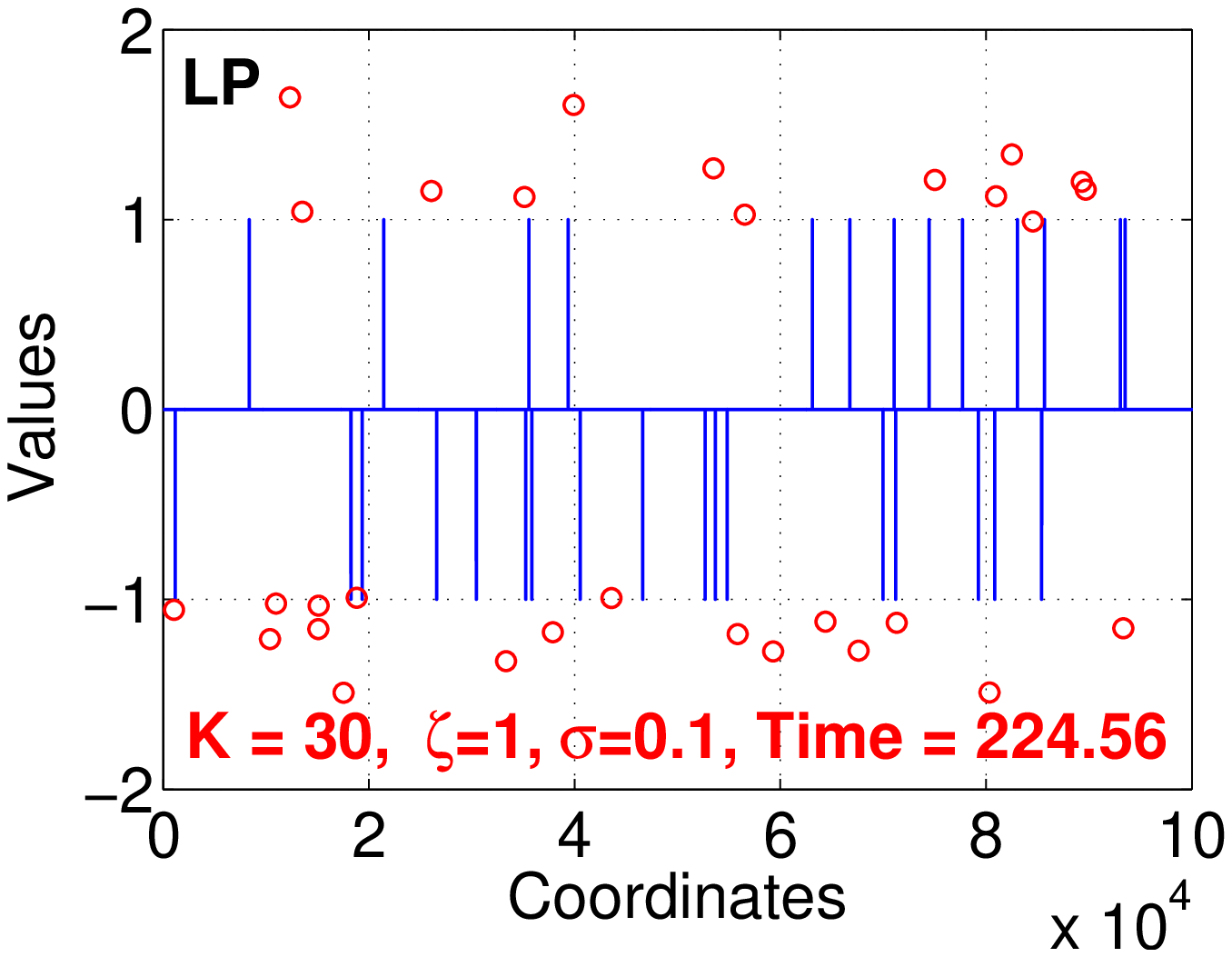}}
\end{center}
\vspace{-0.2in}
\caption{\small Reconstruction results from one simulation, with $N=100000$, $K=30$, $M=M_0$ (i.e., $\zeta=1$), $\sigma=0.1$, and sign signals. With the proposed method, the signal is perfectly reconstructed in one iteration. In comparisons, both OMP and LP perform very poorly.}\label{fig_RecSignB1Sig01}
\end{figure}

\begin{figure}[h!]
\begin{center}
\mbox{
\includegraphics[width=2.5in]{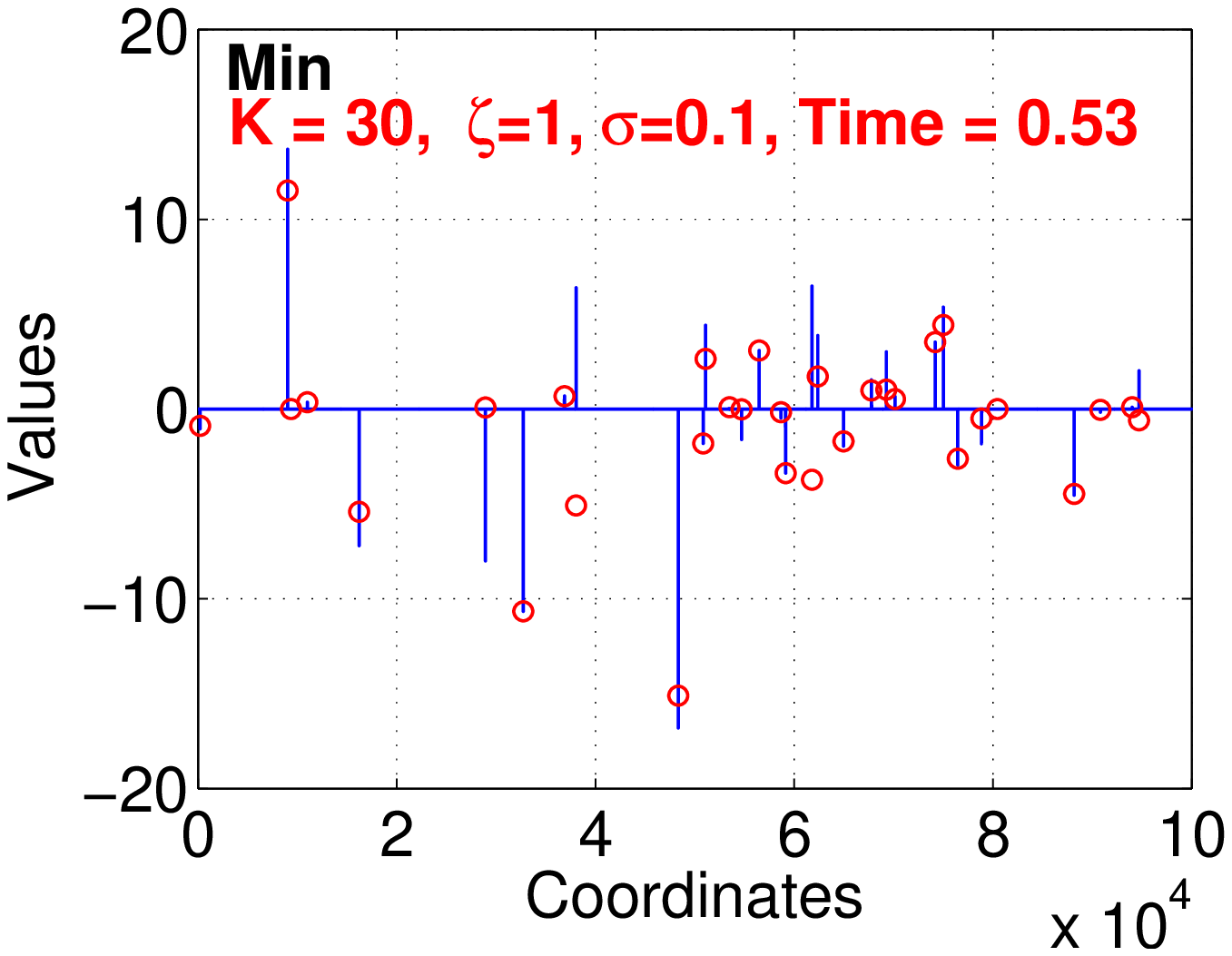}\hspace{0in}
\includegraphics[width=2.5in]{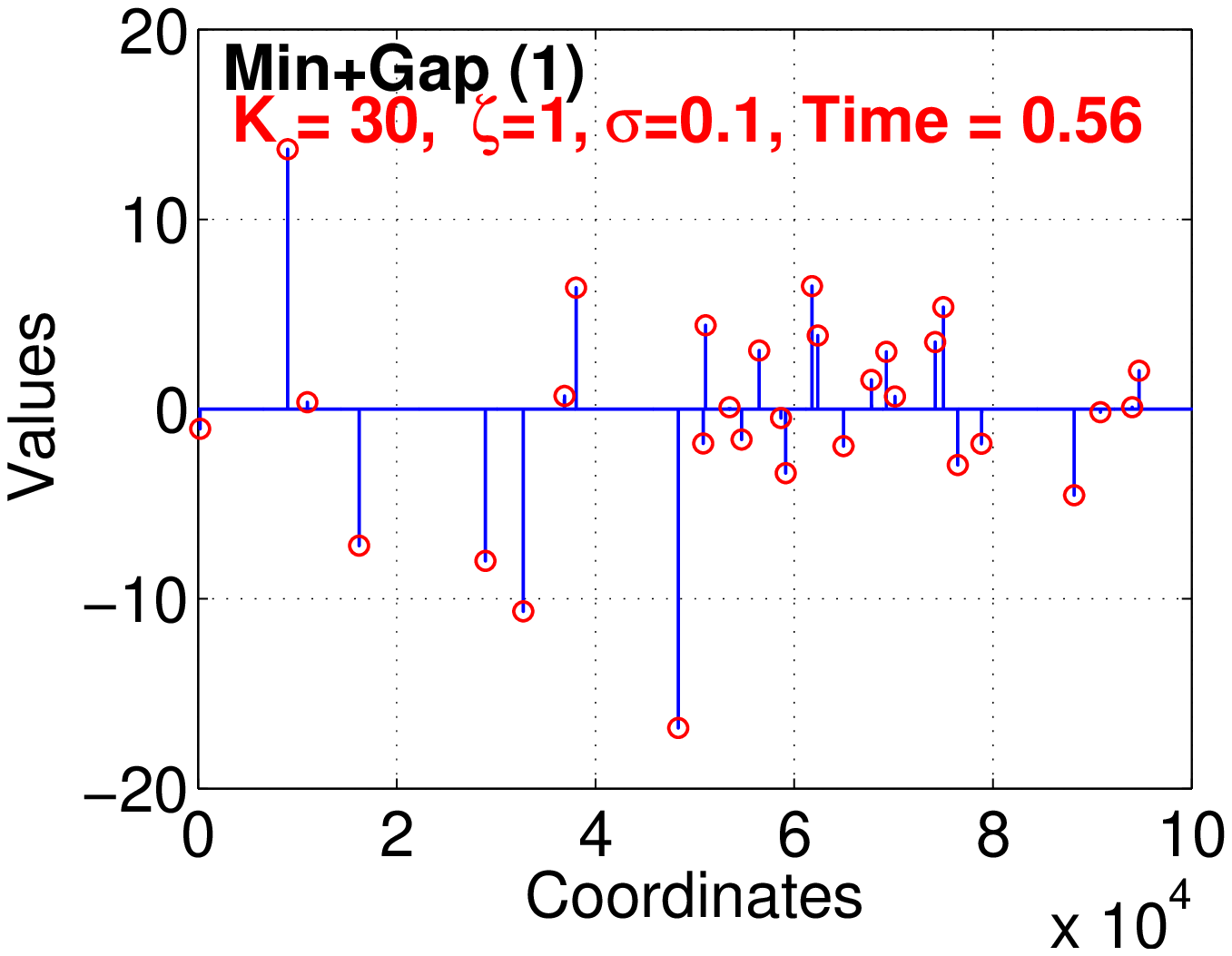}}

\mbox{
\includegraphics[width=2.5in]{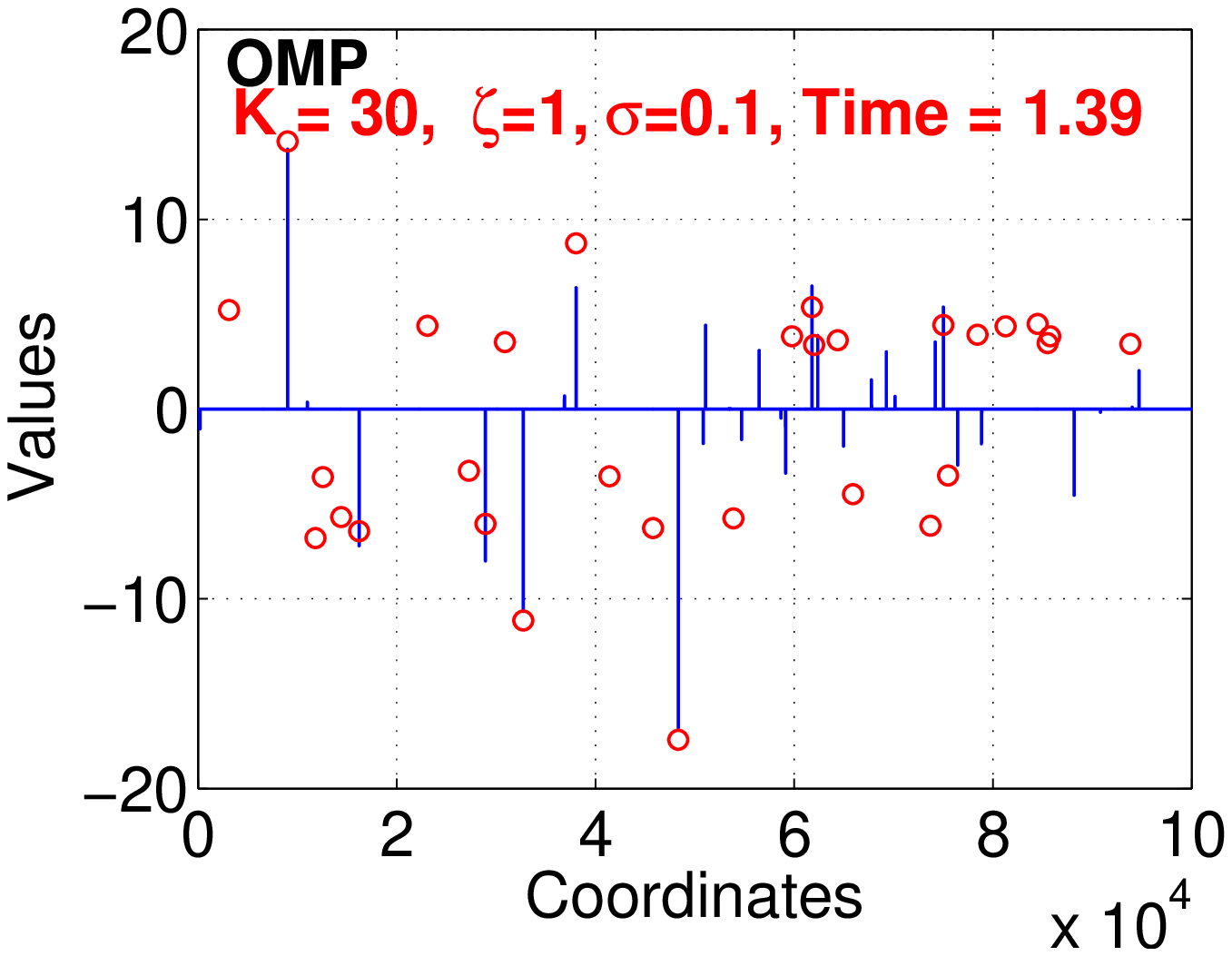}\hspace{0in}
\includegraphics[width=2.5in]{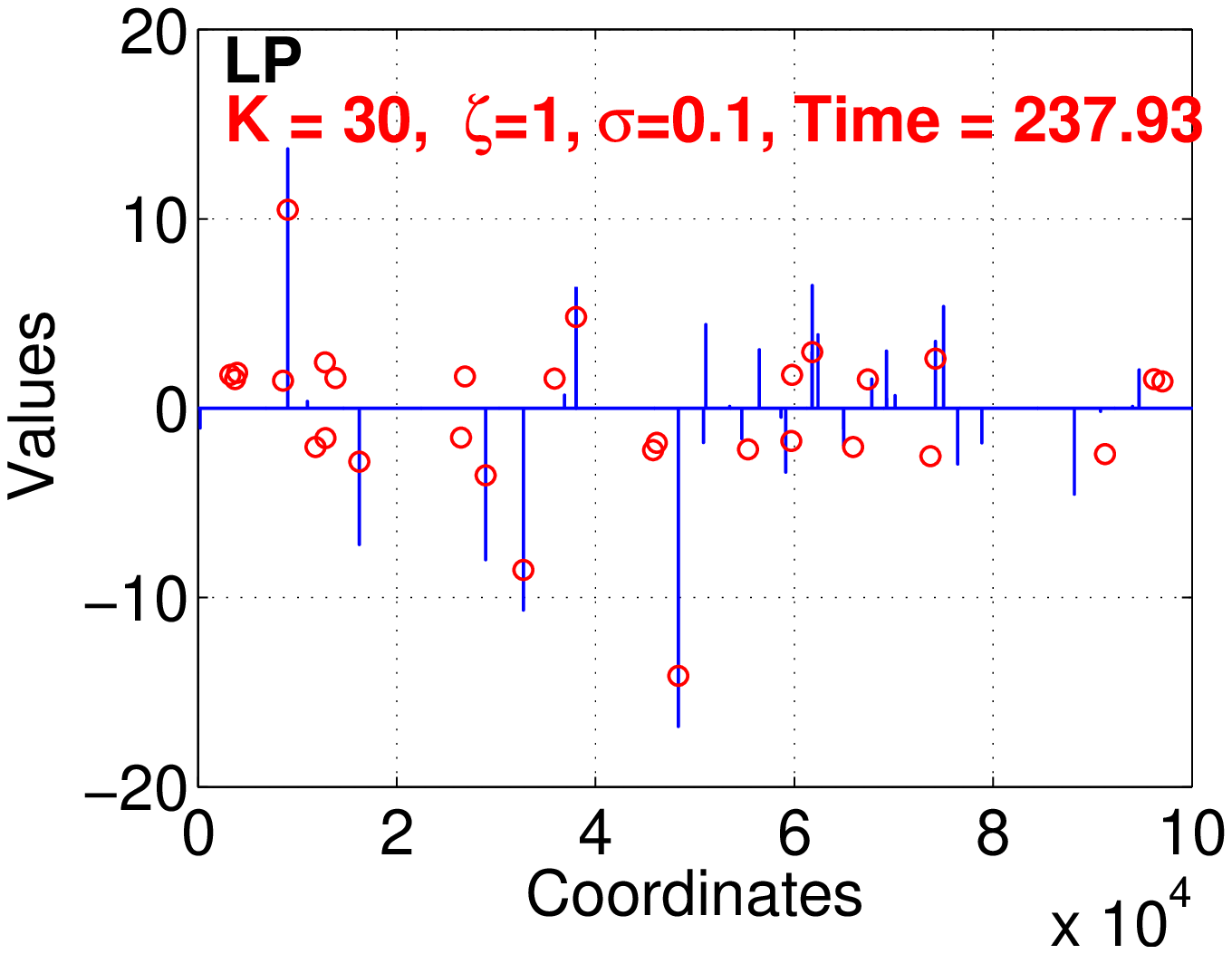}}
\end{center}
\vspace{-0.2in}
\caption{\small Reconstruction results from one simulation, using $N=100000$, $K=30$, $M=M_0$ (i.e., $\zeta=1$), $\sigma=0.1$, and sign signals. Our method (using just one iteration) can still perfectly reconstruct the signal.}\label{fig_RecGausB1Sig01}
\end{figure}

\begin{figure}[h!]
\begin{center}
\mbox{
\includegraphics[width=2.5in]{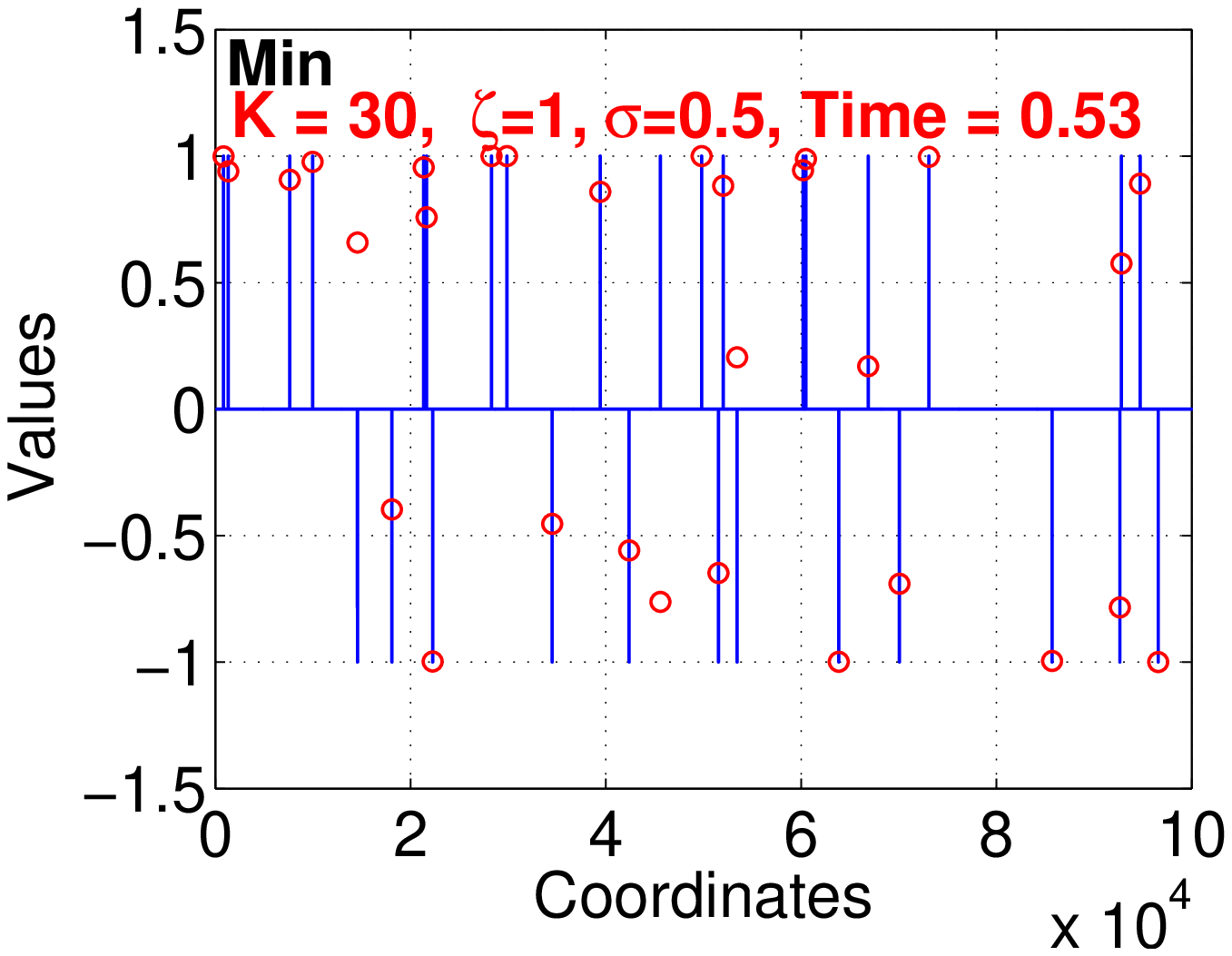}\hspace{0in}
\includegraphics[width=2.5in]{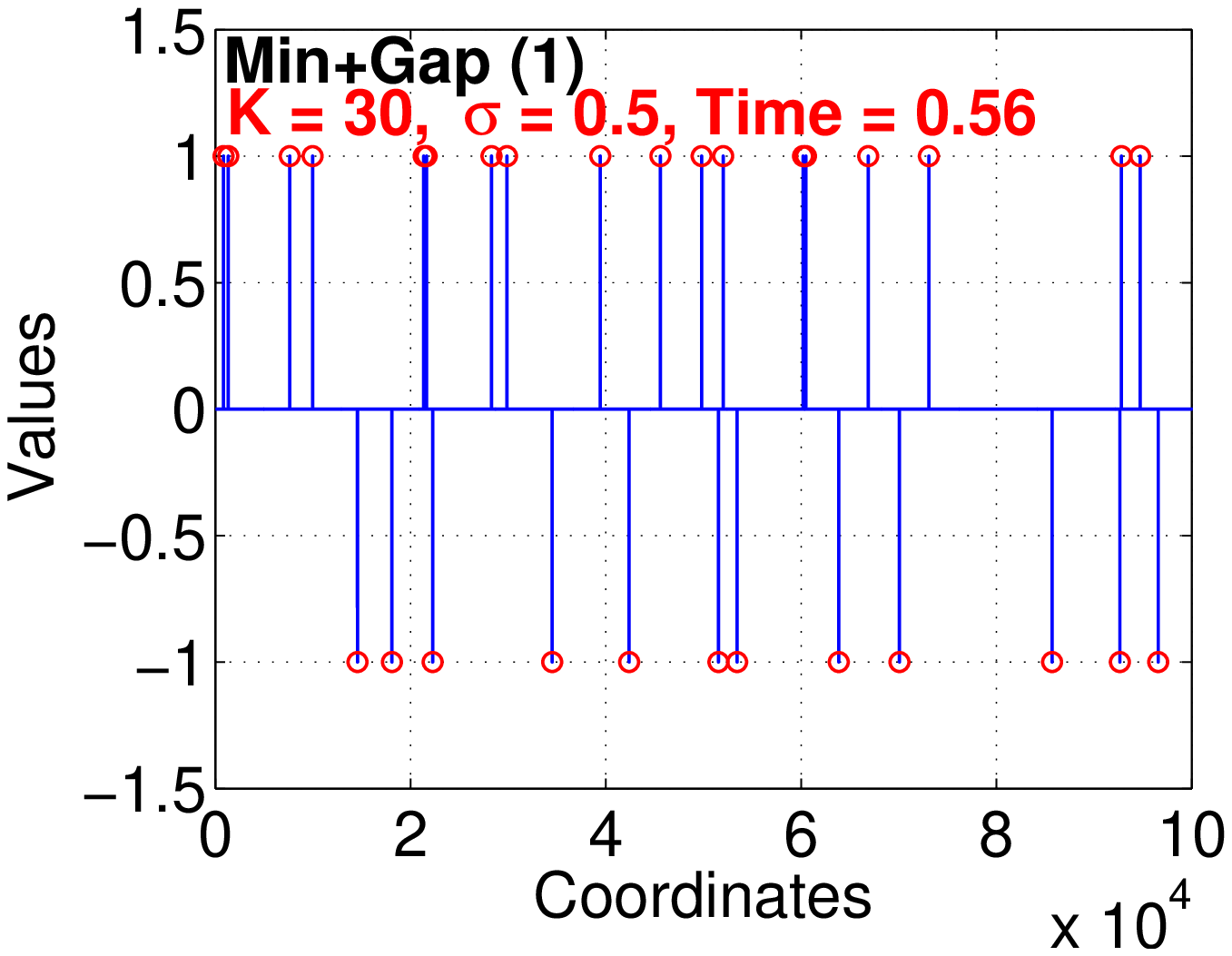}}

\mbox{
\includegraphics[width=2.5in]{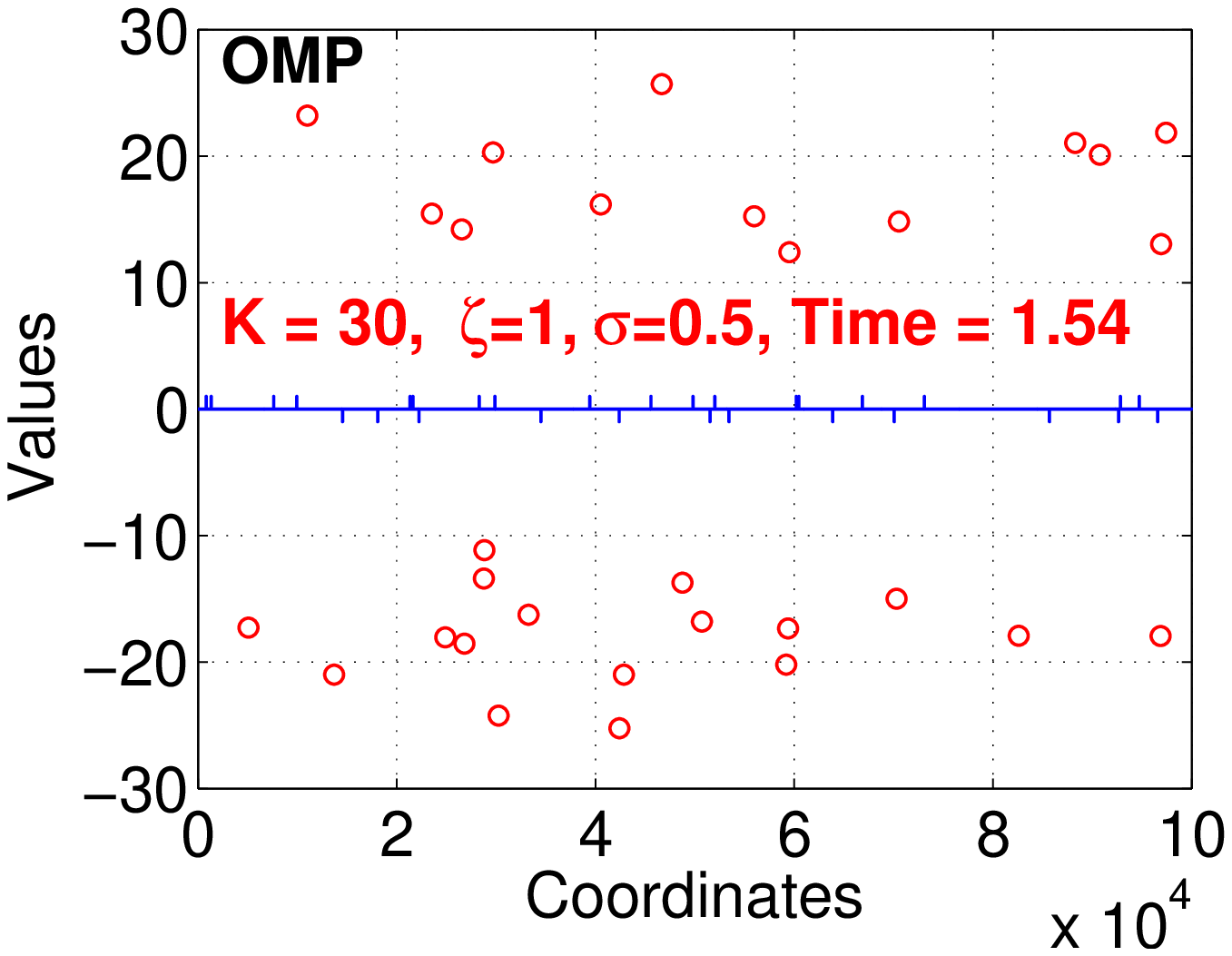}\hspace{0in}
\includegraphics[width=2.5in]{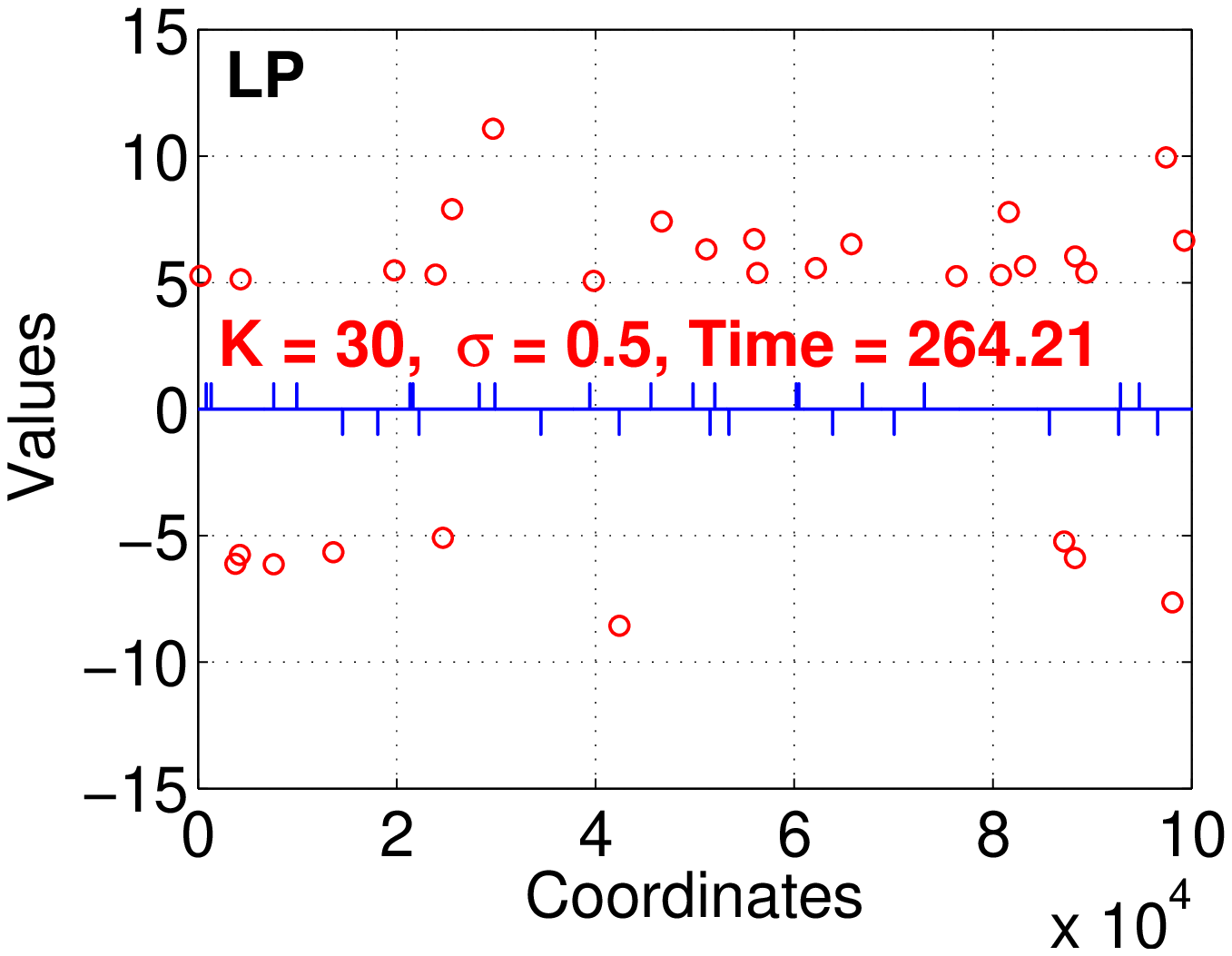}}
\end{center}
\vspace{-0.2in}
\caption{\small Reconstruction results from one simulation, with $N=100000$, $K=30$, $M=M_0$ (i.e., $\zeta=1$), $\sigma=0.5$, and sign signals. Using our method, the signal is perfectly reconstructed with one iteration. In comparisons, both OMP and LP perform poorly.}\label{fig_RecSignB1Sig05}
\end{figure}

\begin{figure}[h!]
\begin{center}
\mbox{
\includegraphics[width=2.5in]{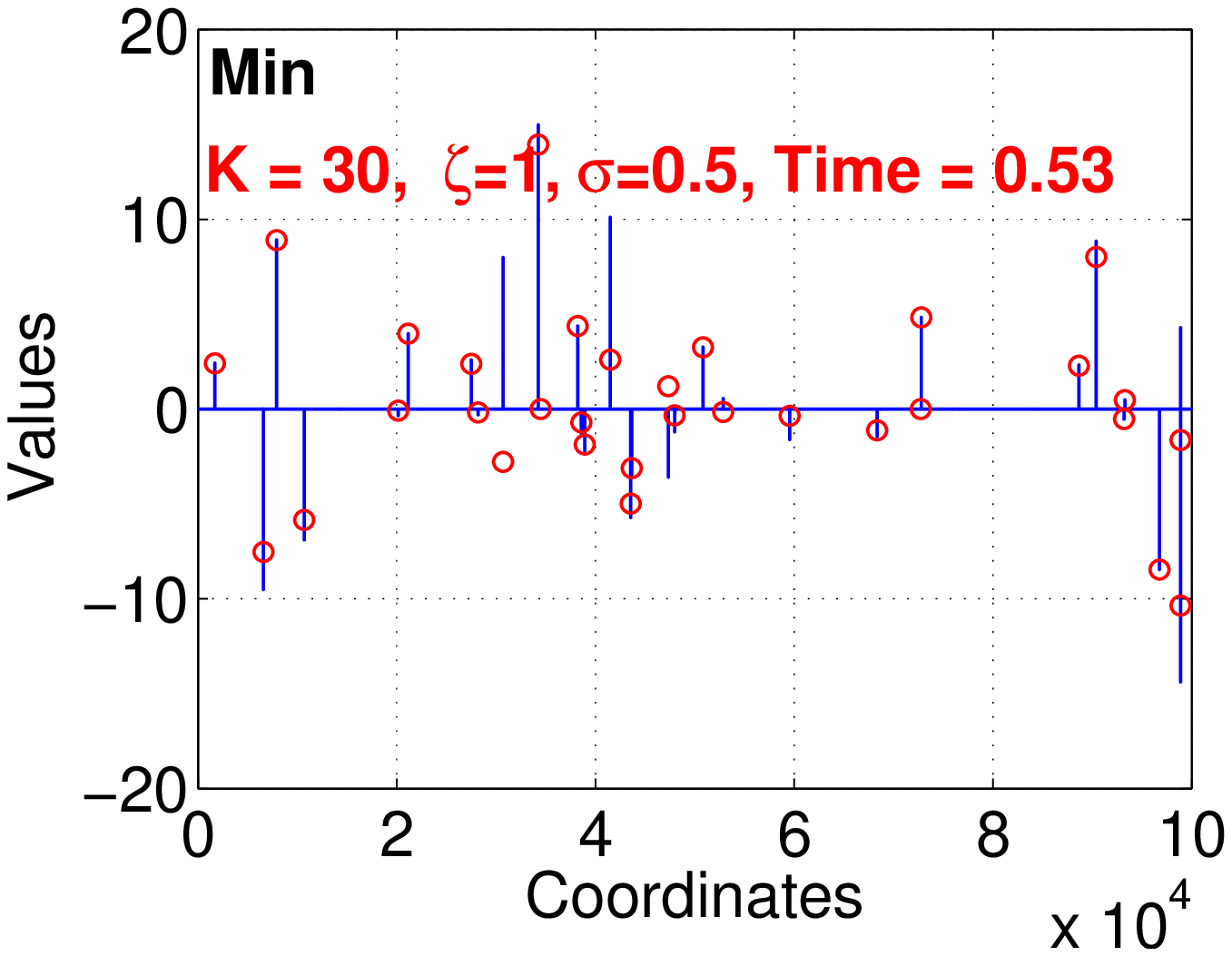}\hspace{0in}
\includegraphics[width=2.5in]{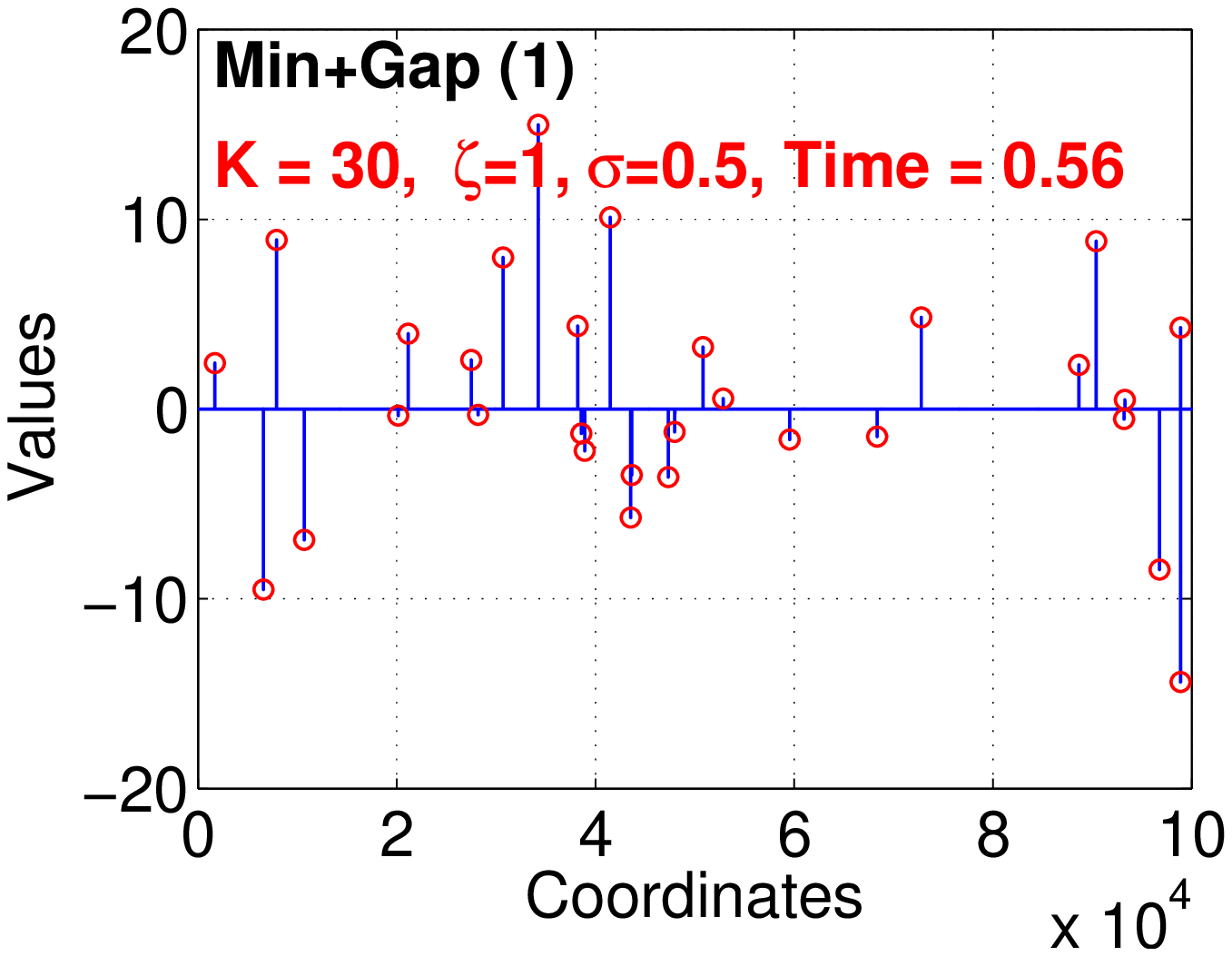}}

\mbox{
\includegraphics[width=2.5in]{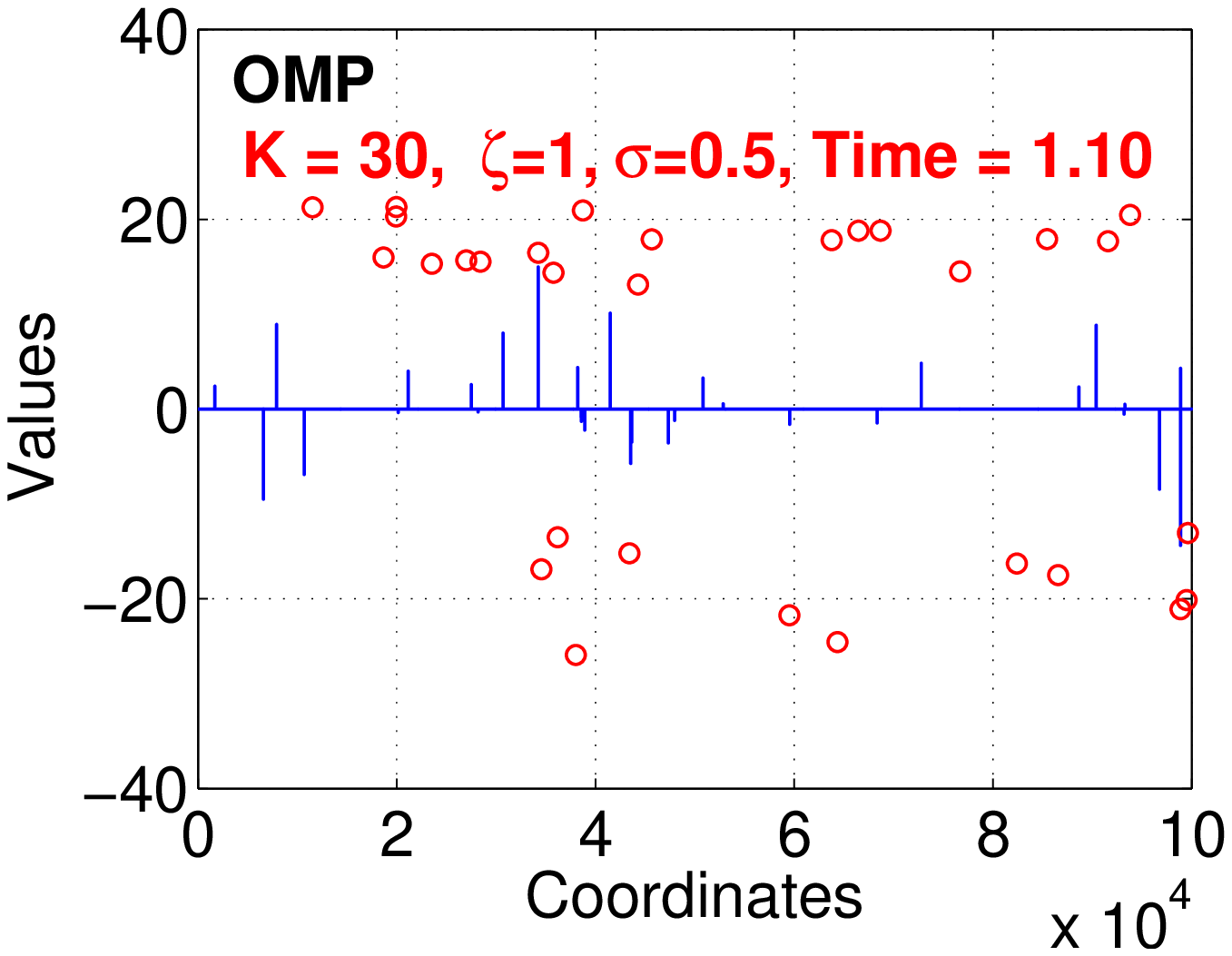}\hspace{0in}
\includegraphics[width=2.5in]{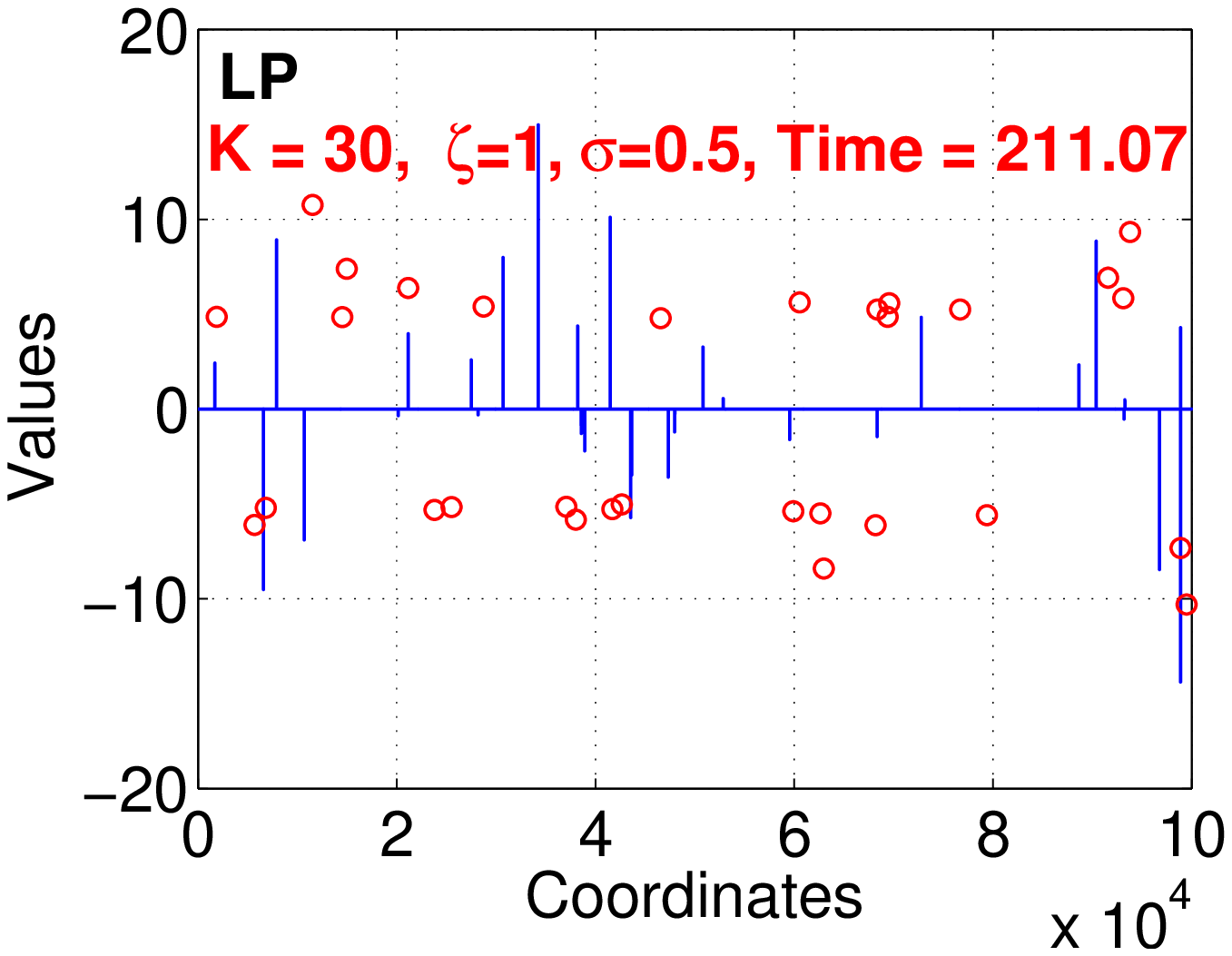}}
\end{center}
\vspace{-0.2in}
\caption{\small Reconstruction results from one simulation, using $N=100000$, $K=30$, $M=M_0$ (i.e., $\zeta=1$), $\sigma=0.5$, and sign signals. Our method (using just one iteration) can  perfectly reconstruct the signal.}\label{fig_RecGausB1Sig05}
\end{figure}

To provide the intuition, we first present a set of experiments with additive noise in Figures~\ref{fig_RecSignB1Sig01} to~\ref{fig_RecGausB1Sig05}.

With $N = 100000$, $K=30$, and $M=M_0$ (i.e., $\zeta =3$), we have seen in the simulations in Sec.~\ref{sec_simu} that all methods perform well, in both Sign and Gaussian signals. When we add additive noises with $\sigma = 0.1$ to the measurements, Figure~\ref{fig_RecSignB1Sig01} and Figure~\ref{fig_RecGausB1Sig01} show that our proposed method still achieves perfect recovery while LP and OMP fail. When we add more measurement noise by using $\sigma = 0.5$ in Figure~\ref{fig_RecSignB1Sig05} and Figure~\ref{fig_RecGausB1Sig05}, we observe that our method again achieves perfect recovery while both OMP and LP fail.

To understand why our method is insensitive to measurement noise, we can examine
\begin{align}
\frac{y_j+n_j}{s_{ij}} = x_i + \theta_i\frac{S_2}{S_1} + \frac{n_j}{S_1}
\end{align}
Without measurement noise, our algorithm utilizes observations with $S_2/S_1\approx 0$ to recover $x_i$, i.e., either $S_1$ is absolutely very large, or $S_1$ is large only relative to $S_2$. Because $S_1$ is extremely heavy-tailed, when $S_2/S_1\approx 0$, it is most likely $|S_1|$ is extremely large in the absolute scale. When $S_1$ is small, $\frac{n_j}{S_1}$ will be large but likely $\frac{S_2}{S_1}$ will be large as well (i.e., the observation would not  be useful anyway). This intuition explains why our method is essentially indifferent to measurement noise.

\subsection{Multiplicative Noise}

An analysis for multiplicative noise turns out to be easy and should also provide a good insight why our method is not sensitive to measurement noise.  For convenience, we consider the following model:
\begin{align}
\tilde{y}_j = \rho_j y_j = \rho_j =\rho_j \sum_{i=1}^N s_{ij} x_i, \ \ \ \rho_j > 0, \ \ j = 1, 2, ..., M
\end{align}
where $\rho_j$'s are assumed to be constants, to simplify the analysis. For example, when $\rho_j = 5$ (or $1/5$), this means the measurement $y_j$ is magnified (or shrunk) by a factor of 5. We assume that we still use the same minimum estimator $\hat{x}_{i,min} = \tilde{y}_t/s_{it}$, where $t = \text{argmin}_j \ \tilde{y}_j/s_{ij}$.

\begin{lemma}\label{lem_fpn}
Assume $\psi/\rho_j^{\frac{\alpha}{1-\alpha}}<1/3$, where   $\psi = \left(\frac{\epsilon}{\theta}\right)^{\frac{\alpha}{1-\alpha}}$ and  $\theta^\alpha = \sum_{i=1}^N |x_i|^\alpha$.  Then
\begin{align}
\mathbf{Pr}\left(|\hat{x}_{i,min}| > \epsilon,x_i=0\right) \leq \prod_{j=1}^M\frac{1}{\left(1+\psi/\rho_j^{\alpha/(1-\alpha)}\right)}.
\end{align}
\textbf{Proof:}\ The proof is analogous to the proof of Theorem~\ref{thm_fp}.\hspace{0.1in}$\hfill\square$

\end{lemma}

Note that $\rho_j^{\alpha/(1-\alpha)}\approx 1$ even for large (or small) $\rho_j$ values. In other words, the false positive error probability is  virtually not affected by the measurement noise.
\clearpage

\section{Combining L0 and L2 Projections}

There have been abundant of studies of compressed sensing using the Gaussian design matrix. It is a natural idea to combine these two types of projections. There are several obvious options.

For example, suppose we can afford to use  $M = M_0 = K\log ((N-K)/\delta)$ measurements to  detect all the nonzero coordinates with essentially no false positives. We can then use additional $K$ (or slightly larger than  $K$) Gaussian measurements to recover the magnitudes of the (candidates of)  nonzero coordinates via one least square. This option is simple and will require $M_0+K$ total measurements.\\

We have experimented with another idea. First, we use $M=M_0/4$  measurements and Alg.~\ref{alg_recovery} with gap estimators and iterations. Because the number of measurements may not be large enough, there will be a small number of undetermined coordinates after the procedure. We can apply additional $K/2$  Gaussian measurements and the LP decoding on the set of the undetermined coordinates. We find this approach also produces excellent recovery accuracy, with about $M_0/4 + K/2$ total measurements.\\

\begin{figure}[h!]
\begin{center}
\mbox{
\includegraphics[width=2.5in]{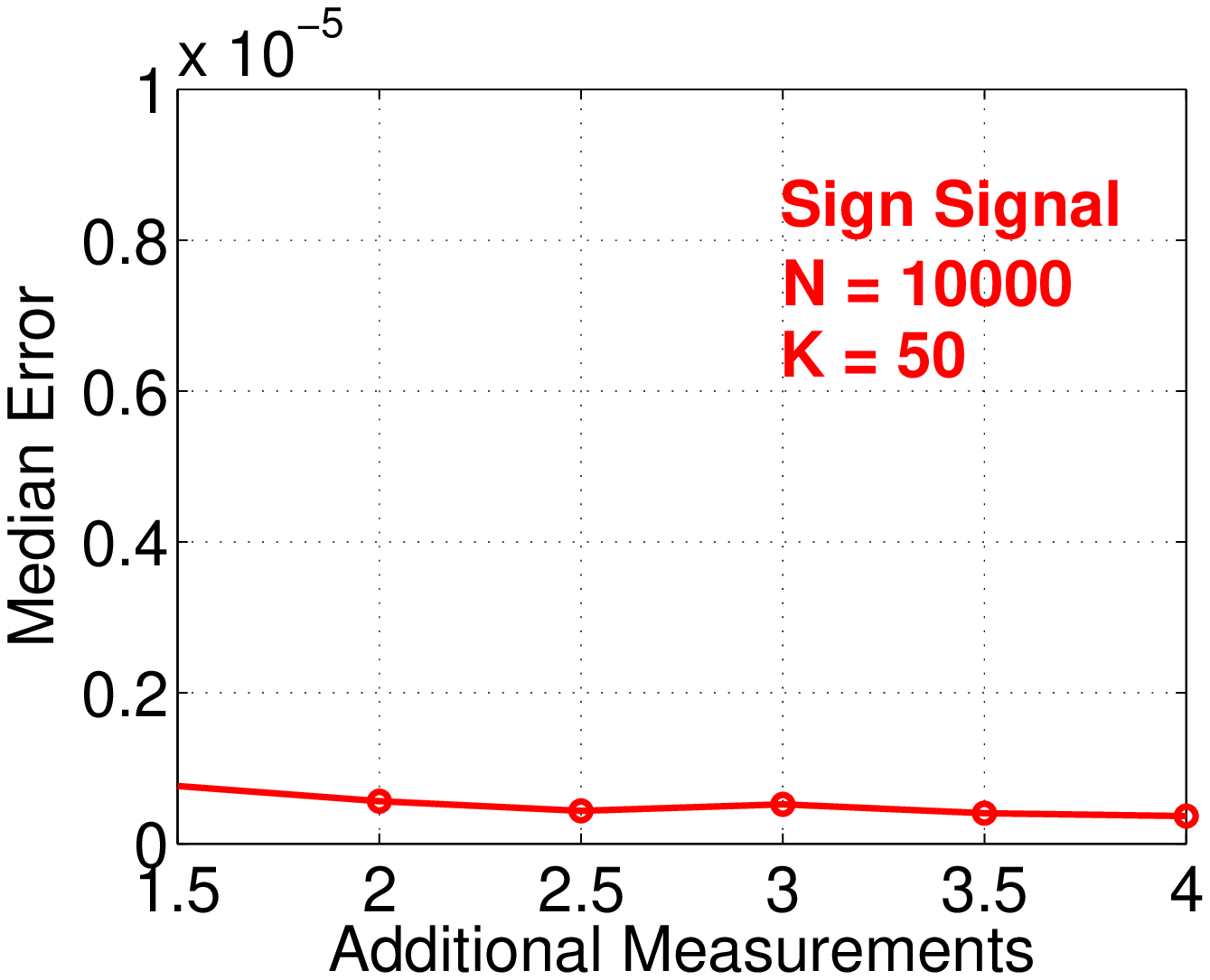}
\includegraphics[width=2.5in]{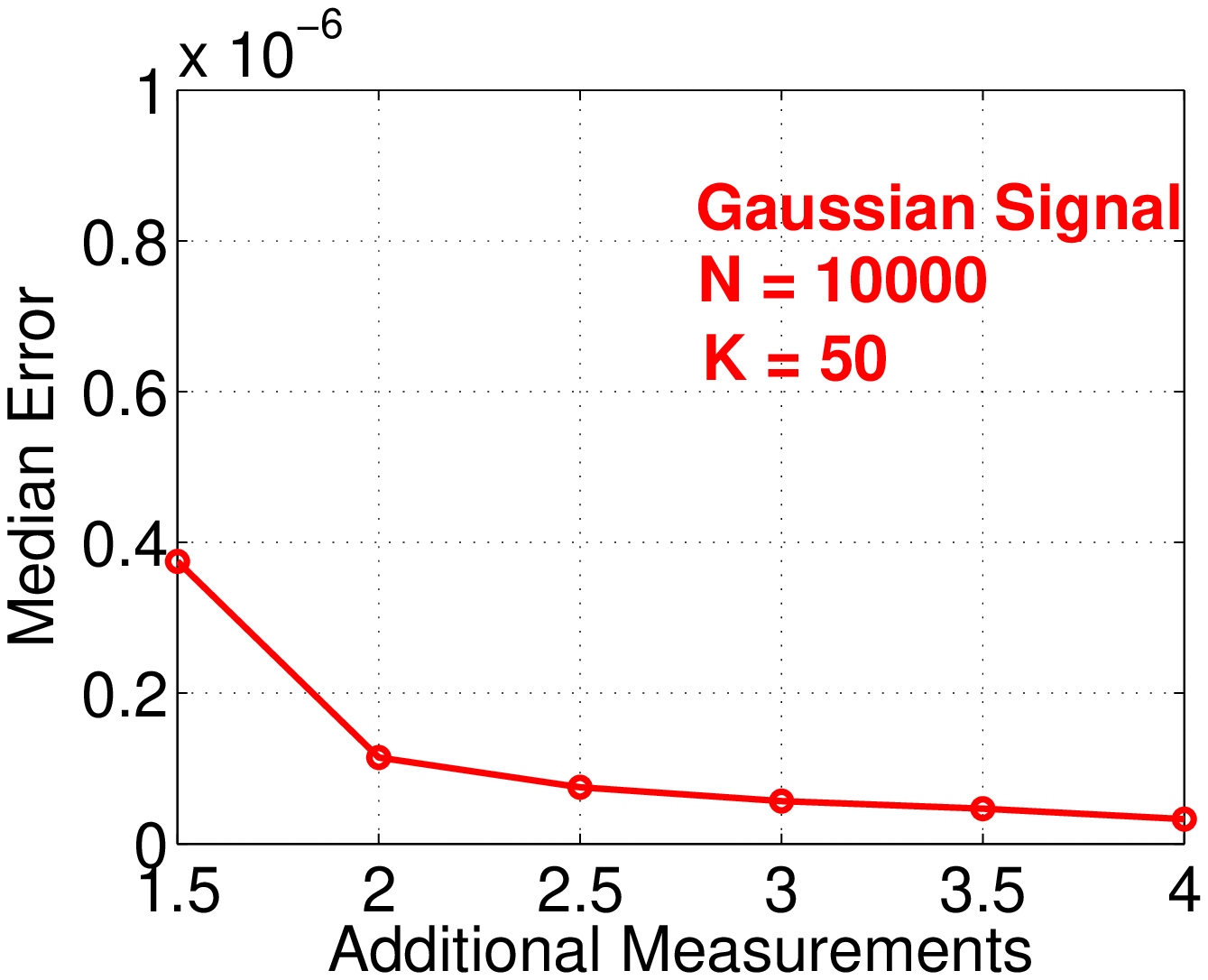}}

\mbox{
\includegraphics[width=2.5in]{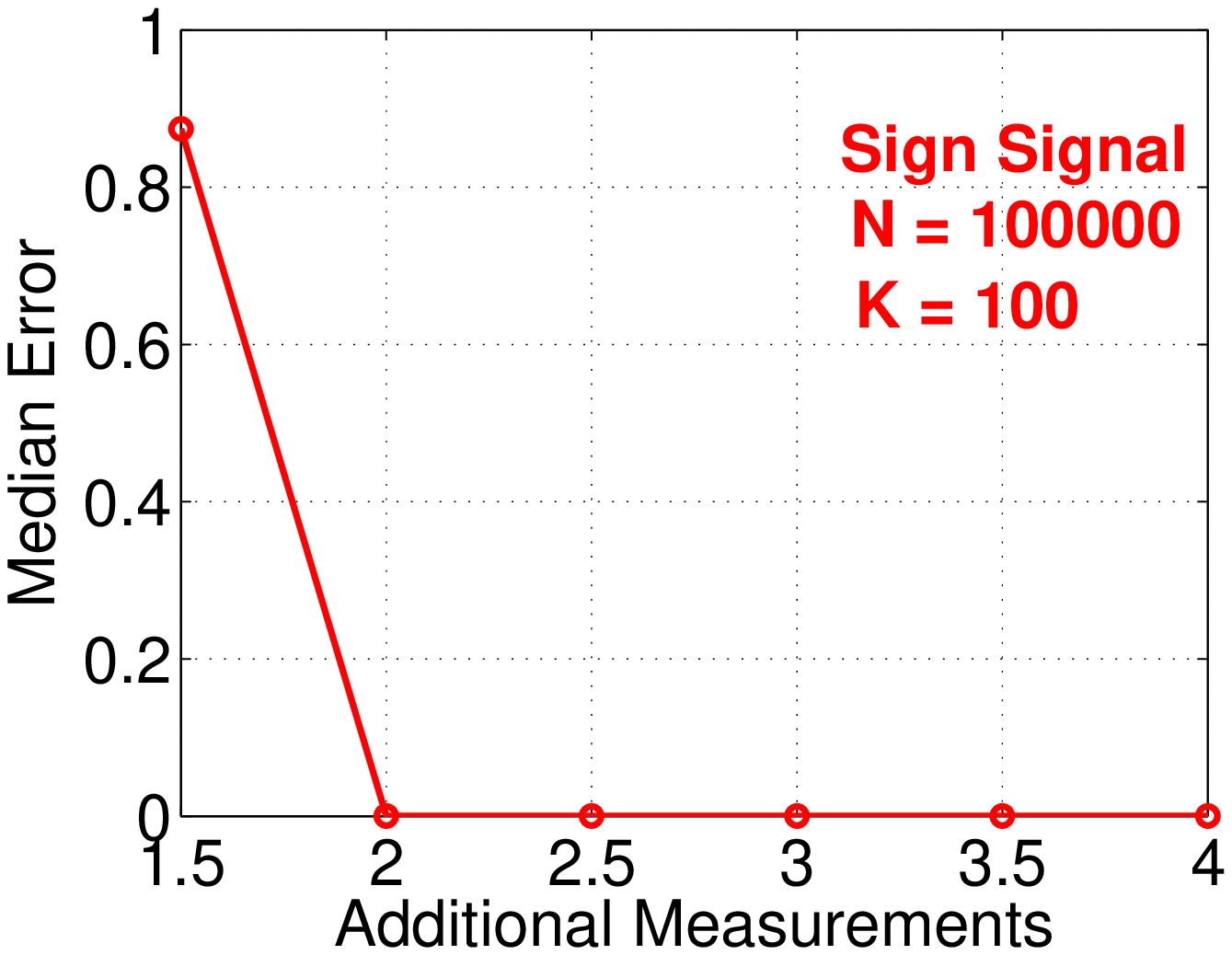}
\includegraphics[width=2.5in]{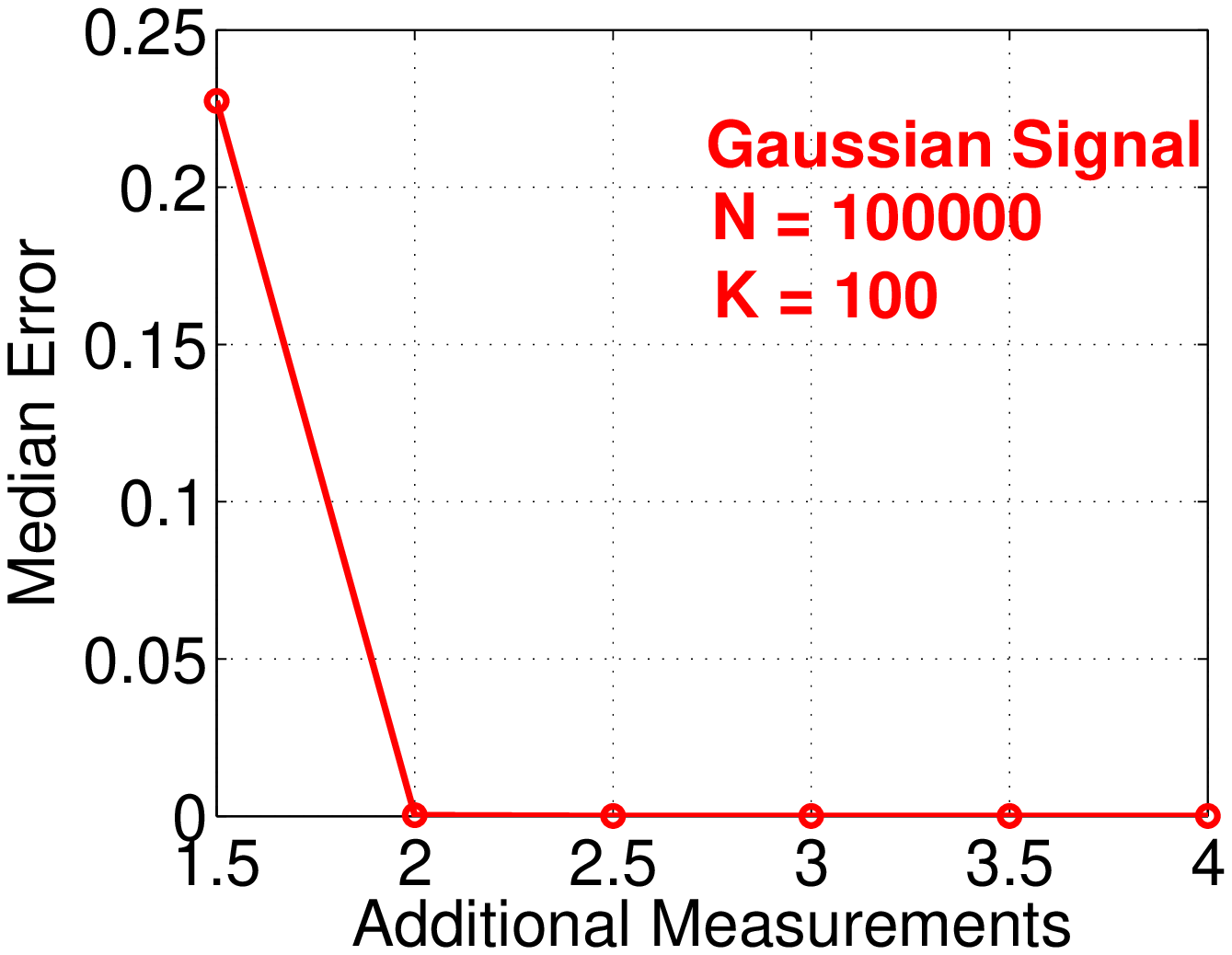}}

\end{center}
\vspace{-0.2in}
\caption{ Median reconstruction errors.  We first apply $M=M_0/2$ (i.e., $\zeta = 2$) measurements and the minimum estimator $\hat{x}_{i,min}$ to detect the nonzero coordinates and then select the top-$T$ coordinates ranked by $|\hat{x}_{i,min}|$, where $T=1.5K, 2K, 2.5K, 3K, 3.5K, 4K$, without using the threshold $\epsilon$. We use additional $T$ Gaussian measurements and apply the LP decoding on these top-$T$ coordinates. }\label{fig_ErrT}
\end{figure}

Here, we present some interesting experimental results on one more idea. That is, we use $M=M_0/2$ measurements and hence there will be a significant number of false positives detected by the minimum estimator $\hat{x}_{i,min}$. Instead of choosing a threshold $\epsilon$, we simply take the top-$T$ coordinates ranked by $|\hat{x}_{i,min}|$. We then use additional $T$ Gaussian measurements and  LP decoding. In Figure~\ref{fig_ErrT}, we let $T=1.5K, 2K, 2.5K, 3K, 3.5K, 4K$.  When $N=10000$ and $K=50$, even using only $T=1.5K$ additional Gaussian measurements produces excellent results. When $N=100000$, we need more( in this case $T=2K$) additional measurements, as one would expect.

\section{Future Work}

We anticipate that this paper is  a start of a new line of  research. For example, we expect that the following research projects (among many others) will be interesting and useful.
\begin{enumerate}
\item \textbf{Very sparse L0 projections.}\hspace{0.1in} Instead of using a dense design matrix $\mathbf{S}$ with entries sampled from i.i.d $S(\alpha,1)$, we can use an extremely sparse matrix by making (e.g.,) $99.9\%$ or more entries be zero. Furthermore, we can sample  nonzero entries from a symmetric $\alpha$-pareto distribution, i.e., a random variable $Z$ with $\mathbf{Pr}\left(|Z|>t\right) = \frac{1}{t^\alpha}$. These efforts will significantly speed up the processing and simplify the hardware design. This is inspired by the work on {\em very sparse stable random projections}~\cite{Proc:Li_KDD07}.
\item \textbf{Correlated projections.}\hspace{0.1in} It might be possible to use multiple projections with different $\alpha$ values to further improve the performance of sparse recovery. Recall that we can generate different (and highly ``correlated'') $\alpha$-stable variables with the same set of uniform $u$ and exponential $w$ variables as in (\ref{eqn_stable_sample}). For example, there is a recent work on using correlated stable projections for entropy estimation~\cite{Proc:Li_Zhang_NIPS12}.
\end{enumerate}

\section{Conclusion}

Compressed sensing has been a highly active area of research, because numerous important applications can be formulated as sparse recovery problems, for example, anomaly detections.  In this paper, we present our first study of using L0 projections for exact sparse recovery. Our practical procedure, which consists of the minimum estimator (for detection),  the gap estimator (for estimation), and the iterative process,  is computationally very efficient. Our algorithm  is able to produce accurate recovery results with smaller number of measurements,  compared to two strong baselines (LP and OMP) using the traditional  Gaussian (or Gaussian-like) design matrix.  Our method utilizes the $\alpha$-stable distribution with $\alpha\approx 0$. In our experiments with Matlab,  in order for interested readers to easily reproduce our results, we  take $\alpha=0.03$ and  find no special storage structure is needed at this value of $\alpha$. Our algorithms are robust against measurement noises. In addition, our algorithm   produces stable  (partial) recovery results with no catastrophic failure even when the number of measurements is very small (e.g., $M\approx K$).

We also analyze an ``idealized'' algorithm by assuming $\alpha\rightarrow0$. For a signal with $K=2$  nonzero coordinates, merely 3 measurements are sufficient for exact recovery. For general $K$, our analysis reveals that about $5K$ measurements are sufficient regardless of the length of the signal vector.

Finally, we anticipate  this work will lead to  interesting new research problems, for example, very sparse L0 projections, correlated projections, etc, to further improve the algorithm.


\newpage

\appendix
\section{Proof of Lemma~\ref{lem_F_lower}}~\label{app_F_lower}

To show,  $\forall t\geq 0$,
\begin{align}\notag
&F_\alpha(t) = E\left(\frac{1}{1+Q_\alpha/t}\right)
 \geq \max\left\{\frac{1/2}{1+1/t},\ \frac{1+(1/t-3)\mathbf{Pr}(Q_\alpha\leq t)/2}{1+1/t}
\right\}
\end{align}
where
\begin{align}\notag
&F_\alpha(t) = \mathbf{Pr}\left(\left|{S_2}/{S_1}\right|^{\alpha/(1-\alpha)}\leq t\right),\\\notag
&\left|{S_2}/{S_1}\right|^{\alpha/(1-\alpha)} = Q_\alpha\frac{w_1}{w_2},\\\notag
&Q_\alpha = Q_\alpha(u_1,u_2) = \left|\frac{q_\alpha(u_2)}{q_\alpha(u_1)}\right|^{\alpha/(1-\alpha)}\\\notag
&q_\alpha(u)  = \frac{\sin(\alpha u)}{\cos^{1/\alpha} u}\left[\cos(u-\alpha u)\right]^{(1-\alpha)/\alpha}
\end{align}

\noindent\textbf{Proof:}\ \ Firstly, since $w_1$ and $w_2$ are independent $exp(1)$ variables, we have
\begin{align}\notag
F_\alpha(t)=E\left(Q_\alpha w_1/w_2\leq t|Q_\alpha\right) = E e^{-Q_\alpha w_1/t} = E\left(\frac{1}{1+Q_\alpha/t}\right)
\end{align}
Note that $\mathbf{Pr}\left(Q_\alpha\leq 1\right ) = 1/2$, and
\begin{align}\notag
F_\alpha(t) = E\left(\frac{1}{1+Q_\alpha/t}\right) \geq \int_0^1\frac{1}{1+z/t}d\mathbf{Pr}(Q_\alpha\leq z)\geq \frac{\mathbf{Pr}\left(Q_\alpha\leq 1\right)}{1+1/t} = \frac{1/2}{1+1/t}
\end{align}
For the other bound, we let $X = \log Q_\alpha$, which is symmetric about 0. This way, we can write $F_\alpha(t) = E\frac{t}{t+e^X}$. Note that $\frac{1}{t+e^X}$ is convex when $X\geq\log t$ (and hence Jensen's inequality applies). For now, we assume $0\leq t\leq 1$ (i.e., $\log t \leq 0$) and obtain
\begin{align}\notag
F_\alpha(t) =& E\frac{t}{t+e^X} = E\left\{\frac{t}{t+e^X}I(X<\log t)\right\} + E\left\{\frac{t}{t+e^X}I(X\geq \log t)\right\}\\\notag
\geq&E\left\{\frac{t}{t+e^X}I(X<\log t)\right\} + E\left\{\frac{t}{t+e^X}I(|X|\leq|\log t|)\right\}\\\notag
\geq&E\left\{\frac{t}{t+t}\mathbf{Pr}(Q_\alpha< t)\right\} + \frac{t\mathbf{Pr}(|X|\leq|\log t|)}{t+e^{E\{X|I(|X|\leq|\log t|)\}}}\\\notag
=& \mathbf{Pr}(Q_\alpha< t)/2 + \frac{t\left(1-2\mathbf{Pr}(Q_\alpha<t)\right)}{t+e^{0}}\\\notag
=&\frac{1+\left(1/t-3\right)\mathbf{Pr}\left(Q_\alpha<t\right)/2}{1+1/t}
\end{align}
In particular, when $0\leq t\leq 1/3$, we have $F_\alpha(t) \geq \frac{1}{1+1/t}$. Also, note that when $t>1$,  $\frac{1/2}{1+1/t}$ is sharper than the other bound.

To prove, for any fixed $t>0$, $\lim_{\alpha\rightarrow 0} F_\alpha(t) = \frac{1}{1+1/t}$, we just need to use dominated convergence theorem and the fact that $Q_\alpha\rightarrow 1$ point-wise.  This completes the proof.

\section{Proof of Lemma~\ref{lem_F_upper}}\label{app_F_upper}
To show, if $0<\alpha\leq1/3$, then
\begin{align}\notag
F_\alpha(t) \leq&C_\alpha  t^{\frac{1-\alpha}{1+\alpha}}\max\{1, t^{\frac{2\alpha}{1+\alpha}}\}
\end{align}
where
\begin{align}\notag
&C_\alpha = \mu_1\mu_2 + \frac{1}{\pi}\left(\mu_2(1-\alpha)\right)^{\frac{1-\alpha}{1+\alpha}}
 \left(\frac{1-\alpha}{\alpha}\right)^{\frac{2\alpha}{1+\alpha}} \left(\frac{1+\alpha}{1-\alpha}\right)\\\notag
&\mu_1 =\frac{1}{\pi} \frac{\Gamma\left(1/(2-2\alpha)\right)\Gamma\left((1-3\alpha)/(2-2\alpha)\right)}{\Gamma\left((2-3\alpha)/(2-2\alpha)\right)}\\\notag
&\mu_2 = 1/\cos\left(\pi\alpha/(2-2\alpha)\right)
\end{align}

\noindent\textbf{Proof:}\ \ We need to first find a good lower bound of $Q_\alpha$, where
\begin{align}\notag
&Q_\alpha =\left|\frac{q_\alpha(u_2)}{q_\alpha(u_1)}\right|^{\alpha/(1-\alpha)} = \frac{|\sin(\alpha u_2)|^{\alpha/(1-\alpha)}\cos(u_2-\alpha u_2)\cos^{1/(1-\alpha)} u_1}{|\sin(\alpha u_1)|^{\alpha/(1-\alpha)}\cos(u_1-\alpha u_1)\cos^{1/(1-\alpha)} u_2}
\end{align}
We will make use of the following inequalities, when $|u|\leq \pi/2$,
\begin{align}\notag
&\alpha|\sin u| \leq |\sin\alpha u|\leq \alpha|u|,\ \ \ \cos(u-\alpha u) \geq \cos u,\ \ \
|\sin(\alpha u)| \leq \alpha |\tan(u)|\\\notag
&\frac{\cos(u-\alpha u)}{(\cos u)^{1/(1-\alpha)}} = \frac{\cos(\alpha u)+\tan(u)\sin(\alpha u)}{(\cos u)^{\alpha/(1-\alpha)}}\leq \frac{1+\alpha|\tan u|}{(\cos u)^{\alpha/(1-\alpha)}}
\end{align}
To see $|\sin(\alpha u)| \leq \alpha |\tan(u)|$, consider, $\forall u\in[0,\pi/2]$, $(\alpha\tan u - \sin(\alpha u))^\prime = \alpha\sec^2u - \alpha \cos(\alpha u)\geq 0$.

We can bound $Q_\alpha$ as follows:
\begin{align}\notag
Q_\alpha \geq& \left|\frac{\sin u_2}{\tan u_1}\right|^{\alpha/(1-\alpha)}\frac{(\cos u_1)^{\alpha/(1-\alpha)}}{1+\alpha|\tan u_1|} \frac{1}{(\cos u_2)^{\alpha/(1-\alpha)}}\\\notag
=&\left|\frac{\tan u_2\cos u_1}{\tan u_1}\right|^{\alpha/(1-\alpha)}\frac{(\cos u_1)^{\alpha/(1-\alpha)}}{1+\alpha|\tan u_1|} \\ \notag
\geq&\frac{\left|\frac{\tan u_2}{\tan u_1}\right|^{\alpha/(1-\alpha)}}{(1+\alpha|\tan u_1|)(1+\tan^2u_1)^{\alpha/(2-2\alpha)}}\\\notag
=&\frac{1}{(1+\alpha|X_1|)(1+X_1^2)^{\alpha/(2-2\alpha)}\left|X_1X_2\right|^{\alpha/(1-\alpha)}}
\end{align}

Because $u_1$  and $u_2$ are i.i.d. $unif(-\pi/2,\pi/2)$, we know that $X_1=\tan u_1$ and $X_2 = 1/\tan u_2$ are i.i.d. standard Cauchy variables. Therefore,
\begin{align}\notag
F_\alpha(t) =& E\left(\frac{1}{1+Q_\alpha/t}\right) \leq E\left(\frac{(1+\alpha|X_1|)(1+X_1^2)^{\alpha/(2-2\alpha)}\left|X_1X_2\right|^{\alpha/(1-\alpha)}}
{(1+\alpha|X_1|)(1+X_1^2)^{\alpha/(2-2\alpha)}\left|X_1X_2\right|^{\alpha/(1-\alpha)}+1/t}\right)\\\notag
\leq&E\left(\frac{(1+\alpha|X_1|)(1+X_1^2)^{\alpha/(2-2\alpha)}\left|X_1\right|^{\alpha/(1-\alpha)}}
{(1+\alpha|X_1|)(1+X_1^2)^{\alpha/(2-2\alpha)}\left|X_1\right|^{\alpha/(1-\alpha)}+1/(t\mu_2)}\right),\hspace{0.5in}(\text{Jensen's Inequality})
\end{align}
where $\mu_2 = E|X_2|^{\alpha/(1-\alpha)}$. Note that $\frac{ax}{ax+b}$ is concave in $x$. Furthermore, for any $z>0$,
\begin{align}\notag
F_\alpha(t) \leq& \int_0^\infty\left(\frac{(1+\alpha x)(1+x^2)^{\alpha/(2-2\alpha)}x^{\alpha/(1-\alpha)}}
{(1+\alpha x)(1+x^2)^{\alpha/(2-2\alpha)}x^{\alpha/(1-\alpha)}+1/(t\mu_2)}\right)\frac{2/\pi}{1+x^2} dx\\\notag
\leq&\int_z^\infty\frac{2/\pi}{1+x^2}dx  + t\mu_2 \int_0^z (1+\alpha x)(1+x^2)^{\alpha/(2-2\alpha)}x^{\alpha/(1-\alpha)} \frac{2/\pi}{1+x^2} dx\\\notag
\leq&\frac{2/\pi}{z}  + t\mu_2\left\{\mu_1+\frac{\alpha}{\pi}\int_0^z (1+x^2)^{\alpha/(2-2\alpha)} \frac{x^{\alpha/(1-\alpha)}}{1+x^2} dx^2\right\}\\\notag
\leq&\frac{2/\pi}{z}  + t\mu_2\left\{\mu_1+\frac{\alpha}{\pi}\int_0^{z^2} (1+x)^{\alpha/(1-\alpha)-1} dx\right\}\\\notag
=&\frac{2/\pi}{z}  + {t\mu_2}\left\{\mu_1+\frac{1-\alpha}{\pi}\left\{ (1+z^2)^{\alpha/(1-\alpha)} -1\right\}\right\}\\\notag
=&t\mu_2\mu_1  + \frac{1}{\pi}\left\{\frac{2}{z}+t\mu_2(1-\alpha)\left\{ (1+z^2)^{\alpha/(1-\alpha)} -1\right\}\right\} \\\notag
\leq &t\mu_2\mu_1  + \frac{1}{\pi}\left\{\frac{2}{z}+{t\mu_2}(1-\alpha) z^{2\alpha/(1-\alpha)}\right\},\hspace{0.2in} \text{if } \frac{\alpha}{1-\alpha}\leq 1
\end{align}
where $\mu_1 = E\left(X_1^2+X_1^4\right)^{\alpha/(2-2\alpha)}$. The next task is to find the $z$ which minimizes this upper bound. Note that $\underset{z\geq 0}{\min}\{2/z+Tz^{2\alpha/(1-\alpha)}\} =T^{\frac{1-\alpha}{1+\alpha}}
 \left(\frac{1-\alpha}{\alpha}\right)^{\frac{2\alpha}{1+\alpha}} \left(\frac{1+\alpha}{1-\alpha}\right)$, attained at $z = \left(\frac{1-\alpha}{T\alpha}\right)^{\frac{1-\alpha}{1+\alpha}}$. Thus, we obtain
\begin{align}\notag
F_\alpha(t) \leq&t\mu_2\mu_1  +\frac{1}{\pi}\left(t\mu_2(1-\alpha)\right)^{\frac{1-\alpha}{1+\alpha}}
 \left(\frac{1-\alpha}{\alpha}\right)^{\frac{2\alpha}{1+\alpha}} \left(\frac{1+\alpha}{1-\alpha}\right)\end{align}
where, using  integral formulas~\cite[3.622.1,3.624.2]{Book:Gradshteyn_07} (assuming $\alpha/(1-\alpha)<0.5$)
\begin{align}\notag
&\mu_2 = E|X_2|^{\alpha/(1-\alpha)} = \frac{2}{\pi}\int_0^{\pi/2} \tan^{\alpha/(1-\alpha)} udu= 1/\cos\left(\pi\alpha/(2-2\alpha)\right)\\\notag
&\mu_1 = E\left(X_1^2+X_1^4\right)^{\alpha/(2-2\alpha)} =\frac{2}{\pi} \int_0^{\pi/2} \frac{\tan^{\alpha/(1-\alpha)} u}{\cos^{\alpha/(1-\alpha)} u} du = \frac{1}{\pi} \frac{\Gamma\left(1/(2-2\alpha)\right)\Gamma\left((1-3\alpha)/(2-2\alpha)\right)}{\Gamma\left((2-3\alpha)/(2-2\alpha)\right)}
\end{align}

Therefore, we can write
\begin{align}\notag
F_\alpha(t) \leq&C_\alpha  t^{\frac{1-\alpha}{1+\alpha}}\max\{1, t^{\frac{2\alpha}{1+\alpha}}\}
\end{align}
where
\begin{align}\notag
C_\alpha = \mu_1\mu_2 + \frac{1}{\pi}\left(\mu_2(1-\alpha)\right)^{\frac{1-\alpha}{1+\alpha}}
 \left(\frac{1-\alpha}{\alpha}\right)^{\frac{2\alpha}{1+\alpha}} \left(\frac{1+\alpha}{1-\alpha}\right)
\end{align}
Moreover, $C_\alpha\rightarrow 1+1/\pi$ as $\alpha\rightarrow 0$, $C_\alpha<1.5$ if $\alpha\leq 0.05$, and $C_\alpha<2$ if $\alpha\leq 0.16$. This completes the proof.

\section{Proof of Lemma~\ref{lem_eta}}\label{app_lem_eta}
Let $k>1$, $\gamma = (1-\alpha)/\alpha$, $1\leq c_0\leq 2$, $z_{i,j} = y_j/s_{ij}$, $t_{i,j} = \left(|z_{i,j} -x_i|/\theta_i\right)^{1/\gamma}$, and $\{[1], [2], ..., [M]\}$ a permutation of $\{1, 2, ..., M\}$ giving $t_{i,[1]} \leq t_{i,[2]}\leq ... \leq t_{i,[M]}$.  To show
\begin{align}\notag
&\mathbf{Pr}\left(\frac{|z_{i,[k+2]}| - |z_{i,[k+1]}|}{|z_{i,[2]} - z_{i,[1]}|}\leq 1, \ \frac{F_\alpha(t_{i,[1]})}{F_\alpha(t_{i,[2]})}\leq (c_0-1)^{1/\gamma}\right)
\leq \eta_{k,\gamma,c_0}\left(1 + \frac{1}{2k}\right)\\\notag
&\eta_{k,\gamma,c_0}= \min\left\{u\in(0,1): c_0\left(1-\left(\frac{u}{2k}\right)^{1/k}\right)^{\gamma} + \left(1-\frac{u}{2k}\right)^{\gamma}\leq 1\right\}
\end{align}

\noindent\textbf{Proof:}\ \ Suppose $F_\alpha(t_{i,[1]})/F_\alpha(t_{i,[2]}) \leq (c_0-1)^{1/\gamma}$. Recall that $F_\alpha(t) = E\frac{t}{t+Q_\alpha}$ with $Q_\alpha\geq 0$. Since $F_\alpha(s)/F_\alpha(t)\leq s/t$ for $s<t$, we know that $t_{i,(1)}^\gamma \geq (c_0-1)t_{i,[2]}^\gamma$; and hence
\begin{align}\notag
|z_{i,[2]} - z_{i,[1]}| = |z_{i,[2]}-x_i - z_{i,[1]}+x_i|\leq|z_{i,[2]}-x_i|+ |z_{i,[1]}-x_i| = \theta_i\left(t_{i,[2]}^\gamma + t_{i,[1]}^\gamma\right) \leq \theta_ic_0t_{i,[2]}^\gamma
\end{align}
Because $|z_{i,[k+2]}| - |z_{i,[k+1]}|\leq |z_{i,[k+2]}-z_{i,[k+1]}| \leq |z_{i,[k+2]}-x_i|+|z_{i,[k+1]}-x_i|$, it follows that
\begin{align}\notag
|z_{i,[k+2]}| - |z_{i,[k+1]}| \leq |z_{i,[2]}-z_{i,[1]}| \Longrightarrow t_{i,[k+2]}^\gamma - t_{i,[k+1]}^\gamma \leq c_0t_{i,[2]}^\gamma
\end{align}
Again, using $F_\alpha(s)/F_\alpha(t)\leq s/t$ for $s<t$ and $c_0>0$, we obtain
\begin{align}\notag
\frac{F_\alpha^\gamma\left(t_{i,[k+1]}\right) + c_0F_\alpha^\gamma\left(t_{i,[2]}\right)}{F_\alpha^\gamma\left(t_{i,[k+2]}\right)} \geq \frac{t_{i,[k+1]}^\gamma + c_0t_{i,[2]}^\gamma}{t_{i,[k+2]}^\gamma}
\end{align}
Therefore, we have
\begin{align}\notag
\frac{|z_{i,[k+2]}| - |z_{i,[k+1]}|}{|z_{i,[2]}-z_{i,[1]}|}\leq 1 \Longrightarrow F_\alpha^\gamma\left(t_{i,[k+2]}\right) - F_\alpha^\gamma\left(t_{i,[k+1]}\right) \leq c_0F_\alpha^\gamma\left(t_{i,[2]}\right)
\end{align}
Consider $a>0$, $b>0$, $a^\gamma + b^\gamma c_0 = 1$. We have
\begin{align}\notag
&\mathbf{Pr}\left\{F_\alpha^\gamma\left(t_{i,[k+1]}\right) + c_0F_\alpha^\gamma\left(t_{i,[2]}\right)\geq F_\alpha^\gamma\left(t_{i,[k+2]}\right)\right\}\\\notag
\leq&\mathbf{Pr}\left\{F_\alpha^\gamma\left(t_{i,[k+1]}\right) \geq a^\gamma F_\alpha^\gamma\left(t_{i,[k+2]}\right)\right\} + \mathbf{Pr}\left\{c_0F_\alpha^\gamma\left(t_{i,[2]}\right)\geq b^\gamma c_0F_\alpha^\gamma\left(t_{i,[k+2]}\right)\right\}\\\notag
=&\mathbf{Pr}\left\{F_\alpha\left(t_{i,[k+1]}\right) / F_\alpha\left(t_{i,[k+2]}\right)  \geq a \right\} + \mathbf{Pr}\left\{F_\alpha\left(t_{i,[2]}\right)/F_\alpha\left(t_{i,[k+2]}\right)\geq b \right\}\\\notag
\end{align}
Note that $F_\alpha\left(t_{i,[j]}\right)$, $j=1, 2, ..., M$  are order statistics of $M$ uniform variables in $unif(0,1)$. This means  $F_\alpha\left(t_{i,[j]}\right)/ F_\alpha\left(t_{i,[k+2]}\right)$ has the $Beta\left(j,k+2-j\right)$ distribution. Thus
\begin{align}\notag
&\mathbf{Pr}\left(F_\alpha\left(t_{i,[k+1]}\right) / F_\alpha\left(t_{i,[k+2]}\right)\geq a \right)\\\notag
 =&\frac{(k+1)!}{(k)!(0)!}\int_a^1 x^{k}(1-x)^{0}dx = k\int_a^1 x^k dx  = \frac{k}{k+1}\left(1-a^{k+1}\right)\\\notag
&\mathbf{Pr}\left(F_\alpha\left(t_{i,[2]}\right) / F_\alpha\left(t_{i,[k+2]}\right)\geq b \right)\\\notag
 =&\frac{(k+1)!}{(1)!(k-1)!}\int_b^1 x^{1}(1-x)^{k-1}dx = (k+1)(k)\int_0^{1-b} (1-x)(x)^{k-1}dx = (1-b)^k(1+kb)
\end{align}
Combining the results, we obtain
\begin{align}\notag
&\mathbf{Pr}\left(\frac{|z_{i,[k+2]}| - |z_{i,[k+1]}|}{|z_{i,[2]} - z_{i,[1]}|}\leq 1, \ \frac{F_\alpha(t_{i,[1]})}{F_\alpha(t_{i,[2]})}\leq (c_0-1)^{1/\gamma}\right)\\\notag
\leq&\mathbf{Pr}\left\{F_\alpha^\gamma\left(t_{i,[k+1]}\right) + c_0F_\alpha^\gamma\left(t_{i,[2]}\right)\geq F_\alpha^\gamma\left(t_{i,[k+2]}\right)\right\}\\\notag
\leq& \frac{k}{k+1}\left(1-a^{k+1}\right) + (1-b)^k(1+kb)
\end{align}
We choose $a = 1-\frac{\eta_k}{2k}$, $b=1-\left(\frac{\eta_k}{2k}\right)^{1/k}$, where
\begin{align}\notag
\eta_k = \eta_{k,\gamma,c_0} =\min\left\{u\in(0,1):\ c_0 \left(1-\left(\frac{u}{2k}\right)^{1/k}\right)^{\gamma}+\left(1-\frac{u}{2k}\right)^\gamma\leq 1\right\}
\end{align}
Therefore,
\begin{align}\notag
&\mathbf{Pr}\left(\frac{|z_{i,[k+2]}| - |z_{i,[k+1]}|}{|z_{i,[2]} - z_{i,[1]}|}\leq 1, \ \frac{F_\alpha(t_{i,[1]})}{F_\alpha(t_{i,[2]})}\leq (c_0-1)^{1/\gamma}\right)\\\notag
\leq& \frac{k}{k+1}\left(1-\left(1-\frac{\eta_k}{2k}\right)^{k+1}\right) + \frac{\eta_k}{2k}\left(1+k-k\left(\frac{\eta_k}{2k}\right)^{1/k}\right)\\\notag
\leq&\frac{k}{k+1}\frac{\eta_k}{2k}\left(k+1-\frac{\eta_k}{2k}\right) + \frac{\eta_k}{2k}\left(1+k-k\left(\frac{\eta_k}{2k}\right)^{1/k}\right)\\\notag
=&\eta_k + \frac{\eta_k}{2k}-\frac{\eta_k^2}{2k(k+1)}-\frac{\eta_k}{2}\left(\frac{\eta_k}{2k}\right)^{1/k}\\\notag
\leq&\eta_k\left(1 + \frac{1}{2k}-\frac{1}{2}\left(\frac{\eta_k}{2k}\right)^{1/k}\right)\\\notag
\leq&\eta_k\left(1 + \frac{1}{2k}\right)
\end{align}

This completes the proof.

\end{document}